%% file: avh.tex
\documentclass{article}

\usepackage{microtype}
\usepackage{graphicx}
\usepackage{subfigure}
\usepackage{booktabs}
\usepackage{nicefrac}
\usepackage{microtype}
\usepackage{graphicx}
\usepackage{subfigure}
\usepackage{caption}
\usepackage{todonotes}
\usepackage{amsthm}
\usepackage{amsmath,amsfonts,amssymb}
\usepackage{bm}
\usepackage{enumitem}
\usepackage{color}
\usepackage{wrapfig}

\usepackage{url}
\usepackage{hyperref}
\usepackage[accepted]{icml2020}

\newcommand{\eg}{\emph{e.g.}}
\newcommand{\ie}{\emph{i.e.}}

\renewcommand{\captionlabelfont}{\footnotesize}
\newtheorem{thm}{Theorem}
\newtheorem{defn}[thm]{Definition}
\newcounter{hyp}
\newtheorem{hypo}[hyp]{Hypothesis}
\DeclareMathOperator*{\argmax}{arg\,max}

\graphicspath{{figs/}}

\begin{document}

\twocolumn[
\icmltitle{Angular Visual Hardness}

\begin{icmlauthorlist}
\icmlauthor{Beidi Chen}{rice}
\icmlauthor{Weiyang Liu}{gatech}
\icmlauthor{Zhiding Yu}{nvidia}
\icmlauthor{Jan Kautz}{nvidia}
\icmlauthor{Anshumali Shrivastava}{rice}\\
\icmlauthor{Animesh Garg}{nvidia,toronto,vector}
\icmlauthor{Anima Anandkumar}{nvidia,caltech}
\end{icmlauthorlist}

\icmlaffiliation{rice}{Rice University}
\icmlaffiliation{gatech}{Georgia Institute of Technology}
\icmlaffiliation{nvidia}{NVIDIA}
\icmlaffiliation{toronto}{University of Toronto}
\icmlaffiliation{vector}{Vector Institute, Toronto}
\icmlaffiliation{caltech}{Caltech}

\icmlcorrespondingauthor{Beidi Chen}{beidi.chen@rice.edu}
\icmlcorrespondingauthor{Weiyang Liu}{wyliu@gatech.edu}
\icmlcorrespondingauthor{Zhiding Yu}{zhidingy@nvidia.com}

\icmlkeywords{Angular Visual Hardness, Uncertainty Estimation, Self-Training}

\vskip 0.3in
]

\printAffiliationsAndNotice{}

\begin{abstract}
Recent convolutional neural networks (CNNs) have led to impressive performance but often suffer from poor calibration. They tend to be overconfident, with the model confidence not always reflecting the underlying true ambiguity and hardness. In this paper, we propose angular visual hardness (AVH), a score given by the normalized angular distance between the sample feature embedding and the target classifier to measure sample hardness. We validate this score with an in-depth and extensive scientific study, and observe that CNN models with the highest accuracy also have the best AVH scores. This agrees with an earlier finding that state-of-art models improve on the classification of harder examples. We observe that the training dynamics of AVH is vastly different compared to the training loss. Specifically, AVH quickly reaches a plateau for all samples even though the training loss keeps improving. This suggests the need for designing better loss functions that can target harder examples more effectively. We also find that AVH has a statistically significant correlation with human visual hardness. Finally, we demonstrate the benefit of AVH to a variety of applications such as self-training for domain adaptation and domain generalization.
\end{abstract}

\input{section1_intro.tex}
\input{section2_related.tex}
\input{section3_discoveries.tex}
\input{section4_connections.tex}
\input{section5_applications.tex}
\input{section6_conclusions.tex}

\bibliography{avh}
\bibliographystyle{icml2020}

\input{section7_appendix.tex}

\end{document}

%% file: section1_intro.tex
\vspace{-0.5cm}
\section{Introduction}
Convolutional neural networks (CNNs) have achieved great progress on many computer vision tasks such as image classification~\cite{he2016deep,krizhevsky2012imagenet}, face recognition~\cite{sun2014deep,sphereface,liu2018learning}, and scene understanding~\cite{zhou2014learning,long2015fully}.
On certain large-scale benchmarks such as ImageNet, CNNs have even surpassed human-level performance~\cite{deng2009imagenet}. Despite the notable progress, CNNs are still far from matching human-level visual recognition in terms of robustness~\cite{goodfellow2014explaining,wang2018iterative}, adaptability~\cite{finn2017model} and few-shot generalizability~\citep{hariharan2017low,liu2019neural}, and could suffer from various biases. For example, ImageNet-trained CNNs are reported to be biased towards textures, and these biases may result in CNNs being overconfident, or prone to domain gaps and adversarial attacks~\citep{geirhos2019imagenet}. 

\begin{figure}[t]
\centering
\hspace{-0.2cm}
\includegraphics[width=0.9 \linewidth]{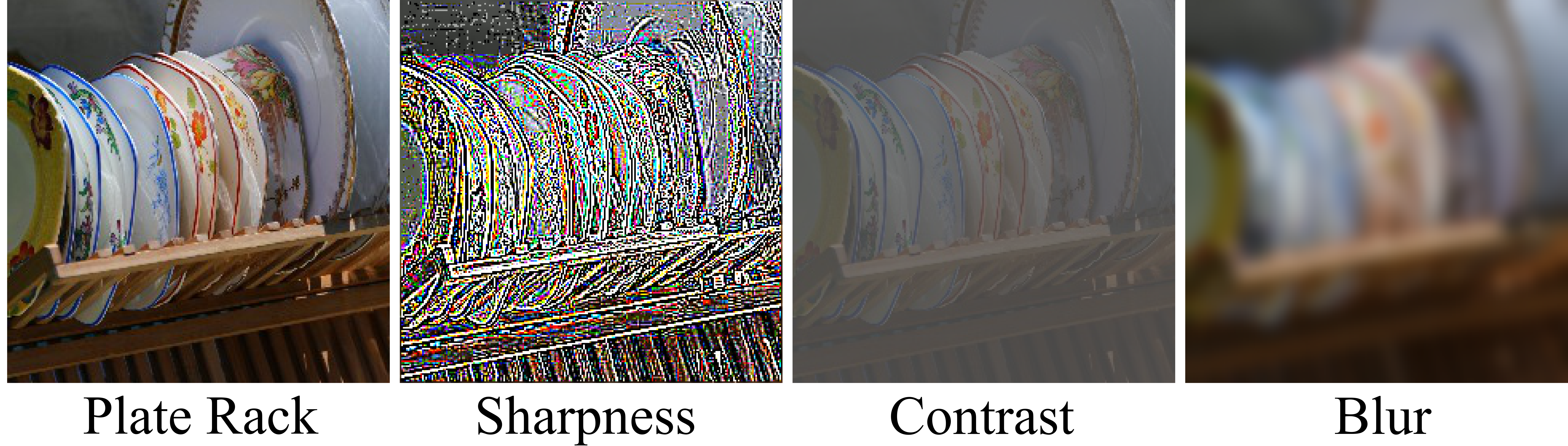}\\

\includegraphics[width=0.9 \linewidth]{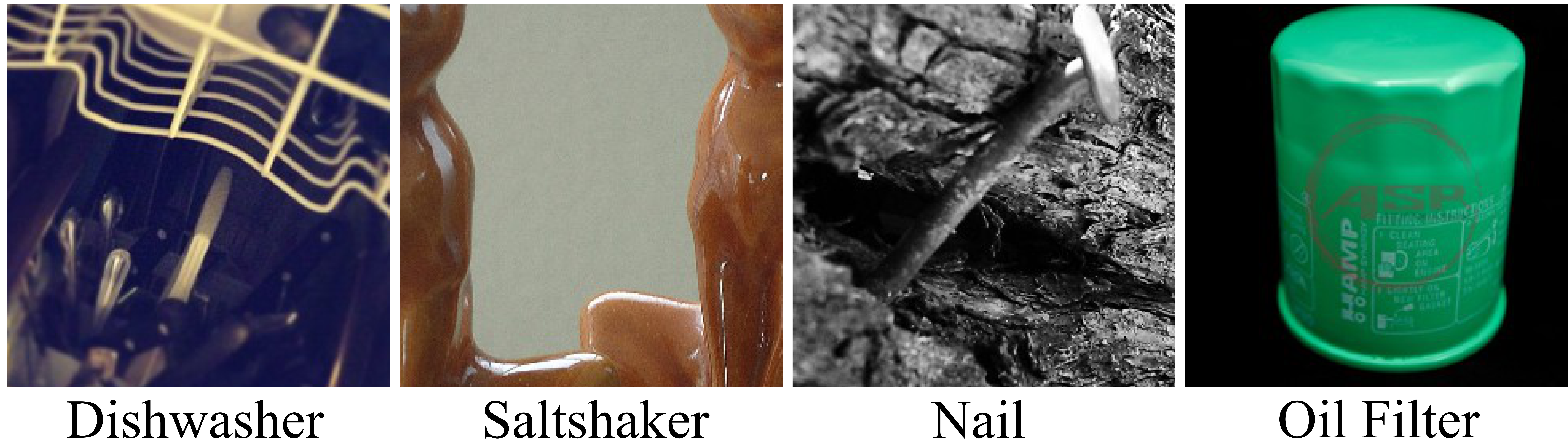}
\vspace{-0.32cm}
\caption{\footnotesize Example images that confuse humans. Top row: images with degradation. Bottom row: images with semantic ambiguity.}
\label{fig:confusion}
\vspace{-0.48cm}
\end{figure}

Softmax score has been widely used as a confidence measure for CNNs but it tends to give over-confident output~\citep{guo2017calibration, li2018reducing}. To fix this issue, one line of work considers confidence calibration from a Bayesian point of view~\cite{NIPS2016_6117,NIPS2017_7219}. Most of these methods tend to focus on the calibration and rescaling of model confidence by matching expected error or ensemble. But how much they are correlated with human confidence is yet to be thoroughly studied. On the other hand, several recent works~\citep{liu2016large,liu2017deep,liu2018decoupled} conjecture that softmax feature embeddings tend to naturally decouple into norms and angular distances that are related to intra-class confidence and inter-class semantic difference. Though inspiring, the conjecture lacks thorough investigation and we make surprising observations partially contradicting to the conjecture on intra-class confidence. This motivates us to conduct rigorous studies for reliable and semantics-related confidence measure.

Human vision is considered much more robust than current CNNs, but this does not mean humans cannot be confused. Many images appear ambiguous or hard for humans due to various image degradation factors such as lighting conditions, occlusions, visual distortions, etc. or due to semantic ambiguity in not understanding the label category, as shown in Figure~\ref{fig:confusion}. It is therefore natural to consider such human ambiguity or visual hardness on images as the gold-standard for confidence measures. However, explicitly encoding human visual hardness in a supervised manner is generally not feasible, since hardness scores can be highly subjective and difficult to obtain. Fortunately, a surrogate for human visual hardness was recently made available on the ImageNet validation set~\citep{recht2019imagenet}. This is based on \textbf{Human Selection Frequency} (HSF) - the average number of times an image gets picked by a crowd of annotators from a pool belonging to certain specified category. We adopt HSF as a surrogate for human visual hardness in this paper to validate our proposed angular hardness measure in CNNs.

\begin{figure}[t]
\centering
\hspace{-0.2cm}
\includegraphics[width=0.55 \linewidth]{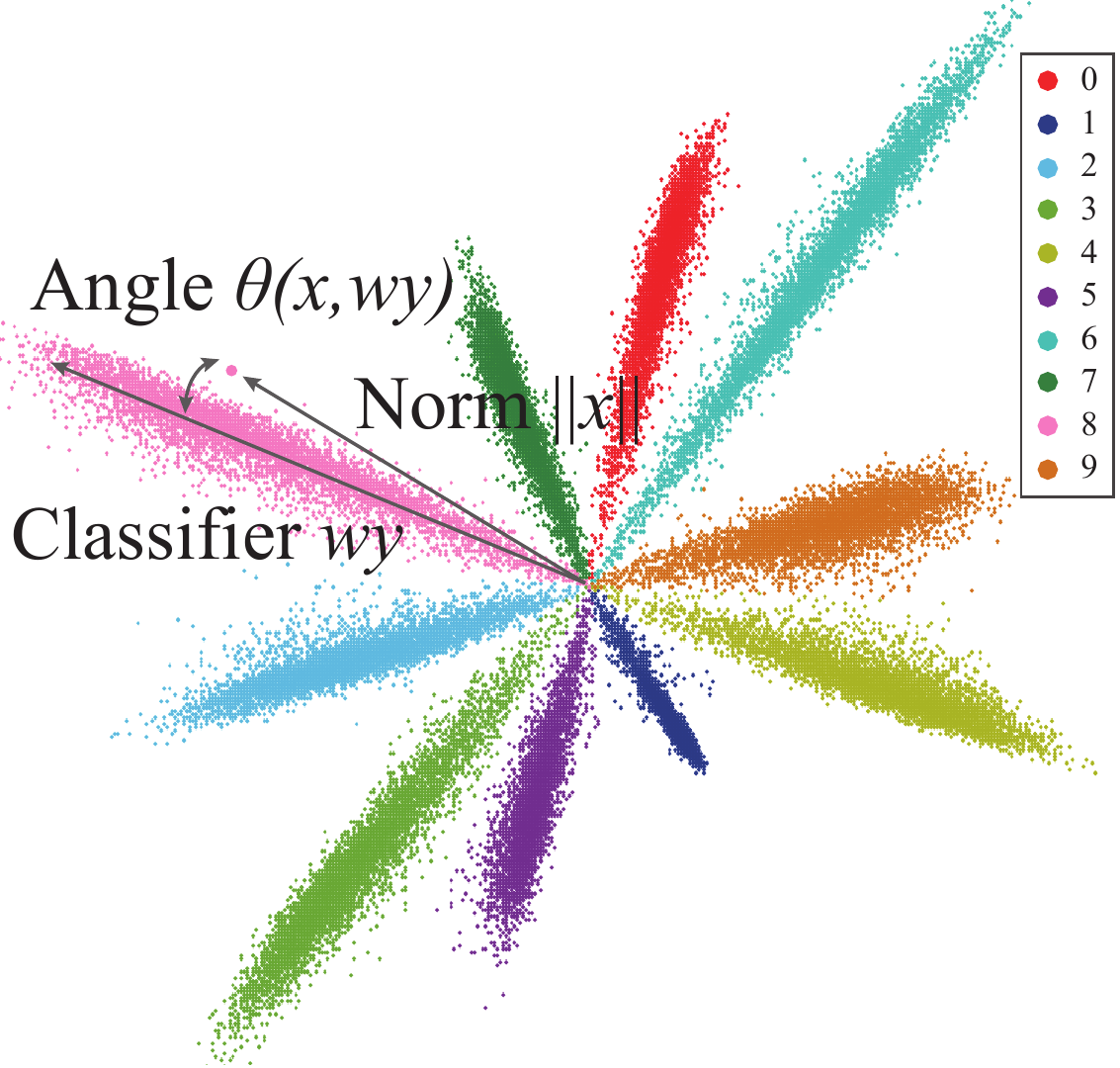}
\vspace{-0.32cm}
\caption{\footnotesize Visualization of embeddings on MNIST by setting their dimensions to 2 in a CNN.}
\label{fig:diagram}
\vspace{-0.48cm}
\end{figure}

\textbf{Contribution: Angular Visual Hardness (AVH).} Given a CNN, we propose a novel score function for measuring sample hardness. It is the normalized angular distance between the image feature embedding and the weights of the target category (See Figure~\ref{fig:diagram} as a toy example). The normalization takes into account the angular distances to other categories. 

We make observations on the dynamic evolution of AVH scores during ImageNet training. We find that AVH plateaus early in training even though the training (cross-entropy) loss keeps decreasing. This is due to the nature of parameterization in softmax loss, of which the minimization goes in two directions: either by aligning the angles between feature embeddings and classifiers or by increasing the norms of feature embeddings. We observe two phases popping up during training: (1) Phase 1, where the softmax improvement is primarily due to angular alignment, and later, (2) Phase 2, where the improvement is primarily due to significant increase in feature-embedding norms. 

The above findings suggest that the AVH can be a robust universal measure of hardness since angular scores are mostly frozen early in training. In addition, they suggest the need to design better loss functions over softmax loss that can improve performance on hard examples and focus on optimizing angles, \eg,~\cite{sphereface,deng2019arcface,wang2018cosface,wang2018additive}. We verify that better models tend to have better average AVH scores, which validates the argument in \citep{recht2019imagenet} that improving on hard examples is the key to improved generalization. We show that AVH has a statistically significant stronger correlation with human selection frequency than widely used confidence measures such as softmax score and embedding norm across several CNN models. This makes AVH a potential proxy of human perceived hardness when such information is not available.
 
Finally, we empirically show the superiority of AVH with its application to self-training for unsupervised domain adaptation and domain generalization. With AVH being an improved confidence measure, our proposed self-training framework renders considerably improved pseudo-label selection and category estimation, leading to state-of-the-art results with significant performance gain over baselines. Our proposed new loss function based on AVH also shows drastic improvement for the task of domain generalization.

%% file: section2_related.tex
\vspace{-1.9mm}
\section{Related Work}
\vspace{-0.65mm}
\textbf{Example hardness measures.}
An automatic detection of examples that are hard for human vision has numerous applications.~\citep{recht2019imagenet} showed that state-of-the-art models perform better on hard examples. This implies that in order to improve generalization, the models need to improve accuracy on hard examples. This can be achieved through various learning algorithms such as curriculum learning~\citep{bengio2009curriculum} and self-paced learning~\citep{kumar2010self} where being able to detect hard examples is crucial. Measuring sample confidence is also important in partially-supervised problems such as semi-supervised learning~\citep{zhu2007semi, zhou2012self}, unsupervised domain adaptation~\citep{chen2011co} and weakly-supervised learning~\citep{tang2017multiple} due to their under-constrained nature. Sample hardness can also be used to identify implicit distribution imbalance in datasets to ensure fairness and remove societal biases~\citep{buolamwini2018gender}.

\begin{figure*}[t]
\vspace{-2mm}
	\begin{center}
	\hspace{-0.5cm}
	\includegraphics[width=0.28 \textwidth]{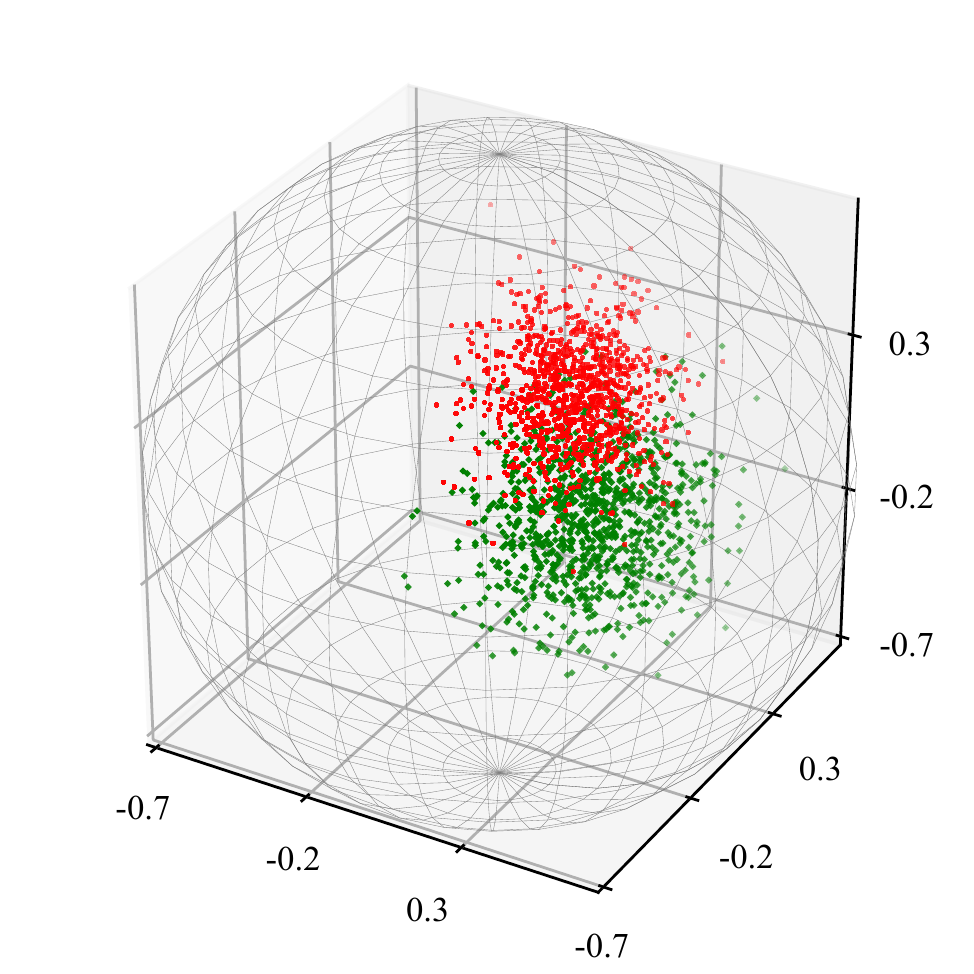}~~~~
	\includegraphics[width=0.28 \textwidth]{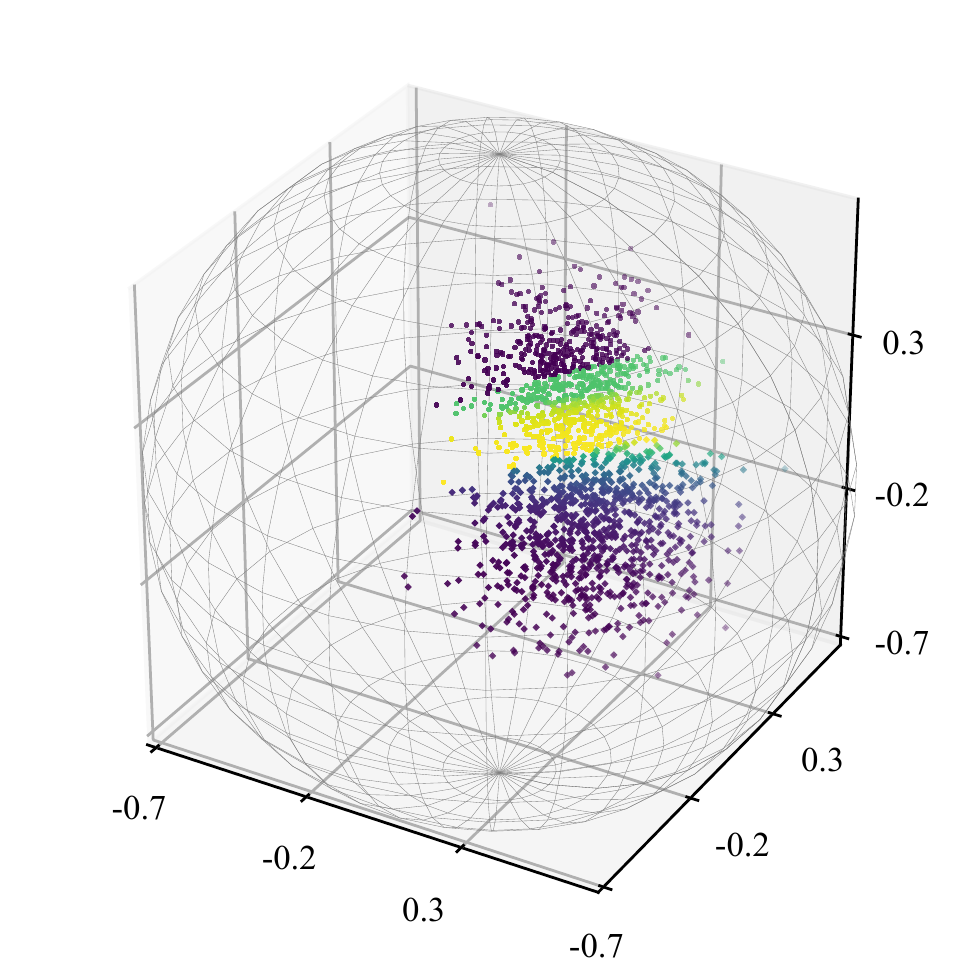}~~~~
	\includegraphics[width=0.28 \textwidth]{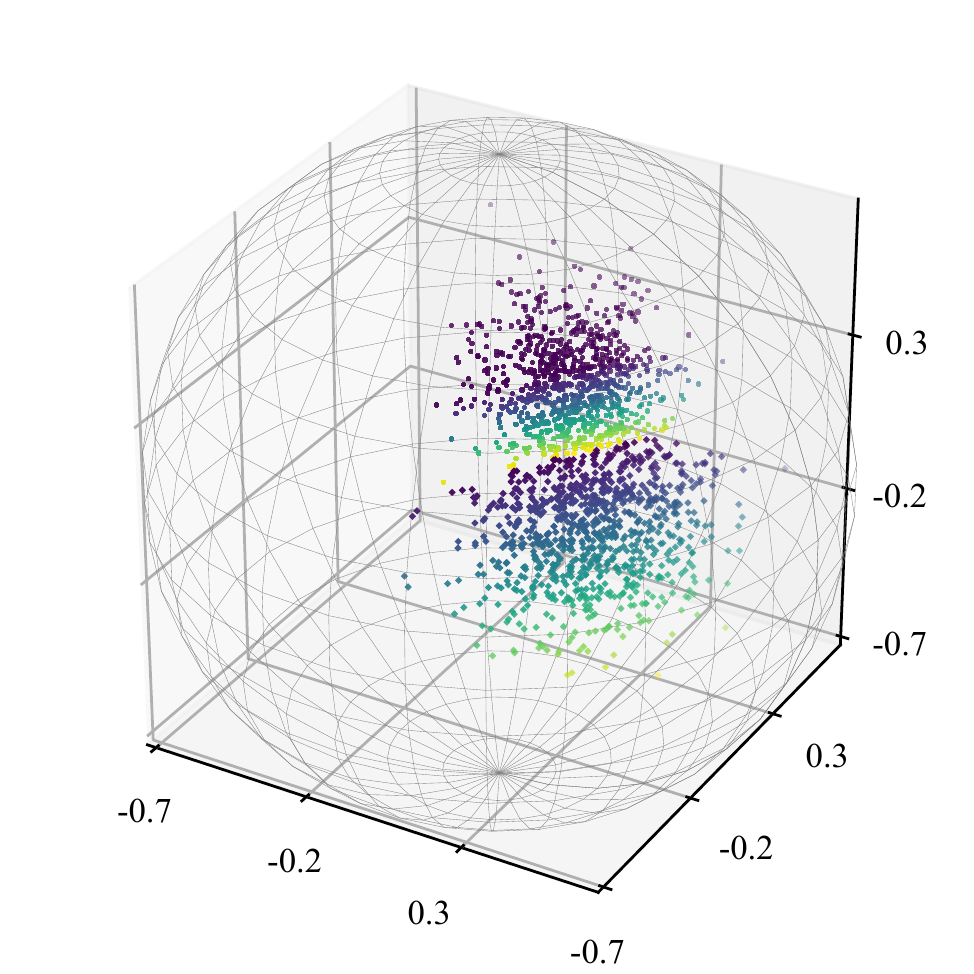}
	\end{center}
 	\vspace{-5mm}
	\caption{\footnotesize Toy example of two overlapping Gaussian distributions (classes) on a unit sphere. Left: samples from the distributions as input to a multi layer perceptron (MLP). Middle: AVH heat map produced by MLP, where samples in lighter colors (higher hardness) are mostly overlapping hard examples. Right: $\ell_2$-norm heat map, where certain non-overlapping samples also have higher values.}
	\label{fig:gaussian}
	\vspace{-2mm}
\end{figure*}

\vspace{-0.3mm}
\textbf{Angular distance in neural networks.}
\cite{zhang2018unreasonable} uses deep features to quantify the semantic difference between images, indicating that deep features contain the most crucial semantic information. It empirically shows that the angular distance between feature maps in deep neural networks is very consistent with the human in distinguishing the semantic difference. \cite{liu2017deep} proposes a hyperspherical neural network that constrains the parameters of neurons on a unit hypersphere and uses angular similarity to replace the inner product similarity. \cite{liu2018decoupled} proposes to decouple the inner product as norm and angle, arguing that norms correspond to intra-class variation, and angles corresponds to inter-class semantic difference. However, this work does not perform in-depth studies to prove this conjecture. Recent research~\cite{liu2018learning,Lin20CoMHE,liu2020orthogonal} comes up with an angle-based hyperspherical energy to characterize the neuron diversity and improve generalization by minimizing this energy.

\vspace{-0.4mm}
%\textbf{Deep model calibration and uncertainty Measure:}
\textbf{Deep model calibration.}
Confidence calibration aims to predict
%is the problem of predicting
probability estimates representative of the true correctness likelihood~\cite{guo2017calibration}. It is well-known that the deep neural networks tend to be mis-calibrated and there has been a rich set of literature trying to solve this problem~\cite{kumar2018trainable, guo2017calibration}. 
%While not claiming causality, they observe connections between model capacity/regularization and model calibration. However, none of them explain the problem from the perspective of training dynamics from different components in CNNs.
While establishing correlation between model confidence and prediction correctness, the connection to human confidence has not been widely studied from a training dynamics perspective.

\vspace{-0.4mm}
\textbf{Uncertainty estimation.}
In uncertainty estimation, two types of uncertainties are often considered: (1) \textit{Aleatoric} uncertainty which captures noise inherent in the observations; (2) \textit{Epistemic} uncertainty which accounts for uncertainty in the model due to limited data~\cite{der2009aleatory}. The latter is widely modeled by Bayesian inference~\cite{kendall2017uncertainties} and its approximation with dropout~\cite{gal2016dropout,gal2017concrete}, but often at the cost of additional computation. The fact that AVH correlates well with Human Selection Frequency indicates its underlying connection to aleatoric uncertainty. This makes it suitable for tasks such as self-training. Yet unlike Bayesian inference, AVH can be naturally computed during regular softmax training, making it convenient to obtain with only one-time training and a drop-in uncertainty measure for most existing neural networks.
% {\color{red}Moreover, most of the work in Bayesian deep learning, which is well-known for its uncertainty estimations, capture either epistemic uncertainty, or aleatoric uncertainty alone ~\cite{gal2016uncertainty}. Though recent work has made great progress on incorporate both uncertainty measure, it comes with significant computational cost~\cite{kendall2017uncertainties}.}

%% file: section3_discoveries.tex
\vspace{-2.25mm}
\section{Discoveries in CNN Training Dynamics}\label{sec:dynamics}
\vspace{-0.95mm}

\textbf{Notation.} Denote $\mathbb{S}^{n}$ as the unit $n$-sphere, \ie,
$\mathbb{S}^{n}=\{\bm{x}\in\mathbb{R}^{n+1}|\Vert \bm{x} \Vert_2=1\}$.  
Below by $\mathcal{A}(\cdot,\cdot)$, we denote the angular distance between two points on $\mathbb{S}^{n}$, \ie, $\mathcal{A}(\bm{u},\bm{v})=\arccos(\frac{\langle\bm{u},\bm{v}\rangle}{\|\bm{u}\|\|\bm{v}\|})$. 
Let $\bm{x}$ be the feature embedding input for the last layer of the classifiers in the pretrained CNNs (\eg, FC-1000 in VGG-19). Let $\mathcal{C}$ be the number of classes for a classification task. Denote $\bm{W} = \{\bm{w}_i| 0 < i \leq \mathcal{C}\}$ as the set of weights for all $\mathcal{C}$ classes in the final layer of the classifier.

\vspace{0.42mm}
\begin{defn}[Model Confidence]
We define Model Confidence on a single sample as the probability score of the true objective class output by the CNN models, $ \frac{e^{\bm{w}_y\bm{x}}}{\sum_{i=1}^{C}e^{\bm{w}_i\bm{x}}}$.
\end{defn}

\begin{defn}[Human Selection Frequency]
We define one way to measure human visual hardness on pictures as Human Selection Frequency (HSF). Quantitatively, given $m$ number of human workers in a labeling process described in~\cite{recht2019imagenet}, if $b$ out of $m$ label a picture as a particular class and that class is the target class of that picture in the final dataset, then HSF is defined as $\frac{b}{m}$.  
\end{defn}

\begin{figure*}[!t]
\vspace{-2.5mm}
	\begin{center}
		\includegraphics[width=0.32 \textwidth]{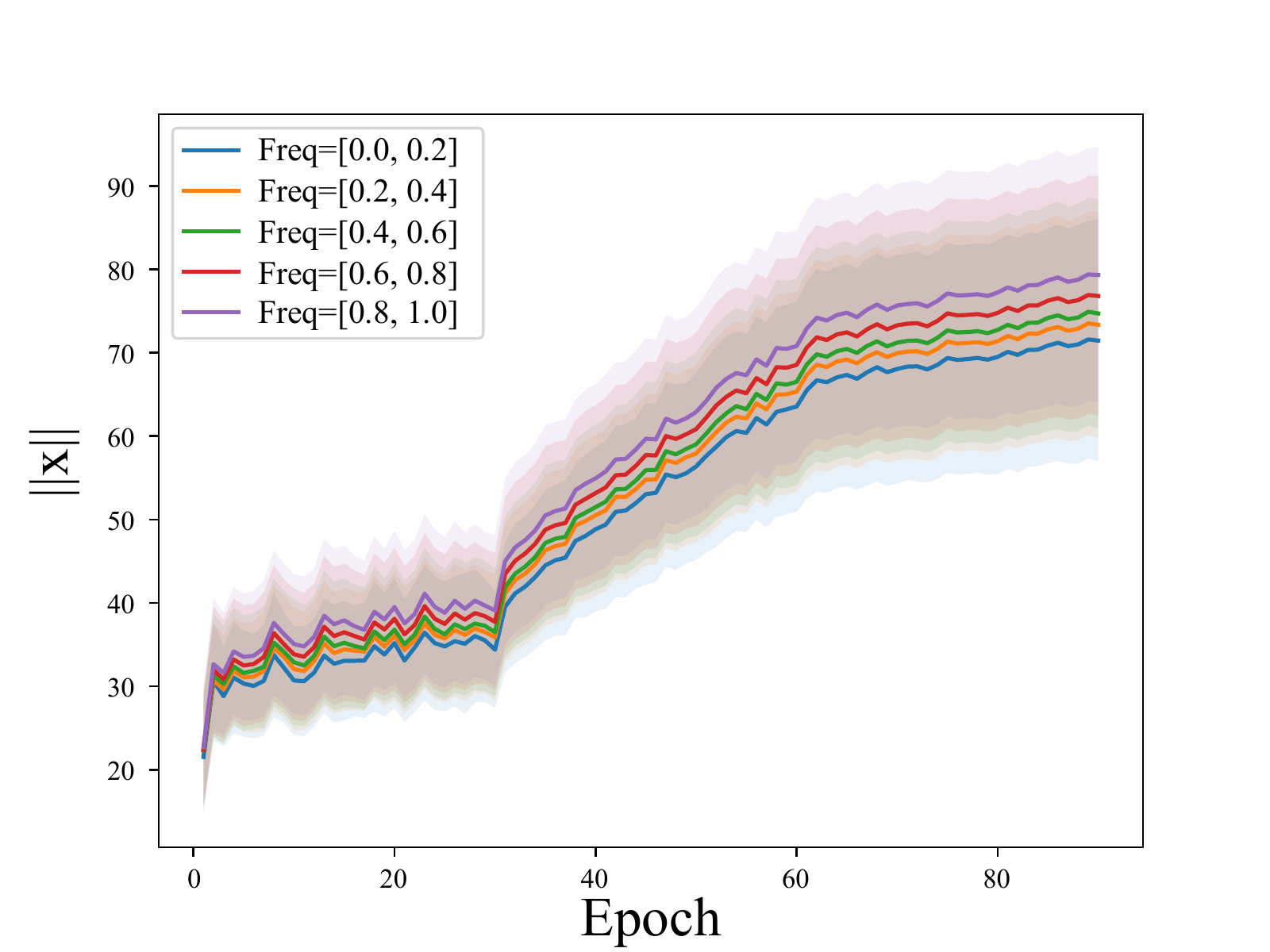}
		\includegraphics[width=0.32 \textwidth]{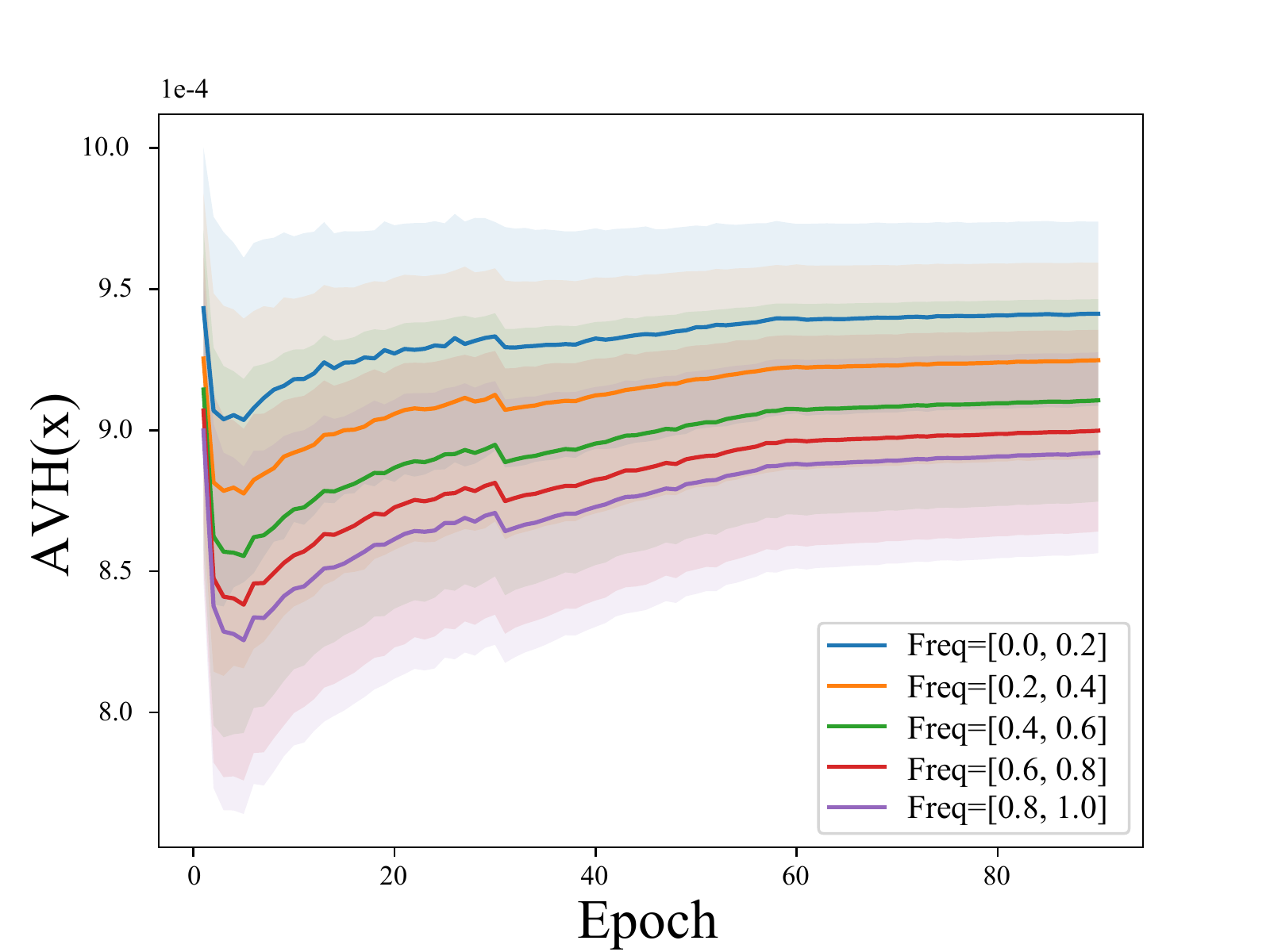}
		\includegraphics[width=0.32 \textwidth]{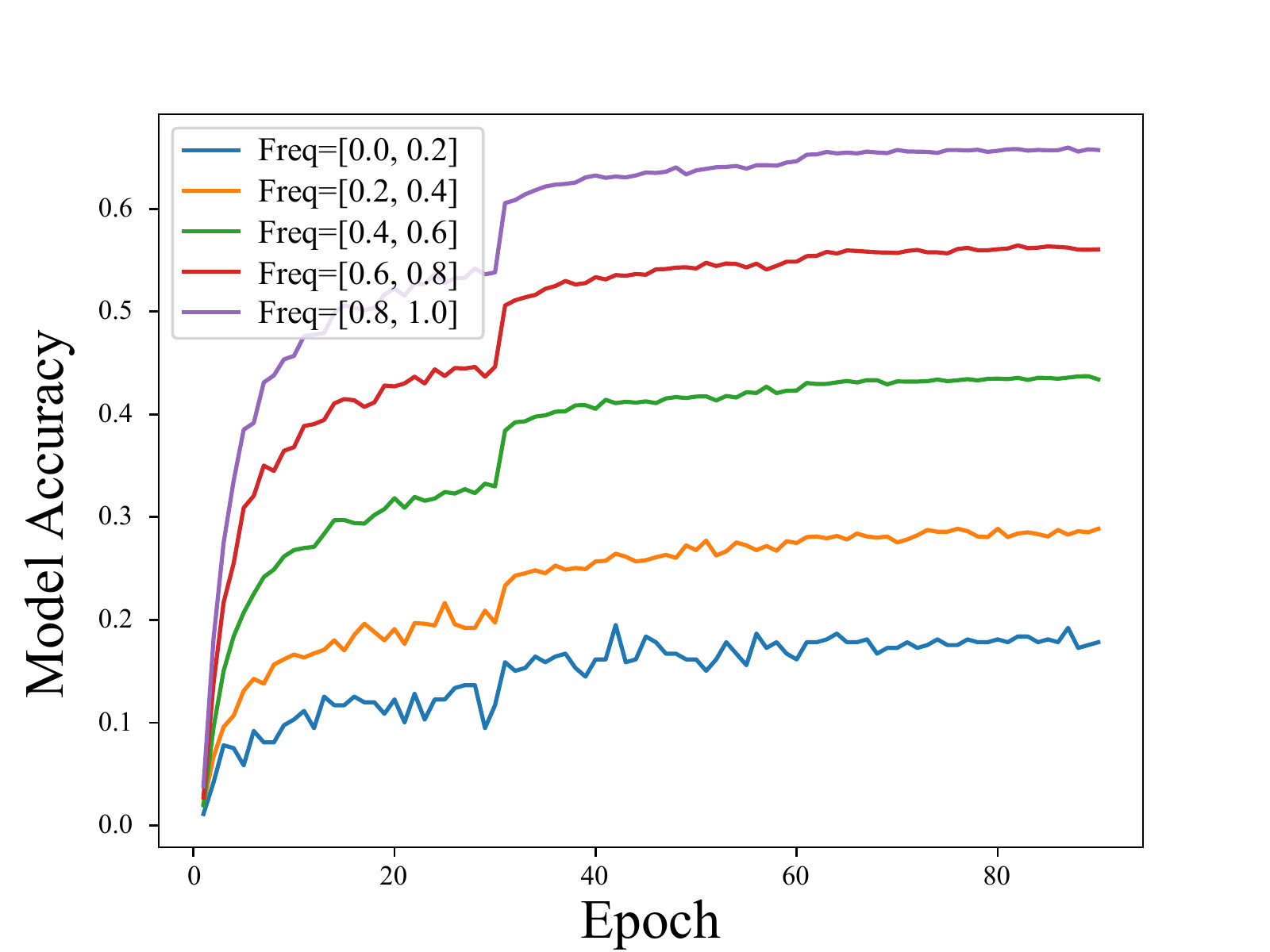}
		
		\includegraphics[width=0.32 \textwidth]{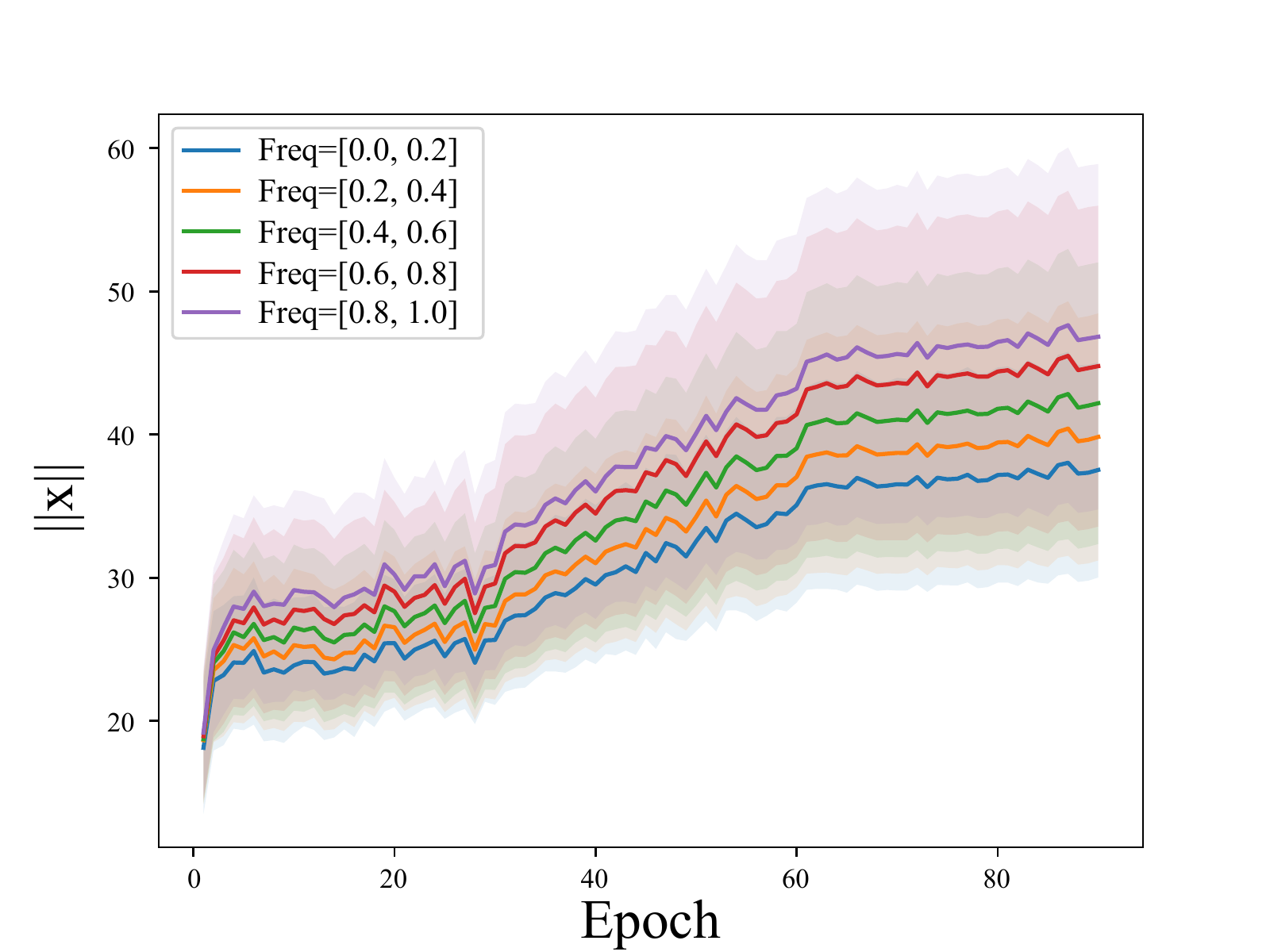}
		\includegraphics[width=0.32 \textwidth]{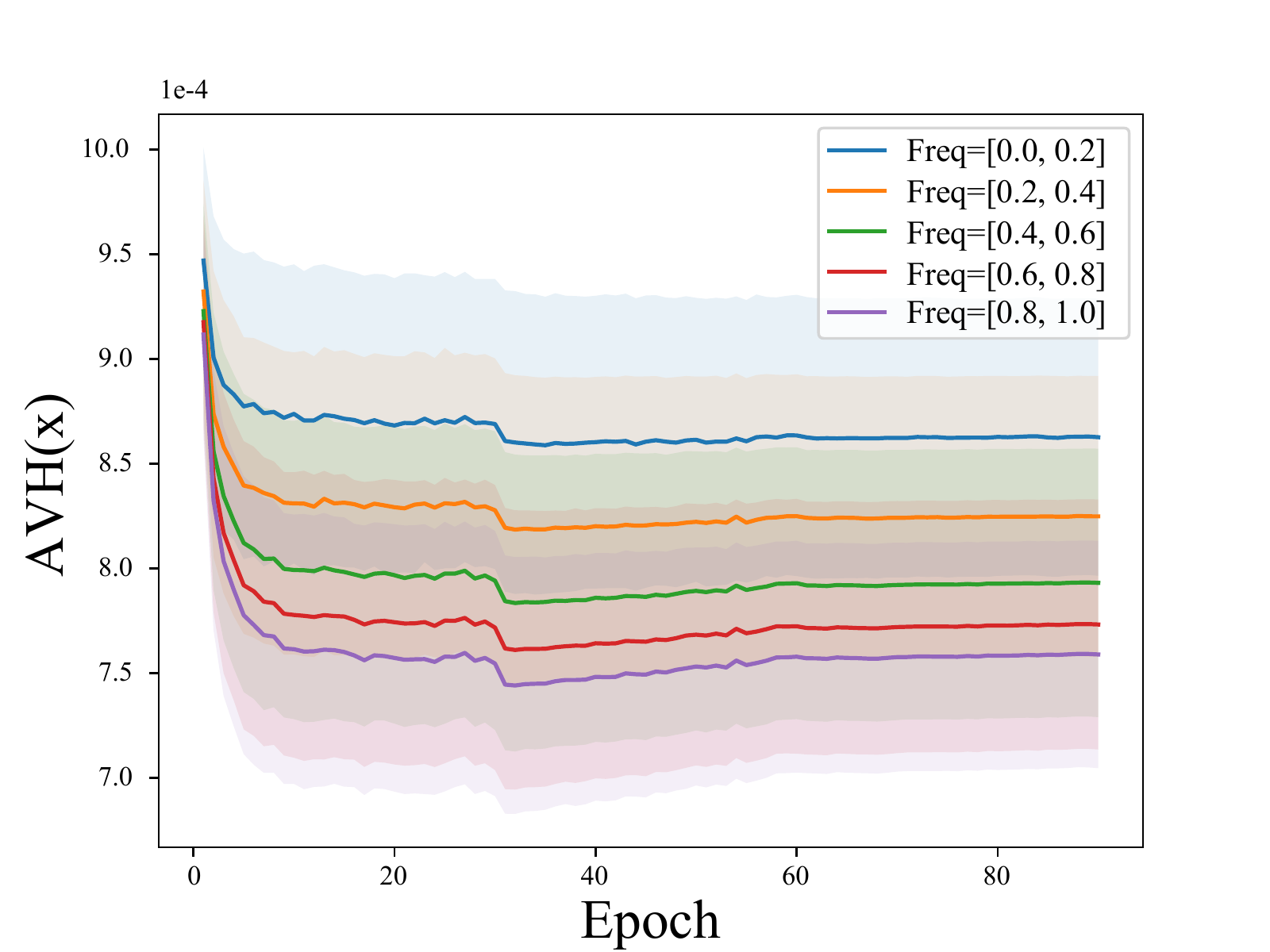}
		\includegraphics[width=0.32 \textwidth]{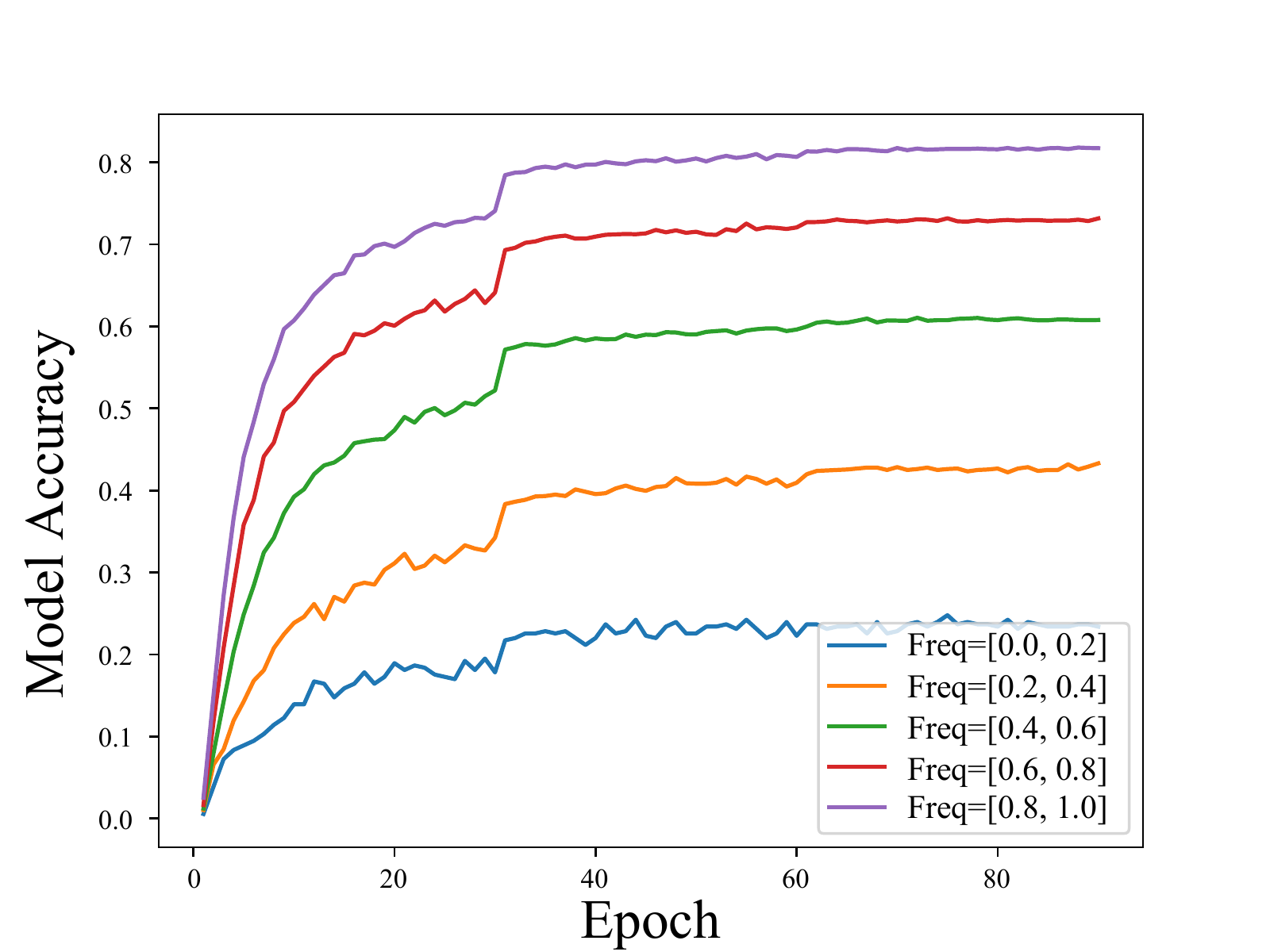}
		
		\includegraphics[width=0.32 \textwidth]{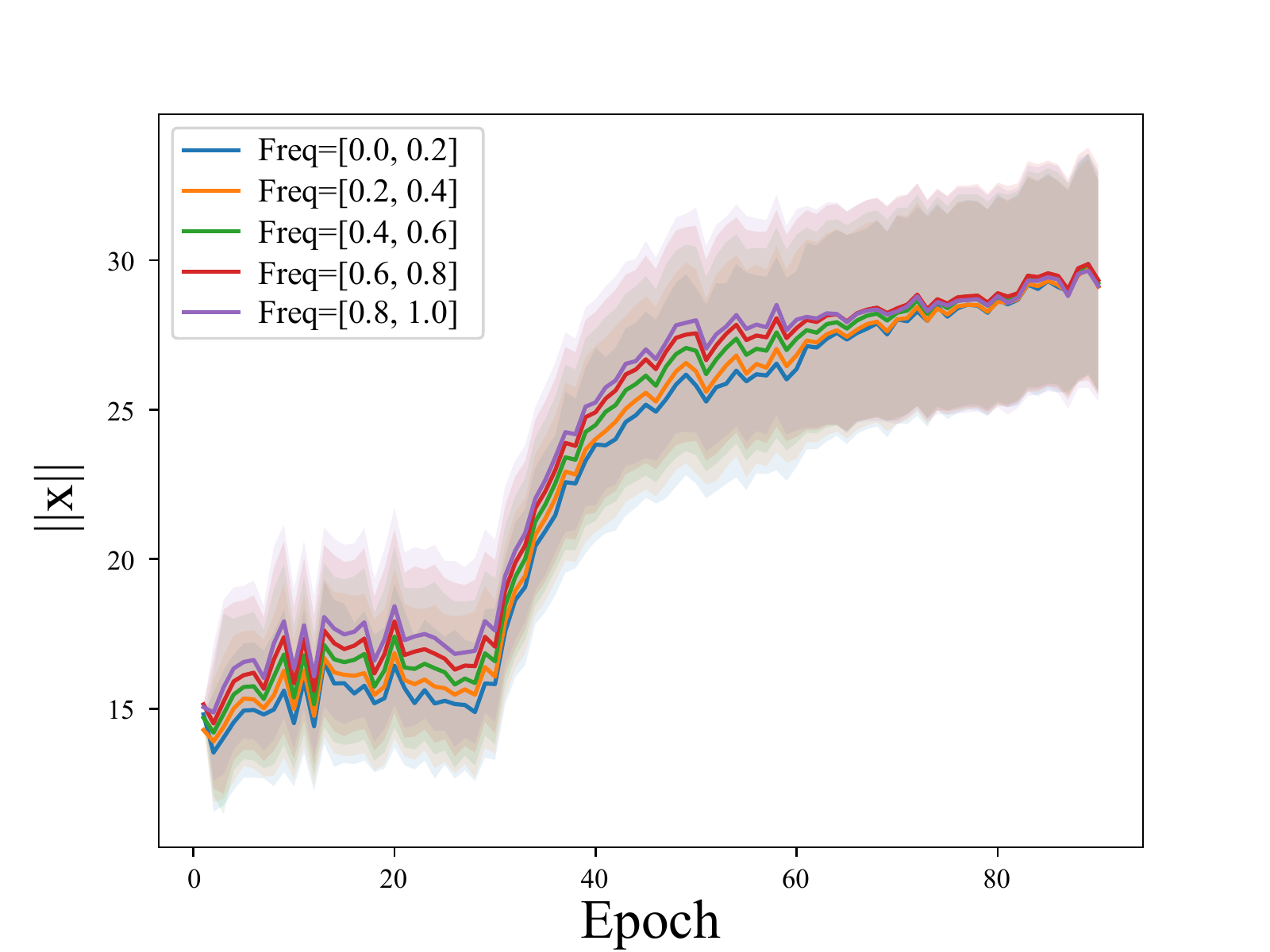}
		\includegraphics[width=0.32 \textwidth]{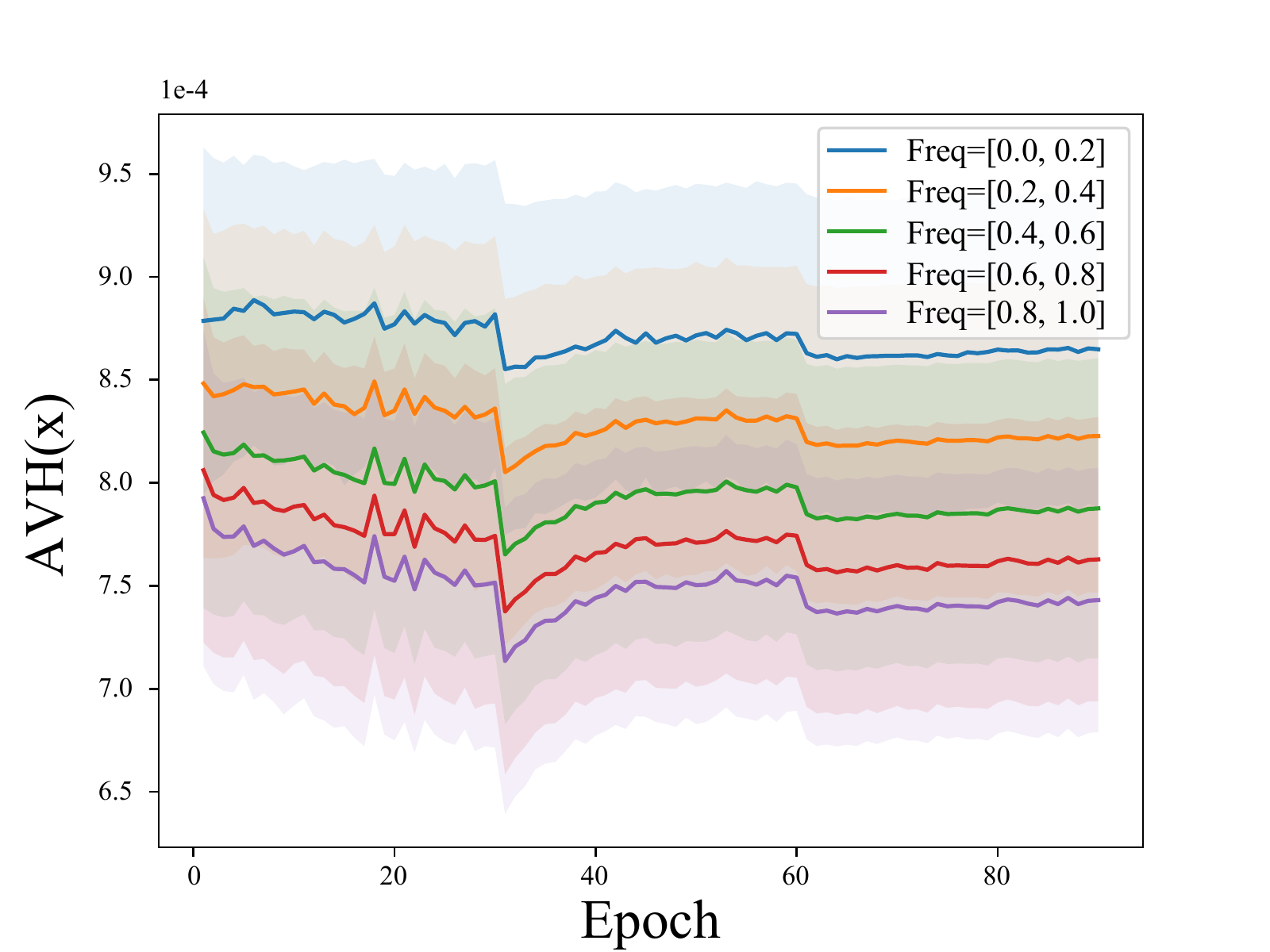}
		\includegraphics[width=0.32 \textwidth]{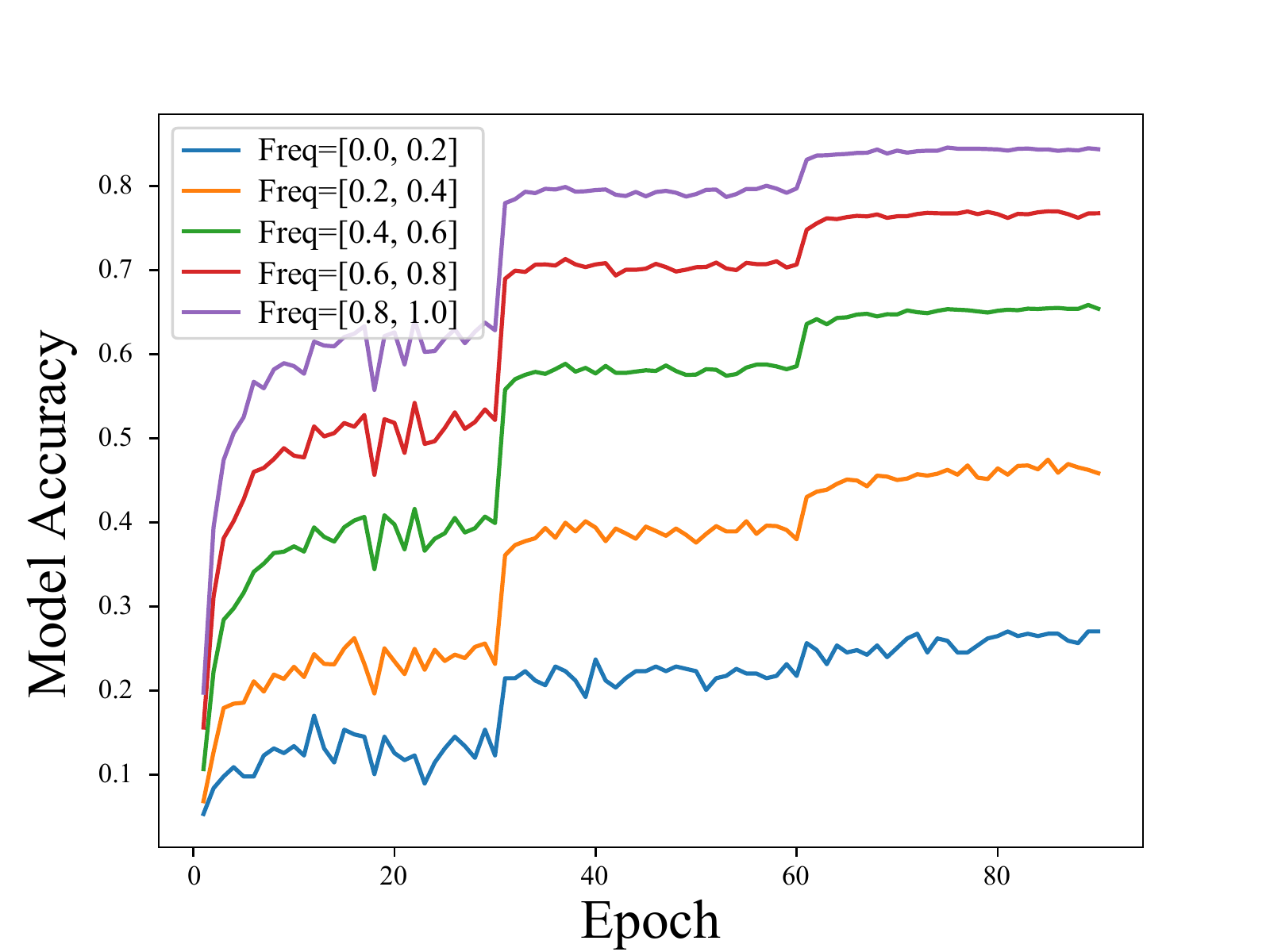}
		
		\includegraphics[width=0.32 \textwidth]{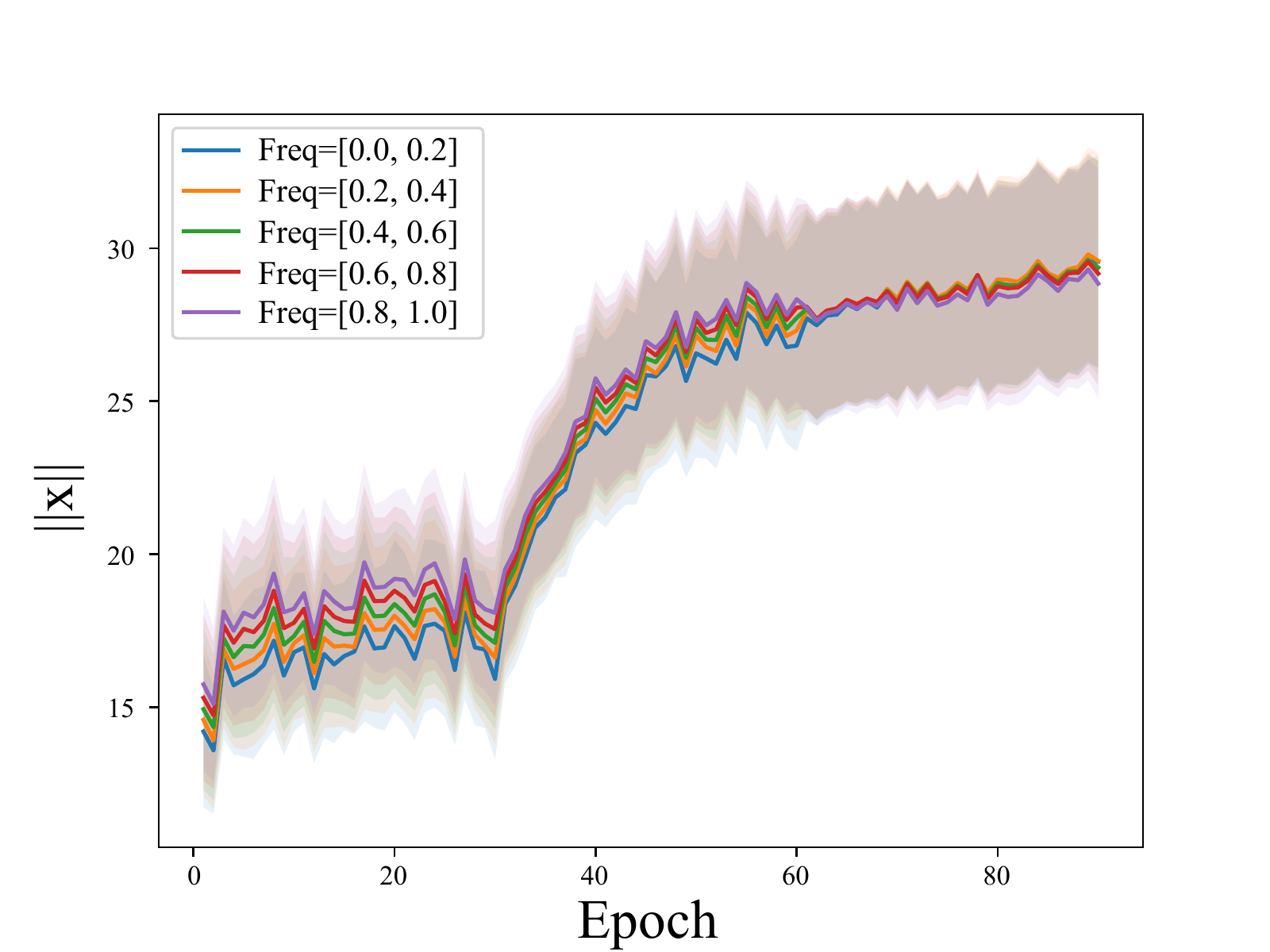}  
		\includegraphics[width=0.32 \textwidth]{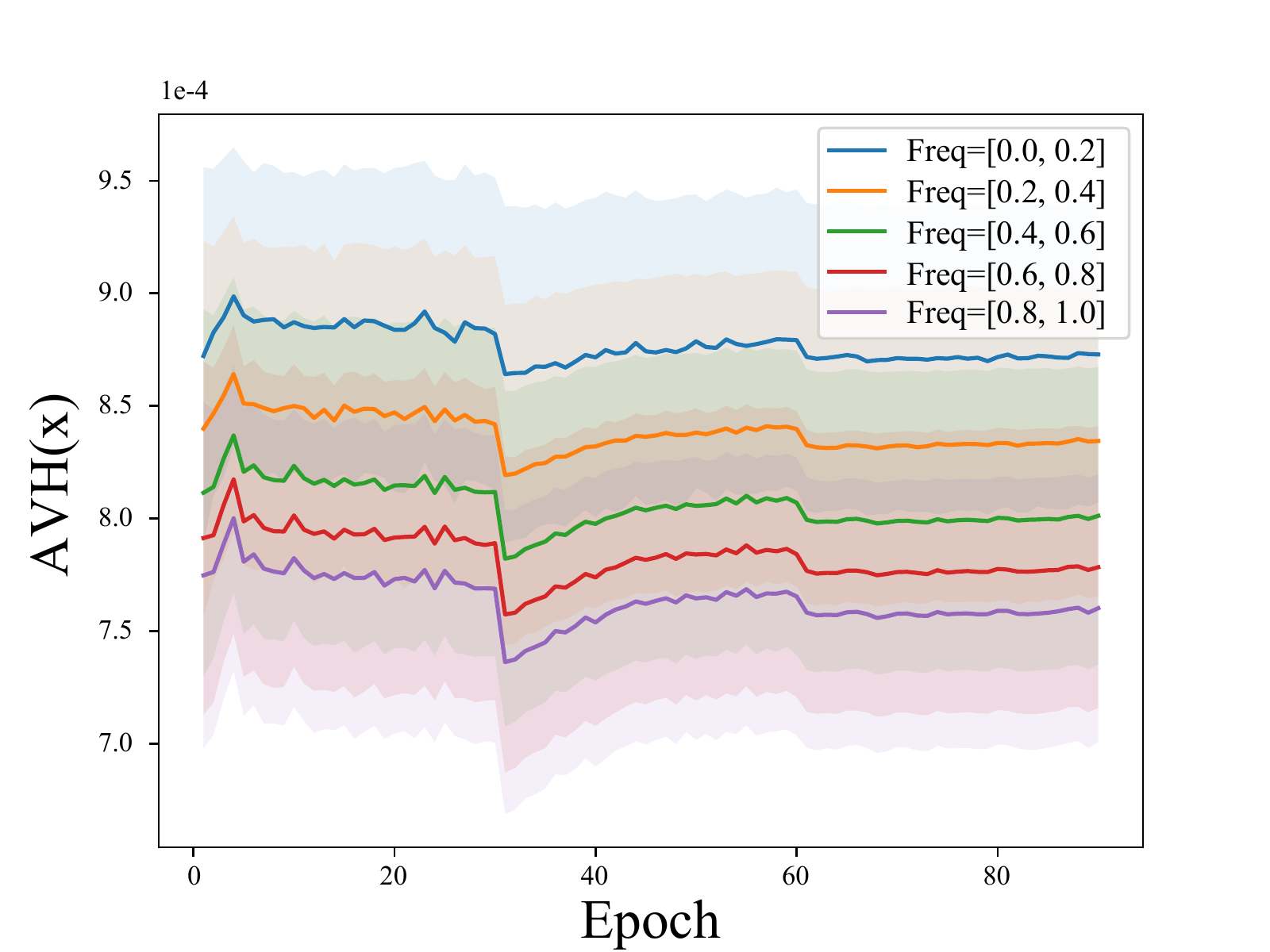}
		\includegraphics[width=0.32 \textwidth]{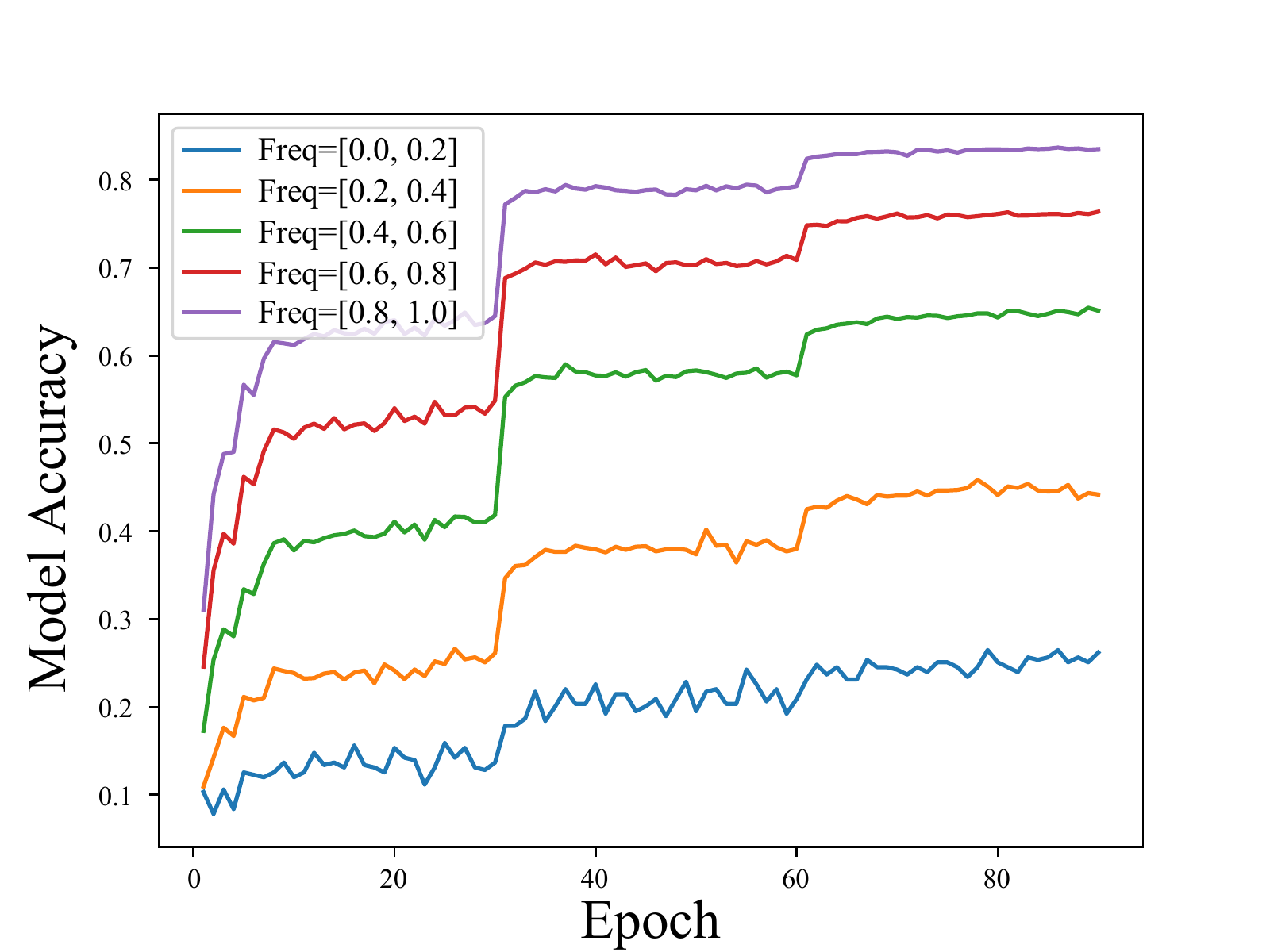}
	\end{center}
	\vspace{-3.5mm}
	\caption{\footnotesize Averaged training dynamics across different Human Selection Frequency levels on ImageNet validation set. Columns from left to right: number of epochs vs. average $\ell_2$ norm, number of epochs vs. average AVH score, and number of epochs vs. model accuracy. Rows from top to bottom: dynamics corresponding to AlexNet, VGG-19, ResNet-50, and DenseNet-121. Shadows in the figures of the first two columns denote the corresponding standard deviations.}\label{fig:change_imagenet}
	\vspace{-1.8mm}
\end{figure*}

\vspace{-3.6mm}
\subsection{Proposal and Intuition}\label{sec:avh}
\vspace{-0.8mm}

\begin{defn}[Angular Visual Hardness]
The AVH score, for any ${(\bm{x}, y)}$, is defined as:

\vspace{-4mm}
\begin{small}
\begin{equation}
    \mathcal{AVH}(\bm{x}) = \frac{\mathcal{A}(\bm{x},\bm{w}_{y})}{\sum_{i=1}^{C} \mathcal{A}(\bm{x}, \bm{w}_i)},
\end{equation}
\end{small}
\vspace{-4mm}

which $\bm{w}_y$ represents the weights of the target class.
\end{defn}\label{def:4}
\vspace{-1.2mm}

\textbf{Theoretical Foundations of AVH.} There are theoretical supports of AVH from both machine learning and vision science perspectives. On the machine learning side, we have briefly discussed above that AVH is directly related to the angle between feature embedding and the classifier weight of ground truth class. \cite{soudry2018implicit} theoretically shows that the logit of ground truth class must diverge to infinity in order to minimize cross-entropy loss to zero under gradient descent. Assuming input feature embeddings have fixed unit-norm, the norm of classifier weight grows to infinity. Similar result is also shown in ~\cite{wei2019improved} where generalization error of a linear classifier is controlled by the output margins normalized with classifier norm. Although the above analyses make certain assumptions, they indicate that norm is a less calibrated variable towards measuring properties of model/data compared to angle. This conclusion is comprehensively validated by our experiments in Section~\ref{sec:dynamics}. On the vision science side, there have been wide studies showing that human vision is highly adapted for extracting structural information~\cite{zhang2018unreasonable,wang2004image}, while the angular distance in AVH is precisely good at capturing such information~\cite{liu2018decoupled}. This also justifies our angular based design as an inductive bias towards measuring human visual hardness.

\vspace{0.65mm}

The AVH score is inspired by the observation from Figure~\ref{fig:diagram} as well as~\citep{liu2018decoupled} that samples from each class concentrate in a convex cone in the embedding space along with some interesting theoretical results that are discussed above. Naturally, we conjecture AVH, a measure with angle or margin information, could be the useful component of softmax score indicating the input sample hardness. We also perform a simulation providing visual intuition of how AVH instead of feature embedding norms corresponds to visually hard examples on two Gaussians in Figure~\ref{fig:gaussian} (simulation details and analyses in Appendix~\ref{app:a}).

\vspace{-2.9mm}
\subsection{Observations and Conjecture}
\label{sec:discovery}
\vspace{-1.4mm}

\textbf{Setup.} We aim to observe the complete training dynamics of models that are trained from scratch on ImageNet instead of the pretrained models. Therefore, we follow the standard training process of AlexNet~\cite{krizhevsky2012imagenet}, VGG-19~\cite{simonyan2014very}, ResNet-50~\cite{he2016deep} and DenseNet-121~\cite{huang2017densely}. For consistency, we train all models for $90$ epochs and decay the initial learning rate by a factor of $10$ every $30$ epochs. The initial learning rate for AlexNet and VGG-19 is $0.01$ and for DenseNet-121 and ResNet-50 is $0.1$. We split all the validation images into $5$ bins, $[0.0, 0.2], [0.2, 0.4], [0.4, 0.6], [0.6, 0.8], [0.8, 1.0]$, based on their HSF respectively. In Appendix~\ref{app:b}, we further provide experimental results on different datasets, such as MNIST, CIFAR10/100, and degraded ImageNet with different contrast or noise level, to better validate our proposal. For all the figures in this section, epoch starts from 1.

\vspace{0.25mm}

Optimization algorithms are used to update weights and biases, \ie, the internal parameters of a model to improve the training loss. Both the angles between the feature embedding and classifiers, and the $L_2$ norm of the embedding can influence the loss. While it is well-known that the training loss or accuracy keeps improving but it is not obvious what would be the dynamics of the angles and norms separately during training. we design the experiments to observe the training dynamics of various network architectures.

\vspace{0.46mm}
\textbf{Observation 1: The norm of feature embeddings keeps increasing during training}. 
\vspace{0.46mm}

The first column of Figure~\ref{fig:change_imagenet} presents the dynamics of averaged $\Vert \bm{x} \Vert_2$ on validation samples with the same range of HSF over $90$ epochs of training. Different figures also cover different network architectures. Note that we are using the validation data for dynamics observation and therefore have never fed them into the model. The average $\Vert \bm{x} \Vert_2$ increases with a small initial slope but it suddenly climbs after 30 epochs when the first learning rate decay happens. The accuracy curve is very similar to that of the average $\Vert \bm{x} \Vert_2$. These observations are consistent in all models and compatible with~\cite{soudry2018implicit} although it is more about the norm of the classifier weights. More interestingly, we find that neural networks with shortcuts (\eg, ResNets and DenseNets) tend to make the norm of the images with different HSF the same, while neural networks without shortcuts (\eg, AlexNet and VGG) tend to keep the gap of norm among the images with different human visual hardness.

\vspace{0.46mm}
\textbf{Observation 2: AVH hits a plateau very early even when the accuracy or loss is still improving}.
\vspace{0.46mm}

Middle row of Figure~\ref{fig:change_imagenet} exhibits the change of average AVH for validation samples in $90$ epochs of training on three models. The average AVH for AlexNet and VGG-19 decreases sharply at the beginning and then starts to bounce back a little bit before converging. However, the dynamics of the average AVH for DenseNet-121 and ResNet-50 are different. They both decrease slightly and then quickly hits a plateau in all three learning rate decay stages. But the common observation is that they all stop improving even when $\Vert \bm{x} \Vert_2$ and model accuracy are increasing. AVH is more important than $\Vert \bm{x} \Vert_2$ in the sense that it is the key factor deciding which class the input sample is classified to. However, optimizing the norm under the current softmax cross-entropy loss would be easier, which cause the plateau of angles for easy examples. However, the plateau for the hard examples can be caused by the limitation of the model itself and we show a simple illustration in Appendix~\ref{app:c}. It shows the necessity of designing loss functions that focus on optimizing angles. 

\textbf{Observation 3: AVH's correlation with Human Selection Frequency consistently holds across models throughout the training process}.

In Figure~\ref{fig:change_imagenet}, we average over validation samples in five HSF bins or five degradation level bins separately , and then compute the average embedding norm, AVH and model accuracies. We can observe that for $\Vert \bm{x} \Vert_2$, the gaps between the samples with different human visual hardness are not obvious in ResNet and DenseNet, while they are quite obvious in AlexNet and VGG. However, for AVH, such AVH gaps are very significant and consistent across every network architecture during the entire training process. Interestingly, even if the network is far from being converged, such AVH gaps are still consistent across different HSF. Also the norm gaps are also consistent. The intuition behind this could be that the angles for hard examples are much harder to decrease and probably never in the region for correct classification. Therefore the corresponding norms would not increase otherwise hurting the loss. It validates that AVH is a consistent and robust measure for visual hardness (and even generalization).

\begin{figure*}[t]
	\begin{center}
		\begin{tabular}{ccc}
		\hspace{-0.5cm}
			\includegraphics[width=0.32 \textwidth]{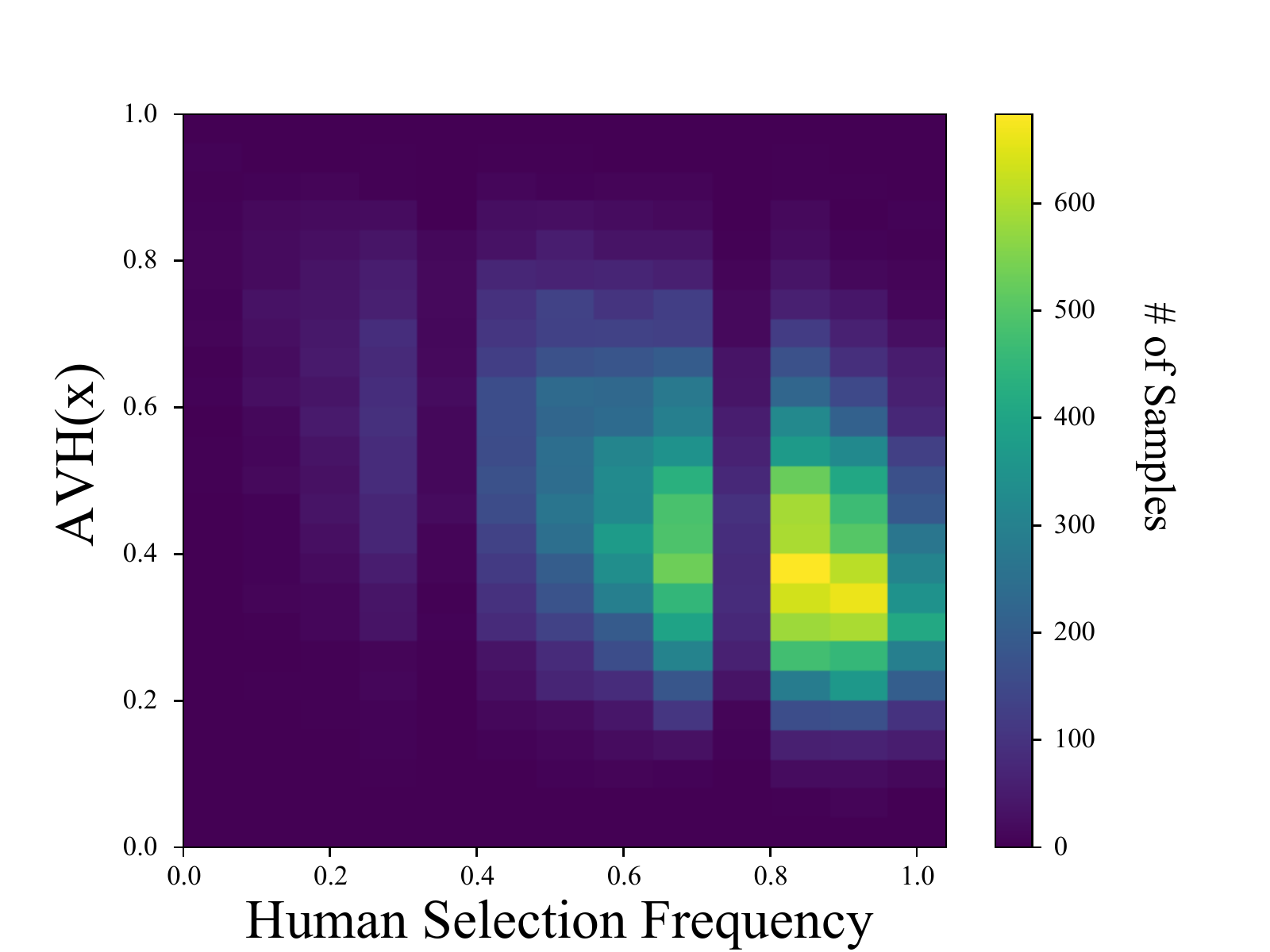} &
			\hspace{-0.5cm}
			\includegraphics[width=0.32 \textwidth]{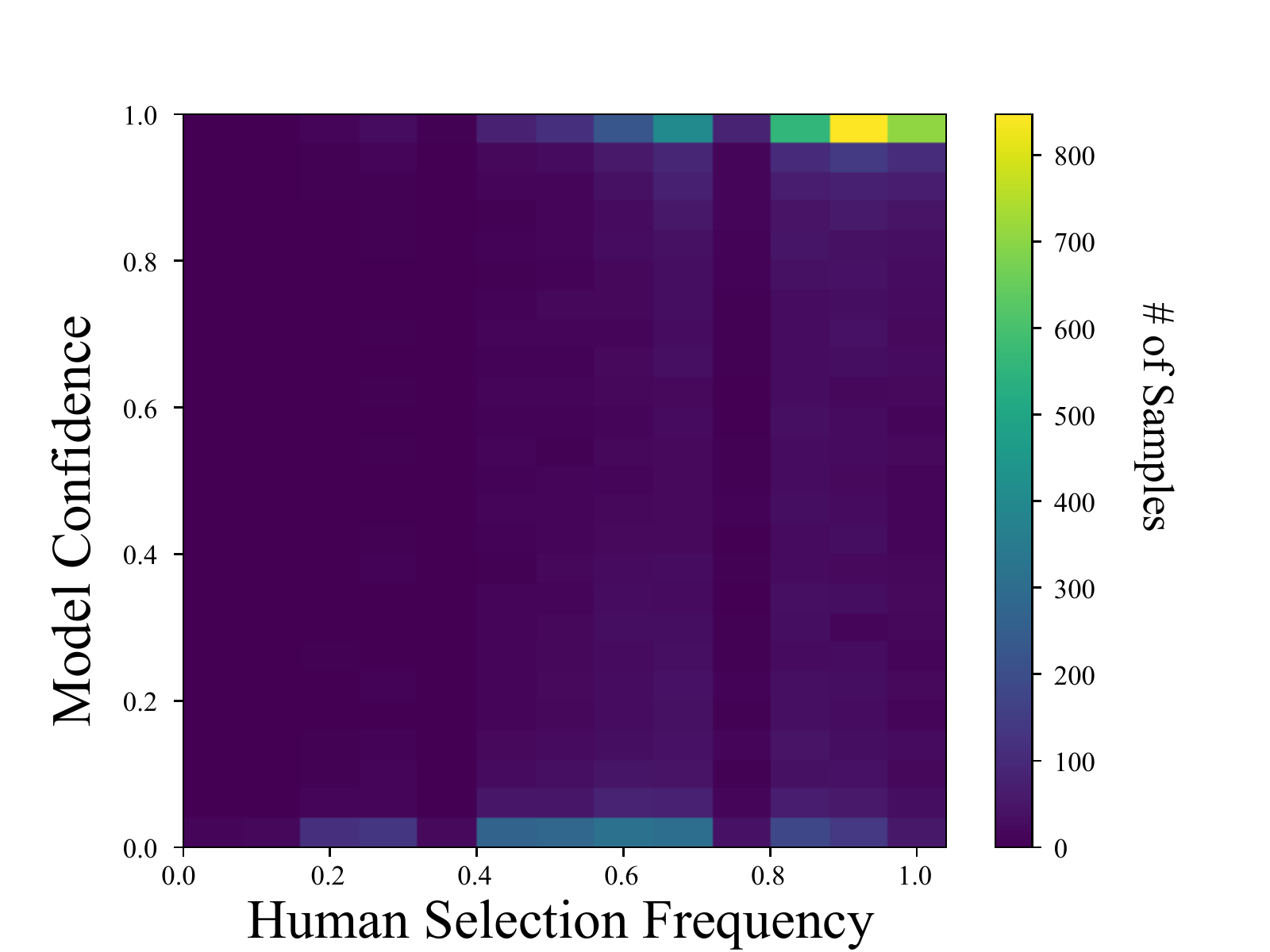} &
			\hspace{-0.5cm}
			\includegraphics[width=0.32 \textwidth]{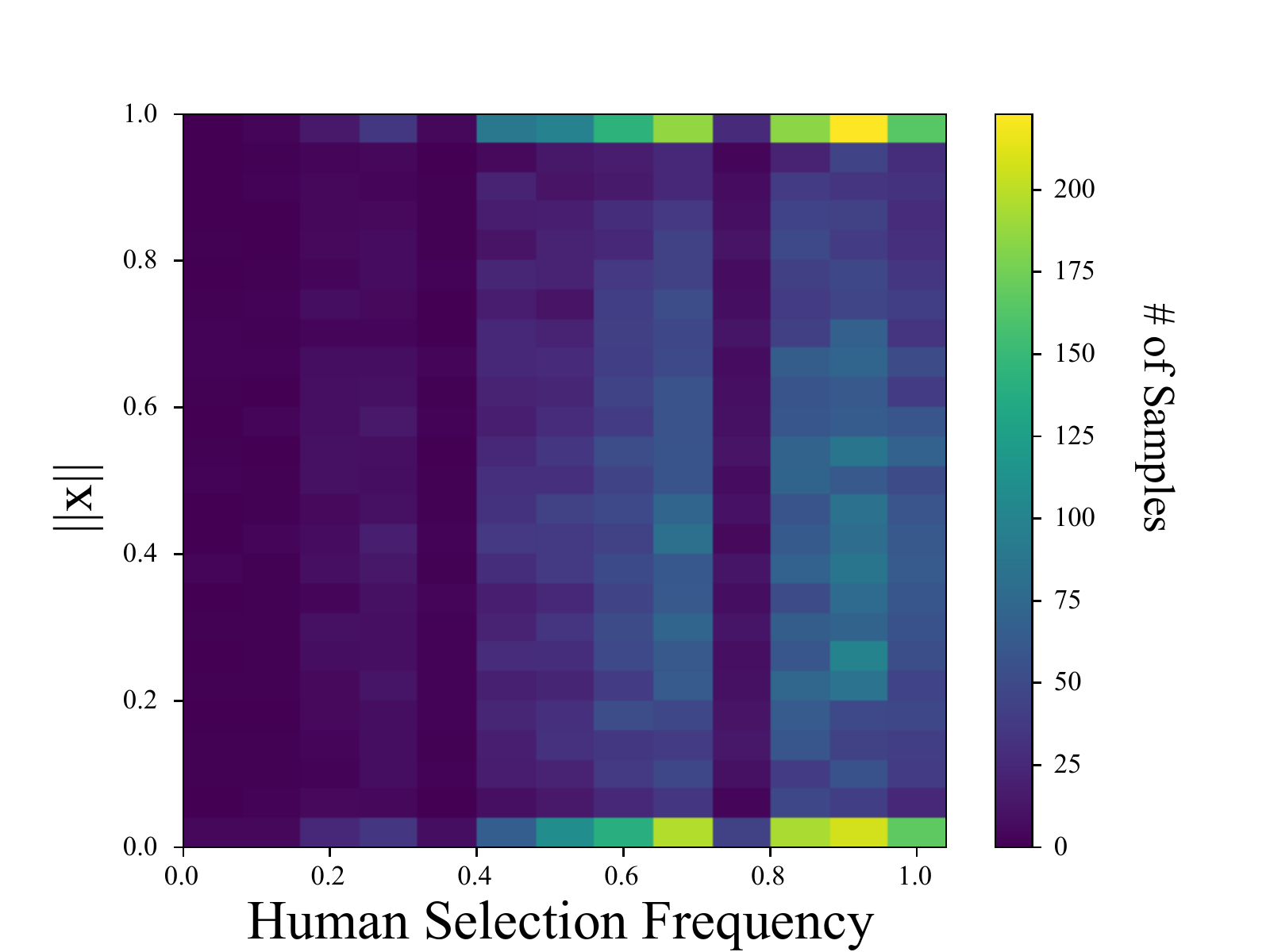}
		\end{tabular}
	\end{center}
 	\vspace{-4mm}
	\caption{\footnotesize The left one presents HSF v.s. $\mathcal{AVH}(\mathbf{x})$, which we can see strong correlation. The second plot presents the correlation between HSF and Model Confidence with ResNet-50. It is not surprising that the density is highest on the right corner. The third one presents HSF v.s. $\Vert \mathbf{x} \Vert_2$. There are no obvious correlation between them. Note that different color indicates the density of samples in that bin.}\label{fig:freq}
	\vspace{-3mm}
\end{figure*}

\begin{table*}[t] 
\scriptsize
\captionsetup{font=small}
\centering
\caption{Spearman's rank correlation coefficients between HSF and AVH, Model Confidence and L2 Norm of the Embedding in ResNet-50 for different visual hardness bin of samples. Note that here we show the absolute value of the coefficient which represents the strength of the correlation. For example, $[0,0.2]$ denotes the samples that have HSF from $0$ to $0.2$.}
\vspace{0.1cm}
\resizebox{0.9\linewidth}{!}{
\begin{tabular}{ c||ccccccc }
\specialrule{.15em}{.05em}{.05em}
&z-score&Total Coef& $[0, 0.2]$& $[0.2, 0.4]$ & $[0.4, 0.6]$ &$[0.6, 0.8]$ &  $[0.8, 1.0]$\\
\specialrule{.15em}{.05em}{.05em}
Number of Samples&- & 29987 &837 &2732 & 6541& 11066& 8811 \\
AVH & 0.377& 0.36 & 0.228 & 0.125& 0.124& 0.103& 0.094 \\ 
Model Confidence&0.337&0.325 & 0.192& 0.122& 0.102& 0.078& 0.056 \\
$\Vert \mathbf{x} \Vert_2$ &- &0.0017 & 0.0013 &0.0007 & 0.0005 &0.0004 &0.0003\\
\specialrule{.15em}{.05em}{.05em}
\end{tabular}
}
\vspace{-0.3cm}
\label{table:correlation}
\end{table*}

\textbf{Observation 4: AVH is an indicator of a model's generalization ability}.

From Figure~\ref{fig:change_imagenet}, we observe that better models (\ie, higher accuracy) have lower average AVH throughout the training process and also across samples under different human visual hardness. For instance, Alexnet is the weakest model, and its overall average AVH and average AVH on each of the five bins are higher than those of the other three models. In addition, we have found that when testing Hypothesis~\ref{hypo:3} for better models, their AVH correlations with HSF are significantly stronger than correlations of Model Confidence. The above observations are aligned with earlier observations from \cite{recht2019imagenet} that better models also tend to generalize better on samples across different levels of human visual hardness. In addition, AVH is potentially a better measure for the generalization of a pretrained model. As shown in ~\cite{sphereface}, the norms of feature embeddings are often related to training data priors such as data imbalance and class granularity~\cite{krizhevsky2012imagenet}. However, when extracting features from unseen classes that do not exist in the training set, such training data prior is often undesired. Since AVH does not consider such feature embedding norms, it potentially presents a better measure towards the open set generalization of a deep network.

\textbf{Conjecture on training dynamics of CNNs}.
From Figure~\ref{fig:change_imagenet} and observations above, we conjecture that the training of CNN has two phases. 1) At the beginning of the training, the softmax cross-entropy loss will first optimize the angles among different classes while the norm will fluctuate and increase very slowly. We argue that it is because changing the norm will not decrease the loss when the angles are not separated enough for correct classification. As a result, the angles get optimized firstly. 2) As the training continues, the angles become more stable and change very slowly while the norm increases rapidly. On the one hand, for easy examples, it is because when the angles get decreased enough for correct classification, the softmax cross-entropy loss can be well minimized by purely increasing the norm. On the other hand, for hard examples, the plateau is caused by that the CNN is unable to decrease the angle to correctly classify examples and thereby also unable to increase the norms (because it may otherwise increase the loss).

%% file: section4_connections.tex
\section{Connections to Human Visual Hardness}
\label{sec:correlation}

\begin{table*}[t] 
	\scriptsize
	\caption{This table presents the Spearman's rank, Pearson, and Kendall's Tau correlation coefficients between Human Selection Frequency and AVH, Model Confidence on ResNet-50, along with significance testings between coefficient pairs. Note that having p-value$<$ 0.05 indicates that the result is statistically significant.}
	\vspace{0.1cm}
	\centering
	\resizebox{0.85\linewidth}{!}{
		\begin{tabular}{ c||cccccc }
			\specialrule{.15em}{.05em}{.05em}
			Type&Coef with AVH & Coef with Model Confidence & $Z_{avh}$& $Z_{mc}$ & Z value & p-value\\
			\specialrule{.15em}{.05em}{.05em}
			Spearman's rank& 0.360&0.325 & 0.377 & 0.337 & 4.85 & $<$ .00001 \\
			Pearson& 0.385& 0.341& 0.406& 0.355& 6.2 & $<$ .00001\\
			Kendall's Tau& 0.257& 0.231& 0.263& 0.235& 3.38 &.0003 \\
			\specialrule{.15em}{.05em}{.05em}
		\end{tabular}
	}\label{table:resnetco}
	\vspace{-0.3cm}
\end{table*}

From Section~\ref{sec:discovery}, we conjecture that AVH has a stronger correlation with Human Selection Frequency - a reflection of human visual hardness that is related to aleatoric uncertainty. In order to validate this claim, we design statistical testings for the connections between Model Confidence, AVH, $\Vert \mathbf{x} \Vert_2$ and HSF. Studying the precise connection or gap between human visual hardness and model uncertainty is usually prohibitive because it is laborious to collect such highly subjective human annotations. In addition, these annotations are application or dataset specific, which significantly reduces the scalability of uncertainty estimation models that are directly supervised by them. This makes yet another motivation to this work since AVH is naturally obtained for free without any confidence supervision. In our case, we only leverage such human annotated visual hardness measure for correlation testing. In this section, We first provide four hypothesis and test them accordingly.  

\begin{hypo}
AVH has a correlation with Human Selection Frequency.
\end{hypo}
\vskip -.3\baselineskip
\textbf{Outcome: Null Hypothesis Rejected}

We use the pre-trained network model to extract the feature embedding $\mathbf{x}$ from each validation sample and also provide the class weights $\mathbf{w}$ to compute $\mathcal{AVH}(\mathbf{x})$. Note that we linearly scale the range of $\mathcal{AVH}(\mathbf{x})$ to $[0,1]$. Table~\ref{table:correlation} shows the overall consistent and stronger correlation between $\mathcal{AVH}(\mathbf{x})$ and HSF (p-value $< 0.001$ rejects the null hypothesis). From the coefficients shown in different bins of sample hardness, we can see that the harder the sample, the weaker the correlation. Also Note that we validate the results across different CNN architectures and found that better models tend to have higher coefficients.

The plot on the left in Figure~\ref{fig:freq} indicates the strong correlation between $\mathcal{AVH}(\mathbf{x})$ and HSF on validation images. One intuition behind this correlation is that the class weights $W$ might correspond to human perceived semantics for each category and thereby $\mathcal{AVH}(\mathbf{x})$ corresponds to human's semantic categorization of an image. In order to test if the strong correlation holds for all models, we perform the same set of experiments on different backbones, including AlexNet, VGG-19 and DenseNet-121. 

\begin{hypo}
Model Confidence has a correlation with Human Selection Frequency.
\end{hypo}
\vskip -.3\baselineskip
\textbf{Outcome: Null Hypothesis Rejected}

An interesting observation in \cite{recht2019imagenet} shows that HSF has strong influence on the Model Confidence. Specifically, examples with low HSF tends to have relatively low Model Confidence. Naturally we examine if the correlation between Model Confidence and HSF is strong. Specifically, all ImageNet validation images are evaluated by the pre-trained models. The corresponding output is simply the Model Confidence on each image. From Table~\ref{table:correlation}, we can first see that it is clear that because p-value is $< 0.001$, Model Confidence does have a strong correlation with HSF. However, the correlation coefficient for Model Confidence and HSF is consistently lower than that of AVH and HSF.

The middle plot in Figure~\ref{fig:freq} presents a two-dimensional histogram for the correlation visualization. The x-axis represents HSF, and the y-axis represents Model Confidence. Each bin exhibits the number of images which lie in the corresponding range. We can observe the high density at the right corner, which means the majority of the images have both high human and model accuracy. However, there is a considerable amount of density on the range of medium human accuracy but either extremely low or high model accuracy.  
One may question that the difference of the correlation coefficient is not large, thereby we also run statistical testing on the significance of the gap, naturally our next step is to test if the difference is significant.

\begin{hypo}\label{hypo:3}
AVH has a stronger correlation to Human Selection Frequency than Model Confidence.
\end{hypo}
\vskip -.3\baselineskip
\textbf{Outcome: Null Hypothesis Rejected}

There are three steps for testing if two correlation coefficient are significantly different. First step is applying Fisher Z-Transformation to both coefficients. The Fisher Z-Transformation is a way to transform the sampling distribution of the correlation coefficient so that it becomes normally distributed. Therefore, we apply fisher transformation for each correlation coefficient: $Z$ score for coefficient of AVH becomes $0.377$ and that of Model Confidence becomes $0.337$. The second step is to compute the $Z$ value of two $Z$ scores. Then we determined the $Z$ value to be $4.85$ from the two above-mentioned $Z$ scores and sample sizes. The last step is that find out the p-value according to the $Z$ table. According to $Z$ table, p-value is 0.00001. Therefore, we reject the null hypothesis and conclude that AVH has statistically significant stronger correlation with HSF than Model Confidence. In later Section~\ref{sec:selftrain}, we also empirically show that such stronger correlation brings cumulative advantages in some applications. 

In Table~\ref{table:resnetco}, besides the Spearman correlation coefficient, we also show the coefficients of Pearson and Kendall Tau. In addition, in Appendix~\ref{app:d}, we run the same tests on four different architectures to test whether the same conclusion holds for different models. Our conclusion is that for all the considered models, AVH correlates significantly stronger than Model Confidence, and the correlation is even stronger for better models. This indicates that besides what we have shown in Section~\ref{sec:dynamics}, AVH is also better aligned with human visual hardness which is related to aleatoric uncertainty.

\begin{hypo}\label{hypo:4}
$\Vert \mathbf{x} \Vert_2$ has a correlation with Human Selection Frequency.
\end{hypo}
\vskip -.3\baselineskip
\textbf{Outcome: Failure to Reject Null Hypothesis}

\cite{liu2018decoupled} conjectures that $\Vert \mathbf{x} \Vert_2$ accounts for intra-class Human/Model Confidence. Particularly, if the norm is larger, the prediction from the model is also more confident, to some extent. Therefore, we conduct similar experiments like previous section to demonstrate the correlation between $\Vert \mathbf{x} \Vert_2$ and HSF. Initially, we compute the $\Vert \mathbf{x} \Vert_2$ for every validation sample for all models. Then we normalize $\Vert \mathbf{x} \Vert_2$ within each class. Table~\ref{table:correlation} presents the results for the correlation test. We omit the results for p-value in the table and report here that they are all much higher than $0.05$, indicating there is no correlation between $\Vert \mathbf{x} \Vert_2$ and HSF. The right plot in Figure~\ref{fig:freq} uses a two-dimensional histogram to show the correlation for all the validation images. Given that the norm has been normalized with each class, naturally, there is notable density when the norm is $0$ or $1$. Except for that, there is no obvious correlation between $\Vert \mathbf{x} \Vert_2$ and HSF. 

We also provide a detailed discussion on the difference between AVH and Model Confidence in Appendix~\ref{sec:extradis}.

%% file: section5_applications.tex
%\vspace{-0.2cm}
\section{Applications}
\subsection{AVH for Self-training and Domain Adaptation}
\label{sec:selftrain}
Unsupervised domain adaptation~\citep{ben2010theory} presents an important transfer learning problem and deep self-training~\citep{lee2013pseudo} recently emerged as a powerful framework to this problem~\citep{saito2017asymmetric, shu2018dirt, zou2018domain, zou2019confidence}. Here we show the application of AVH as an improved confidence measure in self-training that could significantly benefit domain adaptation.

\textbf{Dataset:} We conduct expeirments on the VisDA-17~\citep{peng2017visda} dataset which is a widely used major benchmark for domain adaptation in image classification. The dataset contains a total number of $152, 409$ 2D synthetic images from 12 categories in the source training set, and $55, 400$ real images from MS-COCO~\cite{lin2014microsoft} with the same set of categories as the target domain validation set. We follow the protocol of previous works to train a source model with the synthetic training set, and report the model performance on target validation set upon adaptation.

\textbf{Baseline:} We use class-balanced self-training (CBST)~\citep{zou2018domain} as a state-of-the-art self-training baseline. We also compare our model with confidence regularized self-training (CRST)\footnote{We consider MRKLD+LRENT which is reported to be the highest one in ~\citep{zou2019confidence}.}~\citep{zou2019confidence}, a more recent framework improved over CBST with network prediction/pseudo-label regularized with smoothness. Specifically, our work follows the exact implementation of CBST/CRST. 

Specifically, given the labeled source domain training set $\mathbf{x}_s \in \mathbf{X}_S$ and the unlabeled target domain data $\mathbf{x}_t \in \mathbf{X}_T$, with known source labels $\mathbf{y}_s=(y_s^{(1)},...,y_s^{(K)}) \in \mathbf{Y}_S$ and unknown target labels $\hat{\mathbf{y}}_t=(\hat{y}_t^{(1)},...,\hat{y}_t^{(K)}) \in \hat{\mathbf{Y}}_T$ from $K$ classes, CBST performs joint network learning and pseudo-label estimation by treating pseudo-labels as discrete learnable latent variables with the following loss:

\vspace{-0.6cm}
\begin{small}
\begin{equation}\label{cbst}
\begin{split}
&\underset{\mathbf{w},\hat{\mathbf{Y}}_T}{\mathop{\min }}\,\mathcal{L}_{CB}(\mathbf{w}, \hat{\mathbf{Y}}) = -\sum\limits_{s\in S}{\sum\limits_{k=1}^K{y_s^{(k)}}\log p(k|\mathbf{x}_s;\mathbf{w})}\\[-1mm]
&-\sum\limits_{t\in T}\sum\limits_{k=1}^{K}\hat{y}_{t}^{(k)}\log \frac{p(k|{\mathbf{x}_{t}};\mathbf{w})}{\lambda_k} \text{ s.t.}~~\hat{\mathbf{y}}_t\in \mathbf{E}^K\cup \{\mathbf{0}\},~\forall t
\end{split}
\end{equation}
\end{small}
\vspace{-0.4cm}

where the feasible set of pseudo-labels is the union of $\{\mathbf{0}\}$ and the $K$ dimensional one-hot vector space $\mathbf{E}^K$, and $\mathbf{w}$ and $p(k|\mathbf{x};\mathbf{w})$ represent the network weights and the classifier's softmax probability for class $k$, respectively. In addition, $\lambda_{k}$ serves as a class-balancing parameter controlling the pseudo-label selection of class $k$, and is determined by the softmax confidence ranked at portion $p$ (in descending order) among samples predicted to class $k$. Therefore, only one parameter $p$ is used to determine all $\lambda_k$'s. The optimization problem in (\ref{cbst}) can be solved via minimizing with respect to $\mathbf{w}$ and $\hat{\mathbf{Y}}$ alternatively, and the solver of $\hat{\mathbf{Y}}$ can be written as:

\vspace{-0.7cm}
\begin{small}
\begin{equation}\label{cbst_solver}
\hat{y}_{t}^{(k)*}=\left\{
\begin{aligned}
1, &~\text{if}~k=\argmax_{c}\{\frac {p(c|{\mathbf{x}_{t}};\mathbf{w})}{\lambda_c}\}~ \text{and}~ p(k|\mathbf{x}_t;\mathbf{w})>\lambda_k\\
0, &~\mathrm{otherwise}
\end{aligned}
\right.
\nonumber
\end{equation}
\end{small}
\vspace{-0.5cm}

The optimization with respect to $\mathbf{w}$ is simply network re-training with source labels and estimated pseudo-labels. The complete self-training process involves alternative repeat of network re-training and pseudo-label estimation.

\textbf{CBST+AVH:} We seek to improve the pseudo-label solver with better confidence measure from AVH. We propose the following definition of angular visual confidence (AVC) to represent the predicted probability of class $c$:

\begin{small}
\vspace{-0.3cm}
\begin{equation}\label{avh_prob2}
\mathcal{AVC}(c|\mathbf{x}; \mathbf{w}) = \frac{\pi - \mathcal{A}(\mathbf{x},\mathbf{w}_c)}{\sum_{k=1}^{K} (\pi - \mathcal{A}(\mathbf{x}, \mathbf{w}_k))},
\end{equation}
\vspace{-0.3cm}
\end{small}

and pseudo-label estimation in CBST+AVH is defined as:

\begin{small}
\vspace{-0.3cm}
\begin{equation}\label{avh_solver}
\hat{y}_{t}^{(k)*}=\left\{
\begin{aligned}
1, &~\text{if}~k=\argmax_{c}\{\frac {p(c|{\mathbf{x}_{t}};\mathbf{w})}{\lambda_c}\}~ \\
&\text{and}~ \mathcal{AVC}(k|\mathbf{x}_t;\mathbf{w})>\beta_k\\
0, &~\mathrm{otherwise}
\end{aligned}
\right.
\vspace{-0.3cm}
\end{equation}
\end{small}

where $p(k|{\mathbf{x}_{t}};\mathbf{w})$ is the softmax output of $\mathbf{x}_{t}$. $\lambda_k$ and $\beta_k$ are determined respectively by referring to $p(k|\mathbf{x}_t;\mathbf{w})$ and $\mathcal{AVC}(k|\mathbf{x}_t;\mathbf{w})$ ranked at a particular portion among samples predicted to class $k$, following the same definition of $\lambda_k$ in CBST. In addition, network re-training in CBST+AVH follows the softmax self-training loss in (\ref{cbst}).

One could see that AVH changes the self-training behavior by having improved pseudo-label selection in (\ref{avh_solver}) in terms of $\mathcal{AVC}(k|\mathbf{x}_t;\mathbf{w})>\beta_{k}$.
Specifically, the condition determines which samples are not ignored during self-training based on AVC. With the improved confidence measure that better resembles human visual hardness, this aspect is likely to influence the final performance of self-training. 

\begin{table*}[tbh]
    \scriptsize
	\centering
	\caption{Class-wise and mean classification accuracies on VisDA-17.}
	\vspace{0.1cm}
	\resizebox{\linewidth}{!}{
	\centering
	\begin{tabular}{c||cccccccccccc||c}
		\specialrule{.15em}{.05em}{.05em}
		Method & Aero & Bike & Bus & Car & Horse & Knife & Motor & Person & Plant & Skateboard & Train & Truck & Mean\\
		\specialrule{.15em}{.05em}{.05em}
		Source \cite{saito2018adversarial} & 55.1 & 53.3 & 61.9 & 59.1 & 80.6 & 17.9 & 79.7 & 31.2 & 81.0 & 26.5 & 73.5 & 8.5 & 52.4\\
		MMD \cite{long2015learning} & 87.1 & 63.0 & 76.5 & 42.0 & 90.3 & 42.9 & 85.9 & 53.1 & 49.7 & 36.3 & \textbf{85.8} & 20.7 & 61.1\\
		DANN \cite{ganin2016domain} & 81.9 & 77.7 & 82.8 & 44.3 & 81.2 & 29.5 & 65.1 & 28.6 & 51.9 & 54.6 & 82.8 & 7.8 & 57.4\\ 
		ENT \cite{grandvalet2005semi} & 80.3 & 75.5 & 75.8 & 48.3 & 77.9 & 27.3 & 69.7 & 40.2 & 46.5 & 46.6 & 79.3 & 16.0 & 57.0\\
		MCD \cite{saito2017maximum} & 87.0 & 60.9 & \textbf{83.7} & 64.0 & 88.9 & 79.6 & 84.7 & 76.9 & 88.6 & 40.3 & 83.0 & 25.8 & 71.9\\
		ADR \cite{saito2018adversarial} & 87.8 & 79.5 & \textbf{83.7} & 65.3 & \textbf{92.3} & 61.8 & \textbf{88.9} & 73.2 & 87.8 & 60.0 & 85.5 & {32.3} & 74.8\\  
		% \hline\hline
		\hline
		Source~\citep{zou2019confidence} & 68.7 & 36.7 & 61.3 & \textbf{70.4} & 67.9 & 5.9 & 82.6 & 25.5 & 75.6 & 29.4 & 83.8 &  10.9 & 51.6\\
		CBST~\citep{zou2019confidence} & 87.2 & 78.8 & 56.5 & 55.4 & 85.1 & 79.2 & 83.8 & 77.7 & 82.8 & \textbf{88.8} & 69.0 & \textbf{72.0} & 76.4 \\
		CRST~\citep{zou2019confidence} & 88.0 & 79.2 & 61.0 & 60.0 & 87.5 & 81.4 & 86.3 & 78.8 & 85.6 & 86.6 & 73.9 &   68.8 & 78.1\\
		% \hline\hline
		\hline
		Proposed & \textbf{93.3} & \textbf{80.2} & 78.9 & 60.9 & 88.4 & \textbf{89.7} & \textbf{88.9} & \textbf{79.6} & \textbf{89.5} & 86.8 & 81.5 & 60.0 & \textbf{81.5}\\
		\specialrule{.15em}{.05em}{.05em}
	\end{tabular}
	}
 	\vspace{-0.1cm}
	\label{table:visda17}
\end{table*}

\begin{figure}[t]
\centering
\includegraphics[width=6cm]{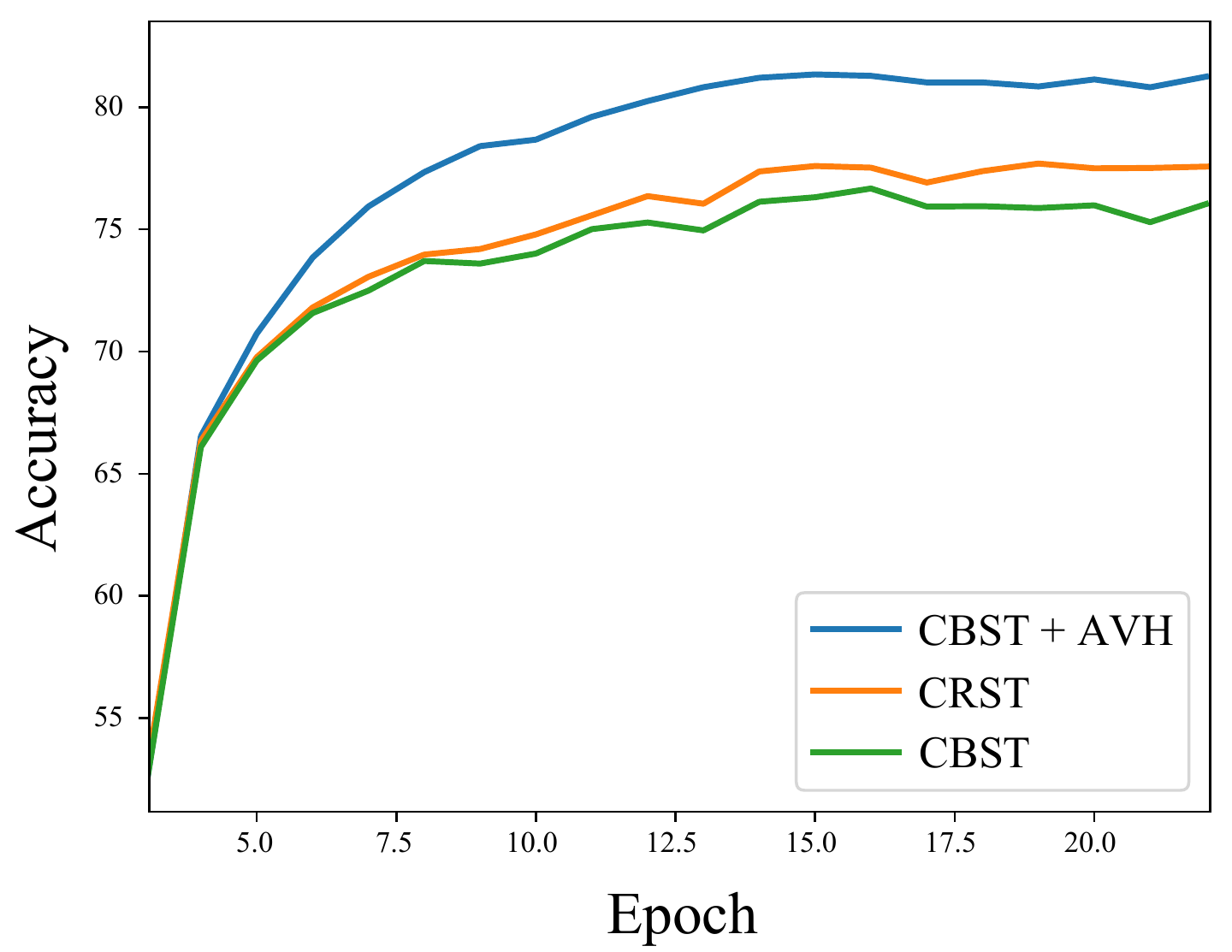}
\vspace{-0.2cm}  
\caption{Adaptation accuracy vs. epoch for different comparing methods on VisDA-17.}
\vspace{-0.1cm}
\label{fig:domain}
\end{figure}

\textbf{Experimental Results:} We present the results of the proposed method in Table~\ref{table:visda17}, and also show its performance with respect to different self-training epochs in Figure~\ref{fig:domain}. One could see that CBST+AVH outperforms both CBST and CRST by a very significant margin. We would like to emphasize that this is a very compelling result under ``apples to apples'' comparison with the same source model, implementation and hyper-parameters.

\textbf{Analysis:} A major challenge of self-training is the amplification of error due to misclassified pseudo-labels. Therefore, traditional self-training methods such as CBST often use model confidence as the measure to select confidently labeled examples. The hope is that higher confidence potentially implies lower error rate. While this generally proves useful, the model tends to focus on the ``less informative'' samples, whereas ignoring the ``more informative'', harder ones near classier boundaries that could be essential for learning a better classifier. More details are in Appendix~\ref{app:F}.

\begin{table}[tbh]
	\vspace{-0.2cm}
	\scriptsize
	\caption{Statistics of the examples selected by CBST+AVH and CBST/CRST.}
	\vspace{0.1cm}
	\centering
	\begin{tabular}{ c||ccccc }
		\specialrule{.15em}{.05em}{.05em}
		Method & TP Rate & AVH (avg) &  Model Confidence& Norm $\Vert x \Vert$\\
		% \hline
		\specialrule{.15em}{.05em}{.05em}
		CBST+AVH &0.844& 0.118  & 0.961 &20.84\\
		CBST/CRST &0.848& 0.117  & 0.976 &21.28\\
		\specialrule{.15em}{.05em}{.05em}
	\end{tabular}
	\label{table:stats}
	\vspace{-0.4cm}
\end{table}

An advantage we observe from AVH is that the improved calibration leads to more frequent sampling of harder samples, whereas the pseudo-label classification on these hard samples generally outperforms softmax results. Table~\ref{table:stats} shows the statistics of examples selected with AVH and model confidence respectively at the beginning of the training process. The true positive rate (TP Rate) for CBST+AVH remains similar to CBST/CRST, indicating AVH overall is not introducing additional noise compare to model confidence. On the other hand, it is observed that the average model confidence of AVH selected samples is lower, indicating there are more selected hard samples that are closer to the decision boundary. It is also observed that the average sample norm by AVH is also lower, confirming the influence of sample norms on ultimate model confidence.

\subsection{AVH-based Loss for Domain Generalization}
\label{sec:DG}
The problem of domain generalization (DG) is to learn from multiple training domains, and extract a domain-agnostic model that can then be applied to an unseen domain. Since we have no assumption on how the unseen domain looks like, the generalization on the unseen domains will mostly depend on the generalizability of the neural network. We use the challenging PACS dataset~\cite{li2017deeper} which consists of Art painting, Cartoon, Photo and Sketch domains. For each domain, we leave it out as the test set and train our models on rest of the three domains.

Specifically, we train a 10-layer plain CNN with the following AVH-based loss (additional details in Appendix~\ref{app:F}):

\begin{small}
\vspace{-0.2cm}
\begin{equation}\label{avh_prob}
\mathcal{L}_{AVH} = \sum_i\frac{\exp\big(s(\pi - \mathcal{A}(\mathbf{x}_i,\mathbf{w}_{y_i}))\big)}{\sum_{k=1}^{K} \exp\big(s(\pi - \mathcal{A}(\mathbf{x}_i, \mathbf{w}_k))\big)},
\end{equation}
\vspace{-0.2cm}
\end{small}

where $s$ is hyperparameter that adjusts the scale of the output logits and implicitly controls the optimization difficulty. This hyperparameter is typically set by cross-validation. Experimental results are reported in Table~\ref{pacs}. With the proposed new loss which directly has an AVH-based design, a simple CNN is outperforming baseline and recent methods that are based on more complex models. In fact, similar learning objectives have also been shown useful in image recognition~\cite{liu2017deep} and face recognition~\cite{wang2017normface,ranjan2017l2}, indicating that AVH is generally effective to improve generalization in various tasks.

\begin{table}
\vspace{-0.3cm}
\setlength{\columnsep}{4pt}
\setlength{\abovecaptionskip}{4pt}
\setlength{\belowcaptionskip}{-10pt}
\centering
\caption{\footnotesize Domain generalization accuracy (\%) on PACS dataset.}
\renewcommand{\captionlabelfont}{\footnotesize}
\vspace{0.1cm}
\resizebox{\linewidth}{!}{
\begin{tabular}{c||ccccc} 
\specialrule{.15em}{.05em}{.05em}
Method & Painting & Cartoon & Photo & Sketch & Avg\\
\specialrule{.15em}{.05em}{.05em}
AlexNet~\cite{li2017deeper} & 62.86 & 66.97& 89.50 & 57.51 & 69.21\\
MLDG~\cite{li2018learning} &66.23 & 66.88 & 88.00 & 58.96&70.01 \\
MetaReg~\cite{balaji2018metareg} & \textbf{69.82} & 70.35 & \textbf{91.07} & 59.26 & \textbf{72.62} \\
Feature-critic~\cite{li2019feature} & 64.89 & \textbf{71.72} & 89.94 & \textbf{61.85} & 72.10\\
\hline
Baseline CNN-10 &66.46 & 67.88 & 89.70 & 51.72 & 68.94\\
CNN-10 + AVH & \textbf{72.02} & 66.42 & \textbf{90.12} & \textbf{61.26} & \textbf{72.46}\\
\specialrule{.15em}{.05em}{.05em}
\end{tabular}
}
\vspace{-0.2cm}
\label{pacs}
\end{table}

%% file: section6_conclusions.tex
\vspace{-1.6mm}
\section{Concluding Remarks}
\vspace{-0.7mm}
We propose a novel measure for CNN models known as Angular Visual Hardness. Our comprehensive empirical studies show that AVH can serve as an indicator of generalization abilities of neural networks, and improving SOTA accuracy entails improving accuracy on hard examples. AVH also has a significantly stronger correlation with Human Selection Frequency. We empirically show the advantage of AVH over Model Confidence in self-training for domain adaptation task and loss function for domain generalization task. AVH can be useful in other applications such as deep metric learning, fairness, knowledge transfer, etc. and we plan to investigate them in the future (discussions in Appendix~\ref{app:g}).

\newpage

\section*{Acknowledgements}
Work done during internship at NVIDIA. We would like to thank Shiyu Liang, Yue Zhu and Yang Zou for the valuable discussions that enlighten our research. We are also grateful to the anonymous reviewers for their constructive comments that significantly helped to improve our paper. Weiyang Liu is partially supported by Baidu scholarship and NVIDIA GPU grant. This work was supported by NSF-1652131, NSF-BIGDATA 1838177, AFOSR-YIPFA9550-18-1-0152, Amazon Research Award, and ONR BRC grant for Randomized Numerical Linear Algebra.

%% file: section7_appendix.tex
\onecolumn

\begin{appendix}
	
\icmltitle{Appendix: Angular Visual Hardness}

\section{Simulation Details}\label{app:a}

\begin{figure*}[h]
	\vspace{-2mm}
	\begin{center}
		\hspace{-0.5cm}
		\includegraphics[width=0.28\textwidth]{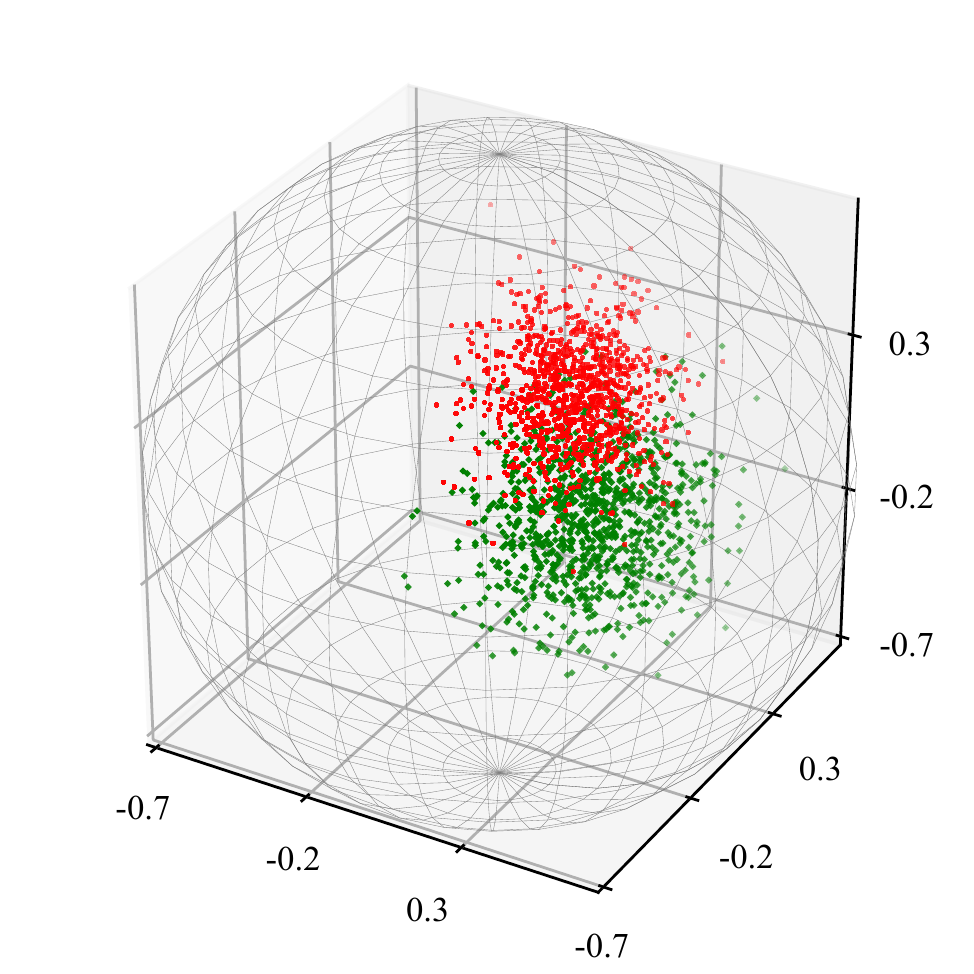}~~~~
		\includegraphics[width=0.28\textwidth]{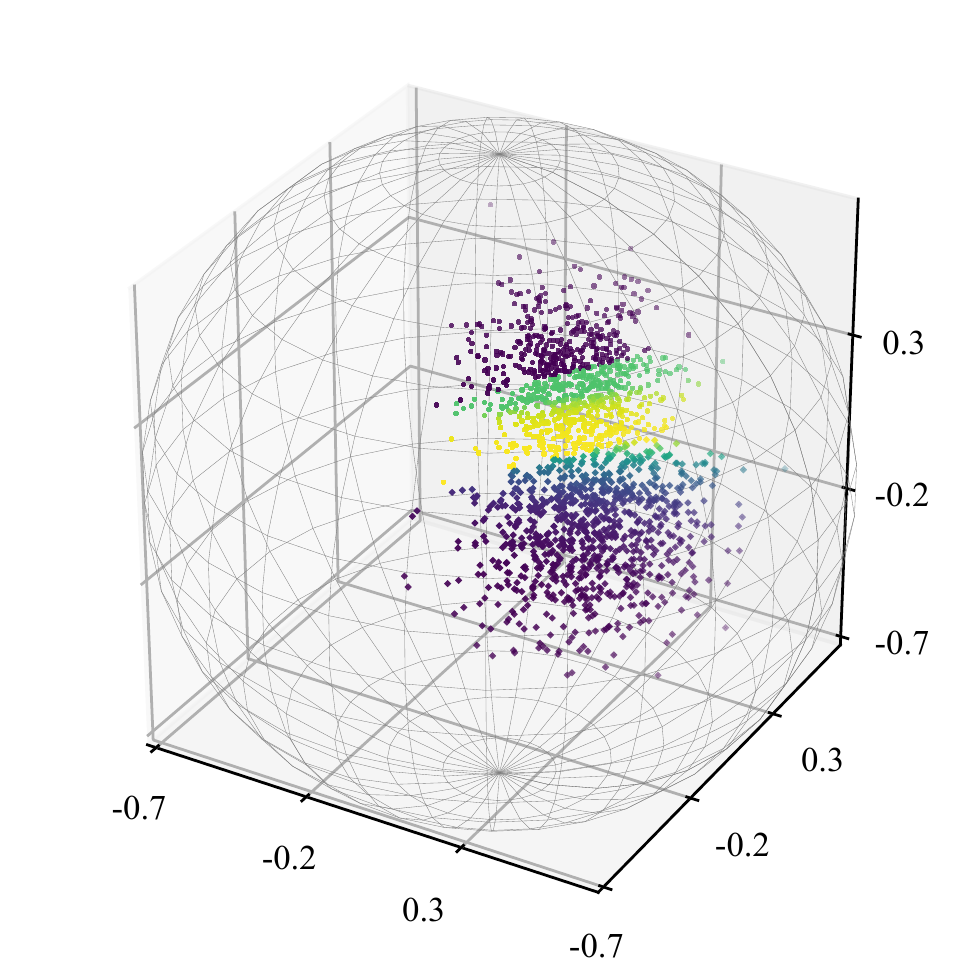}~~~~
		\includegraphics[width=0.28\textwidth]{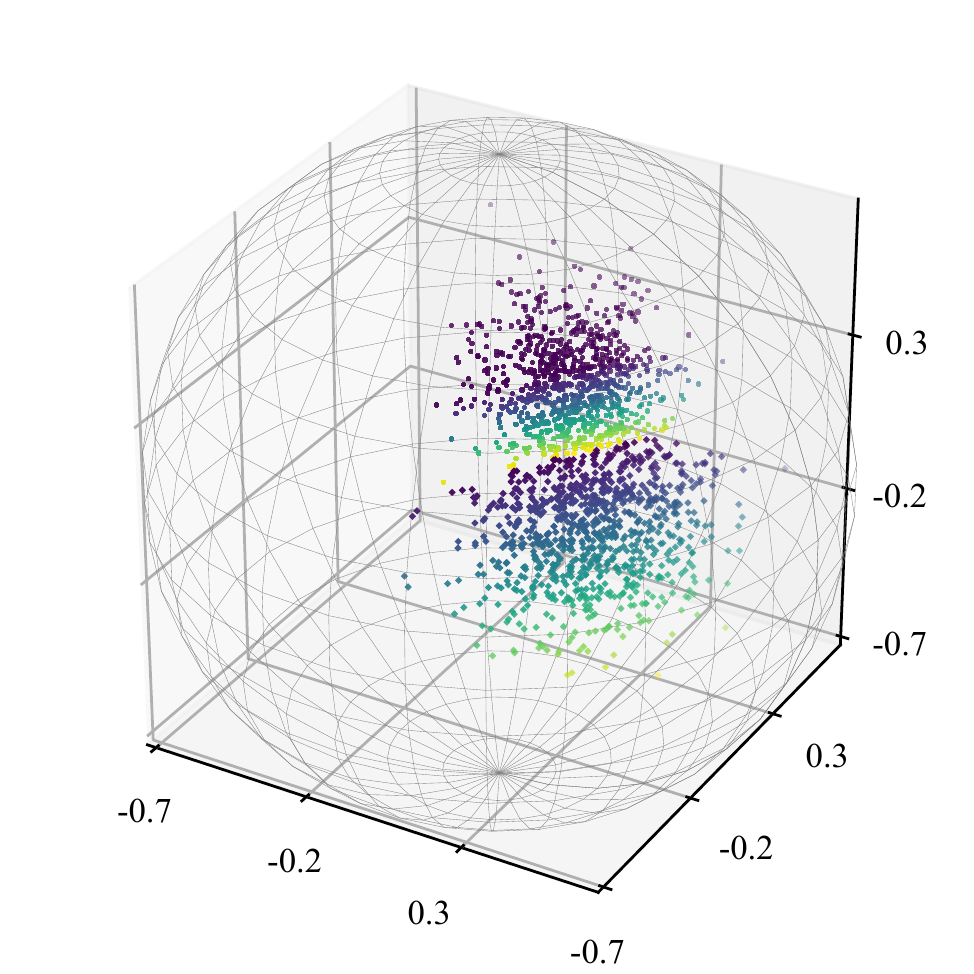}
	\end{center}
	\caption{\footnotesize Toy example of two overlapping Gaussian distributions (classes) on a unit sphere. Left: samples from the distributions as input to a multi layer perceptron (MLP). Middle: AVH heat map produced by MLP, where samples in lighter colors (higher hardness) are mostly overlapping hard examples. Right: $\ell_2$-norm heat map, where certain non-overlapping samples also have higher values.}
	\label{fig:gaussian2}
	\vspace{-2mm}
\end{figure*}

\textbf{Gaussian Simulation Plot:} We generate 2000 3-d random vectors from two multivariate normal distribution (1000 for each) and normalize to unit norm, shown in red and green color on the left plot in Figure~\ref{fig:gaussian2}. Then these data points are passed as the inputs to a simple multi layer perceptron classification model with one $3 \times 2$ hidden layer. Upon convergence, we compute the AVH scores for each data point. The middle image shows the visualization of AVH scores for all data points, with lighter color representing higher AVH scores. It is obvious that AVH scores for points lying on the intersection of two clusters are higher, which agrees with the intuition that those are hard examples. We also compute the $\ell_2$ norm of the feature embeddings shown in the right plot. One can see there is less correlation with hard examples.

\begin{figure*}[h]
	\vspace{-2mm}
\begin{center}
\includegraphics[width=0.28\textwidth]{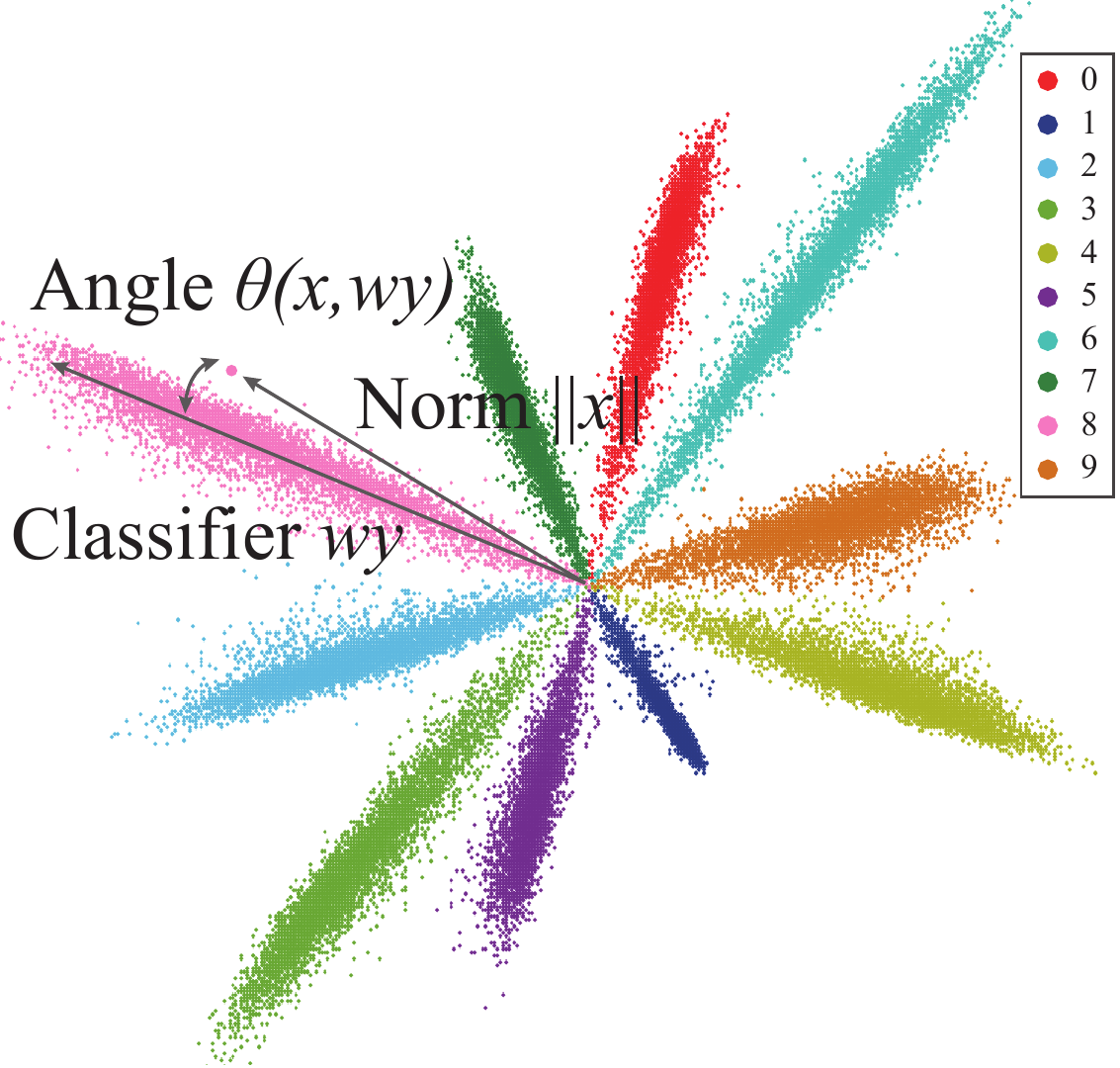}
\caption{\footnotesize Visualization of embeddings on MNIST by setting their dimensions to 2 in a CNN.}\label{fig:diagram1}
\vspace{-0.3cm}
\end{center}
	\vspace{-2mm}
\end{figure*}

\textbf{MNIST Simulation Plot:} We train MNIST with a very simple CNN model which the dimension of the embedding (right before the classifier) is 2. Figure~\ref{fig:diagram1} shows the visualization of those 2D embeddings.   
\clearpage

\section{Additional Results of Training Dynamics}
\label{app:b}
\subsection{Additional Results on ImageNet}
\label{sec:degra}

\begin{figure*}[h!]
	\vspace{-3mm}
	\begin{center}
		\begin{tabular}{ccc}
		\includegraphics[width=0.3 \textwidth]{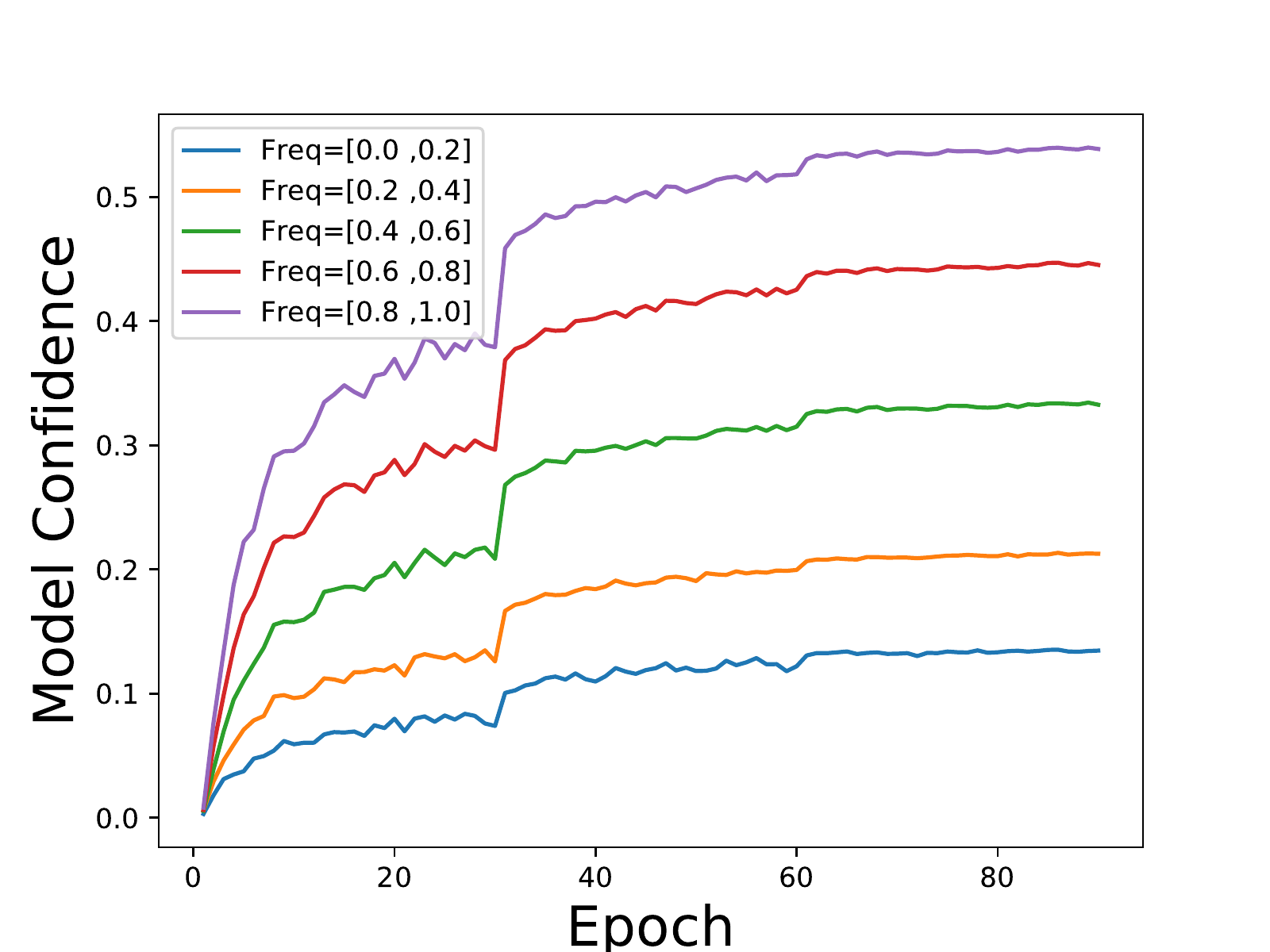} & 
		\includegraphics[width=0.3 \textwidth]{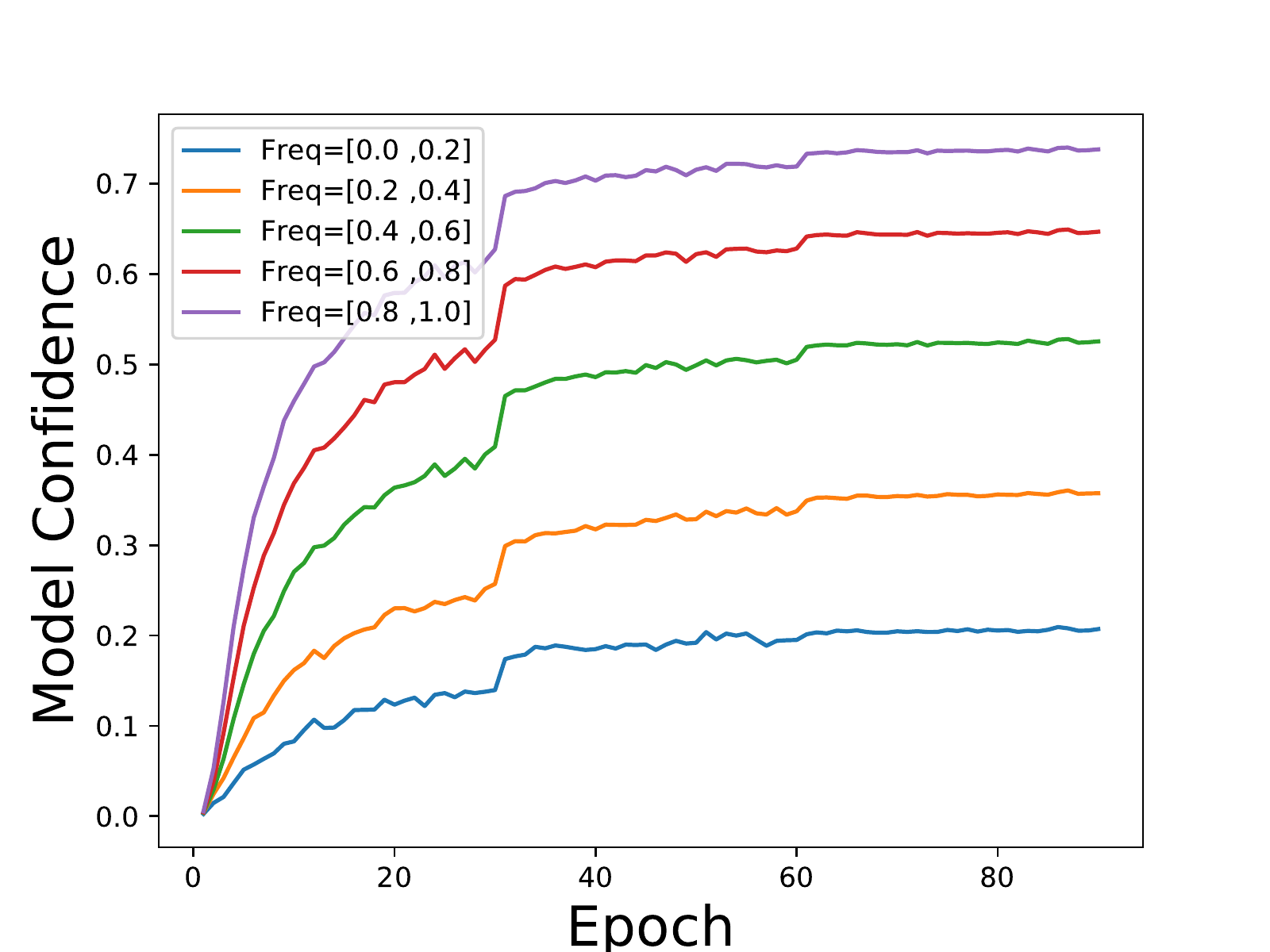} & 
		\includegraphics[width=0.3 \textwidth]{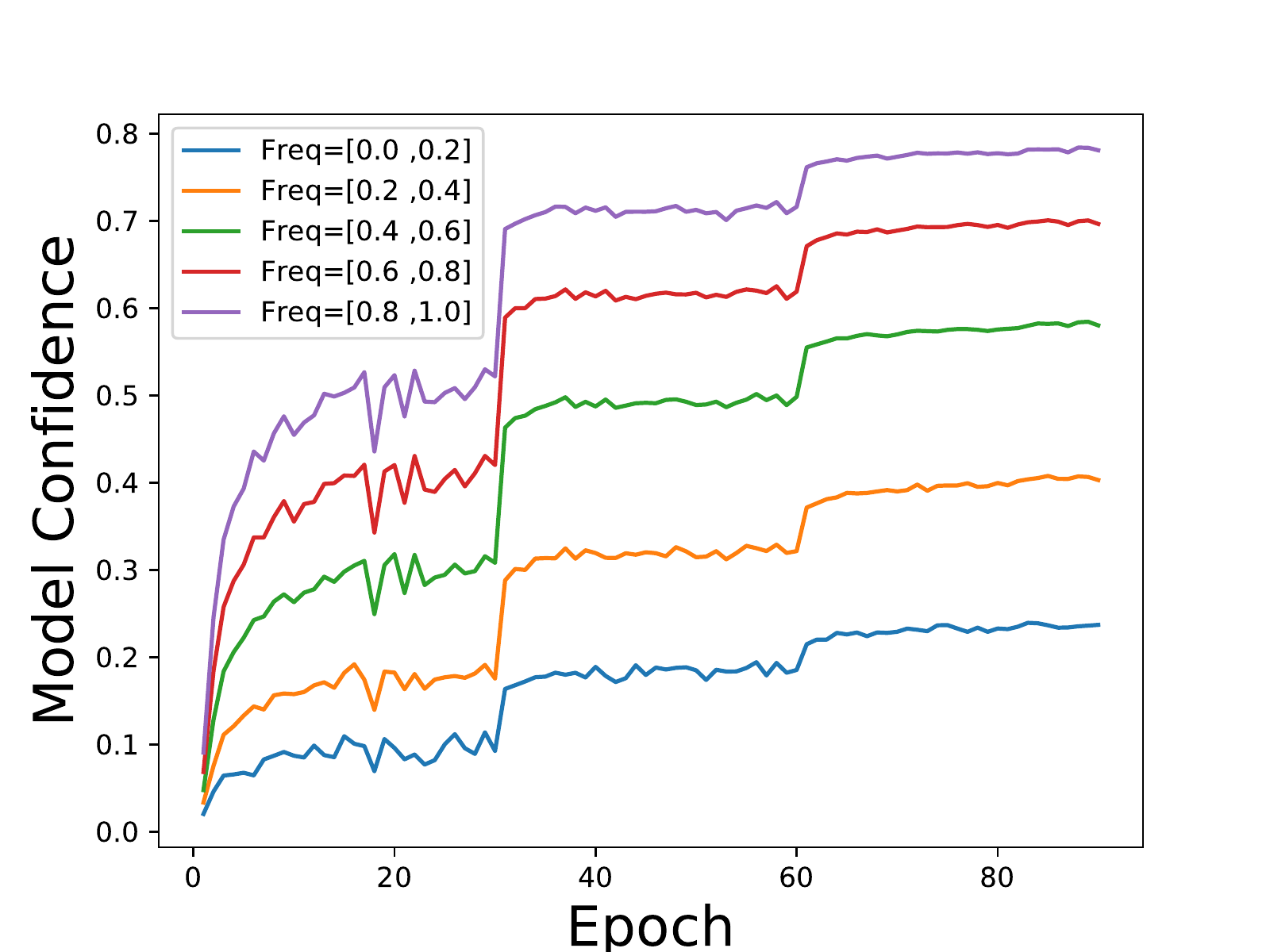}\\
		\end{tabular}
	\end{center}
	\caption{\footnotesize Number of epochs vs. Model Confidence. Results from left to right correspond to AlexNet, VGG-19 and ResNet-50.}\label{fig:softmax}
\end{figure*}

\textbf{Model Confidence:} Figure~\ref{fig:softmax} shows the training dynamics of the model confidence corresponding to AlexNet, VGG-19, and ResNet-50.

\begin{figure*}[h!]
	\begin{center}
		\begin{tabular}{ccc}
			\includegraphics[width=0.3 \textwidth]{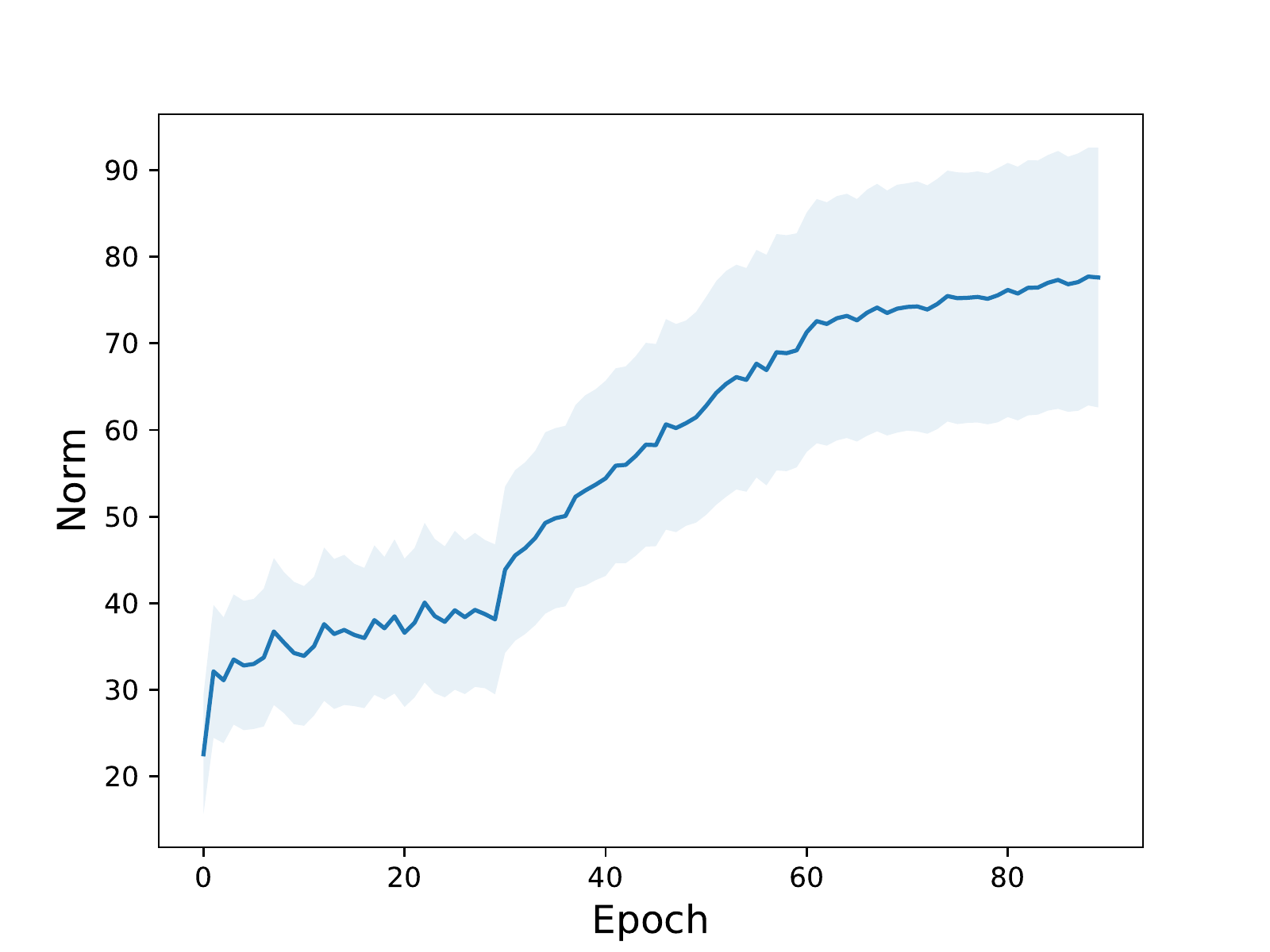} & 
			\includegraphics[width=0.3 \textwidth]{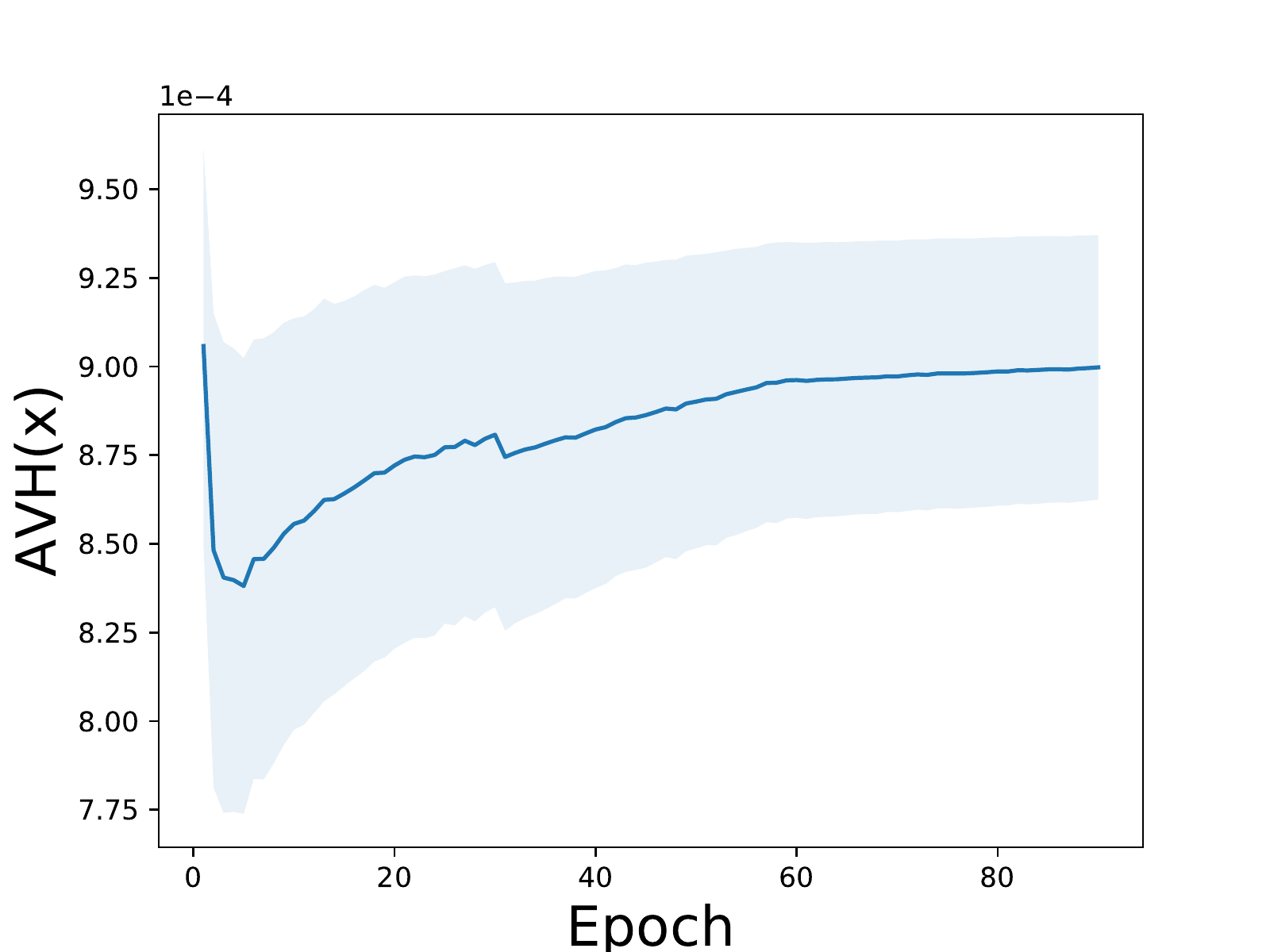} & 
			\includegraphics[width=0.3 \textwidth]{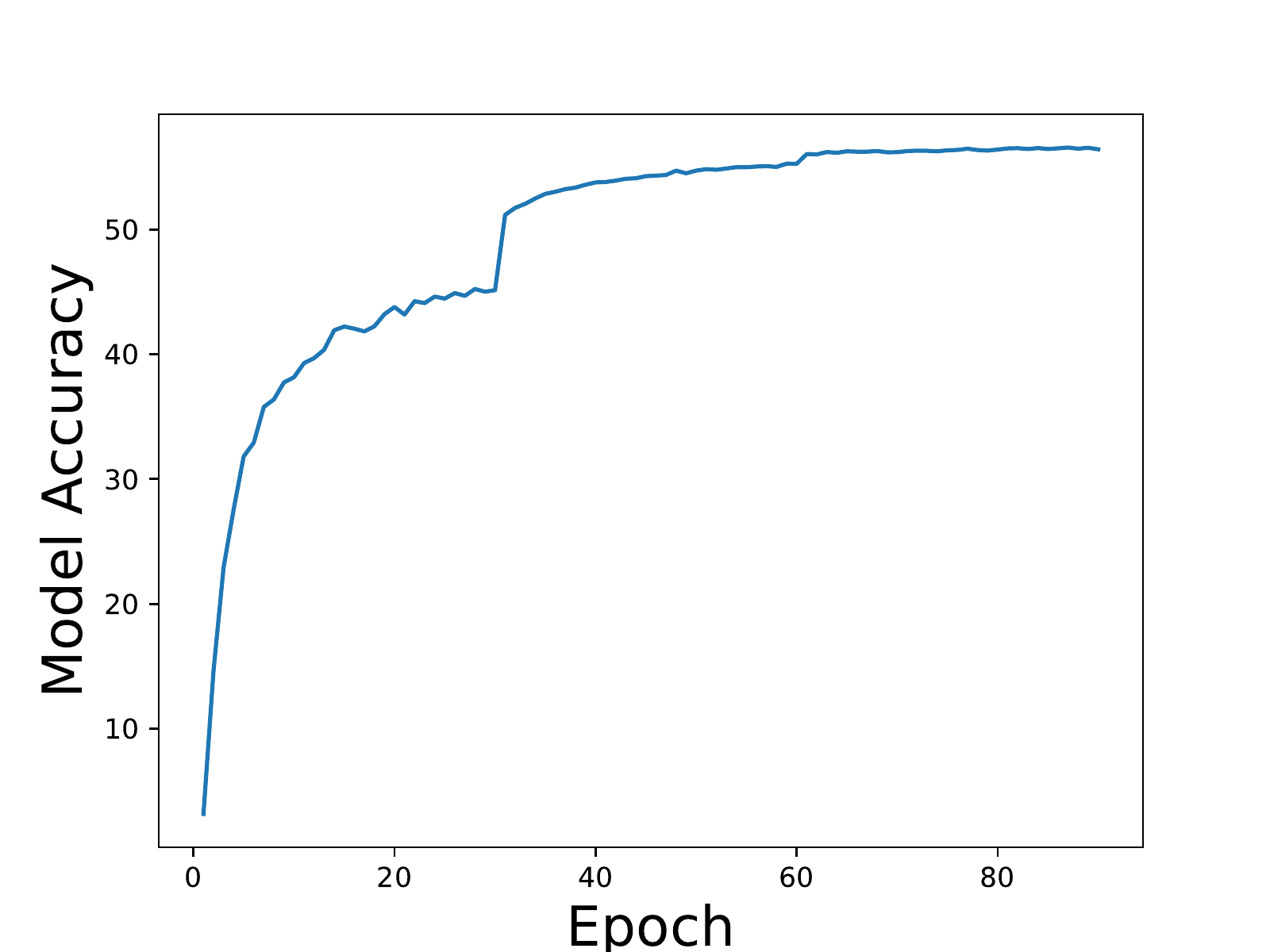} \\
			\includegraphics[width=0.3 \textwidth]{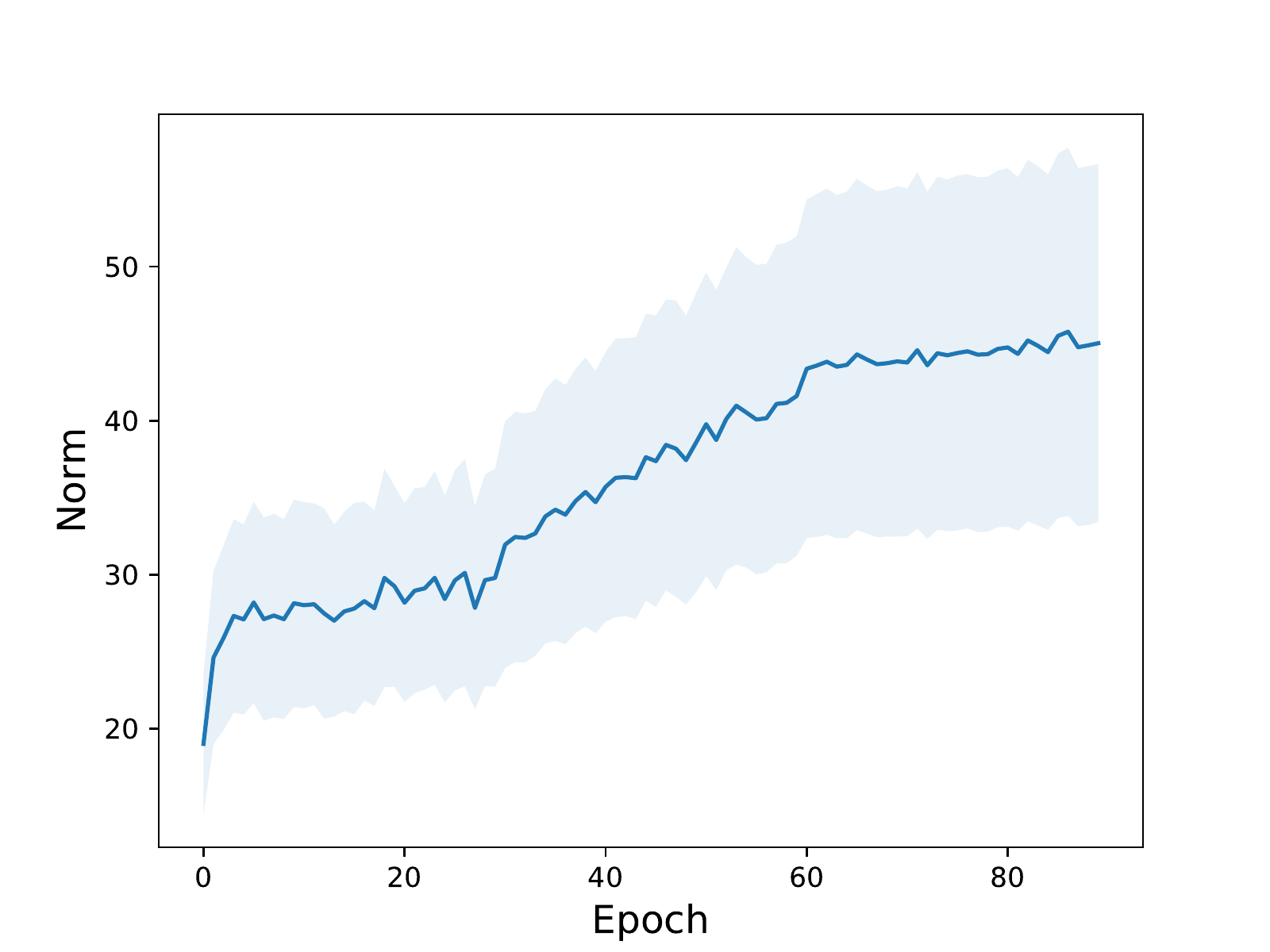} & 
			\includegraphics[width=0.3 \textwidth]{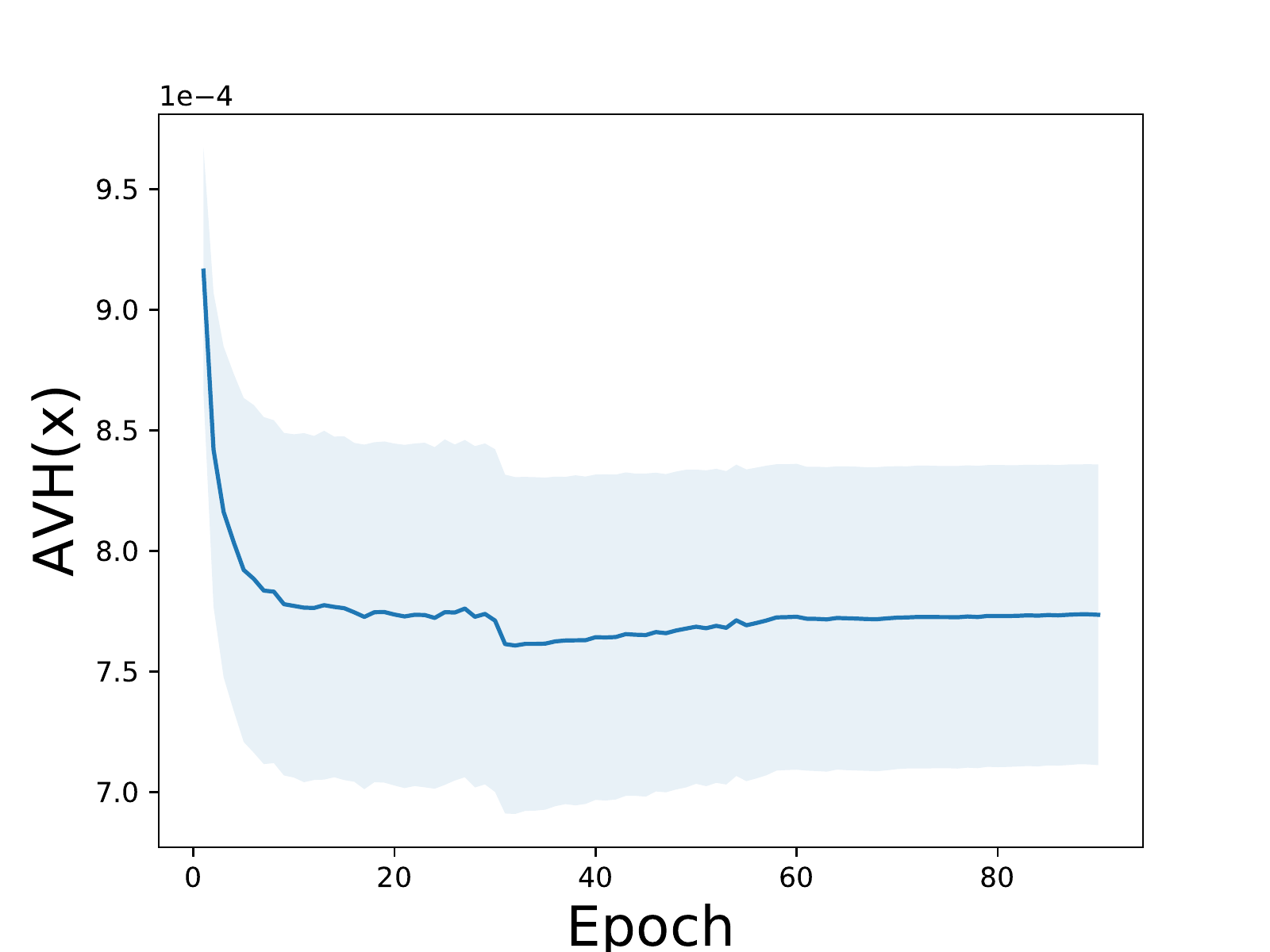} & 
			\includegraphics[width=0.3 \textwidth]{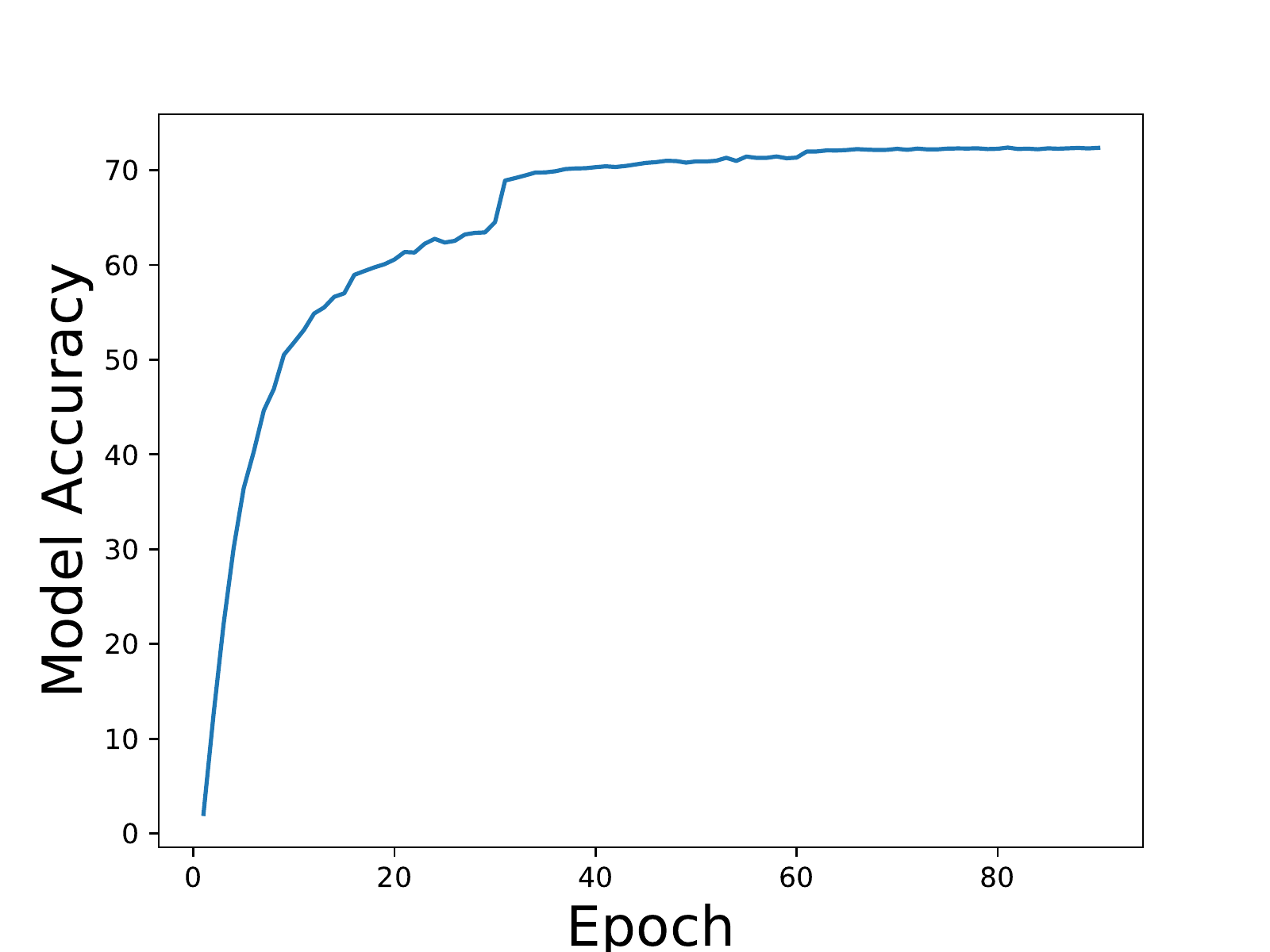} \\
			\includegraphics[width=0.3 \textwidth]{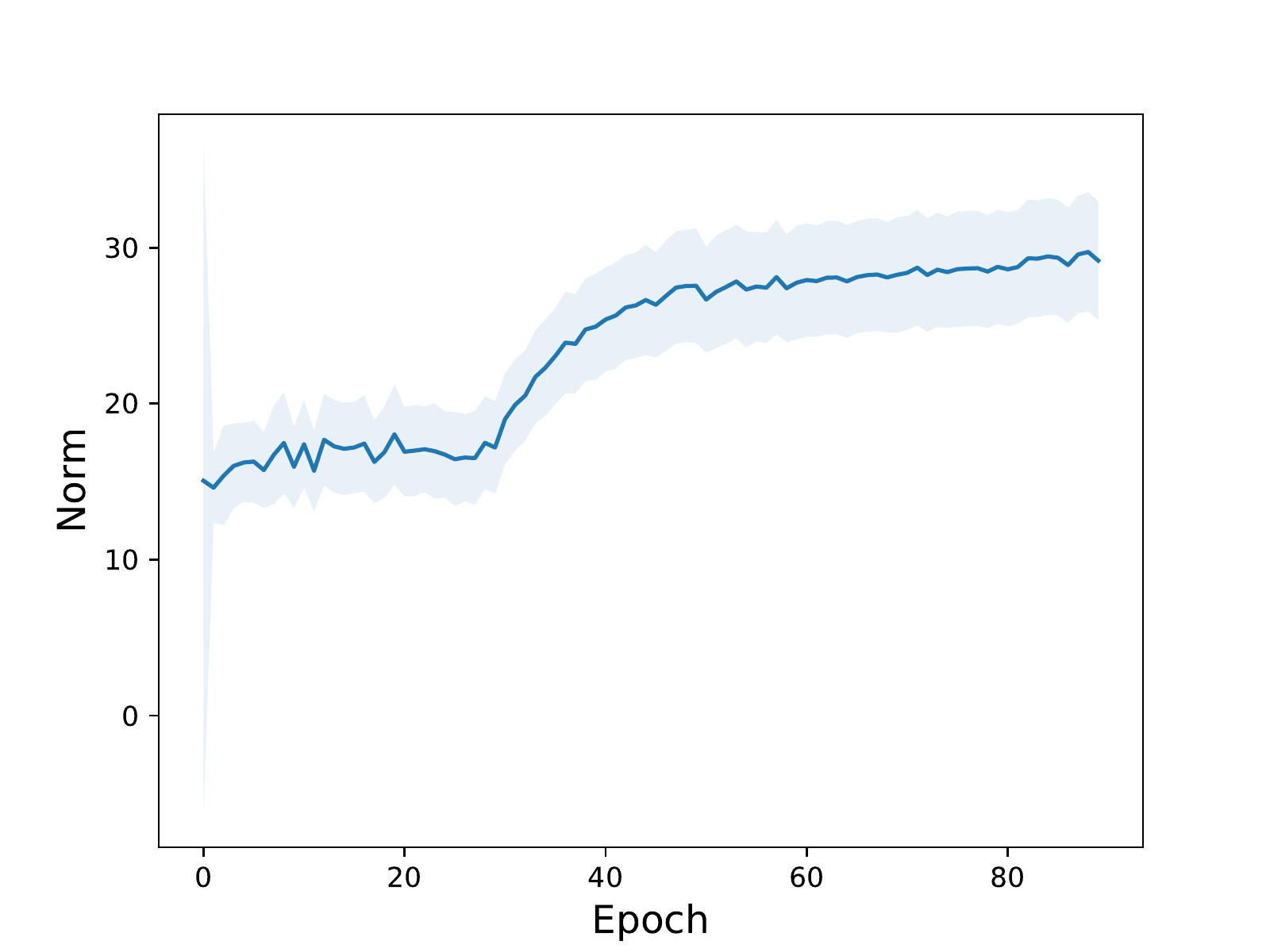} & 
			\includegraphics[width=0.3 \textwidth]{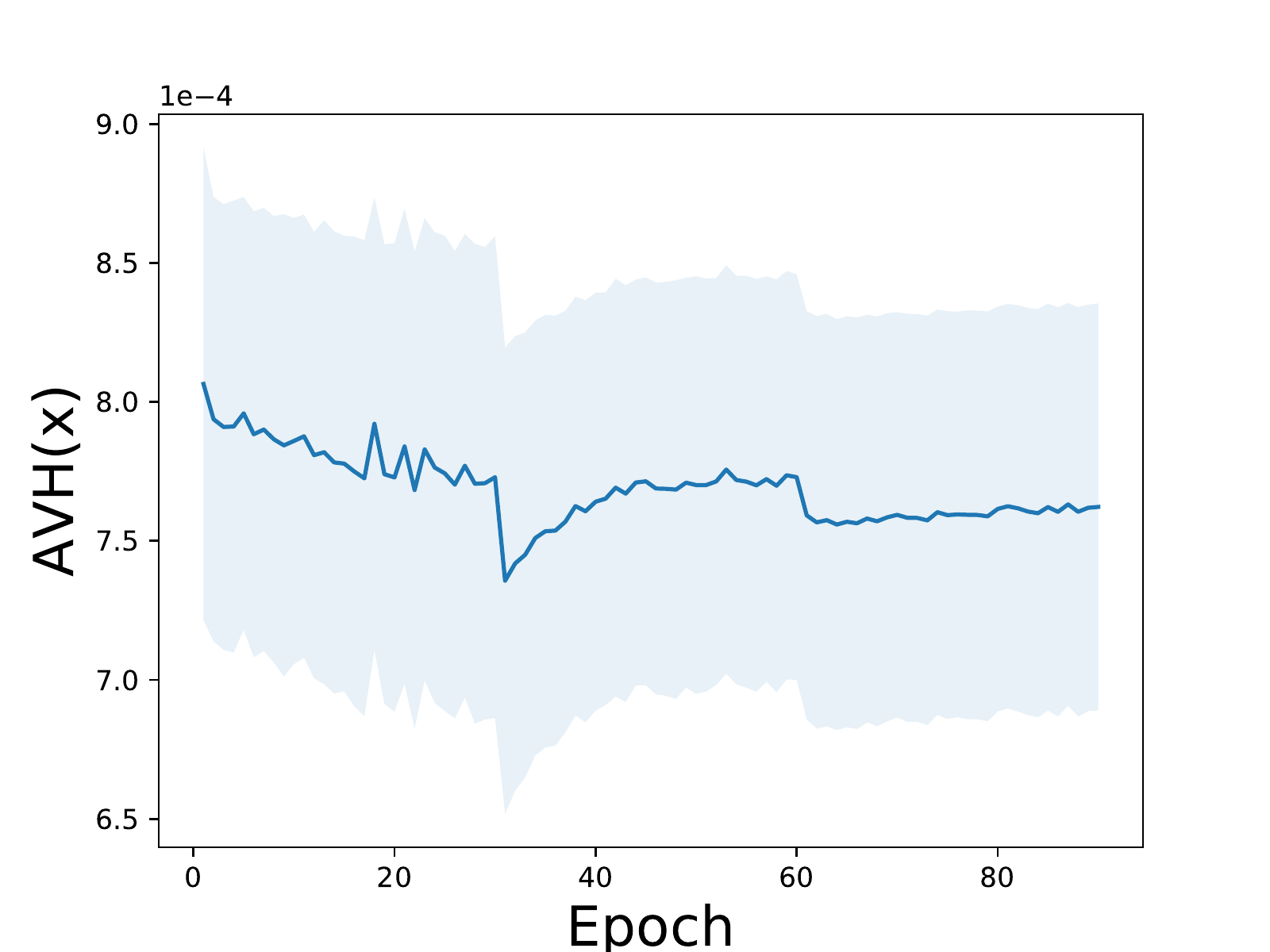} & 
			\includegraphics[width=0.3 \textwidth]{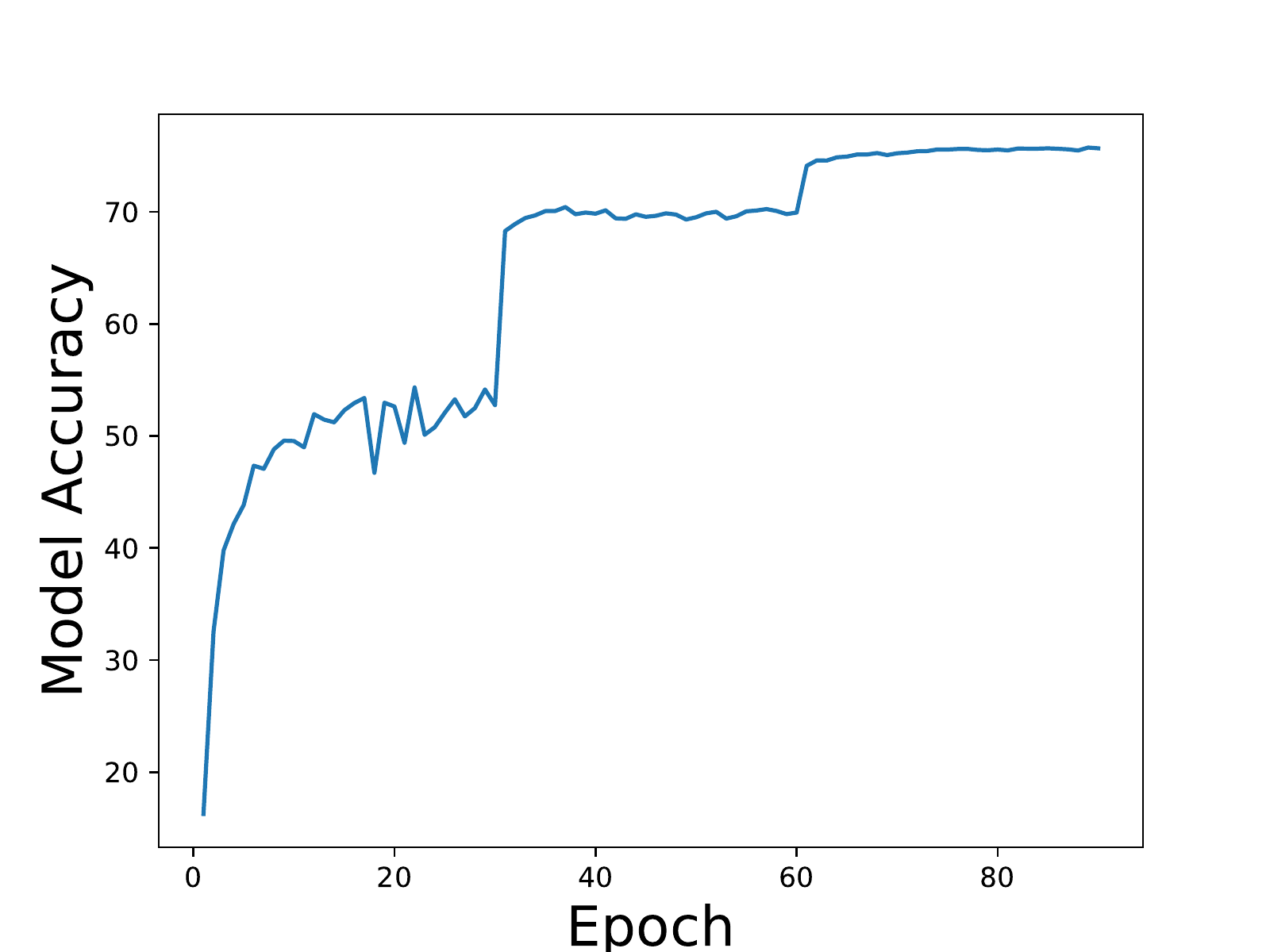}\\
			\includegraphics[width=0.3 \textwidth]{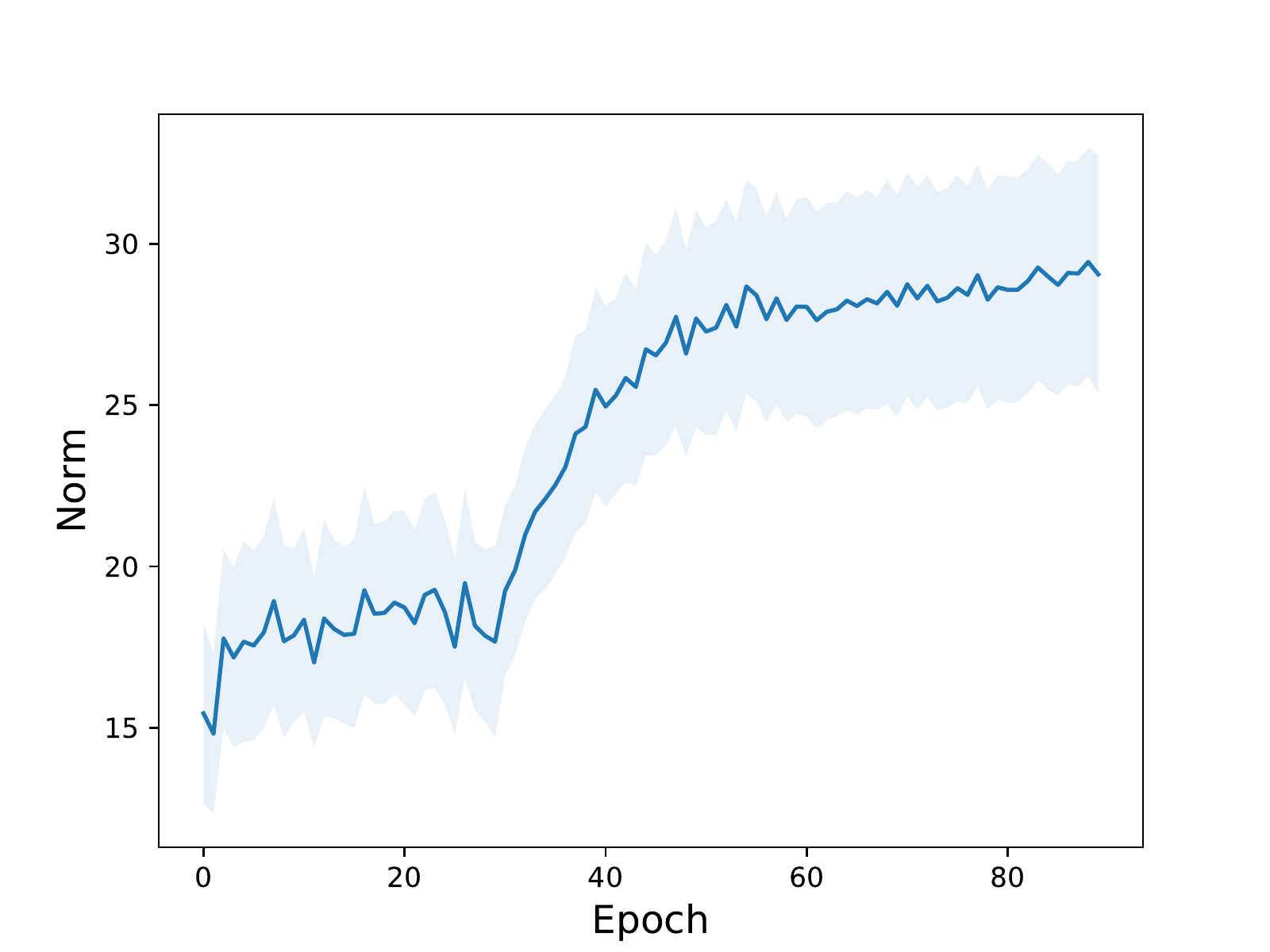} & 
			\includegraphics[width=0.3 \textwidth]{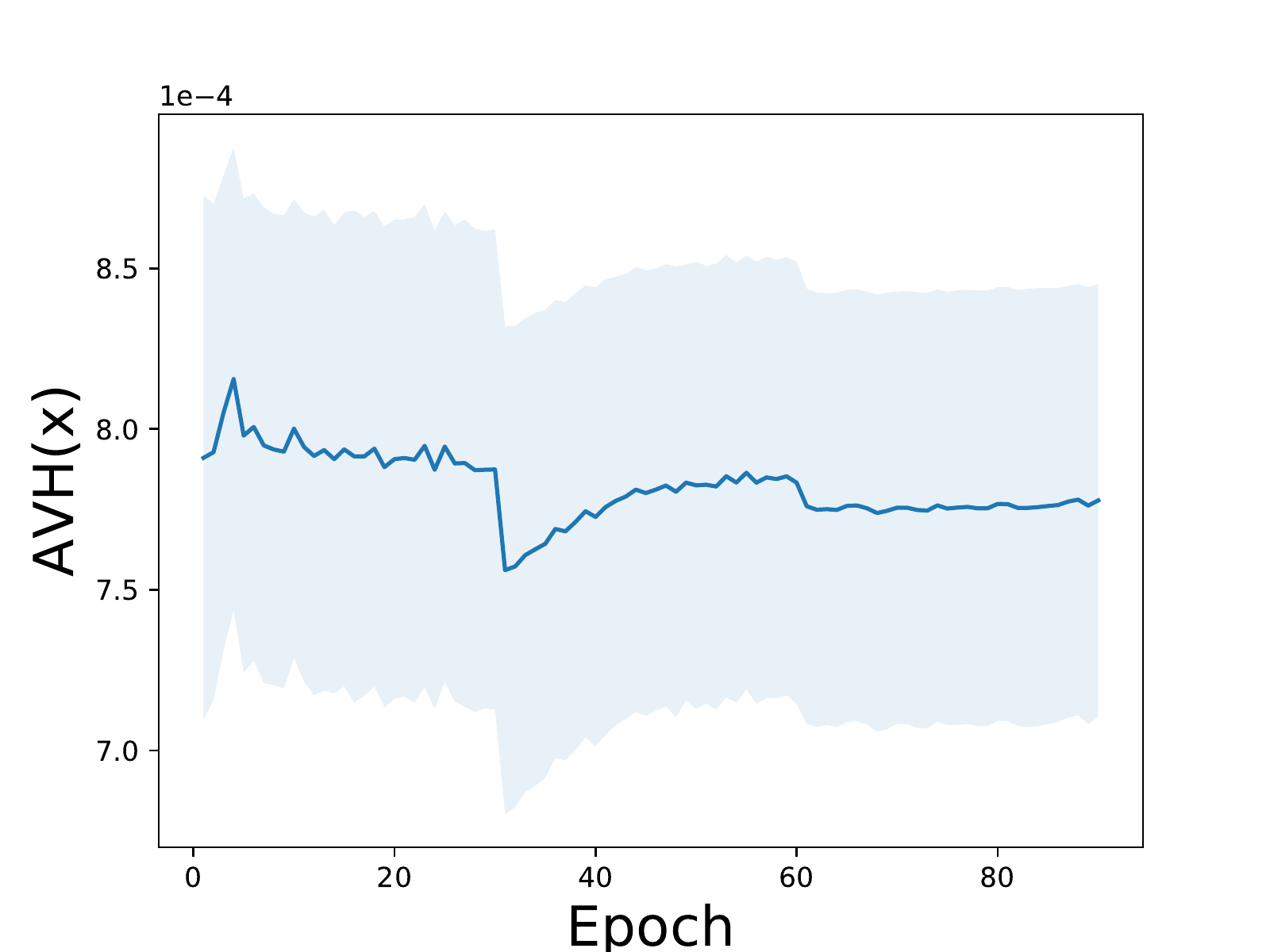} & 
			\includegraphics[width=0.3 \textwidth]{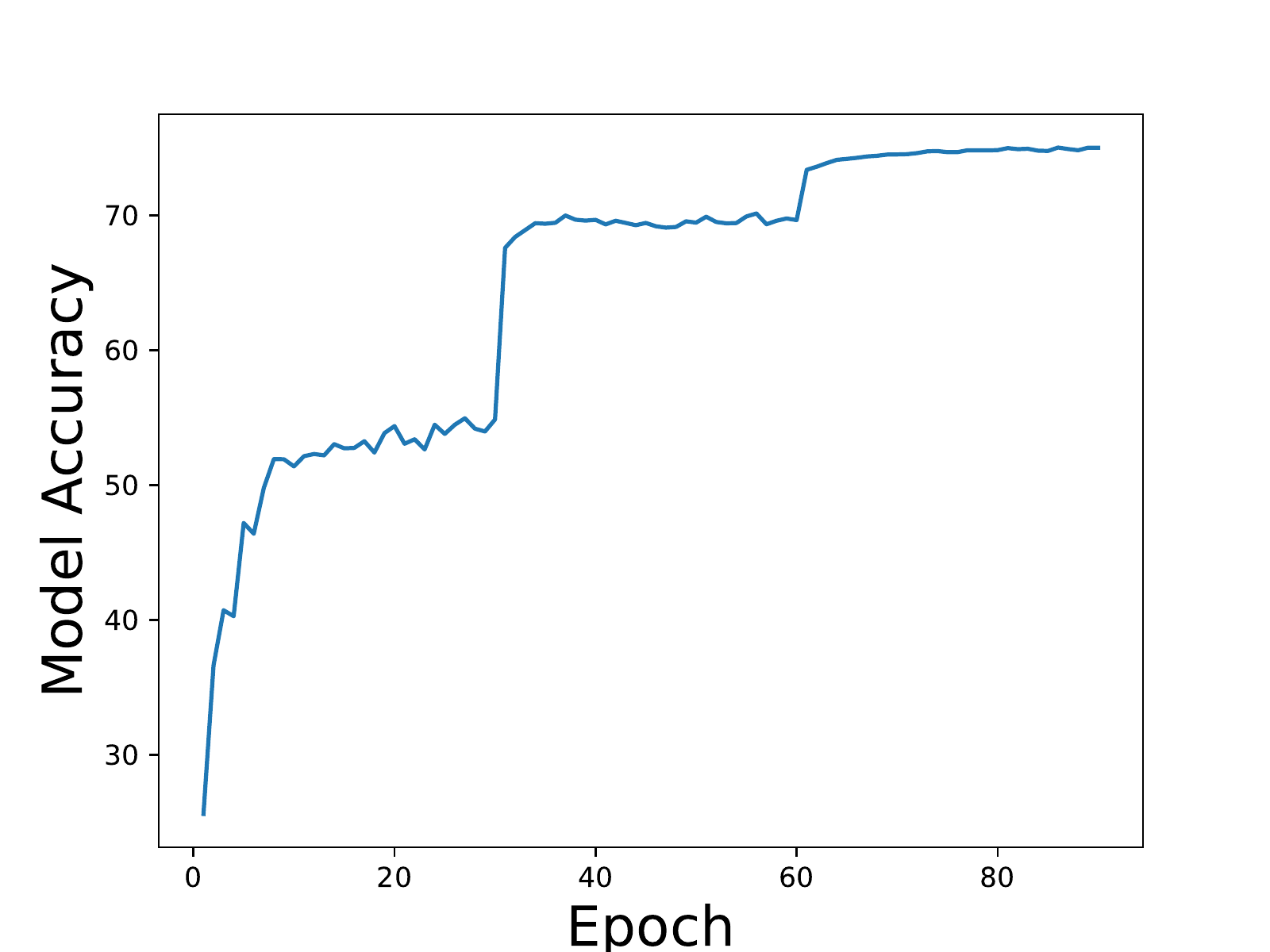}
		\end{tabular}
	\end{center}
	\caption{\footnotesize Averaged training dynamics on ImageNet validation set. Columns from left to right: number of epochs vs. average $\ell_2$ norm, number of epochs vs. average AVH score, and number of epochs vs. model accuracy. Rows from top to bottom: dynamics corresponding to AlexNet, VGG-19, ResNet-50, and DenseNet-121.}\label{fig:average}
\end{figure*}

\textbf{Averaged training dynamics:} In Figure~\ref{fig:average}, we plot the average embedding norm, AVH and model accuracies for AlexNet, VGG-19, ResNet-50 and DenseNet-121 over the validation samples.

\textbf{Image degradation: }
Because CNNs and humans can achieve similar accuracy on large-scale benchmark dataset such as ImageNet, a number of works have investigated similarities and differences between CNNs and human vision~\cite{Martin_Cichy_2017, Kheradpisheh_2016, Dodge_2017a, dekel2017human, Pramod_2016, alex2017eigendistortions}. Since human annotation data is relatively hard to obtain, researchers have proposed an alternative measure of visual hardness on images based on image degradation~\cite{lindsay2013human}. This involves adding noise or changing image properties such as contrast, blurriness, and brightness.~\cite{geirhos2018generalisation} employed psychological studies to validate the degradation method as a way to measure human visual hardness. It should be noted that the artificial visual hardness introduced by degradation is a different concept from the natural visual hardness. The hardness based on degradation only reflects the hardness of a single original image with various of transformations, while natural visual hardness based on the ambiguity of human perception across a distribution of natural images. In the following additional experiments, we also consider different level of degradation as the surrogate of human visual hardness besides Human Selection Frequency.

\begin{defn}[Image Degradation Level]
We define another way to measure human visual hardness on pictures as Image Degradation Level. We consider two degradation methods in this paper, decreasing contrast and adding noise. Quantitatively, Image Degradation Level for decreasing contrast is directly the contrast level.
Image Degradation Level for adding noise is the amount of pixel-wise additive uniform noise.
\end{defn}

\begin{figure*}[h!]
	\begin{center}
		\begin{tabular}{ccc}
			\includegraphics[width=0.3 \textwidth]{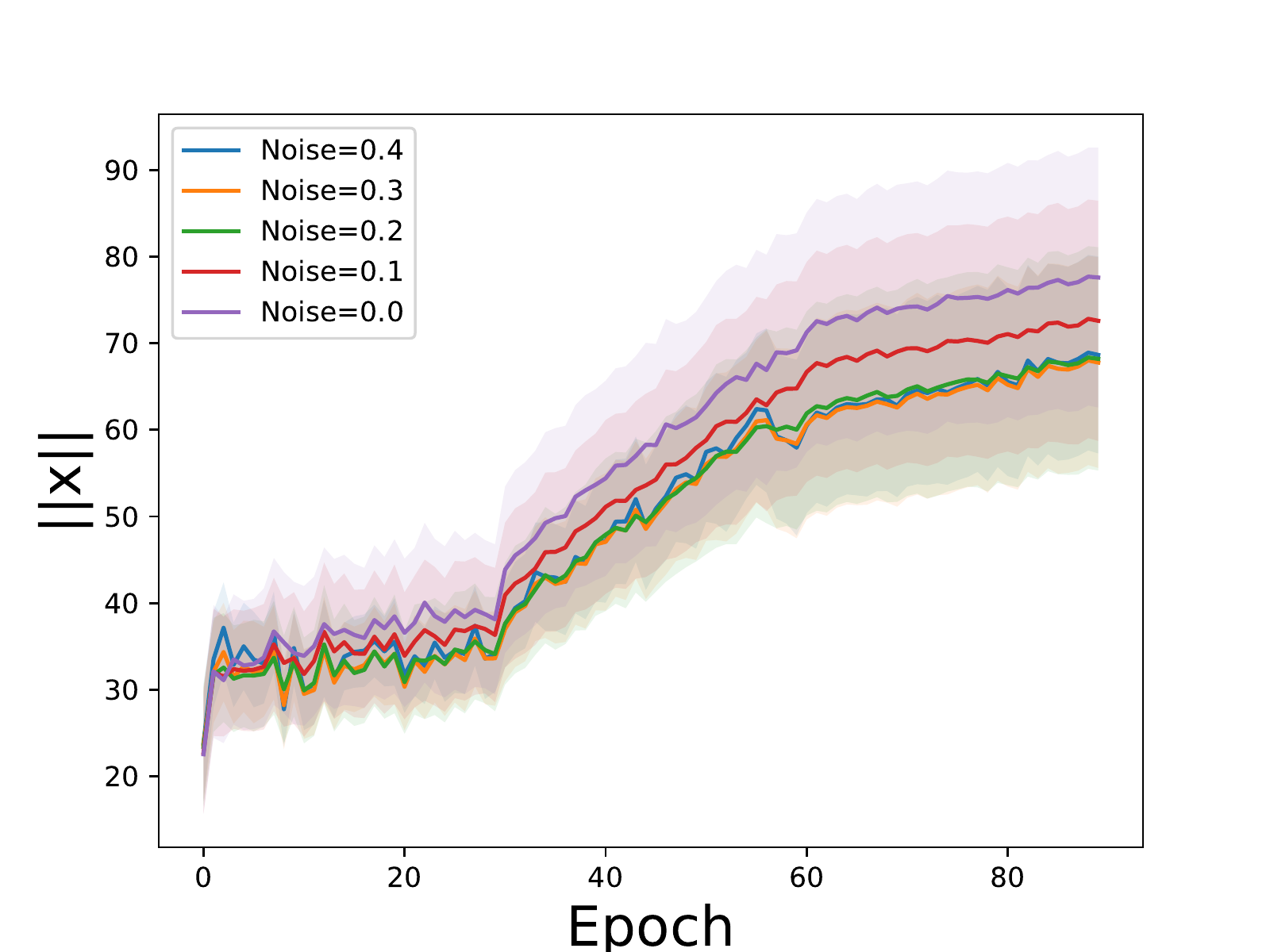} & 
			\includegraphics[width=0.3 \textwidth]{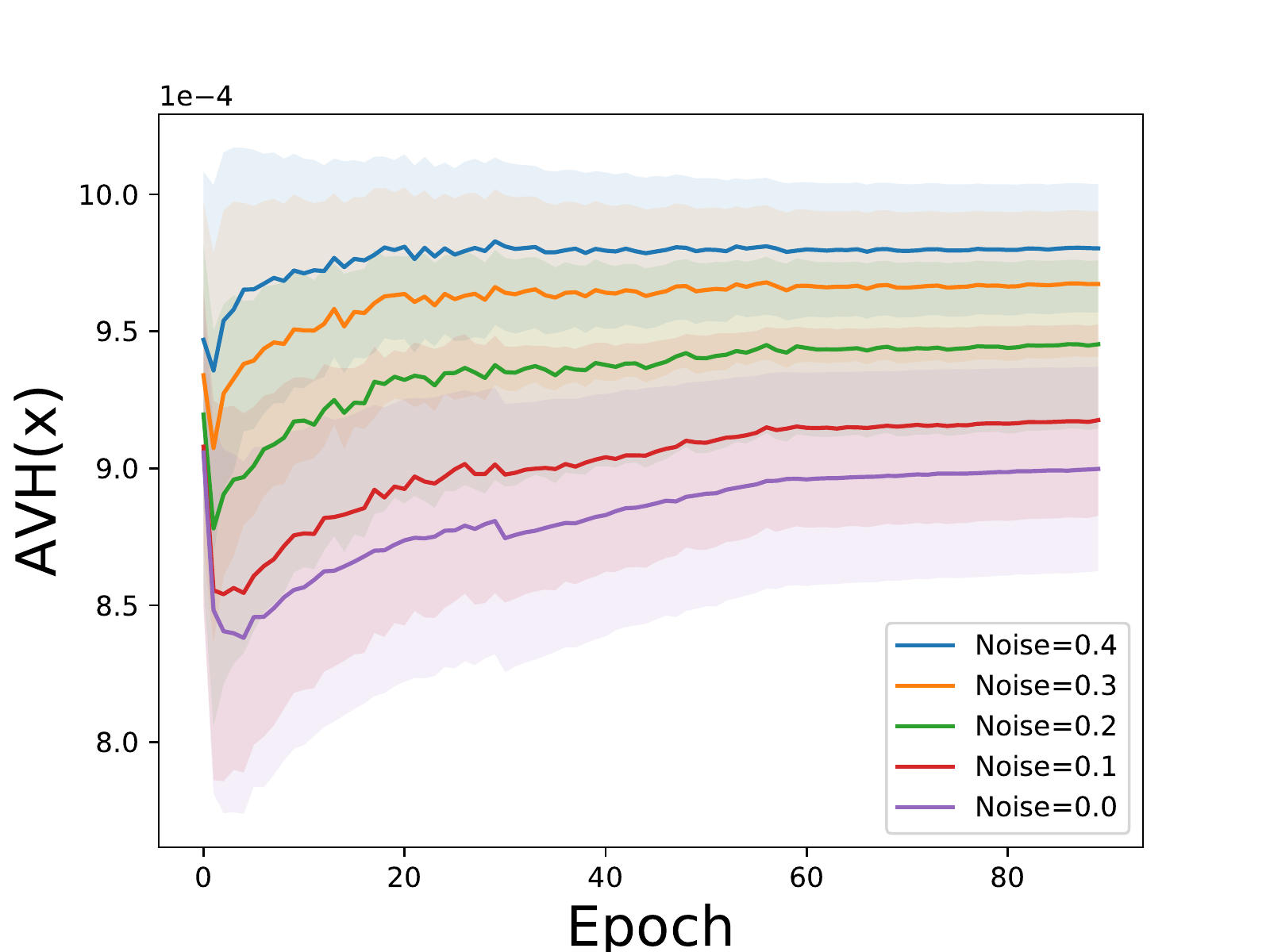} & 
			\includegraphics[width=0.3 \textwidth]{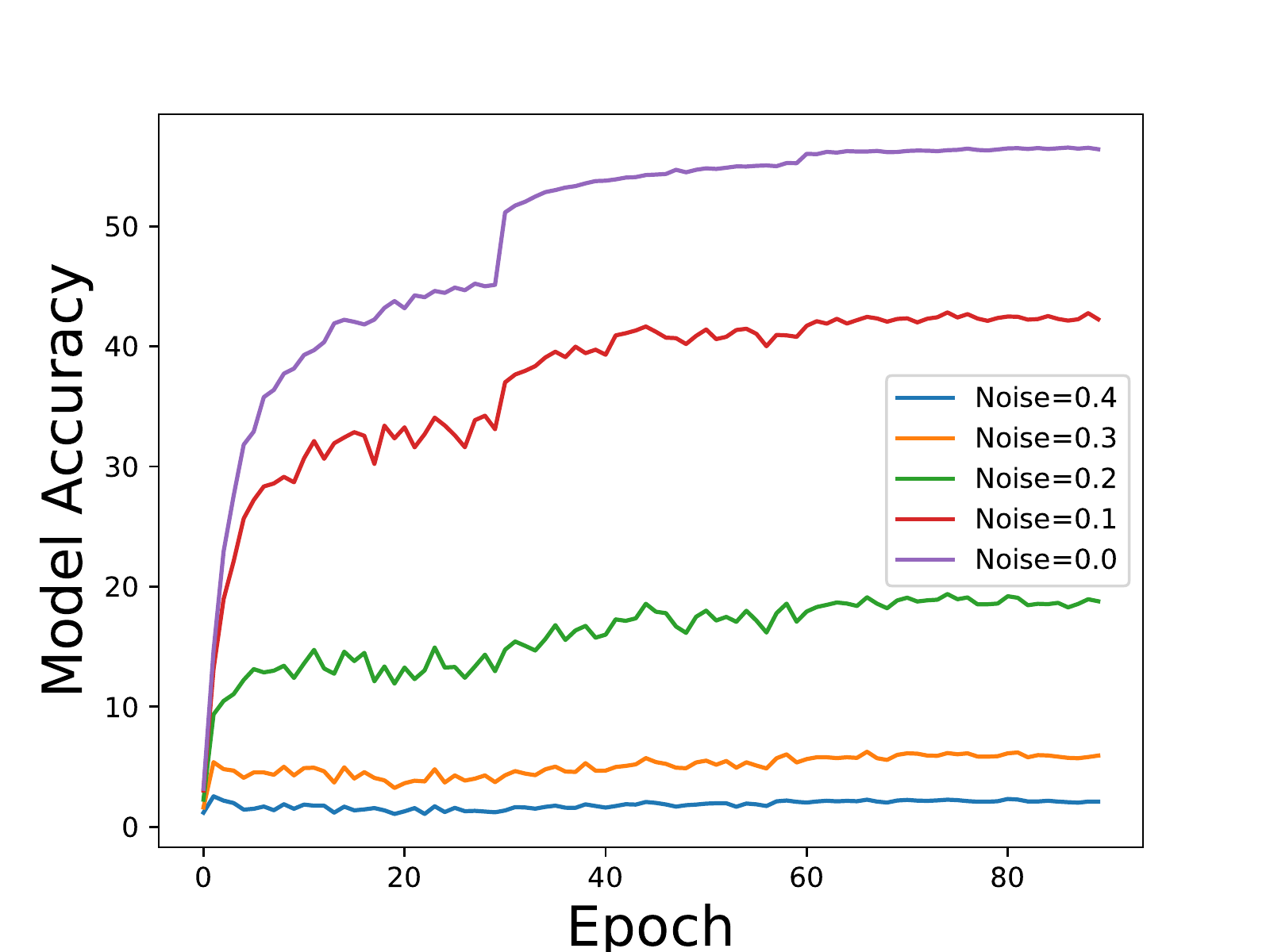} \\
			\includegraphics[width=0.3 \textwidth]{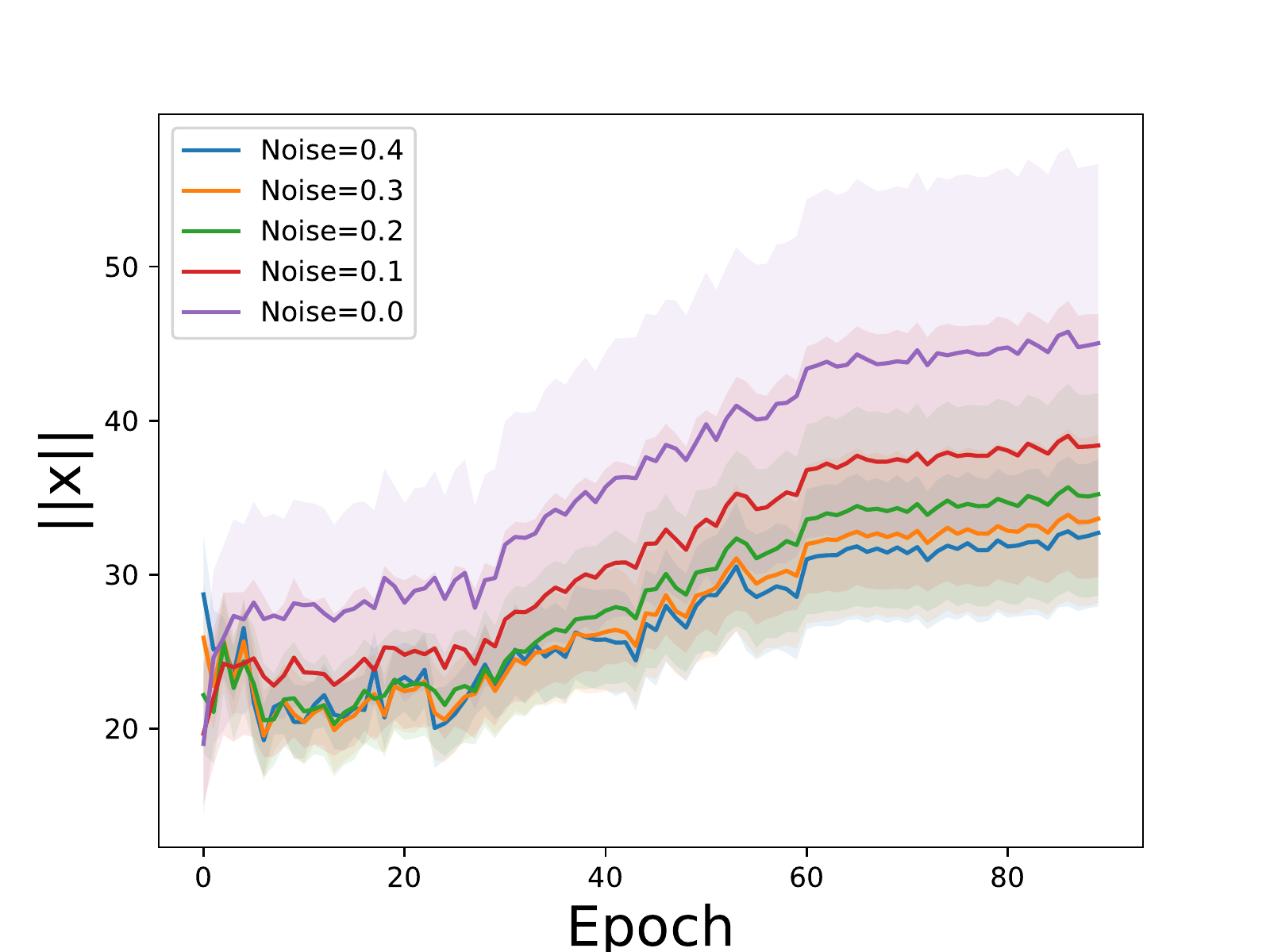} & 
			\includegraphics[width=0.3 \textwidth]{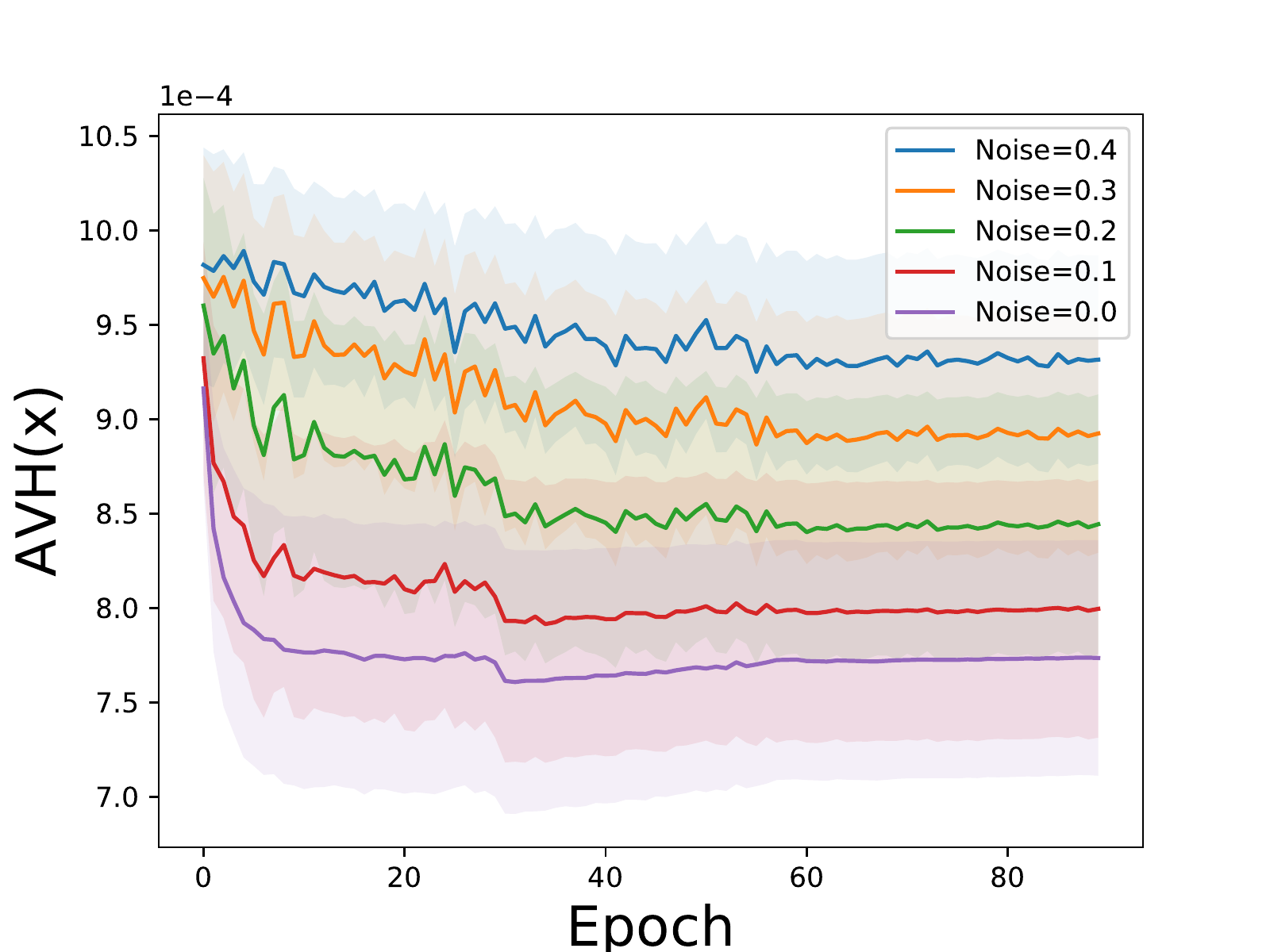} & 
			\includegraphics[width=0.3 \textwidth]{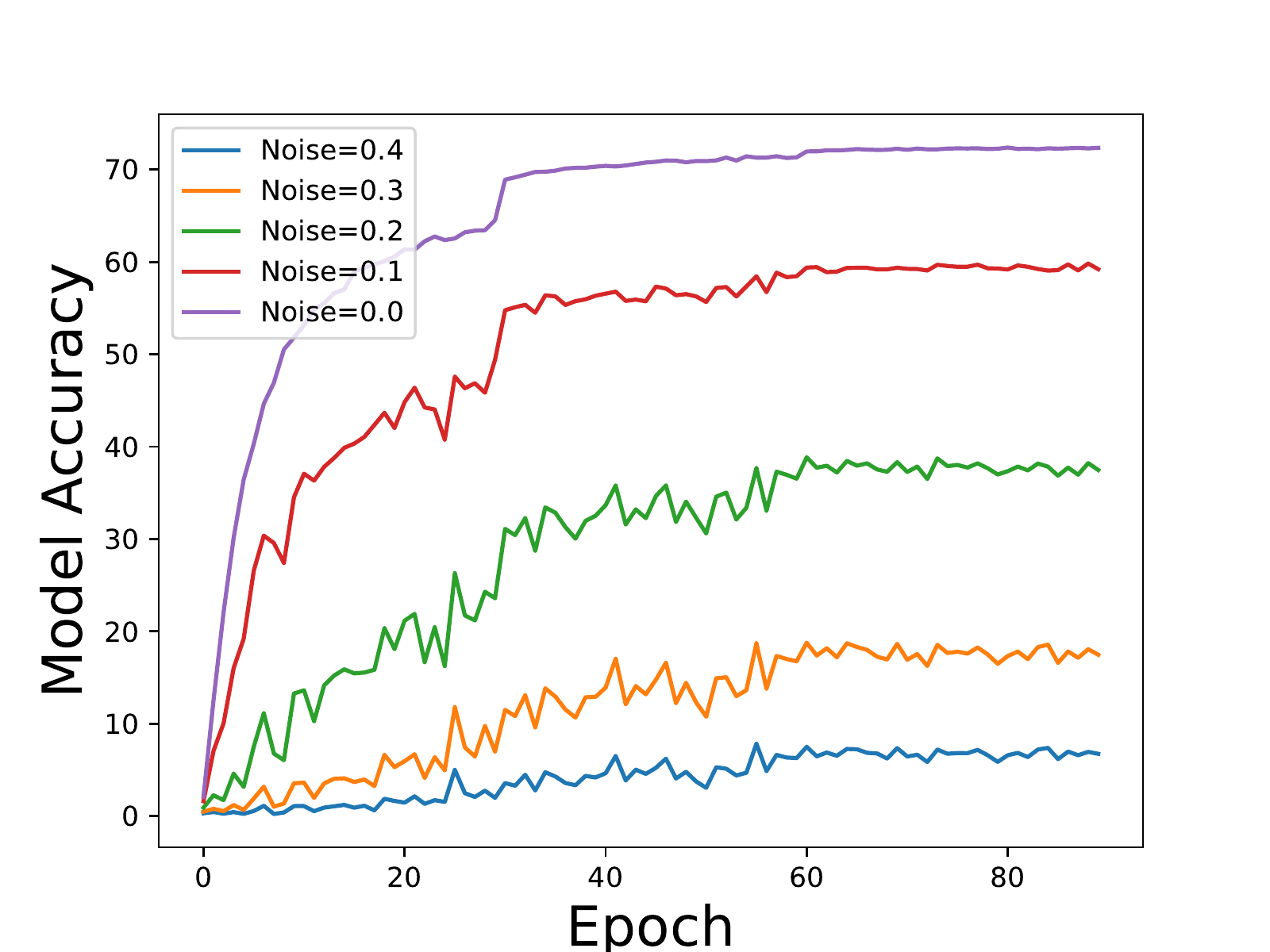} \\
			\includegraphics[width=0.3 \textwidth]{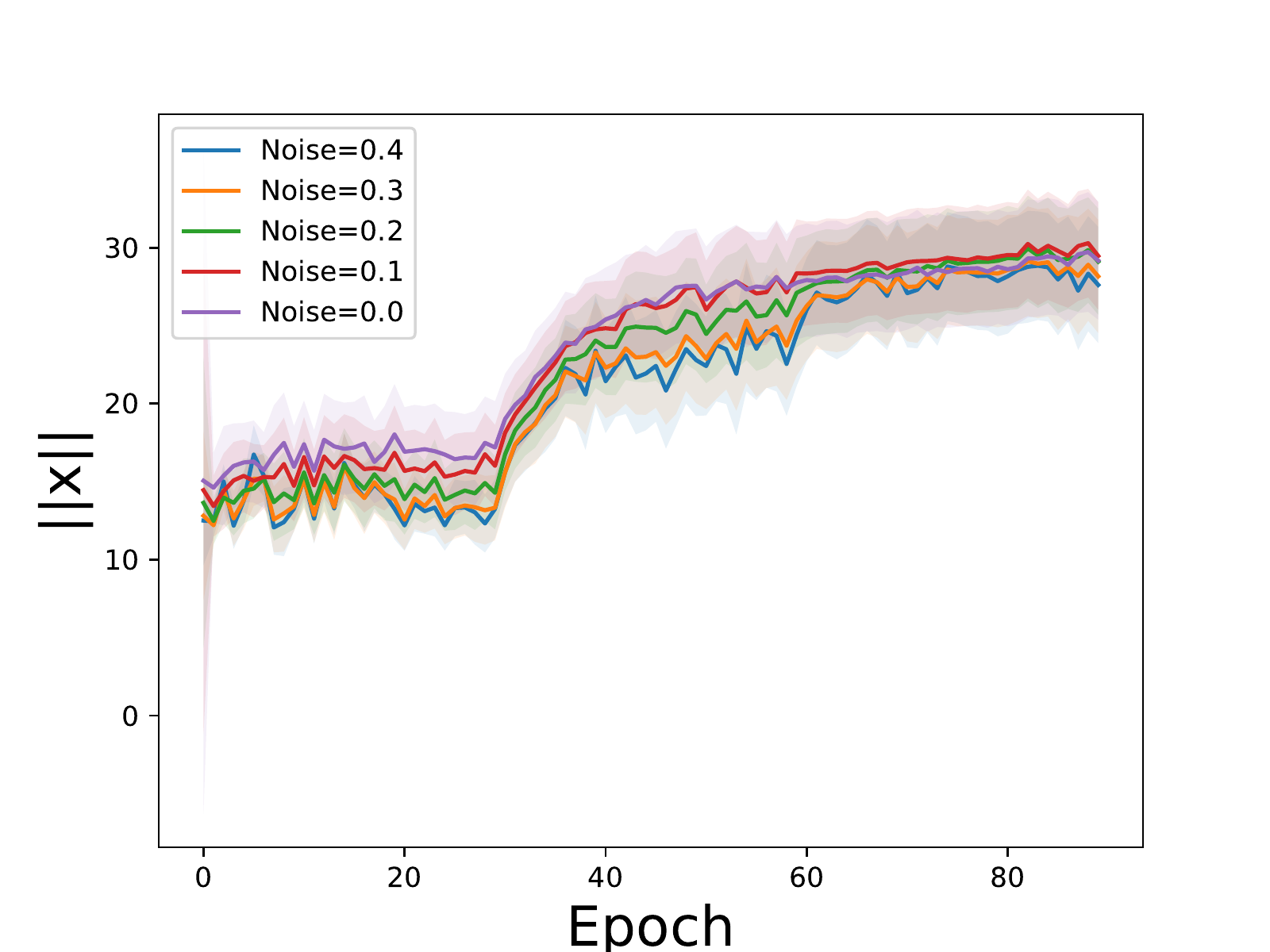} & 
			\includegraphics[width=0.3 \textwidth]{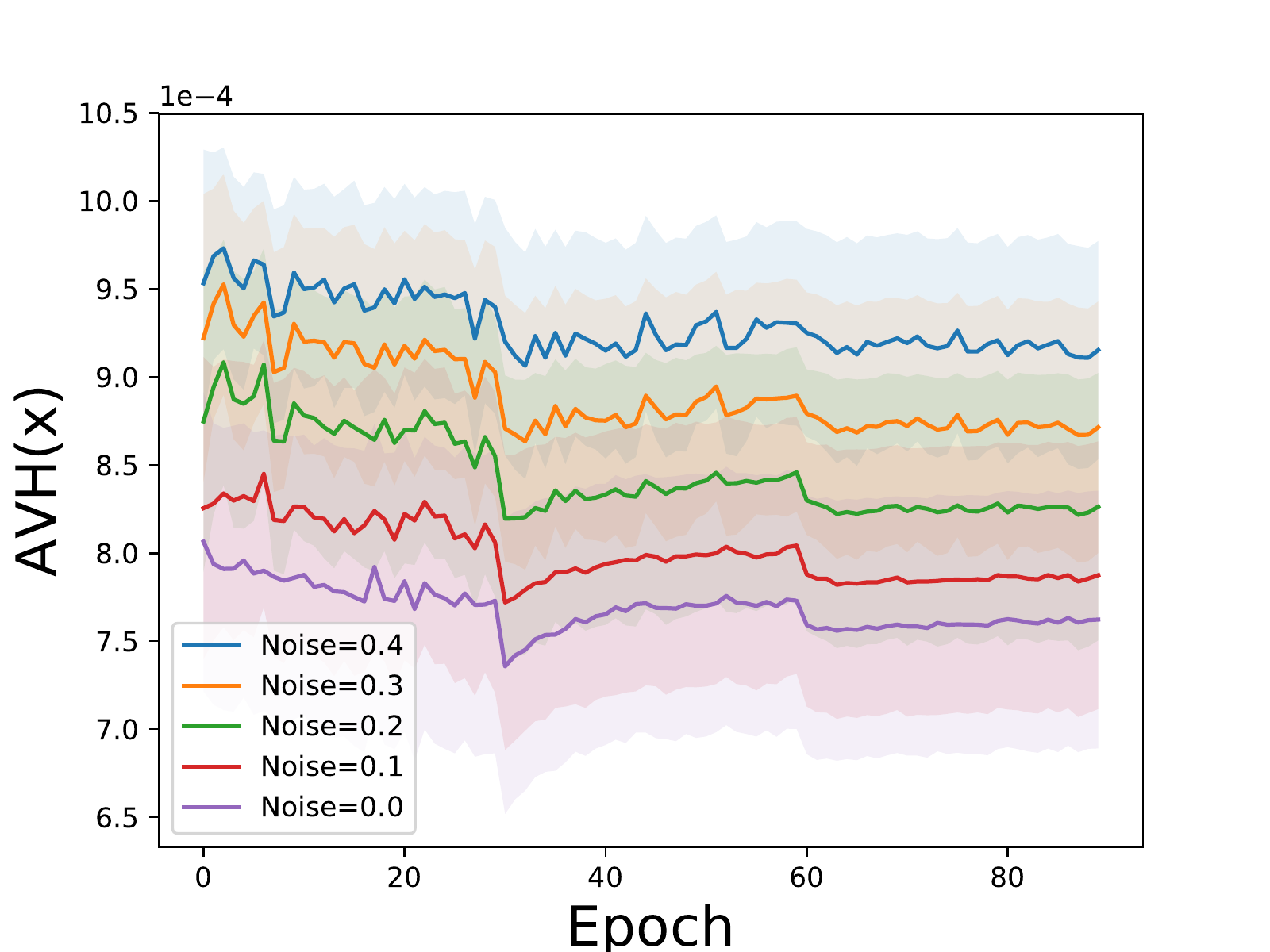} & 
			\includegraphics[width=0.3 \textwidth]{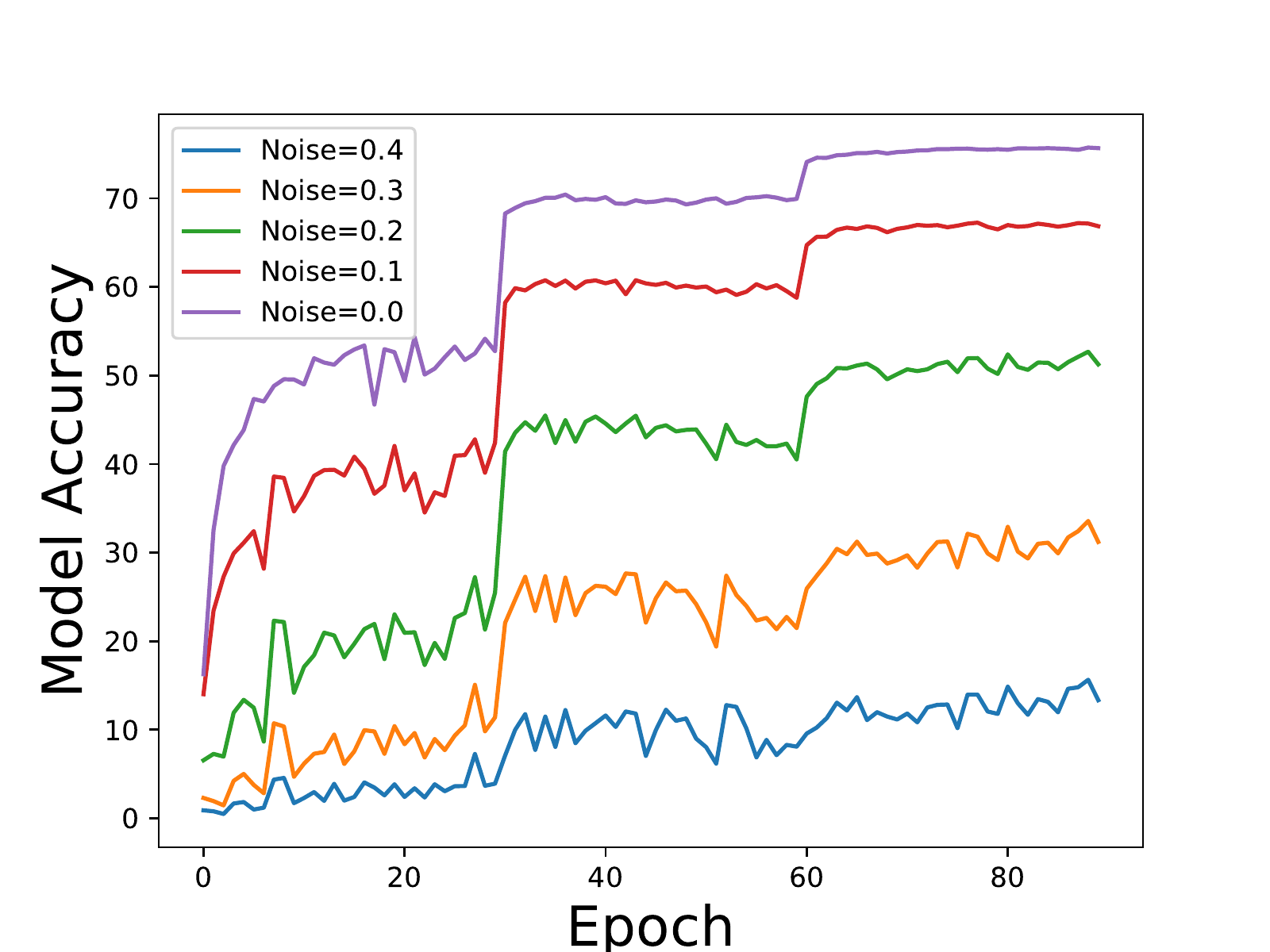} \\
			\includegraphics[width=0.3 \textwidth]{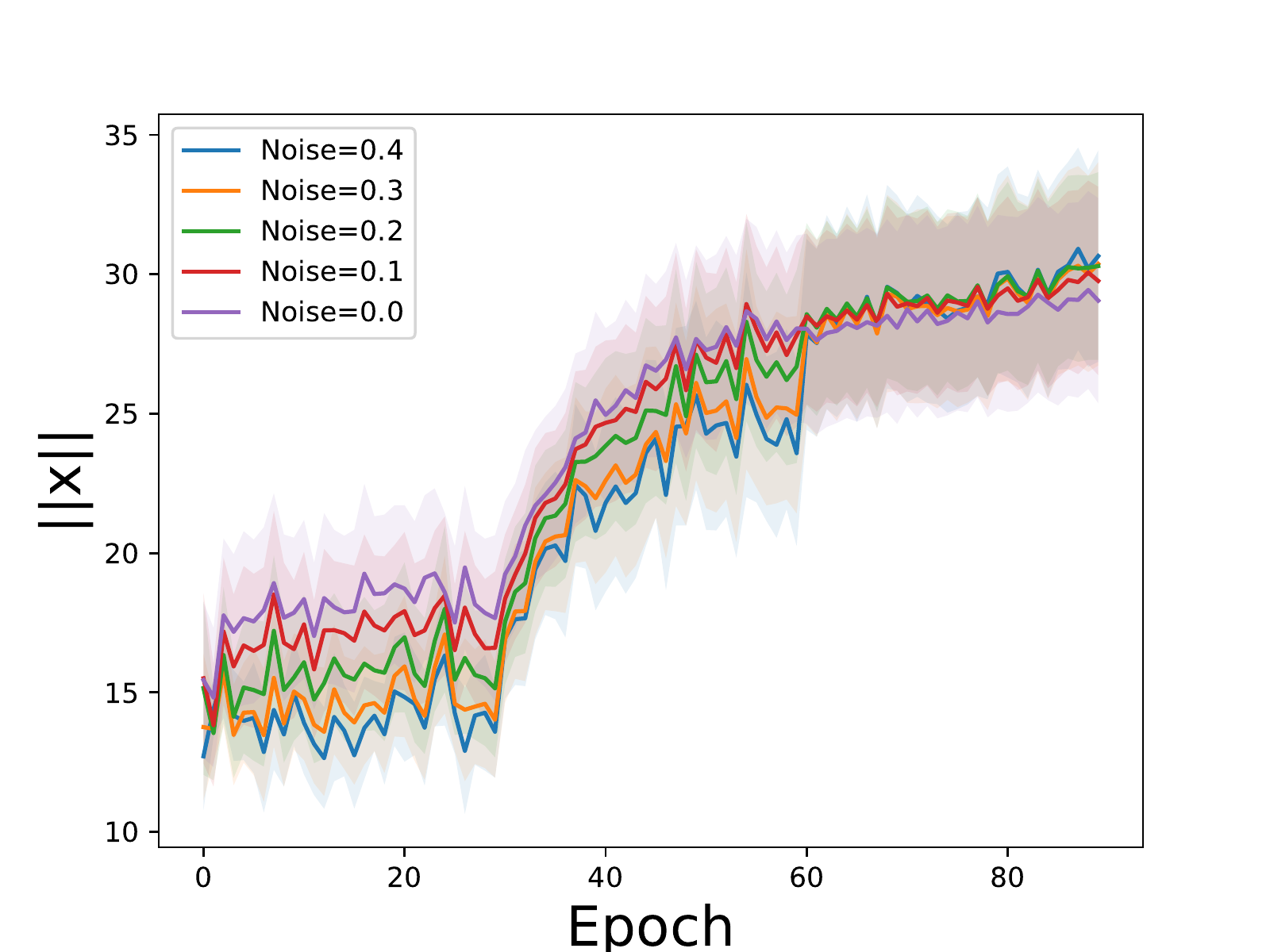} & 
			\includegraphics[width=0.3 \textwidth]{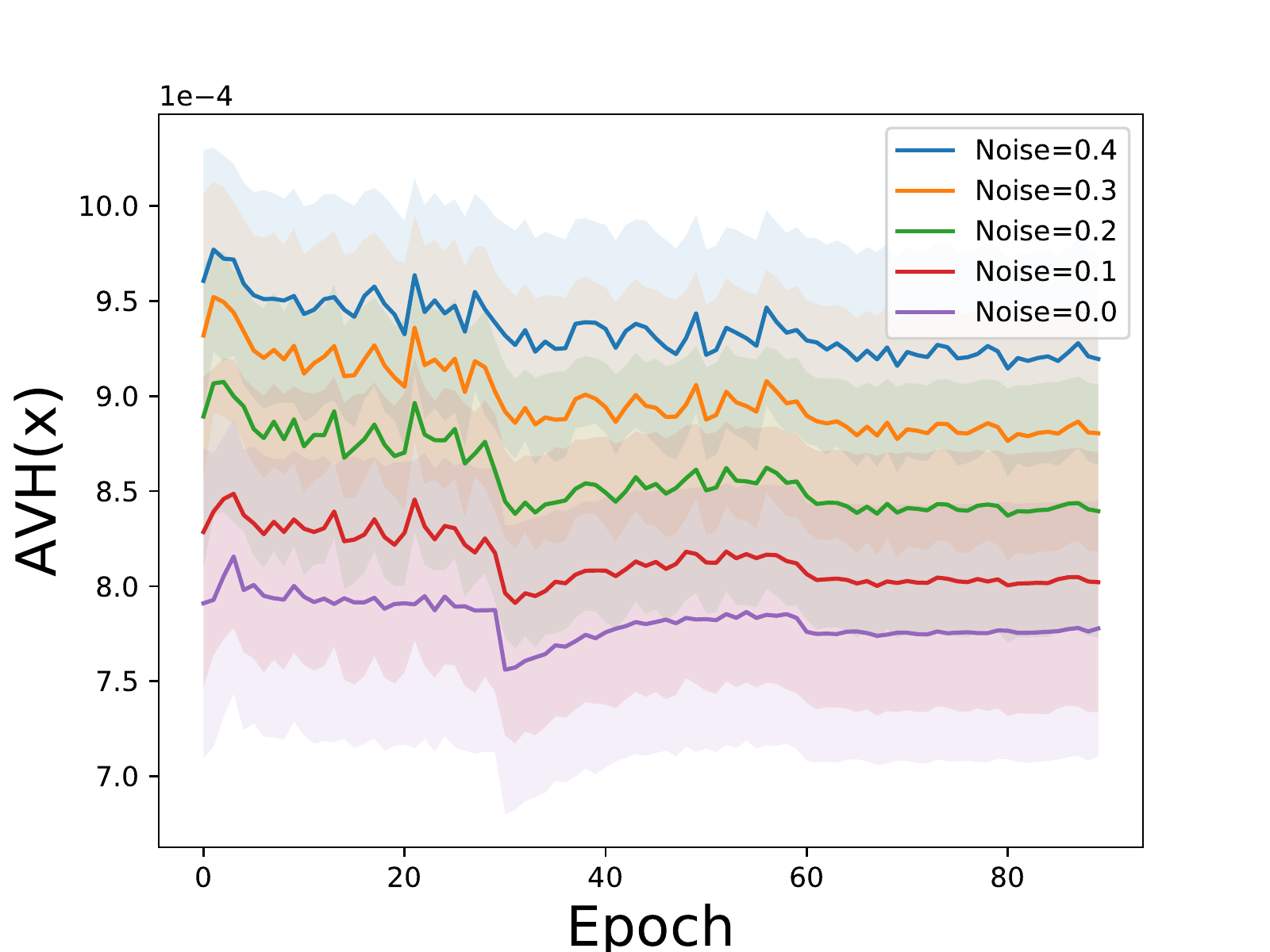} &
			\includegraphics[width=0.3 \textwidth]{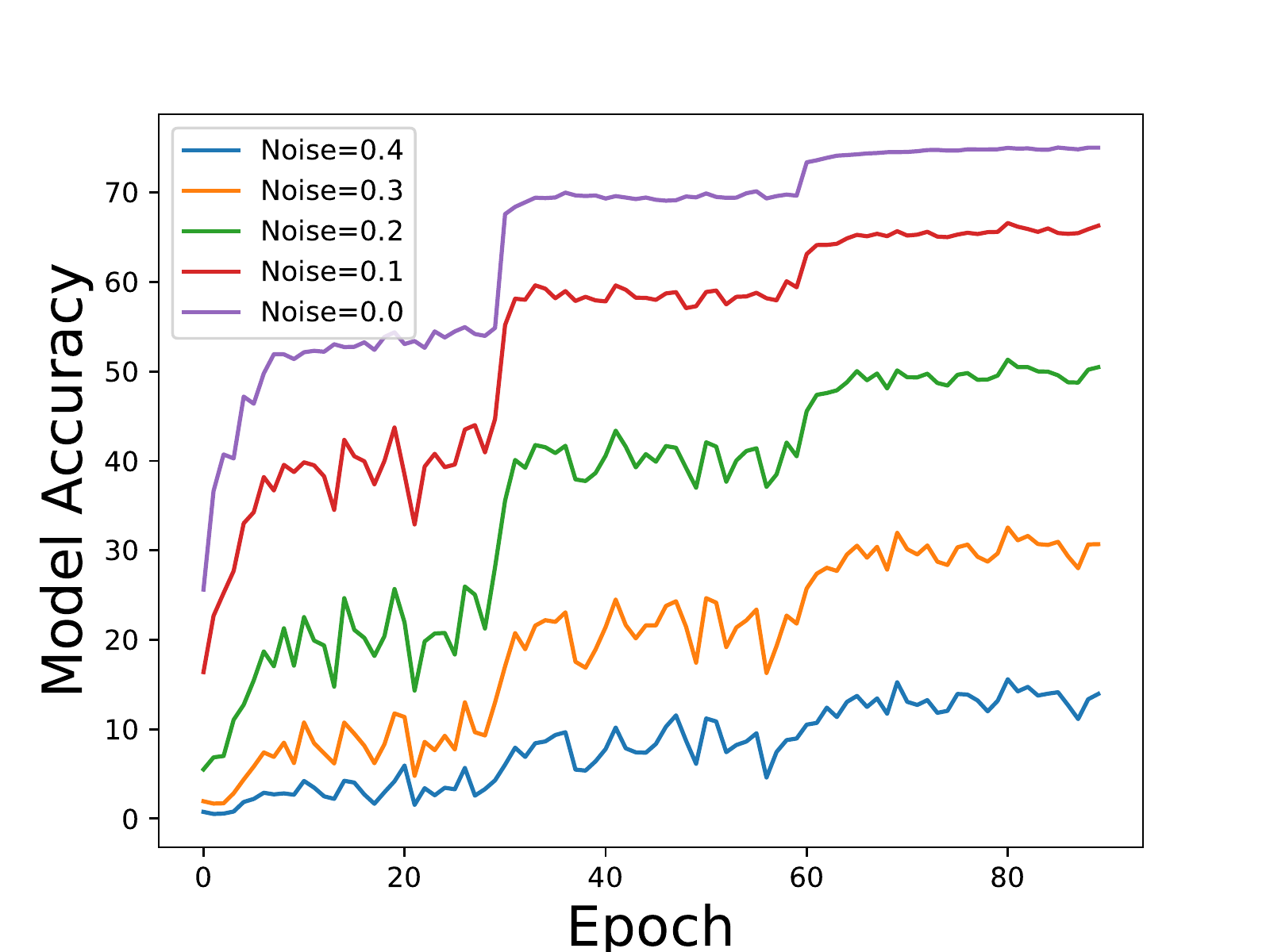}
		\end{tabular}
	\end{center}
	\caption{\footnotesize Averaged training dynamics across different noise degradation levels. Columns from left to right: number of epochs vs. average $\ell_2$ norm, number of epochs vs. average AVH score, and number of epochs vs. model accuracy. Rows from top to bottom: dynamics corresponding to AlexNet, VGG-19, ResNet-50, and DenseNet-121.}\label{fig:degradation_noise}
\end{figure*}

\clearpage
\textbf{Dynamics across noise degradation levels: } In Figure~\ref{fig:degradation_noise}, we illustrate the averaged training dynamics on the ImageNet validation set across five image noise degradation levels - [0.4, 0.3, 0.2, 0.1, 0.0].

\clearpage

\begin{figure*}[h!]
	\begin{center}
		\begin{tabular}{ccc}
			\includegraphics[width=0.3 \textwidth]{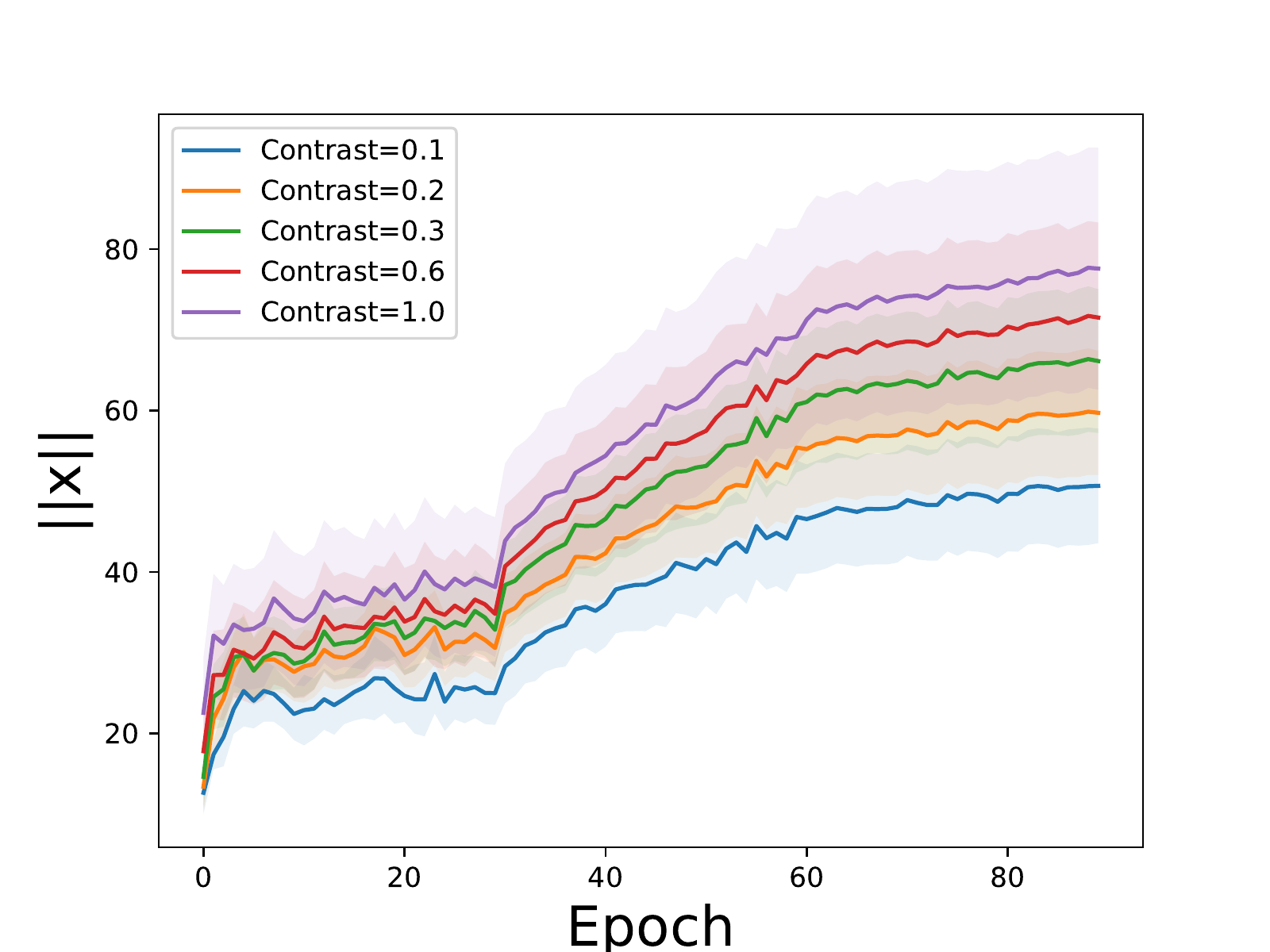} & 
			\includegraphics[width=0.3 \textwidth]{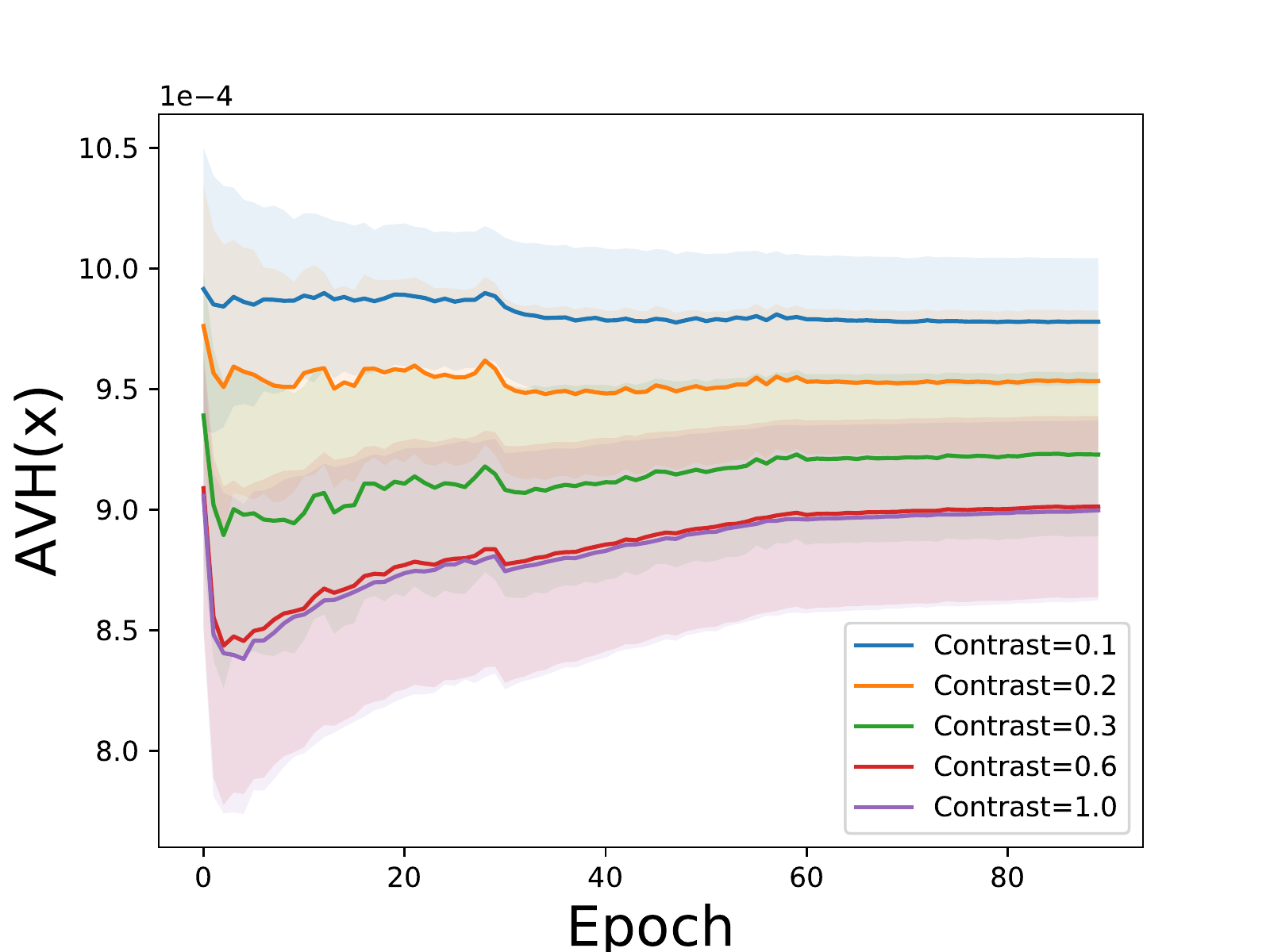} & 
			\includegraphics[width=0.3 \textwidth]{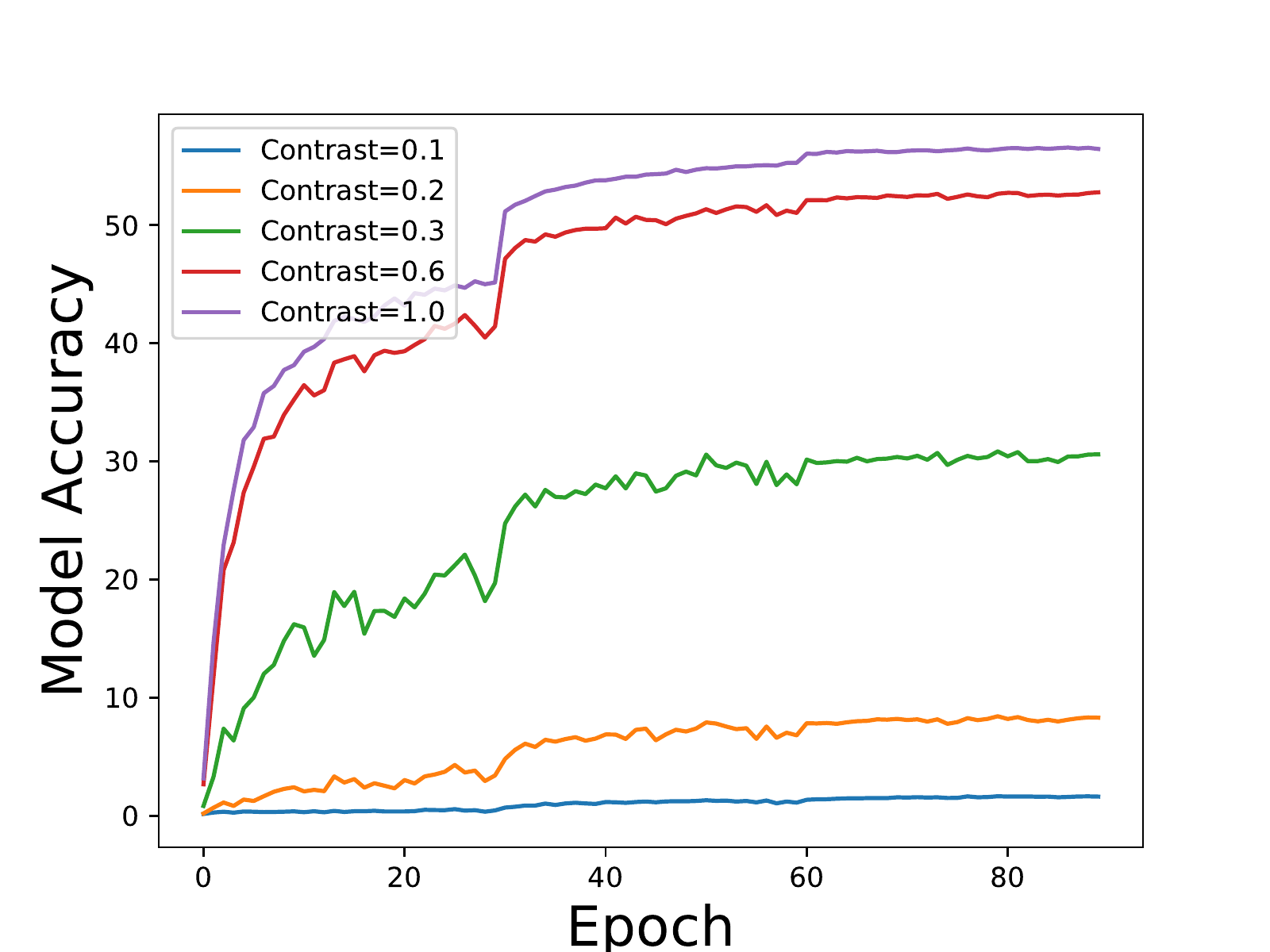} \\
			\includegraphics[width=0.3 \textwidth]{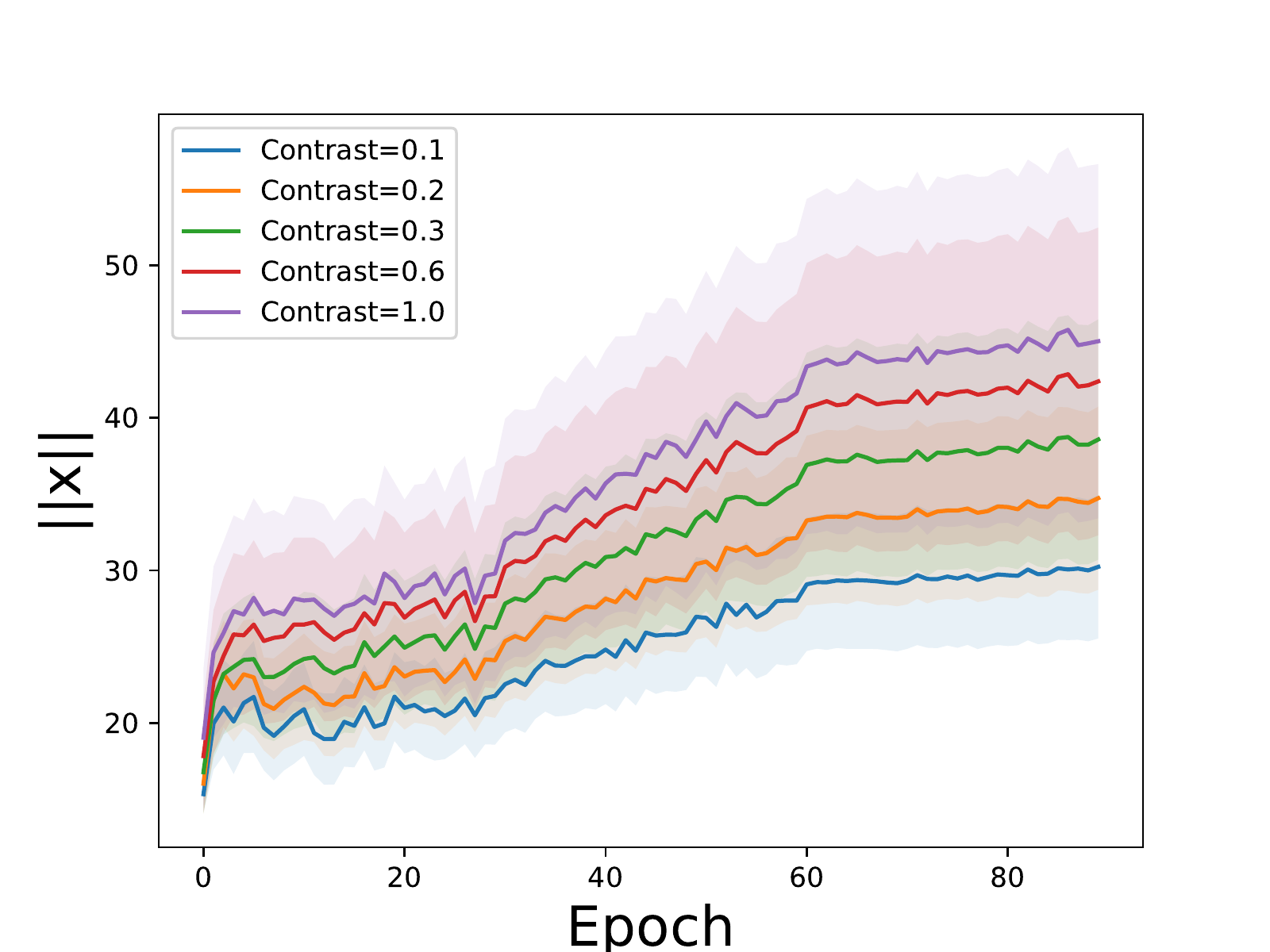} & 
			\includegraphics[width=0.3 \textwidth]{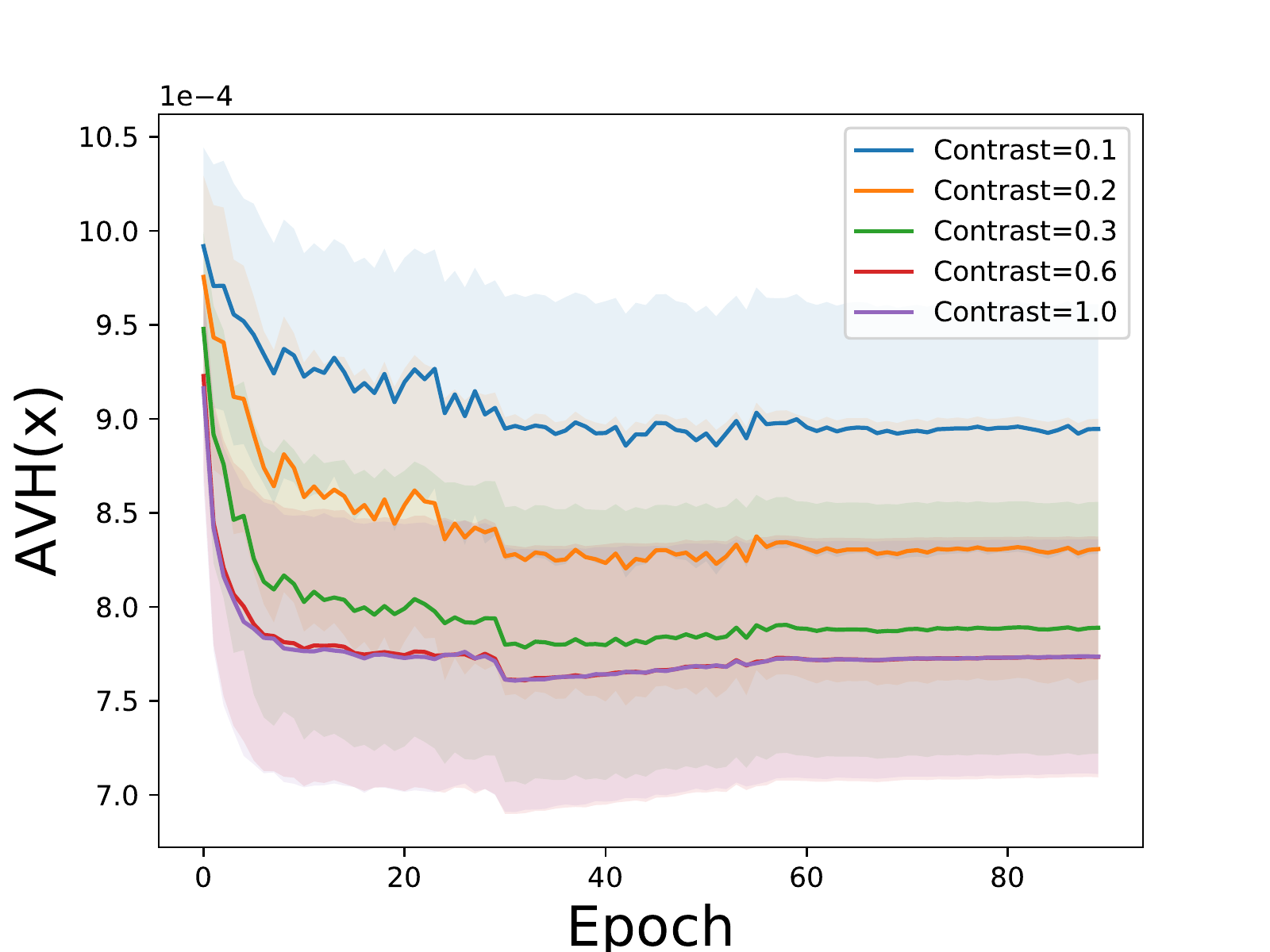} & 
			\includegraphics[width=0.3 \textwidth]{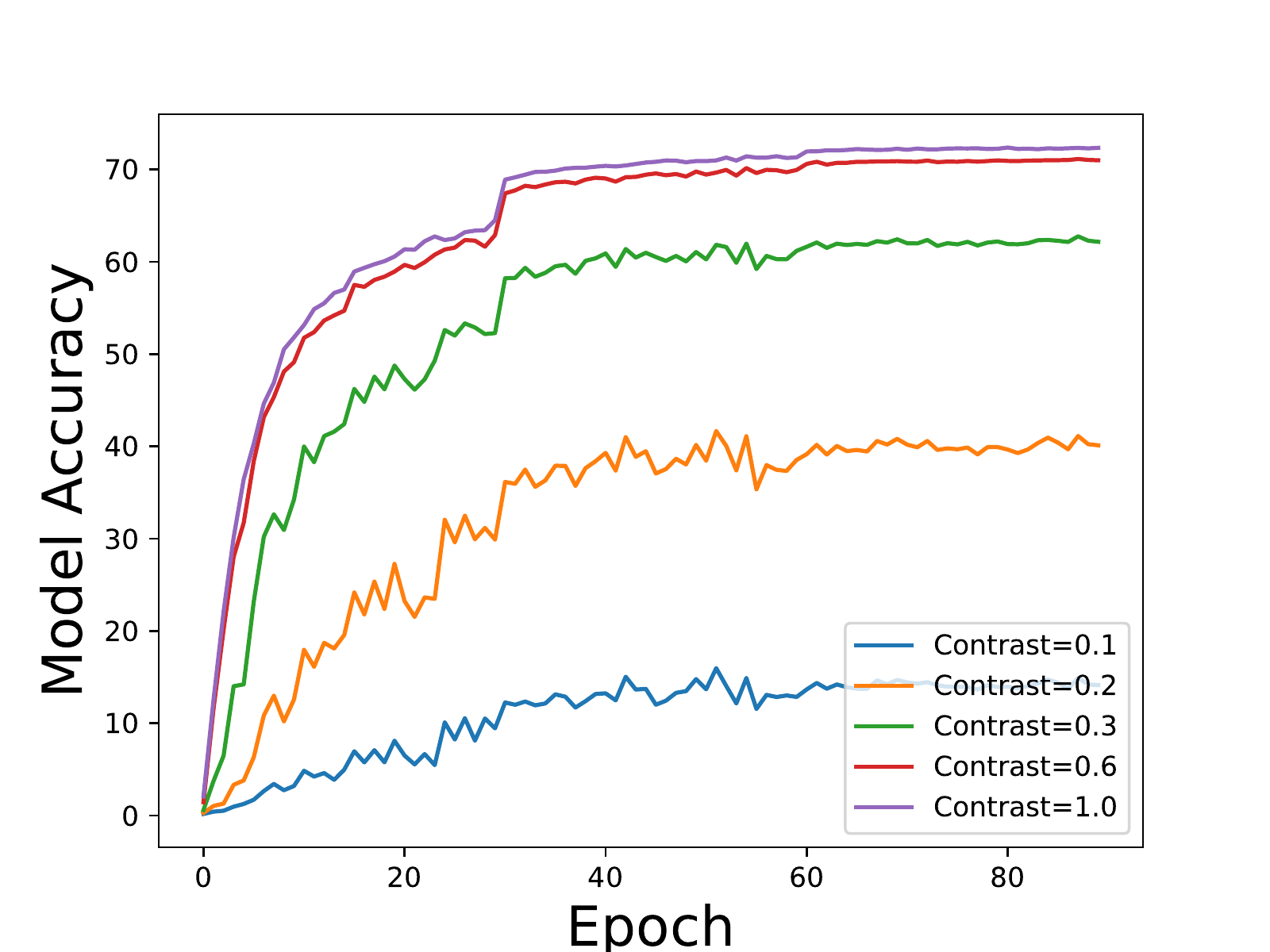} \\
			\includegraphics[width=0.3 \textwidth]{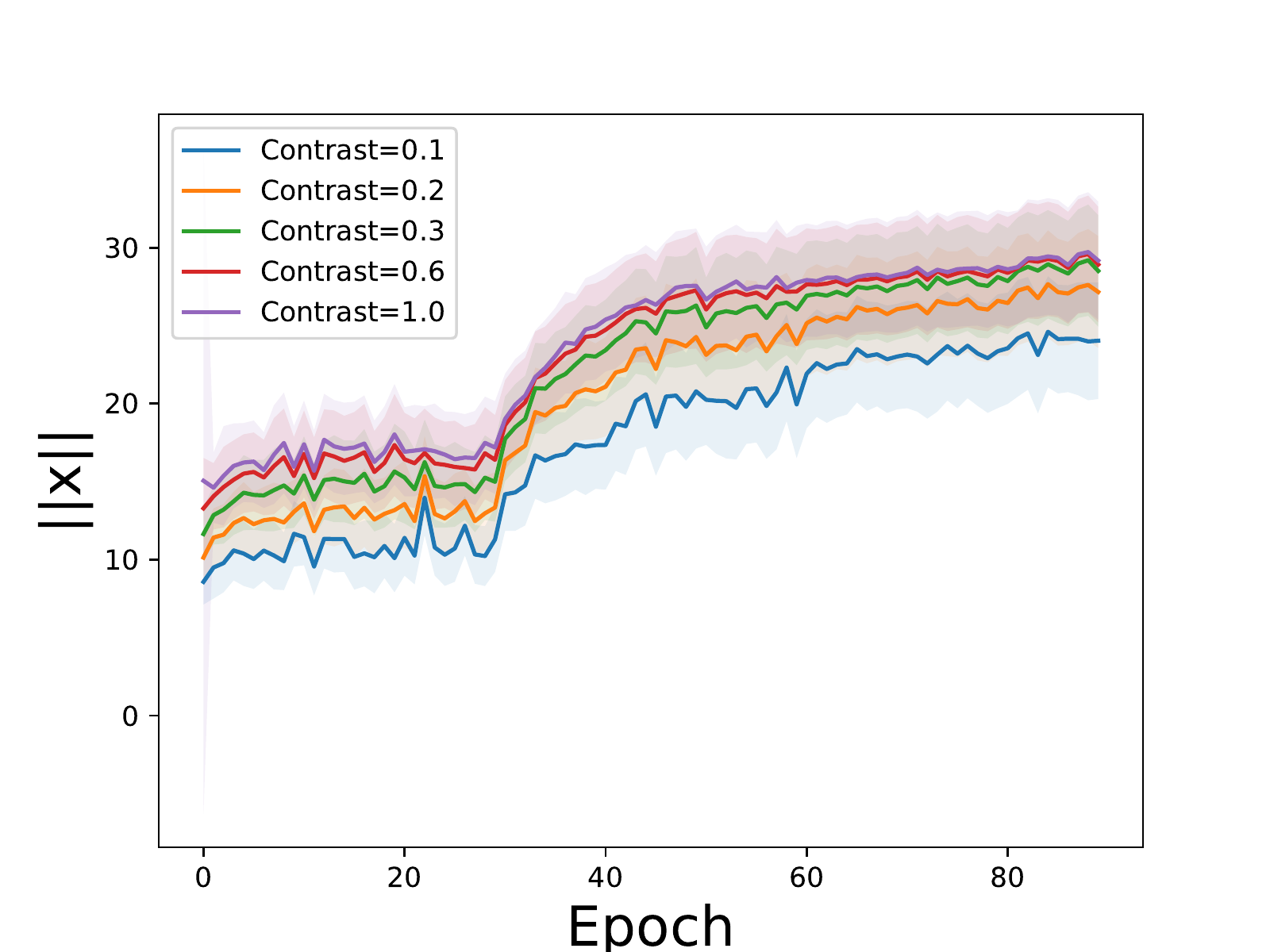} & 
			\includegraphics[width=0.3 \textwidth]{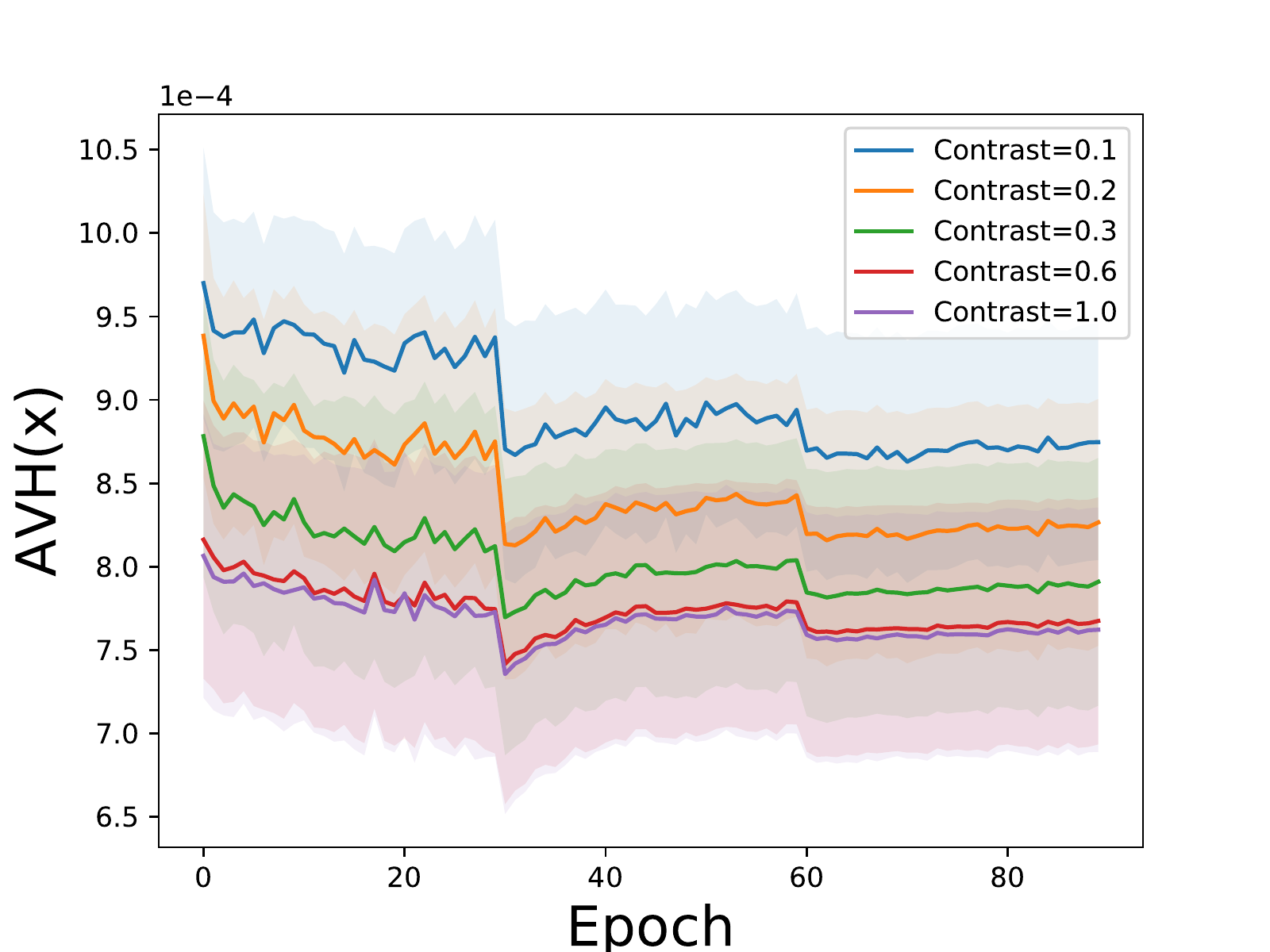} & 
			\includegraphics[width=0.3 \textwidth]{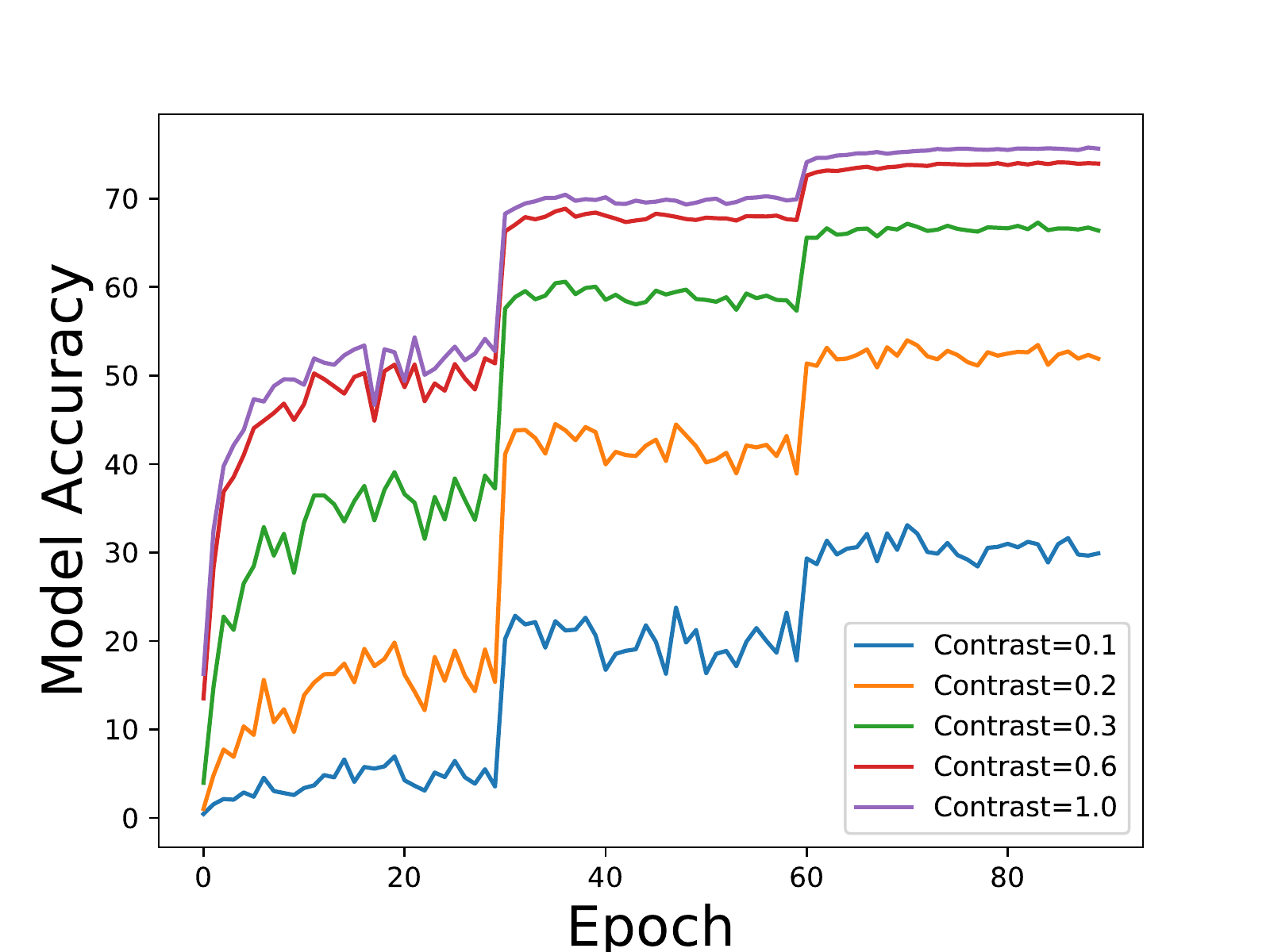} \\
			\includegraphics[width=0.3 \textwidth]{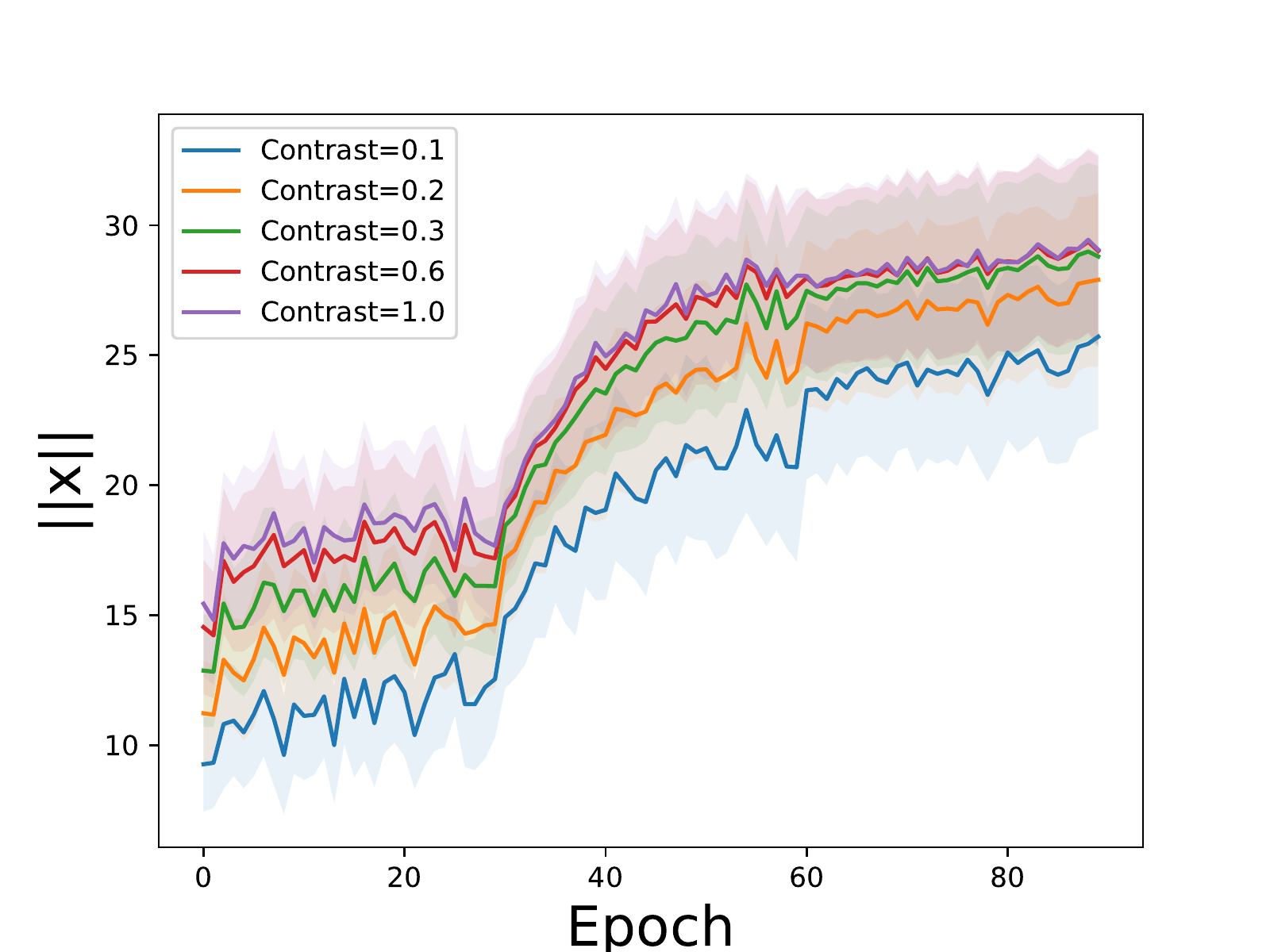} & 
			\includegraphics[width=0.3 \textwidth]{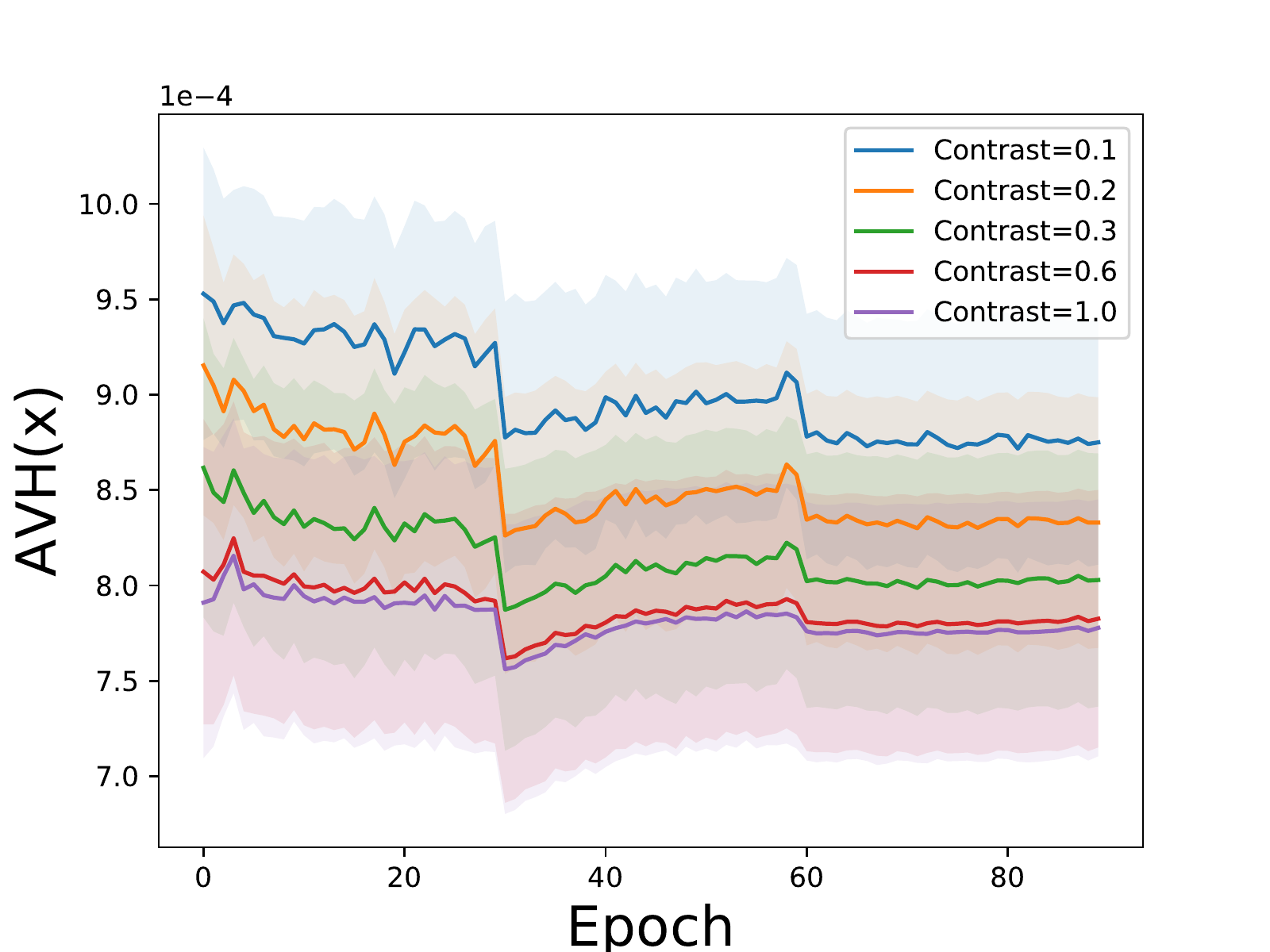} &
			\includegraphics[width=0.3 \textwidth]{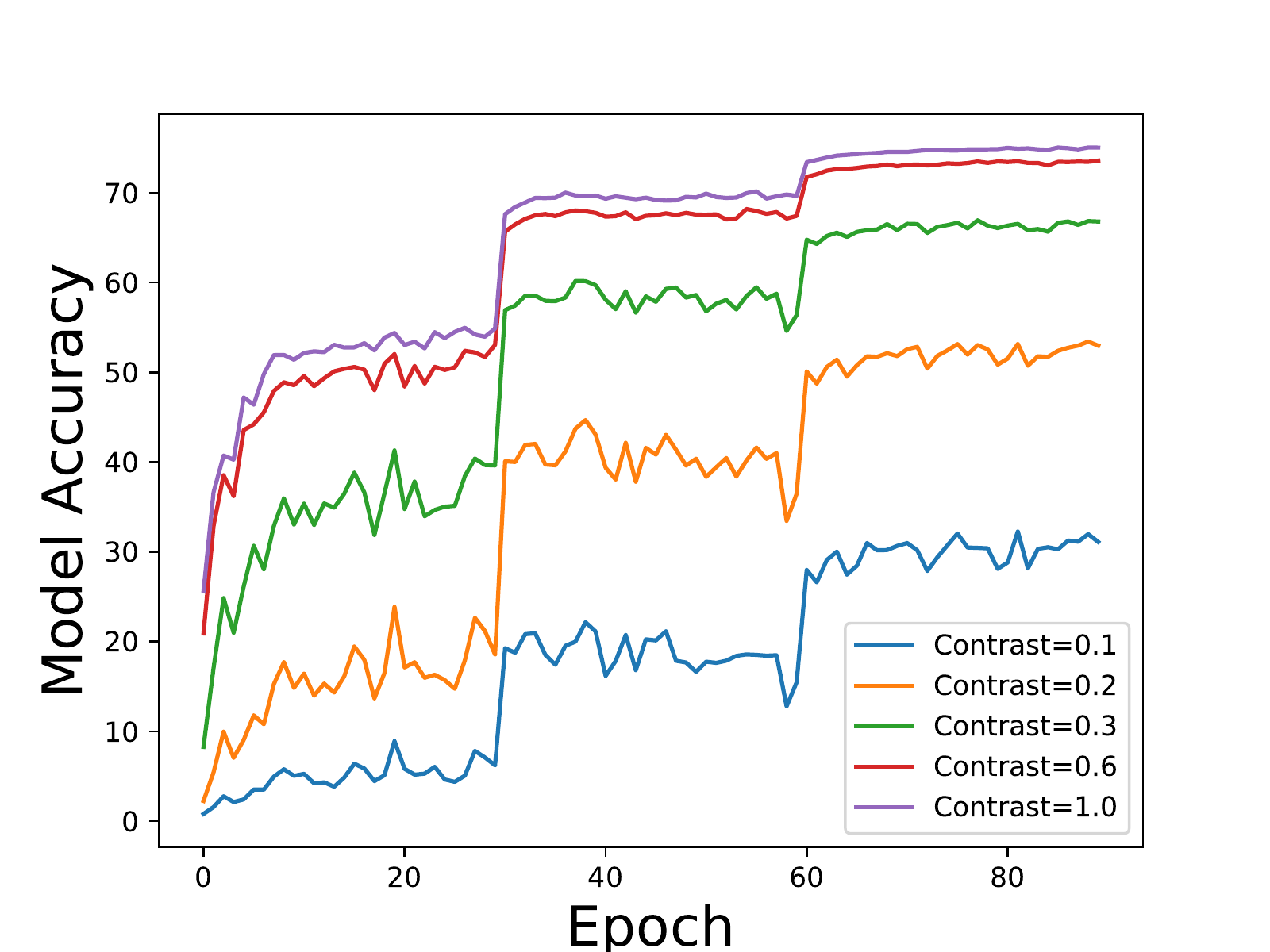}
		\end{tabular}
	\end{center}
	\caption{\footnotesize Averaged training dynamics across different contrast degradation levels. Columns from left to right: number of epochs vs. average $\ell_2$ norm, number of epochs vs. average AVH score, and number of epochs vs. model accuracy. Rows from top to bottom: dynamics corresponding to AlexNet, VGG-19, ResNet-50, and DenseNet-121.}\label{fig:degradation_contrast}
\end{figure*}

\textbf{Dynamics across contrast degradation levels: } In Figure~\ref{fig:degradation_contrast}, we illustrate the averaged training dynamics on the ImageNet validation set across five image contrast degradation levels - [0.1, 0.2, 0.3, 0.6, 1.0].

One can see that the observations from Section~\ref{sec:dynamics} in the main paper also hold on this set of experiments.

\clearpage
\subsection{Additional Results on CIFAR-10, CIFAR-100 and MNIST}
\label{sec:extracifar}
\begin{figure*}[h!]
	\begin{center}
		\begin{tabular}{ccc}
			\includegraphics[width=0.3 \textwidth]{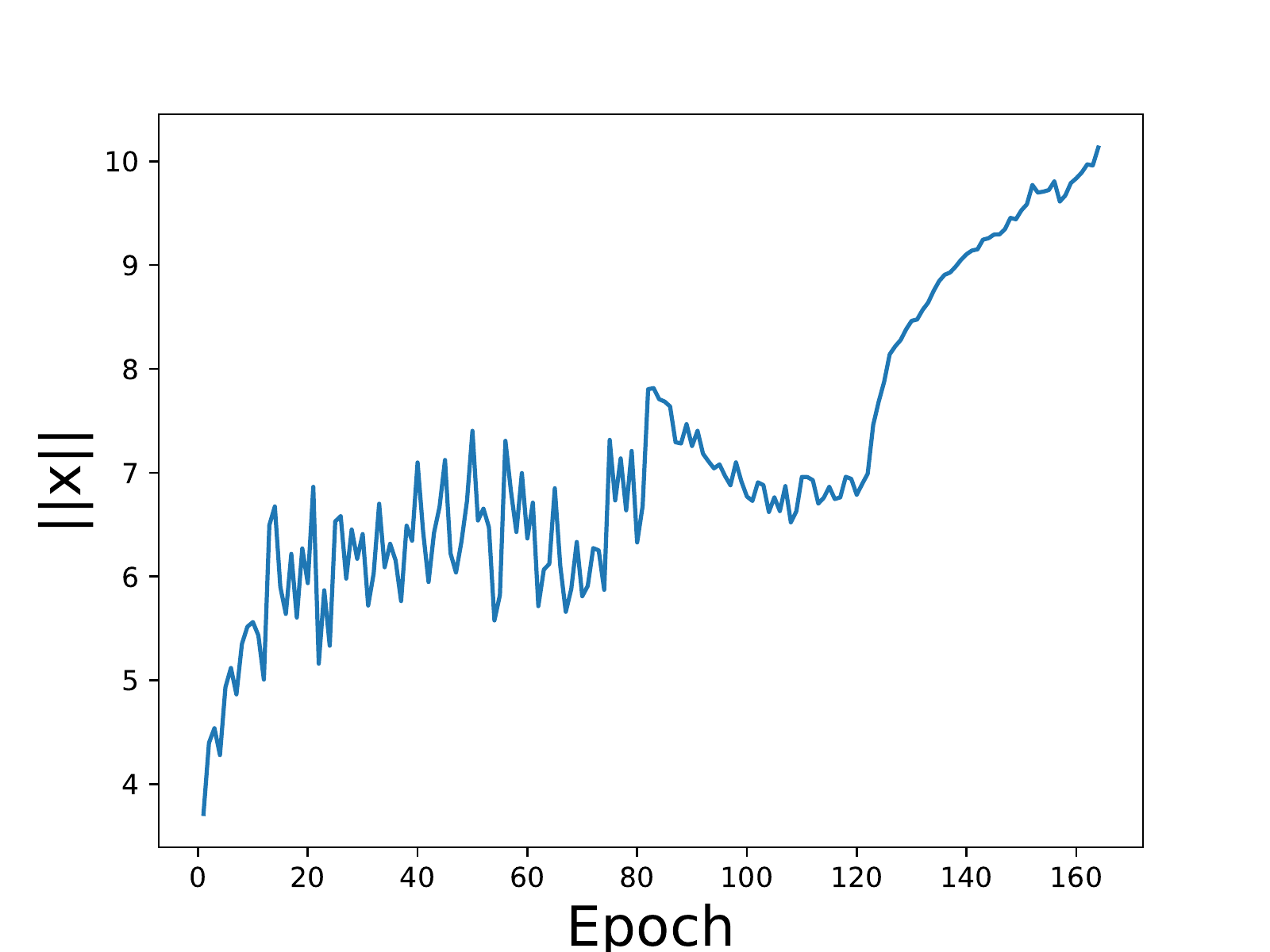} & 
			\includegraphics[width=0.3 \textwidth]{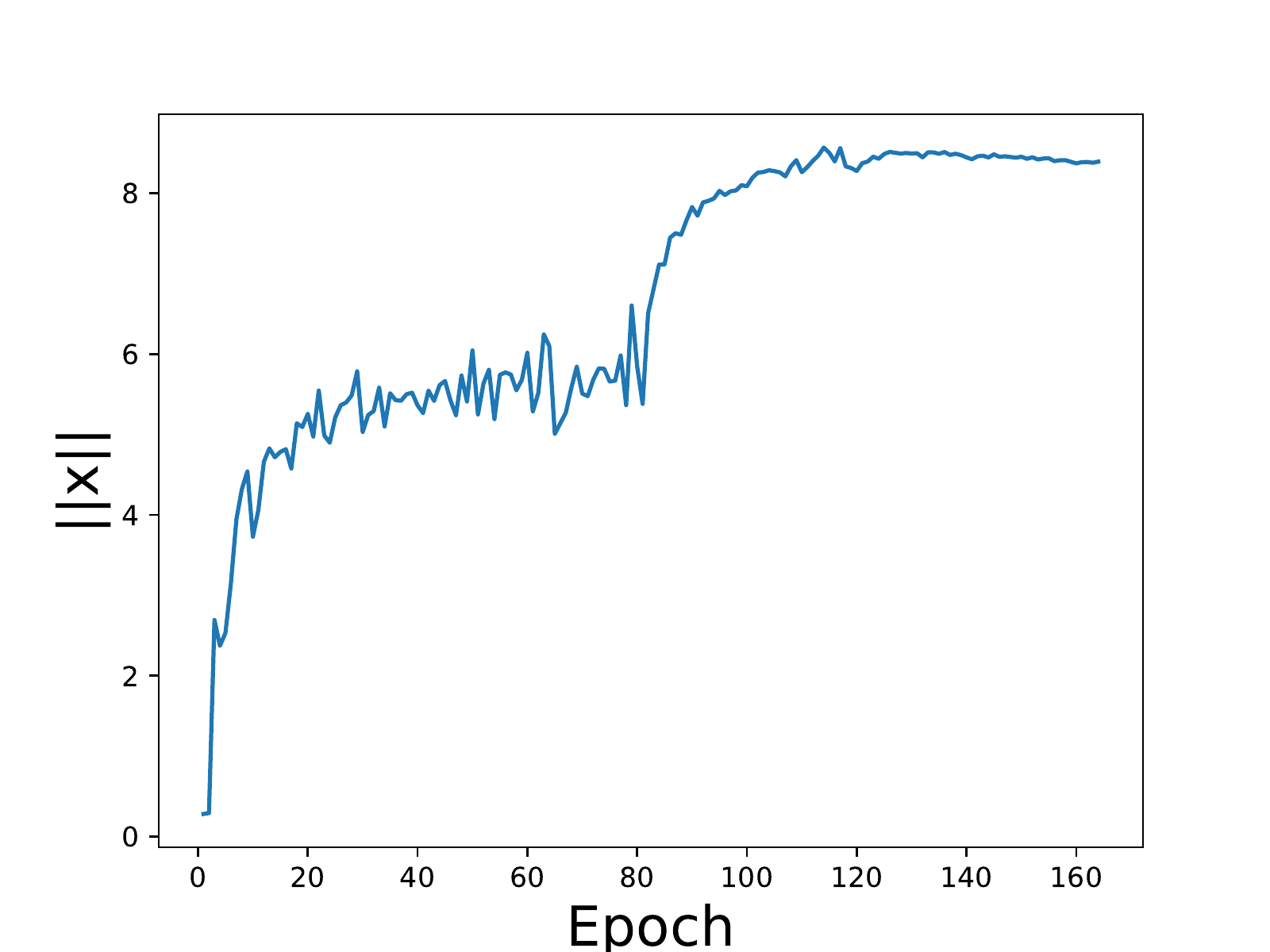} &
			\includegraphics[width=0.3 \textwidth]{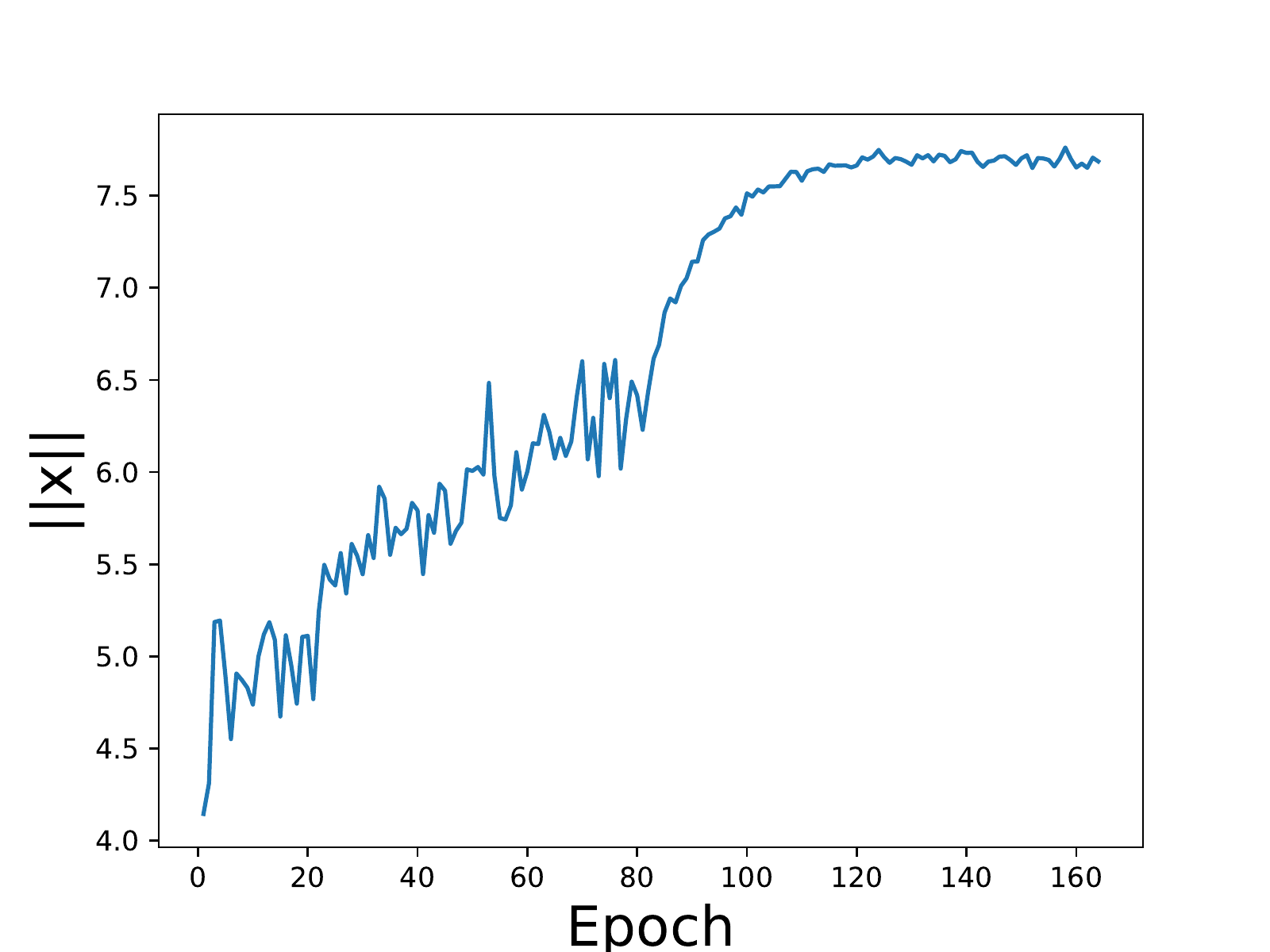} \\
			\includegraphics[width=0.3 \textwidth]{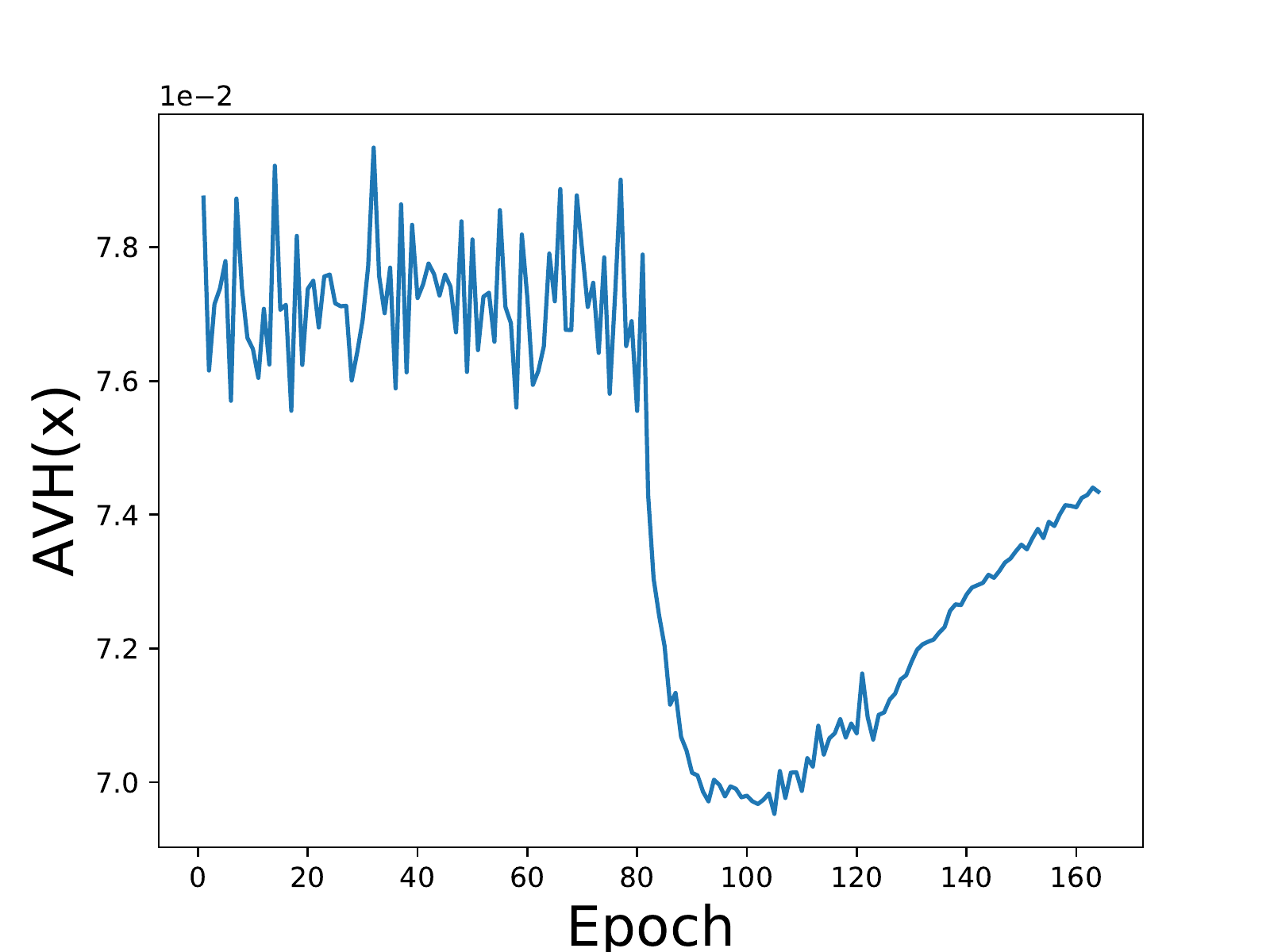} & 
			\includegraphics[width=0.3 \textwidth]{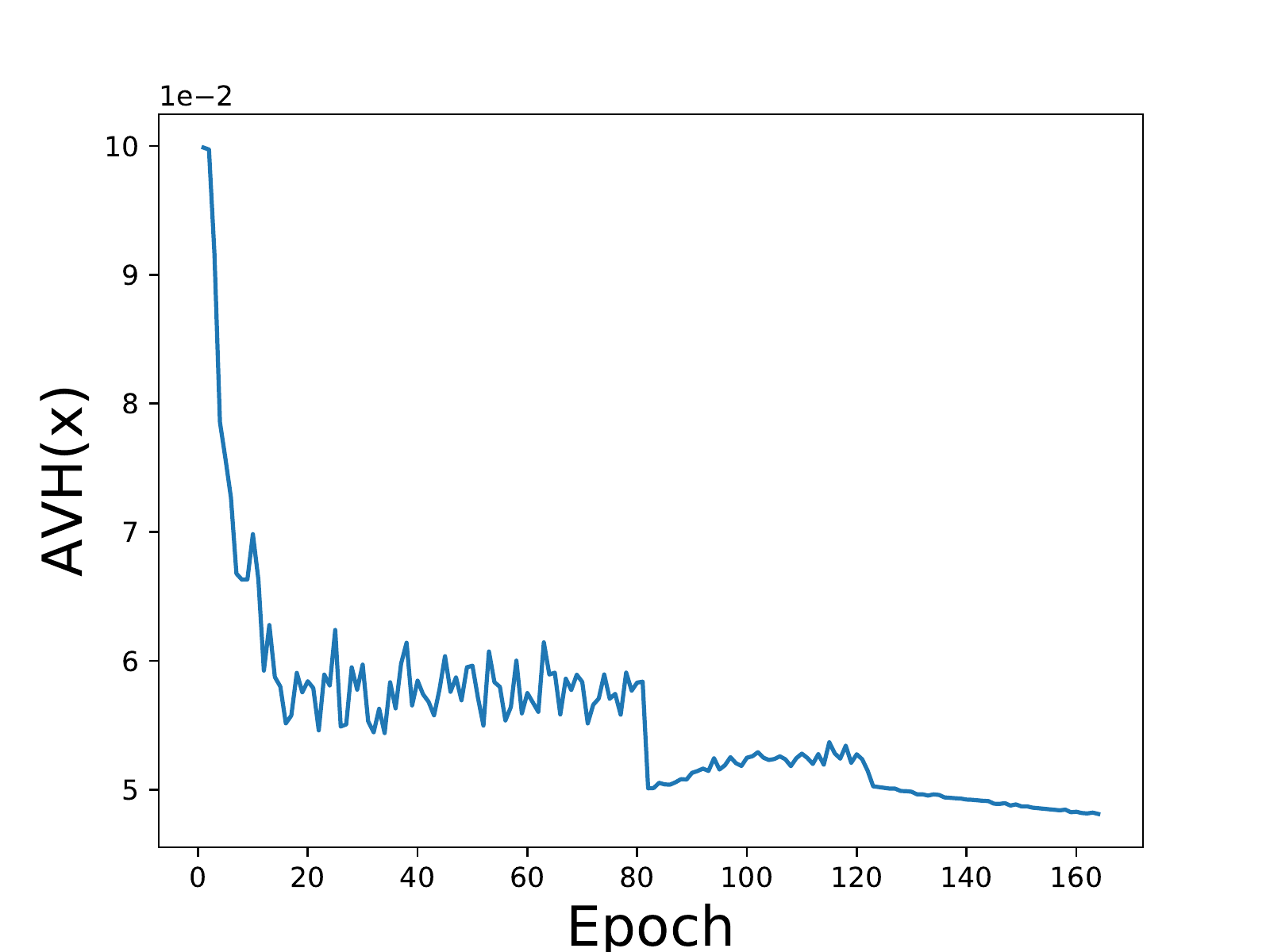} & 
			\includegraphics[width=0.3 \textwidth]{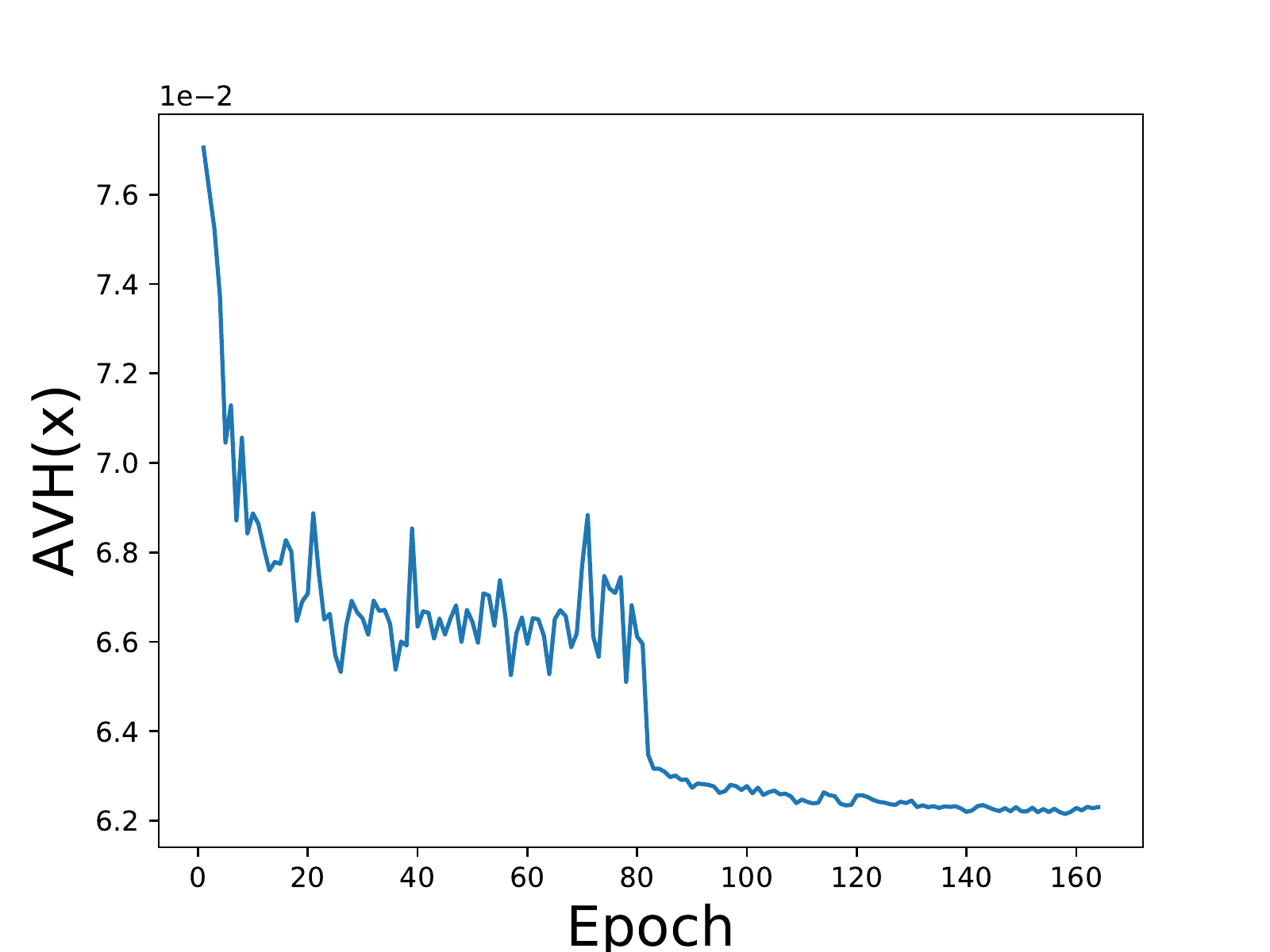}
		\end{tabular}
	\end{center}
	\caption{\footnotesize The top three plots show the number of Epochs v.s. Average $\ell_2$ norm across CIFAR-10 validation samples. The bottom three plots represent number of Epochs v.s. Average AVH(x). From left to right, we use AlexNet, VGG-19 and ResNet-50.}\label{fig:cifar10}
\end{figure*}

\begin{figure*}[h!]
	\begin{center}
		\begin{tabular}{ccc}
			\includegraphics[width=0.3 \textwidth]{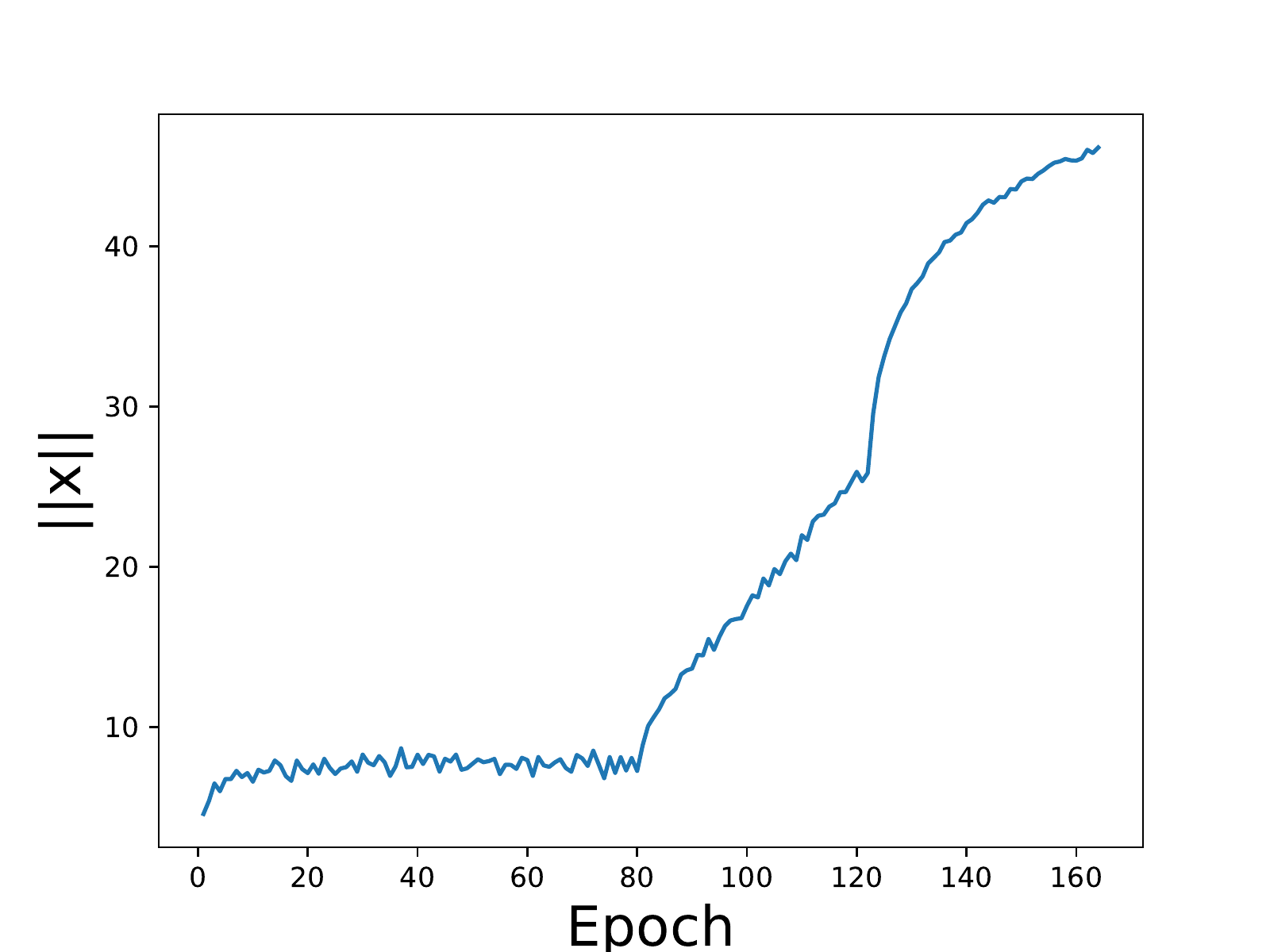} & 
			\includegraphics[width=0.3 \textwidth]{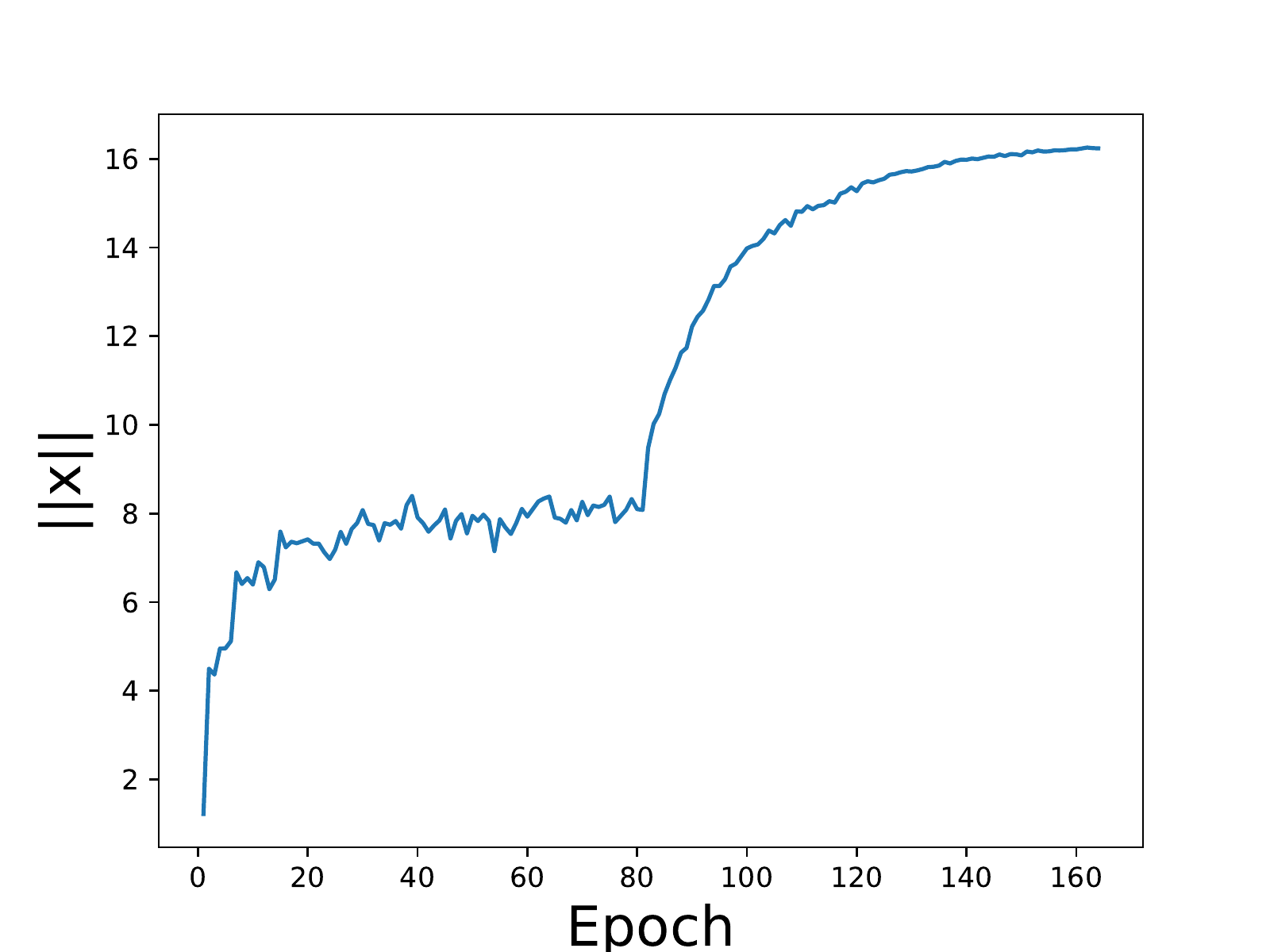} &
			\includegraphics[width=0.3 \textwidth]{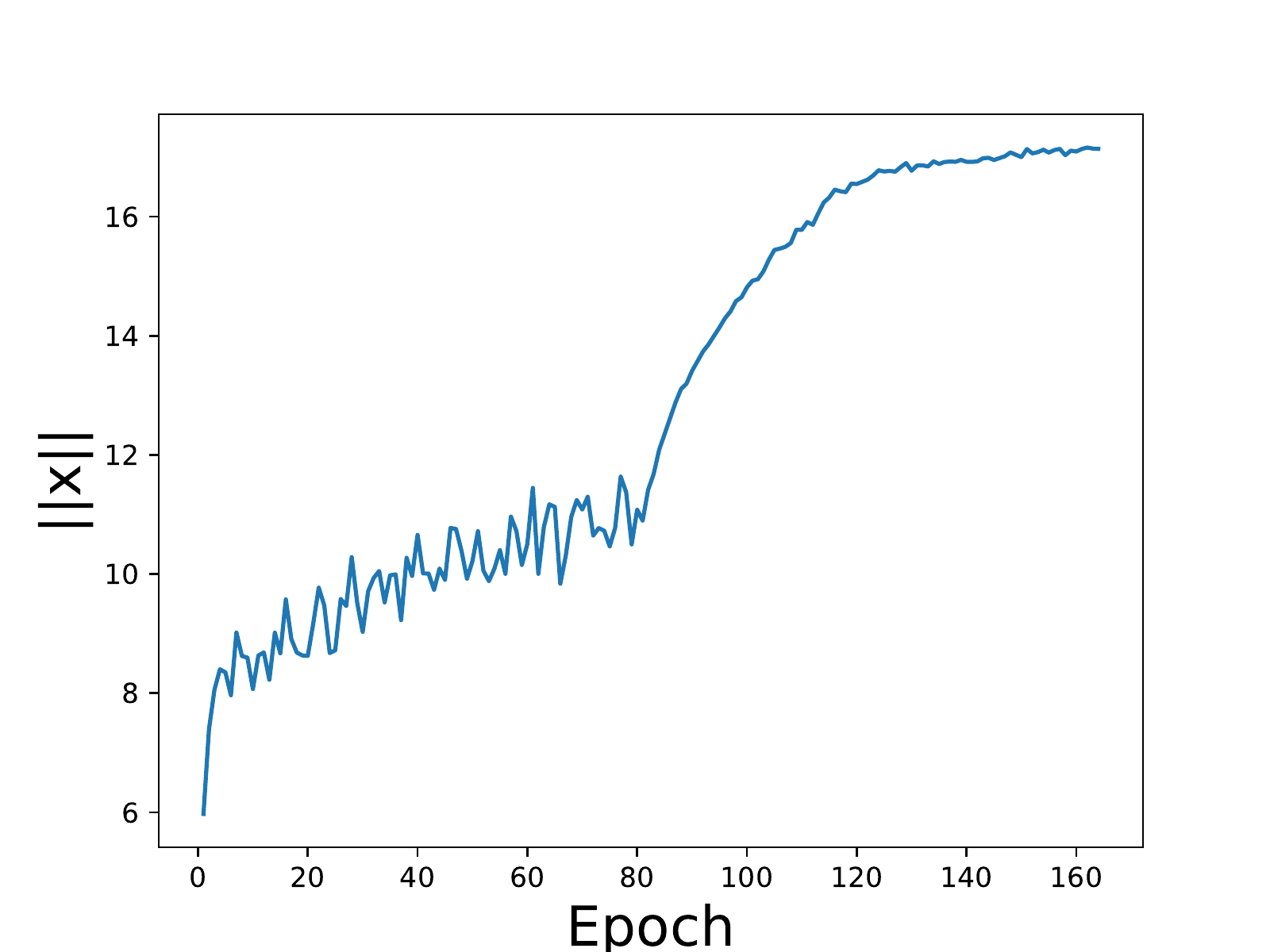} \\
			\includegraphics[width=0.3 \textwidth]{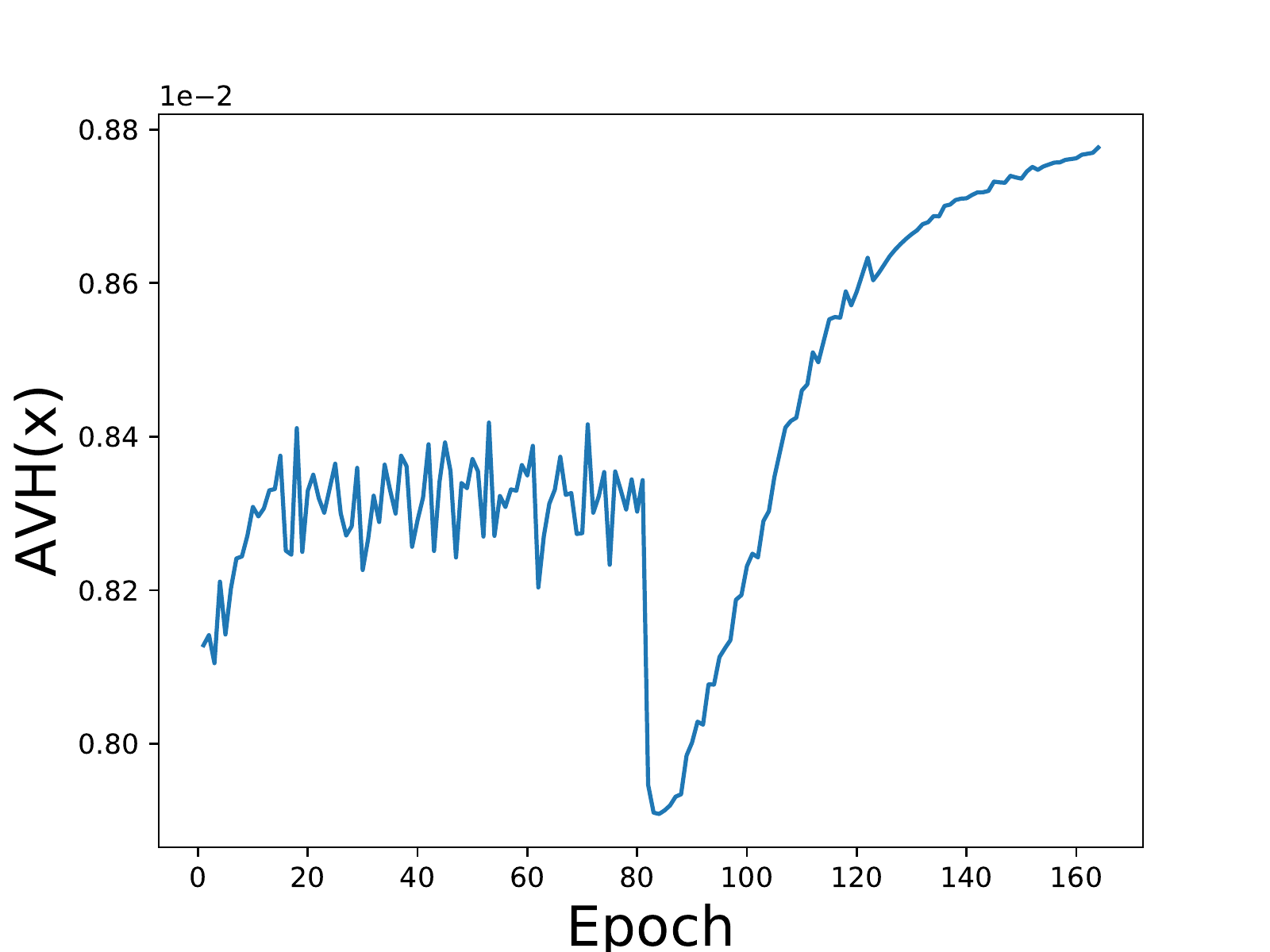} & 
			\includegraphics[width=0.3 \textwidth]{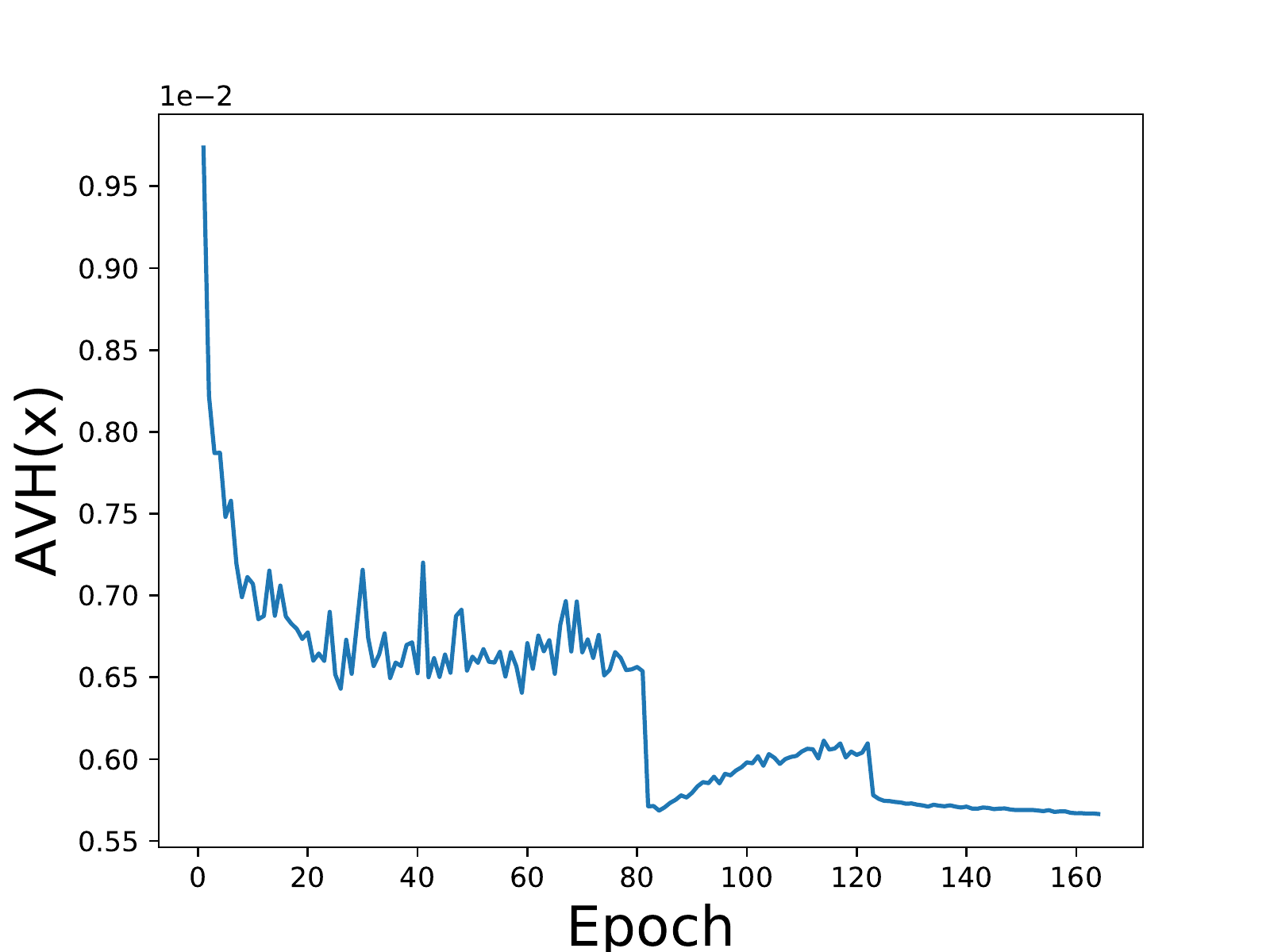} & 
			\includegraphics[width=0.3 \textwidth]{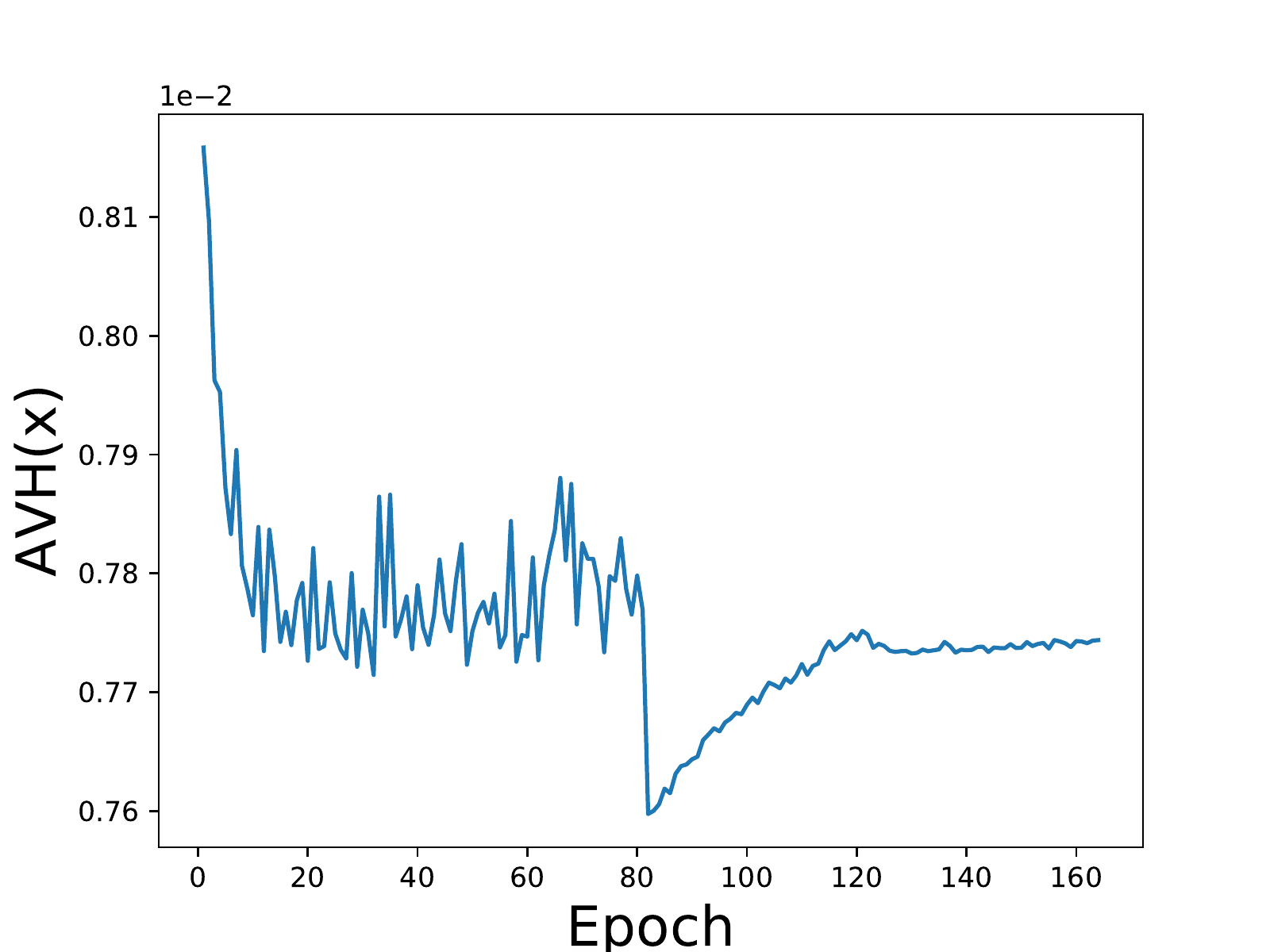}
		\end{tabular}
	\end{center}
	\caption{\footnotesize The top three plots show the number of Epochs v.s. Average $\ell_2$ norm across CIFAR-100 validation samples. The bottom three plots represent number of Epochs v.s. Average AVH(x). From left to right, we use AlexNet, VGG-19 and ResNet-50.}\label{fig:cifar100}
\end{figure*}

Figure~\ref{fig:cifar10} and~\ref{fig:cifar100} show the dynamics of average $\ell_2$ norm of the embeddings and average AVH(x) on CIFAR-10 and CIFAR-100 datasets respectively. We can observe the similar phenomenons we have discussed in section 3. It further supports our theoretical foundation from~\cite{soudry2018implicit} that gradient descent converges to the same direction as maximum margin solutions irrelevant to the $\ell_2$ norm of classifier weights or feature embeddings. 

\begin{figure*}[h]
	\begin{center}
		\begin{tabular}{cc}
			\includegraphics[width=0.4 \textwidth]{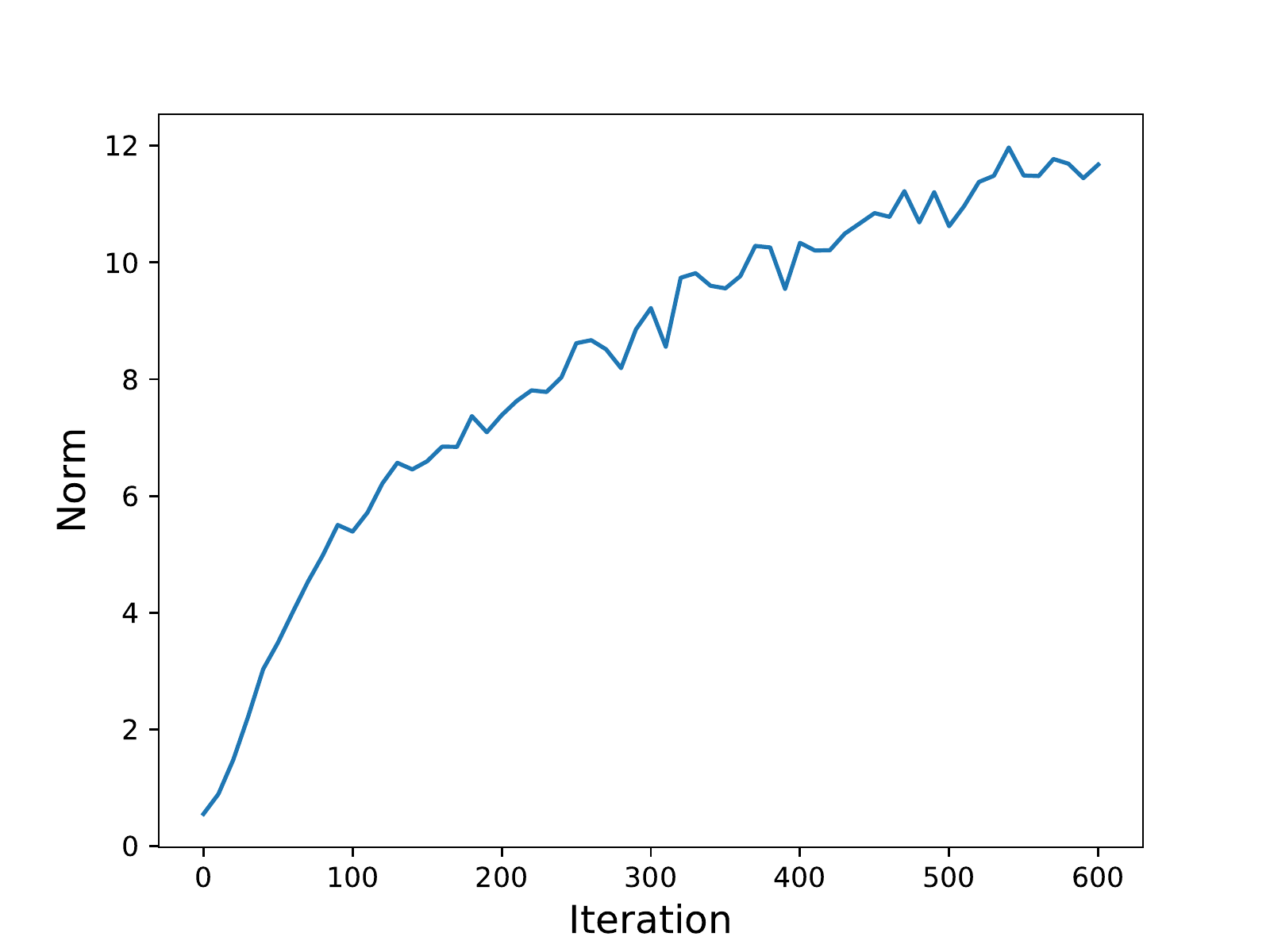} & 
			\includegraphics[width=0.4 \textwidth]{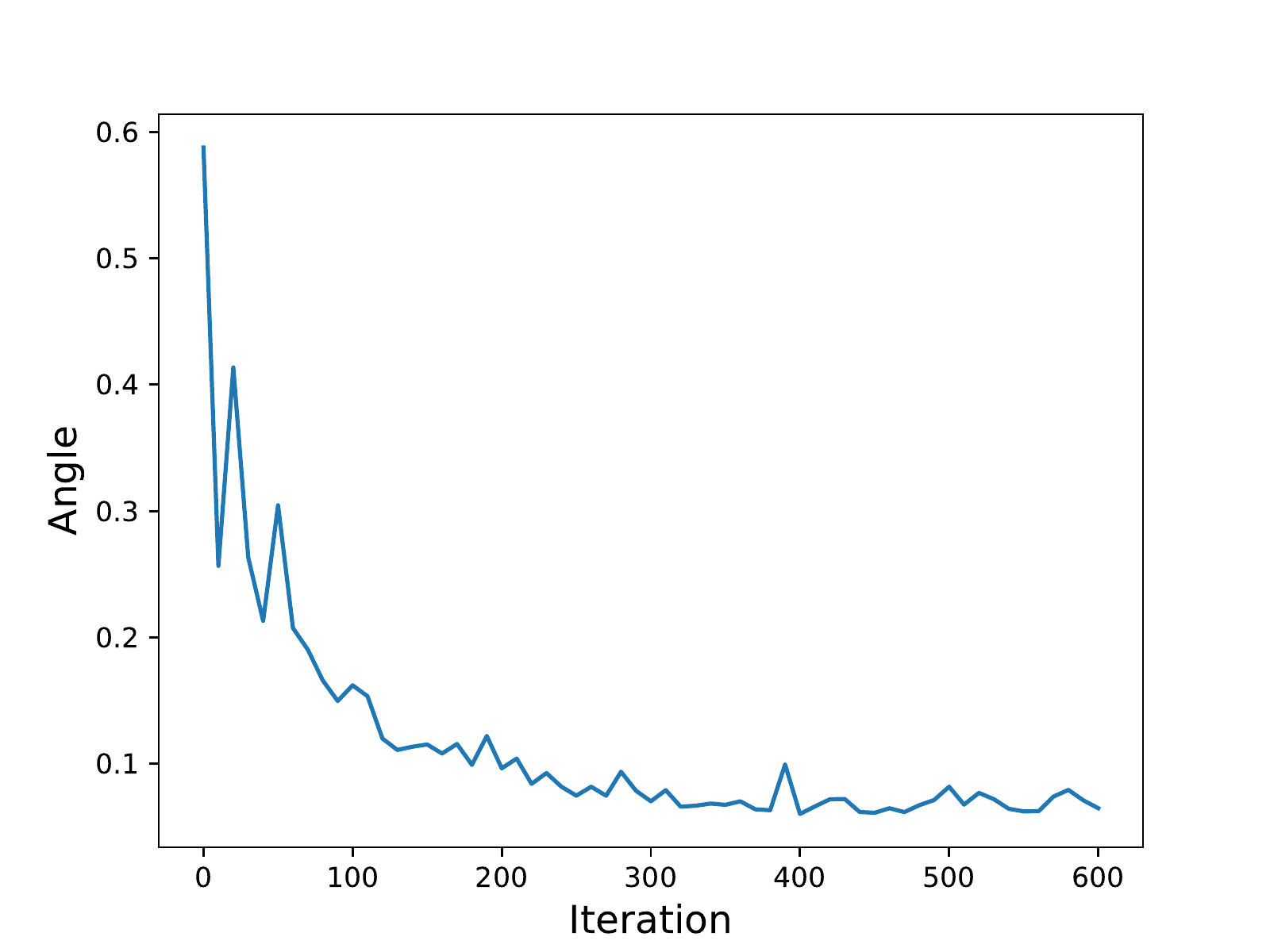}
		\end{tabular}
	\end{center}
	\caption{\footnotesize Average $\ell_2$ norm and angle of the embedding across all testing samples v.s. iteration number.}\label{fig:change}
\end{figure*}

\clearpage
Figure~\ref{fig:change} illustrates how the average norm of the feature embedding and AVH between feature and class embedding for testing samples vary in $60$ iterations during the training process on MNIST. The average norm increases with a large initial slope but it flattens slightly after $10$ iterations. On the other hand, the average angle decreases sharply at the beginning and then becomes almost flat after $10$ iterations.

\section{Additional Discussions for Observations in Training Dynamics}\label{app:c}
Observation 2 in section 3 describes that AVH hits a plateau very early even when the accuracy or loss is still improving. AVH is more important than $\Vert \mathbf{x} \Vert_2$ in the sense that it is the key factor deciding which class the input sample is classified to. 

However, optimizing the norm under the current softmax cross-entropy loss would be easier for easy examples. Let us consider a simple binary classification case where the softmax score for class 1 is \begin{equation}\label{eq:sm}
\frac{\exp(\bm{w}_1\bm{x})}{\sum_i\exp(\bm{w}_i\bm{x})}=\frac{\exp(\|\bm{w}_1\|\|\bm{x}\|\cos(\theta_{\bm{w}_1,\bm{x}}))}{\sum_i\exp(\|\bm{w}_i\|\|\bm{x}\|\cos(\theta_{\bm{w}_i,\bm{x}}))}    
\end{equation}
where $\bm{w}_i$ is the classifier weights of class $i$, $\bm{x}$ is the input deep feature and $\theta_{\bm{w}_i,\bm{x}}$ is the angle between $\bm{w}_i$ and $\bm{x}$. To simplify, we assume the norm of $\bm{w}_1$ and $\bm{w}_2$ are the same, and then the classification result is based on the angle now. For easy examples, during early stage of the training, $\theta_{\bm{w}_1,\bm{x}}$ quickly becomes smaller than $\theta_{\bm{w}_2,\bm{x}}$ and the network will classify the sample $\bm{x}$ as class 1. However, in order to further minimize the cross-entropy loss after making $\theta_{\bm{w}_1,\bm{x}}$ smaller than $\theta_{\bm{w}_2,\bm{x}}$, the network has a trivial solution: increasing the feature norm $\|\bm{x}\|$ instead of further minimizing the $\theta_{\bm{w}_1,\bm{x}}$. It is obviously a much more difficult task to minimize $\theta_{\bm{w}_1,\bm{x}}$ rather than increasing $\|\bm{x}\|$. Therefore, the network will tend to increase the feature norm $\|\bm{x}\|$ to minimize the cross-entropy loss, which is equivalent to maximizing the Model Confidence in class 1. In fact, this also matches our empirical observation that the feature norm keeps increasing during training. Moreover, this also matches our empirical result that AVH easily gets saturated while Model Confidence can keep improving. For hard examples, after some time of training, the feature norms are unavoidable also increasing (although slower than those of easy examples). We can see from equation~\ref{eq:sm} that when $\|\bm{x}\|$ is very large and $\cos(\theta_{\bm{w}_i,\bm{x}})$ is very small, improving the angle becomes much harder because for a bit improvement on these examples, the model needs to sacrifice a lot for those easy ones. For the case of value of $\cos(\theta_{\bm{w}_i,\bm{x}})$ is around the decision boundary, a little change to AVH can cause a lot improvement on loss and accuracy and thereby we can still observe the change of accuracy and loss while AVH plateaus. More details about why $\|\bm{x}\|$ might be harmful in the training process is in Appendix~\ref{sec:extradis}.

\clearpage

\section{Additional Experiments for Connections to Human Visual Hardness}
\label{app:d}
\subsection{Additional Results for Correlation Testings}
\label{sec:correlationextra}
In order to run rigorous correlation testings, besides computing the Spearman coefficient, we provide additional results on Pearson and Kendall Tau correlation coefficients. Moreover, we show results for all four architectures, AlexNet, VGG-19, ResNet-50 and DenseNet-121 in Table~\ref{table:alexnet}, ~\ref{table:vgg}, ~\ref{table:resnet} and~\ref{table:densenet} respectively to support our claims in section 4.

\begin{table*}[h!] 
\scriptsize
\caption{This table presents the Spearman's rank correlation coefficients between Human Selection Frequency and AVH, Model Confidence on AlexNet. Note that we show the absolute value of the coefficient which represents the strength of the correlation. Z value is computed by Z scores of both coefficients. p-value$<$ 0.05 indicates that the result is statistically significant.}
\vspace{0.1cm}
\centering
\begin{tabular}{ c||cccccc }
\specialrule{.15em}{.05em}{.05em}
Type &Coef with AVH & Coef with Model Confidence & $Z_{avh}$& $Z_{mc}$ & Z value &p-value\\
\specialrule{.15em}{.05em}{.05em}
Spearman's rank&0.339 &0.325 & 0.352& 0.337 &1.92 &  0.027\\
Pearson& 0.324&0.31& 0.336 & 0.320& 1.90 & 0.028\\
Kendall's Tau& 0.244&0.23 & 0.249 & 0.234 &1.81 & 0.035\\
\specialrule{.15em}{.05em}{.05em}
\end{tabular}
\label{table:alexnet}
\end{table*}

\begin{table*}[h!] 
\scriptsize
\caption{This table presents the Spearman's rank correlation coefficients between Human Selection Frequency and AVH, Model Confidence on VGG-19. Note that we show the absolute value of the coefficient which represents the strength of the correlation. Z value is computed by Z scores of both coefficients. p-value$<$ 0.05 indicates that the result is statistically significant. }
\vspace{0.1cm}
\centering
\begin{tabular}{ c||cccccc }
\specialrule{.15em}{.05em}{.05em}
&Coef with AVH & Coef with Model Confidence & $Z_{avh}$& $Z_{mc}$ & Z value &p-value\\
\specialrule{.15em}{.05em}{.05em}
Spearman's rank& 0.349&0.335 & 0.364 & 0.348 &1.94 &0.026 \\
Pearson& 0.358& 0.343& 0.374& 0.357&2.09 &0.018 \\
Kendall's Tau& 0.244& 0.229& 0.249 &, 0.233 & 1.94 & 0.026 \\
\specialrule{.15em}{.05em}{.05em}
\end{tabular}
\label{table:vgg}
\end{table*}

\begin{table*}[h!] 
\scriptsize
\caption{This table presents the Spearman's rank correlation coefficient between Human Selection Frequency and AVH, Model Confidence on ResNet-50. Note that we show the absolute value of the coefficient which represents the strength of the correlation.  Z value is computed by Z scores of both coefficients. p-value$<$ 0.05 indicates that the result is statistically significant.}
\vspace{0.1cm}
\centering
\begin{tabular}{ c||cccccc }
\specialrule{.15em}{.05em}{.05em}
&Coef with AVH & Coef with Model Confidence & $Z_{avh}$& $Z_{mc}$ & Z value & p-value\\
\specialrule{.15em}{.05em}{.05em}
Spearman's rank& 0.360&0.325 & 0.377 & 0.337 & 4.85 & $<$ .00001 \\
Pearson& 0.385& 0.341& 0.406& 0.355& 6.2 & $<$ .00001\\
Kendall's Tau& 0.257& 0.231& 0.263& 0.235& 3.38 &.0003 \\
\specialrule{.15em}{.05em}{.05em}
\end{tabular}
\label{table:resnet}
\end{table*}

\begin{table*}[h!] 
\scriptsize
\caption{This table presents the Spearman's rank correlation coefficients between Human Selection Frequency and AVH, Model Confidence in DenseNet-121. Note that we show the absolute value of the coefficient which represents the strength of the correlation. Z value is computed by Z scores of both coefficients. p-value $<$ 0.05 indicates that the result is statistically significant.}
\vspace{0.1cm}
\centering
\begin{tabular}{ c||cccccc }
\specialrule{.15em}{.05em}{.05em}
&Coef with AVH & Coef with Model Confidence & $Z_{avh}$& $Z_{mc}$ & Z value &p-value\\
\specialrule{.15em}{.05em}{.05em}
Spearman's& 0.367&0.329 & 0.4059& 0.355 &6.2 &$<$ .00001\\
Pearson&0.390& 0.347& 0.412 & 0.362& 6.09 &$<$ .00001\\
Kendall's Tau& 0.262& 0.234& 0.268&0.238& 3.65 &.0001 \\
\specialrule{.15em}{.05em}{.05em}
\end{tabular}
\label{table:densenet}
\end{table*}

\clearpage
\subsection{Additional Plots for Hypothesis Testings}
\label{sec:extradegra}

\begin{figure*}[h!]
	\begin{center}
		\begin{tabular}{ccc}
			\includegraphics[width=0.3 \textwidth]{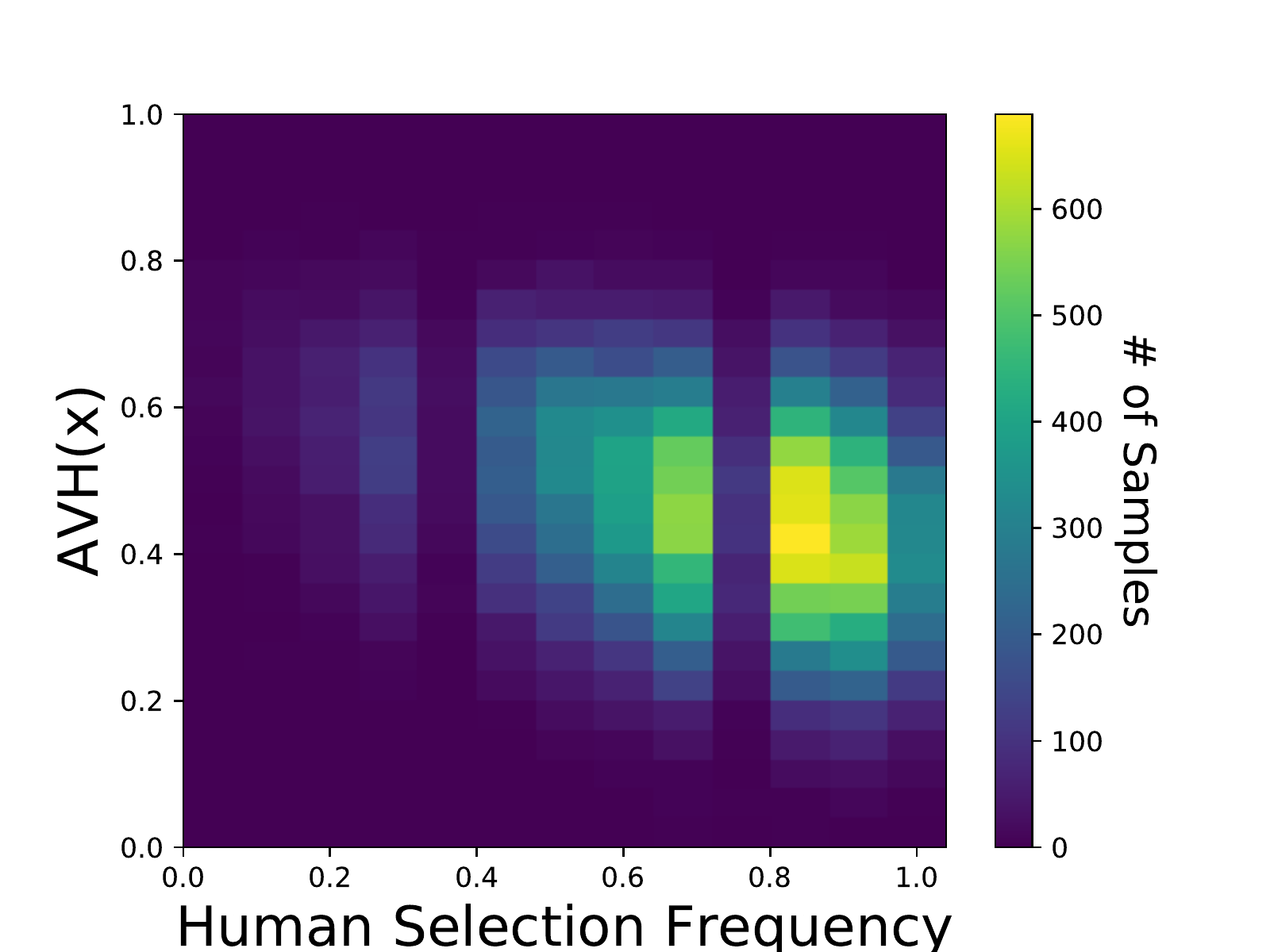} & 
			\includegraphics[width=0.3 \textwidth]{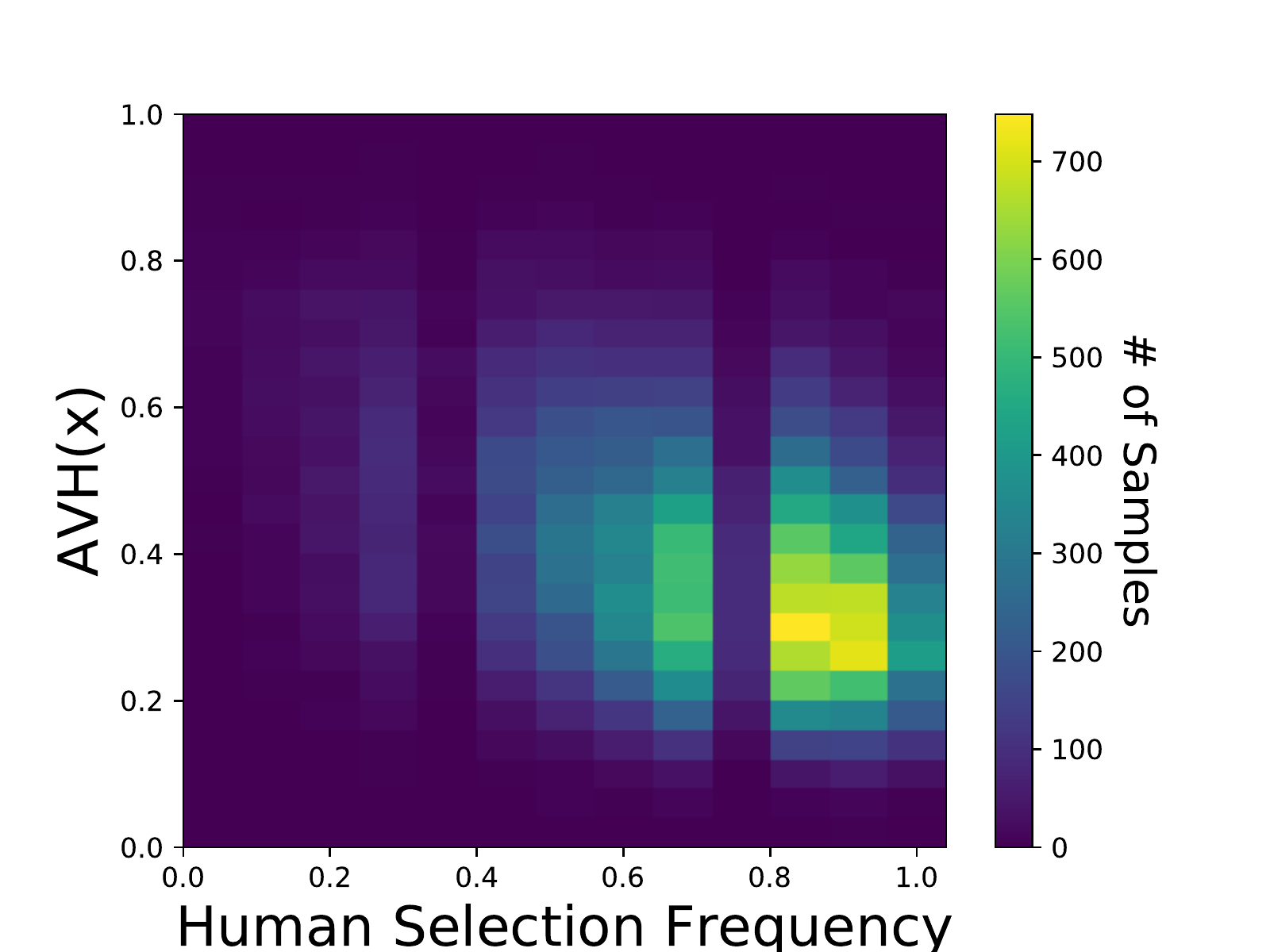} & 
			\includegraphics[width=0.3 \textwidth]{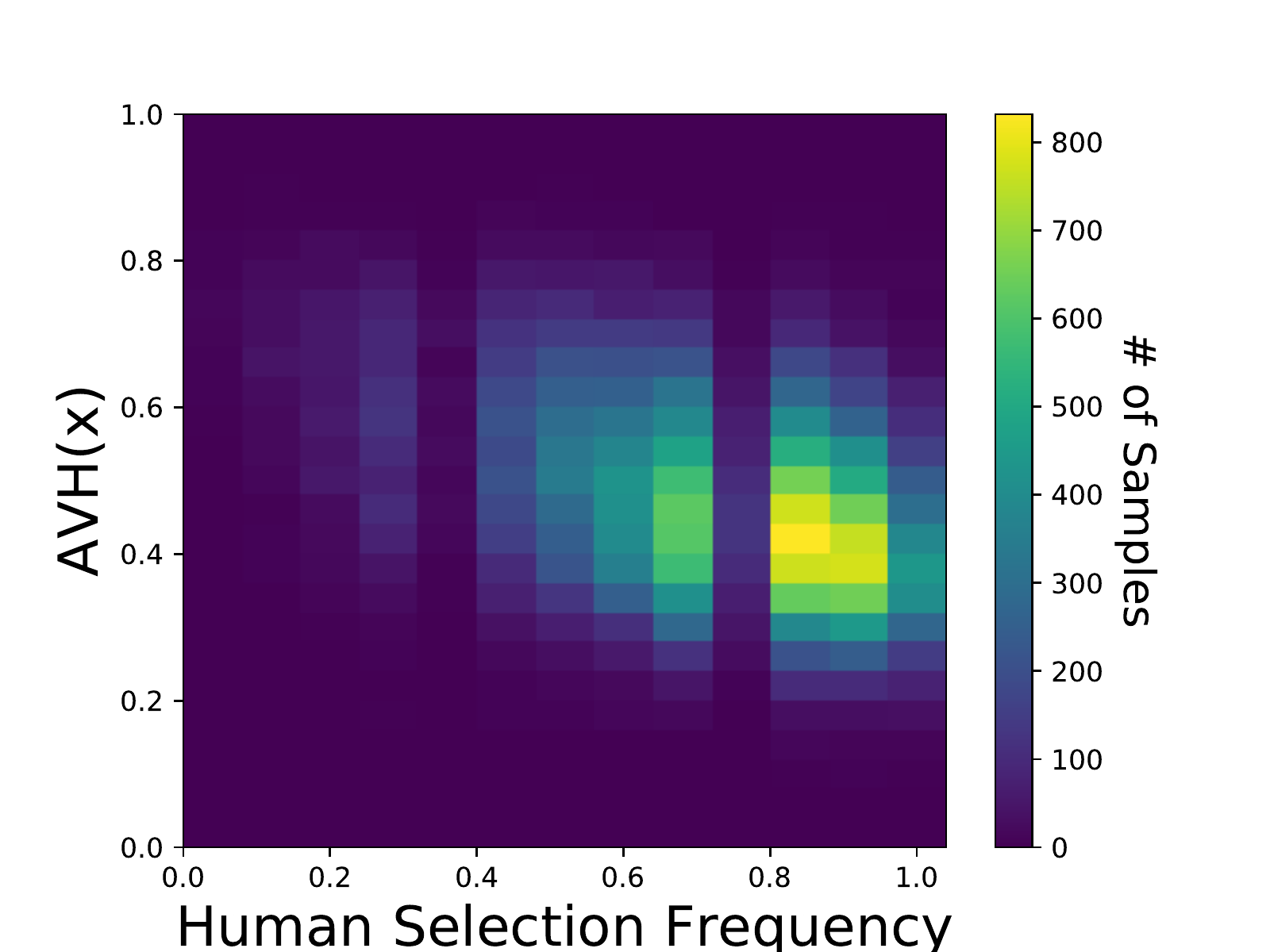} 
		\end{tabular}
	\end{center}
	\caption{\footnotesize The three plots present the correlation between Human Selection Frequency and AVH using AlexNet, VGG-19 and DeseNet121.}\label{fig:angle}
\end{figure*}

\textbf{Additional plots for Section~\ref{sec:correlation}:} Figure~\ref{fig:angle} presents the correlation between Human Selection Frequency and AVH using AlexNet, VGG-19 and DenseNet-121.

\begin{figure*}[h!]
	\begin{center}
		\includegraphics[width=0.246\textwidth]{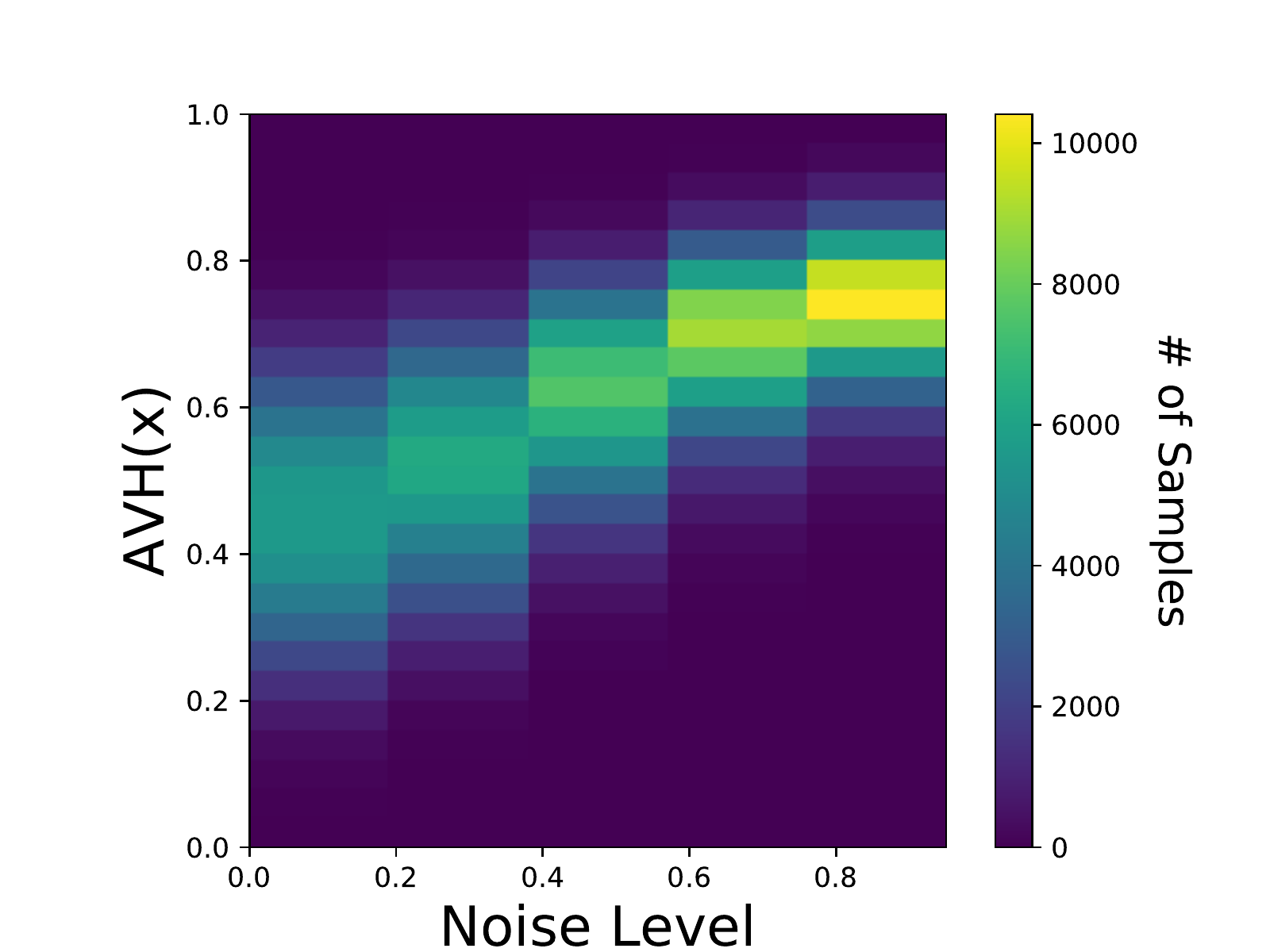}
		\includegraphics[width=0.246\textwidth]{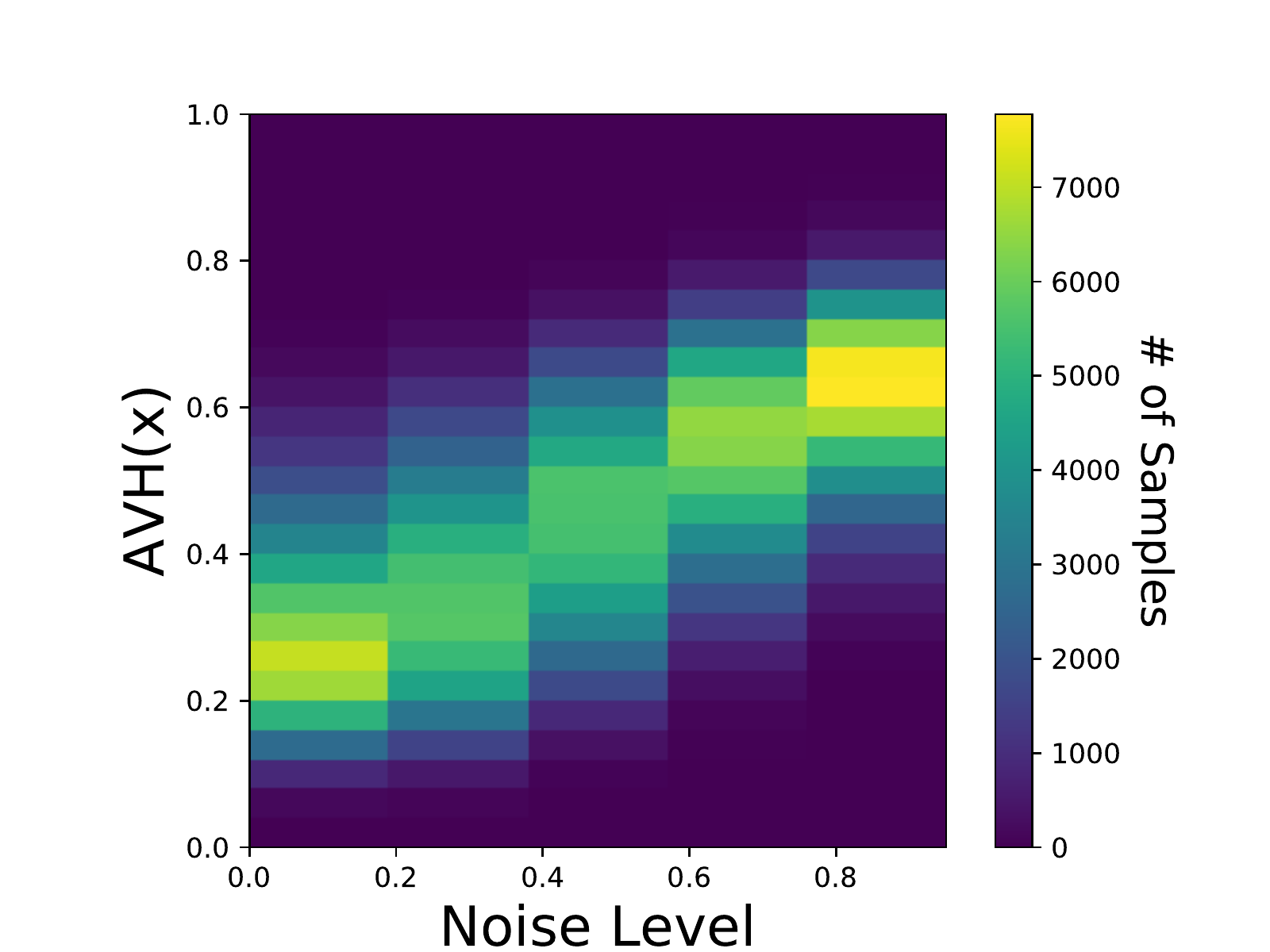}
		\includegraphics[width=0.246\textwidth]{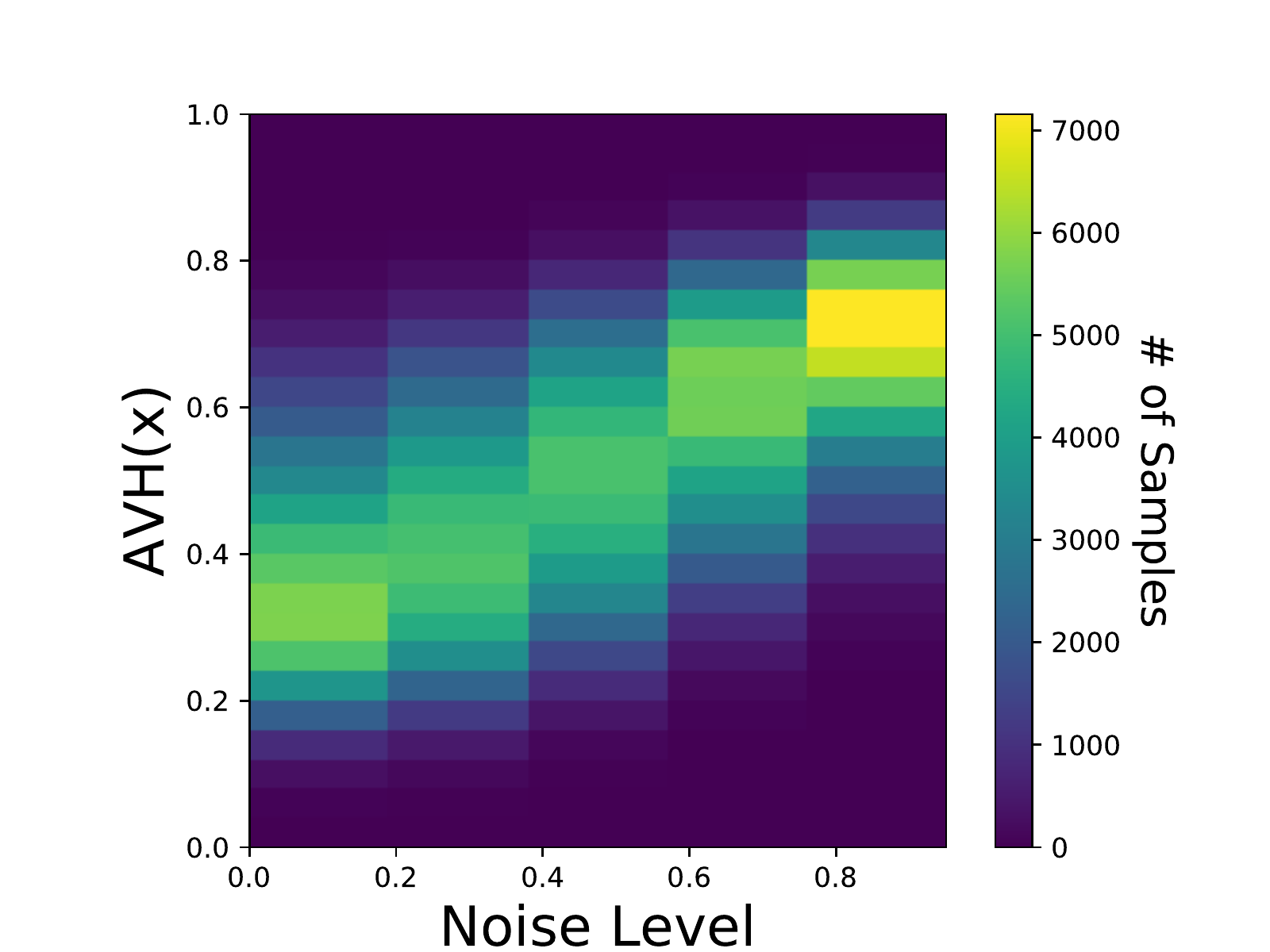}
		\includegraphics[width=0.246\textwidth]{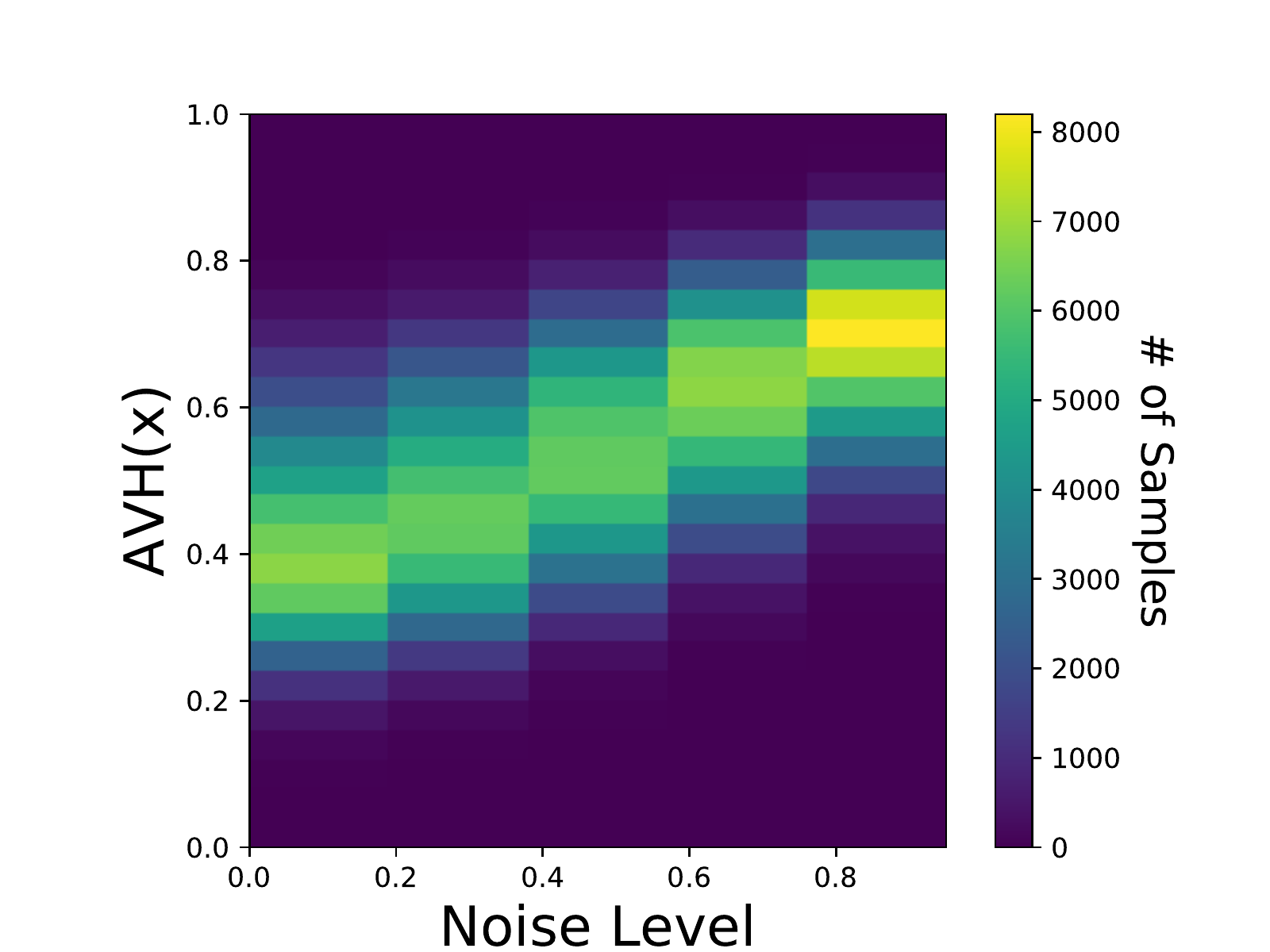}
	\end{center}
	\caption{\footnotesize Correlation between noise degradation levels and AVH scores on AlexNet, VGG-19, ResNet-50 and DenseNet-121. Note that the larger the noise level is, the harder a human can recognize the image.}\label{fig:uniform}
\end{figure*}

\begin{figure*}[h!]
	\begin{center}
		\includegraphics[width=0.246\textwidth]{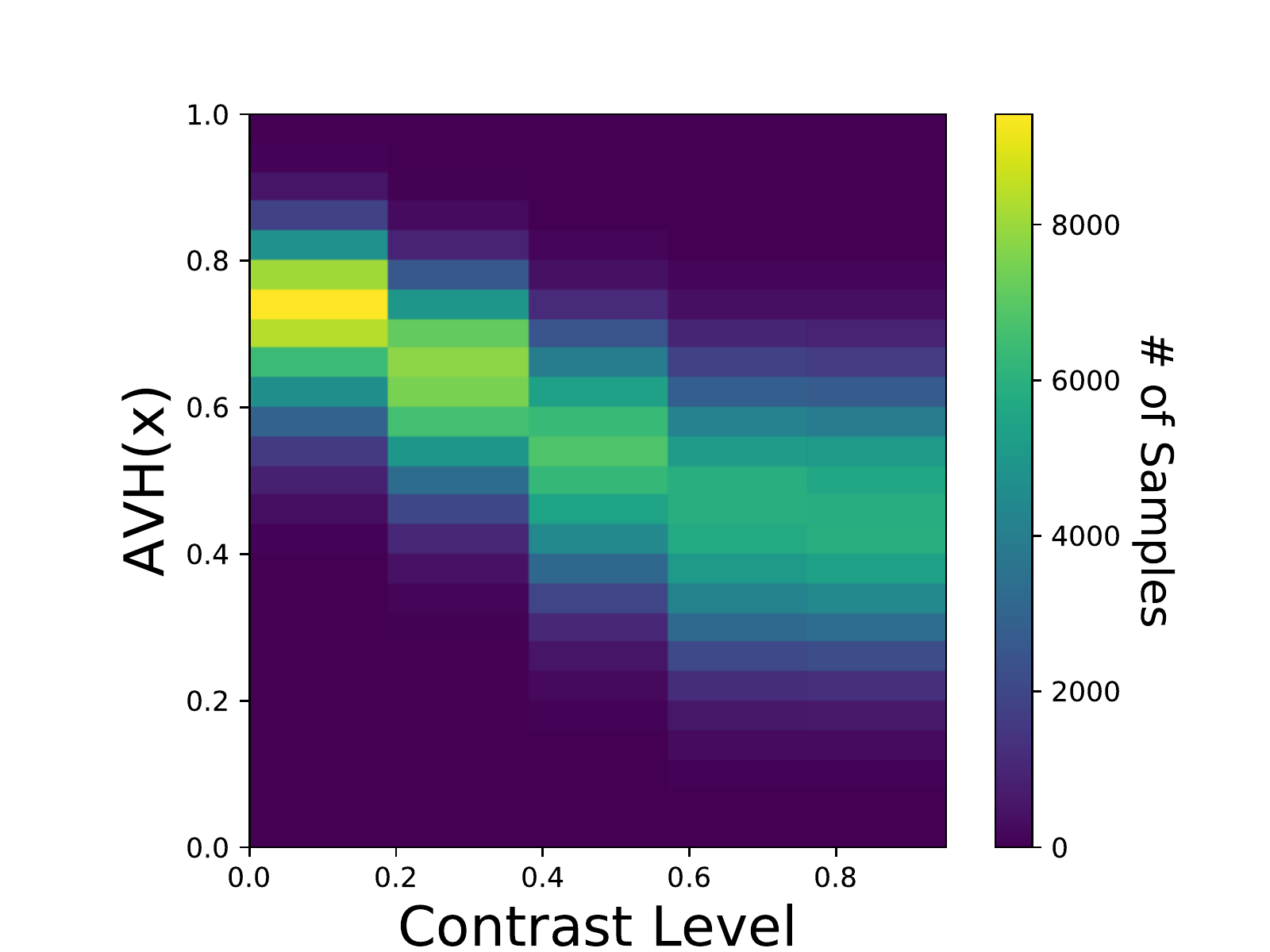}
		\includegraphics[width=0.246\textwidth]{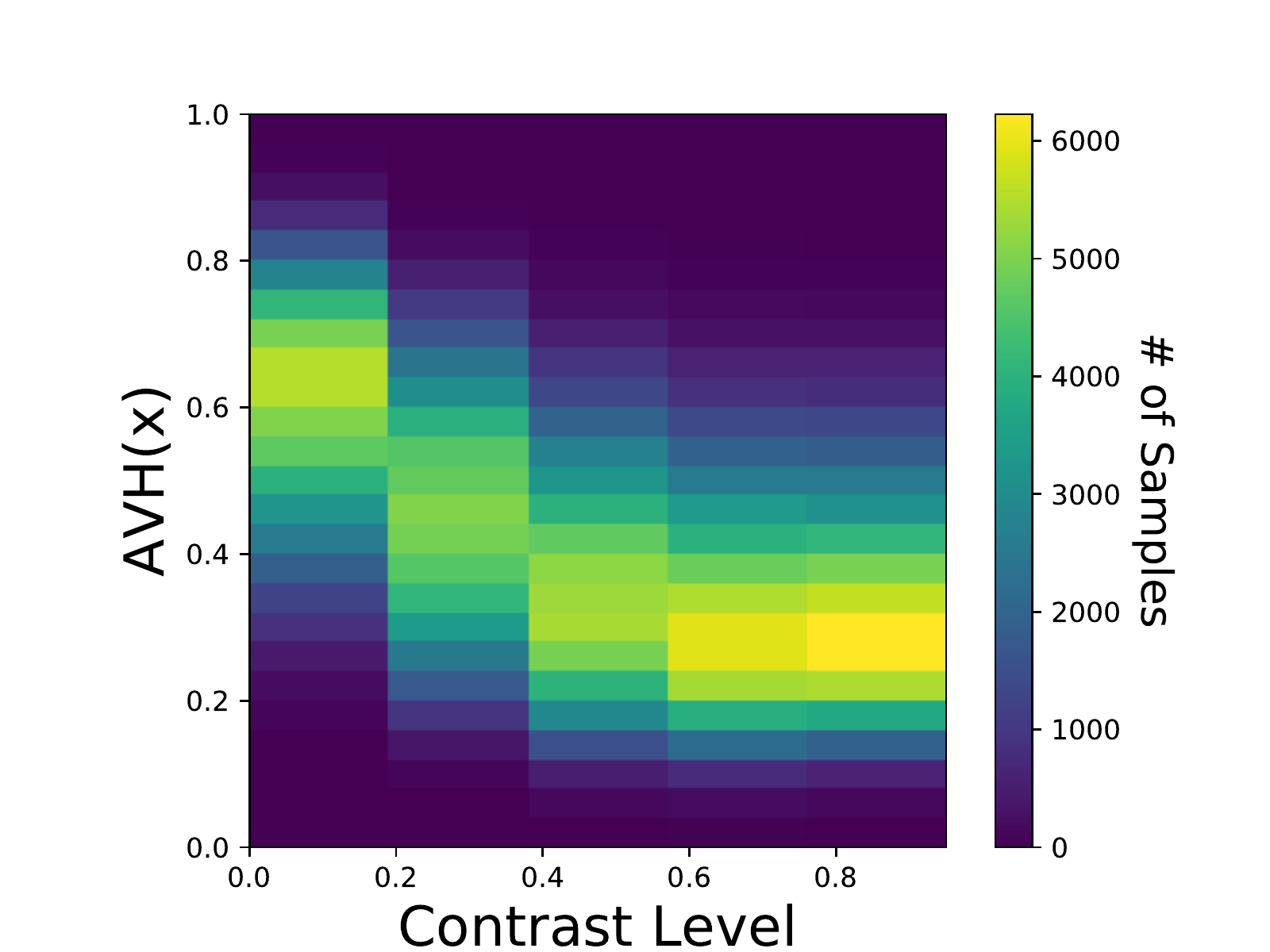}
		\includegraphics[width=0.246\textwidth]{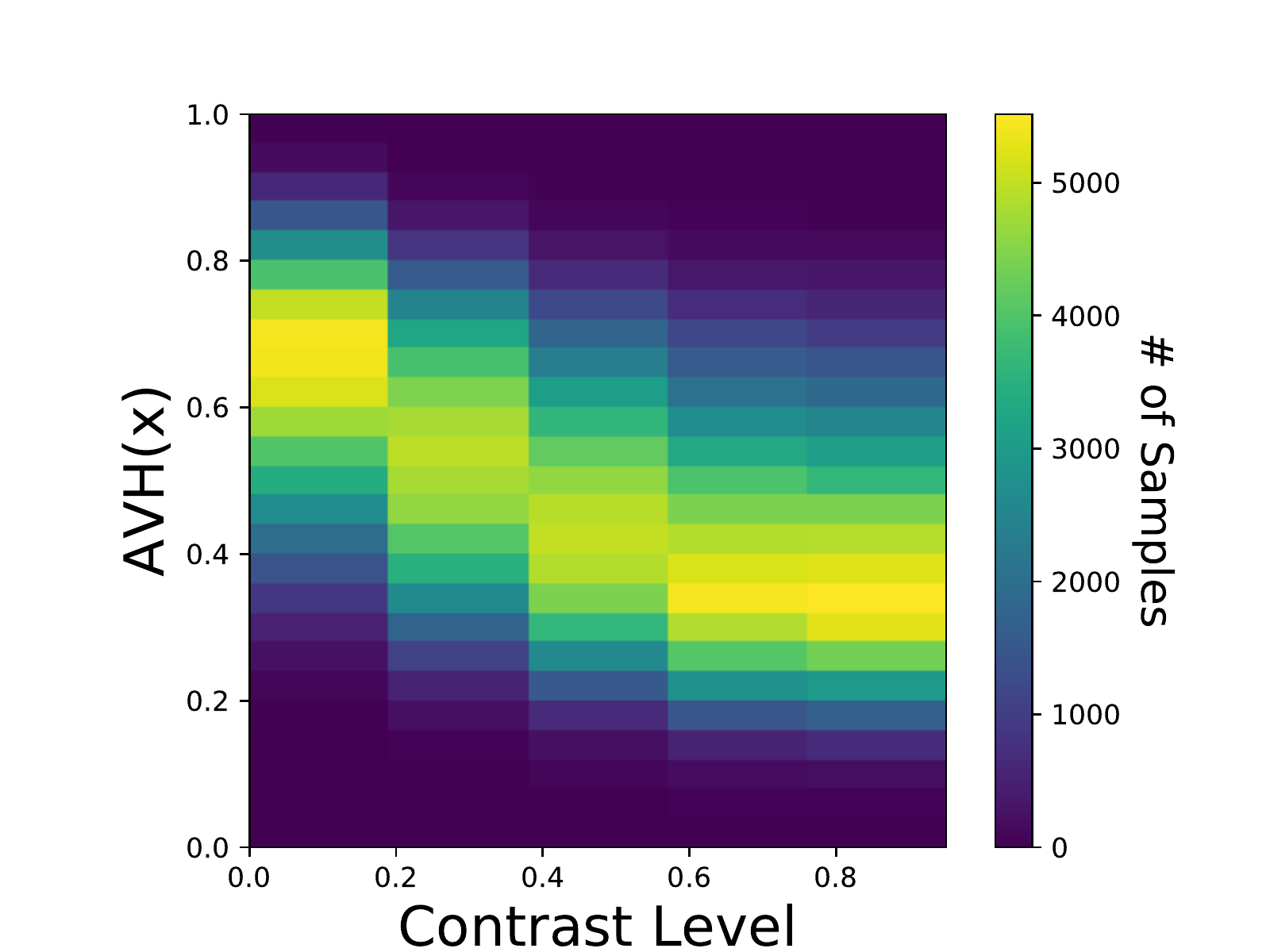}
		\includegraphics[width=0.246\textwidth]{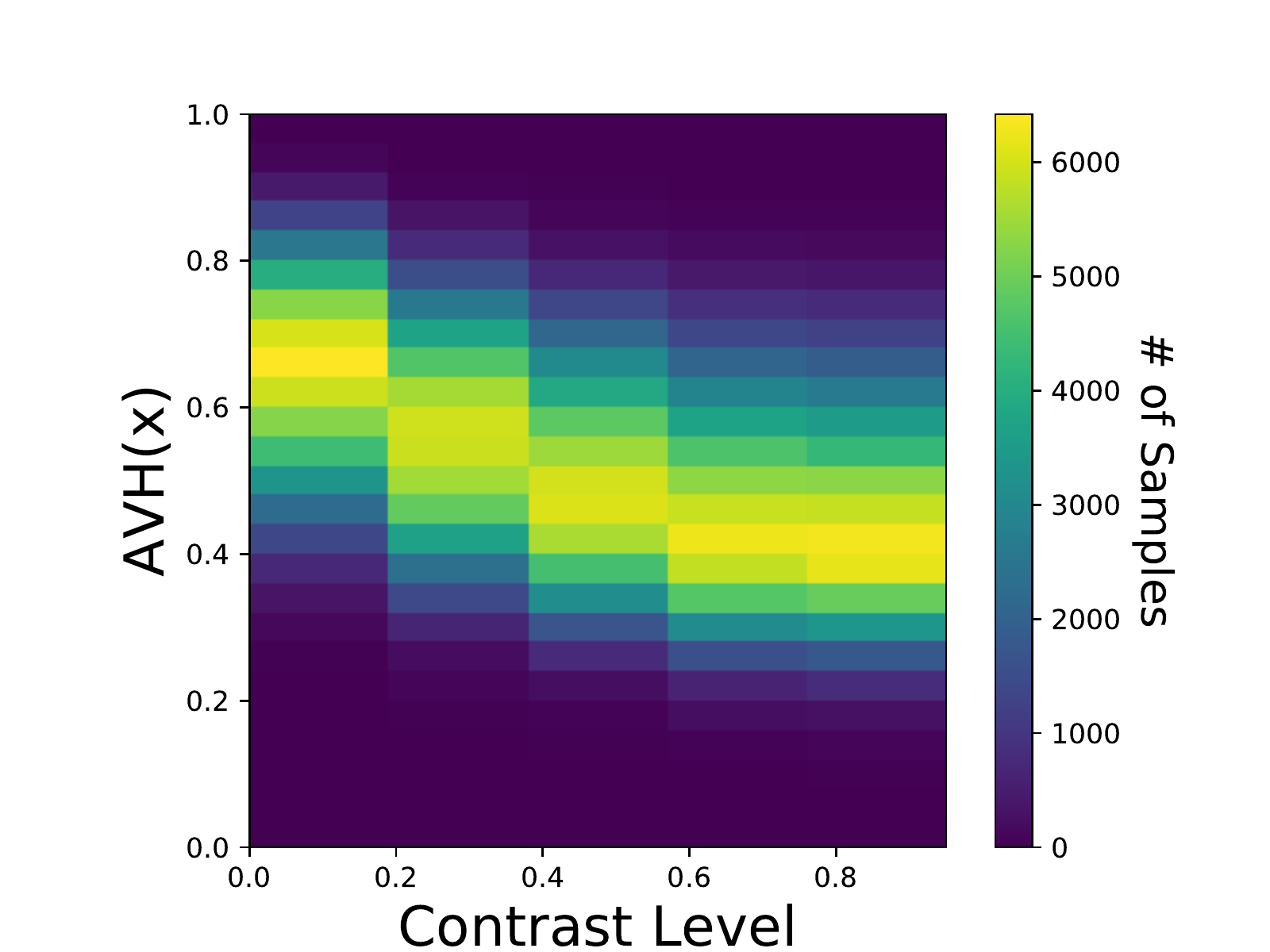}
	\end{center}
	\caption{\footnotesize Correlation between contrast degradation levels and AVH scores on AlexNet, VGG-19, ResNet-50 and DenseNet-121. Note that the larger the contrast Level is, the easier a human can recognize the image.}\label{fig:contrast}
\end{figure*}

\textbf{Correlation between AVH and image degradation:} In order to test if the results in Figure~\ref{fig:freq} from the main paper also hold on proxies other than human visual hardness (image degradation level), we perform the similar experiments but on the augmented ImageNet validation set. Figure~\ref{fig:uniform} shows the correlation between $\mathcal{AVH}(\mathbf{x})$ and different noise degradation levels, while the plots in Figure~\ref{fig:contrast} shows the correlation between $\mathcal{AVH}(\mathbf{x})$ and different contrast degradation levels. Along with Figure~\ref{fig:angle}, these results all indicate that $\mathcal{AVH}(\mathbf{x})$ is a reliable measure of Human Visual Hardness.

\clearpage

\begin{figure*}[h!]
	\begin{center}
		\begin{tabular}{ccc}
			\includegraphics[width=0.3 \textwidth]{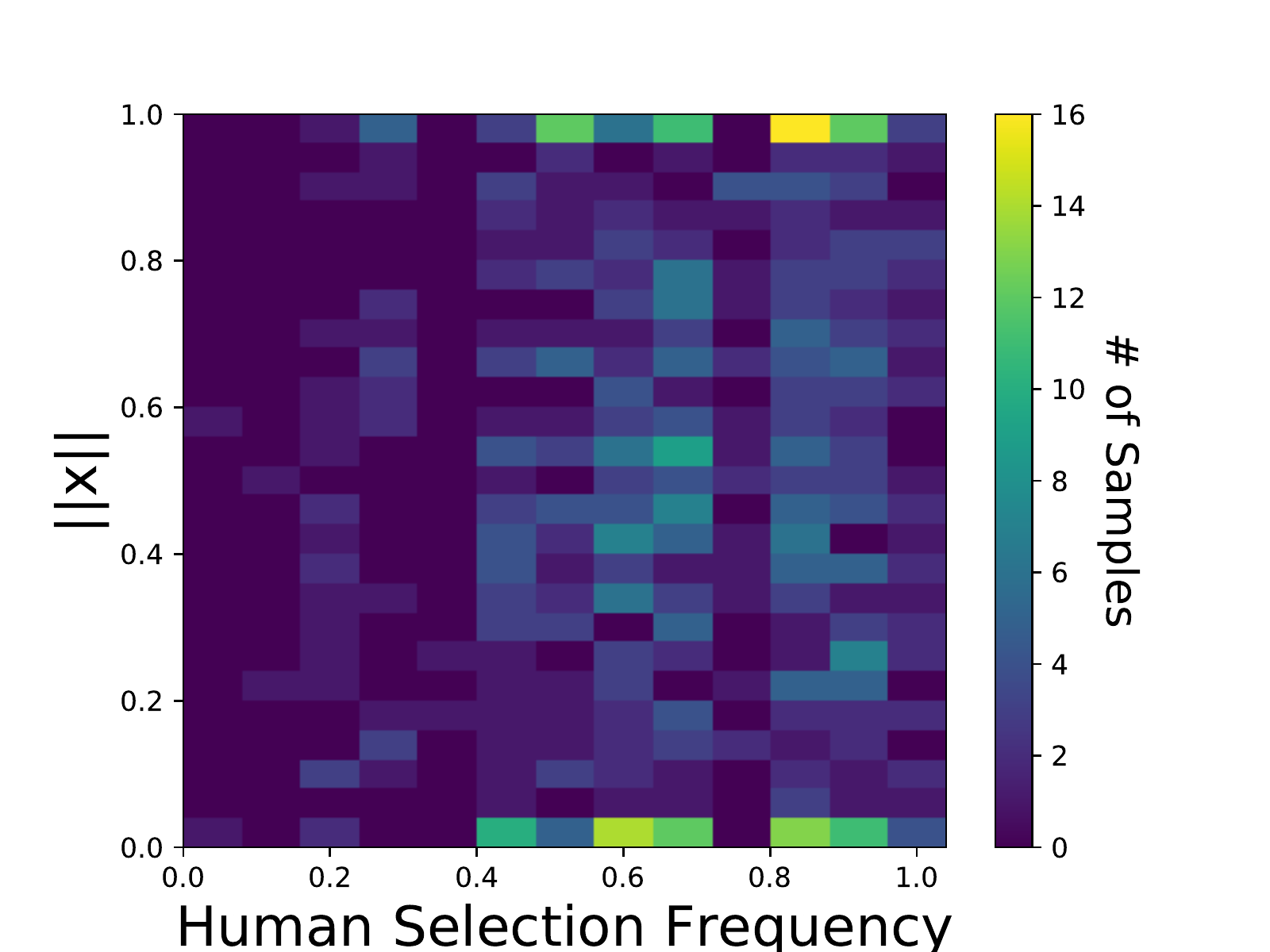} & 
			\includegraphics[width=0.3 \textwidth]{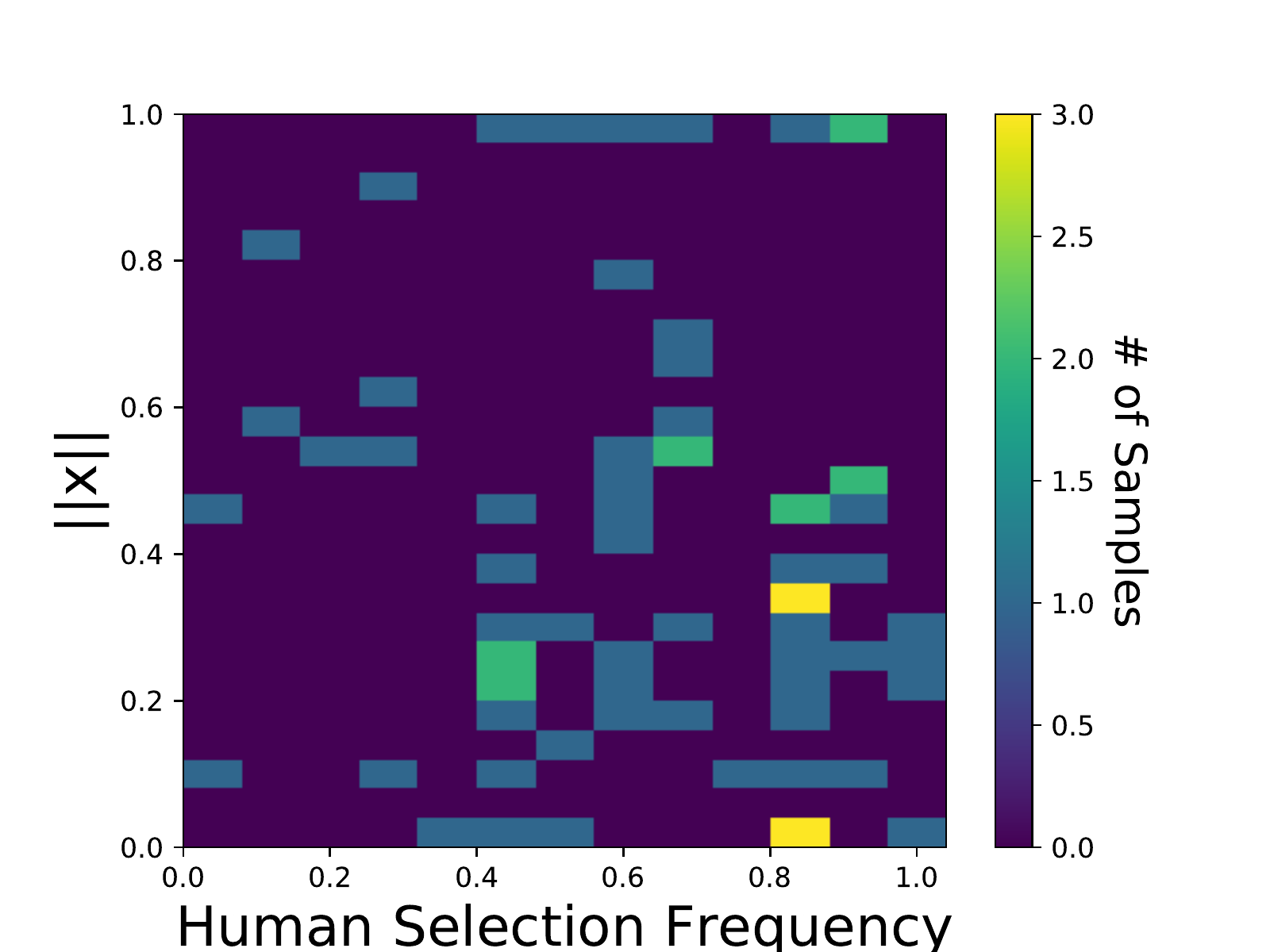} &
			\includegraphics[width=0.3 \textwidth]{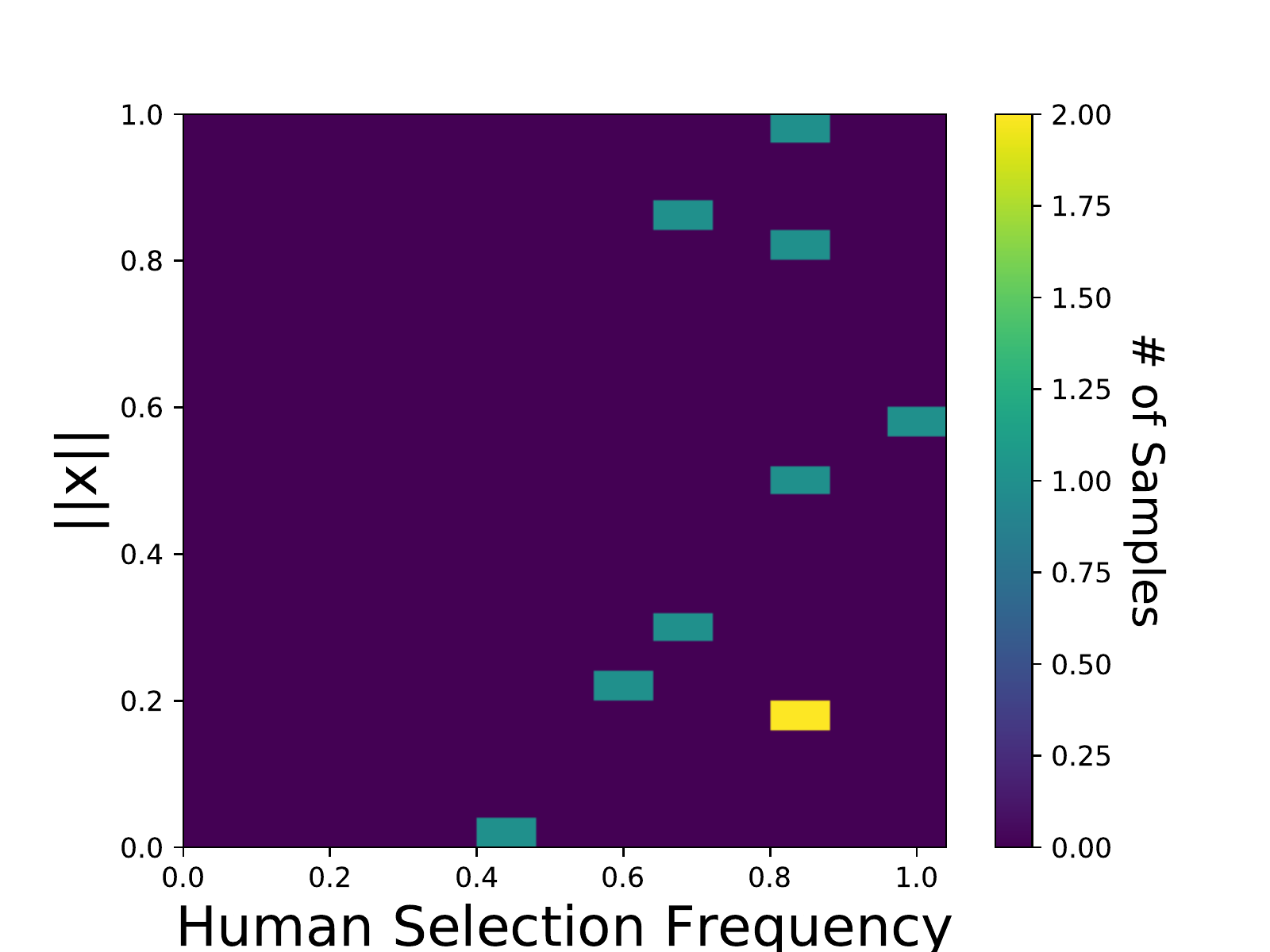} 
		\end{tabular}
	\end{center}
	\caption{\footnotesize $\ell_2$ norm of the embedding vs. Human Selection Frequency under different class granularities (according to WordNet hierarchy). From left to right, there are 58, 7, 1 classes respectively. Human Selection Frequency is therefore computed based on the new class granularity.}\label{fig:angle_freq}
\end{figure*}

\textbf{Additional plots for Hypothesis~\ref{hypo:4}:} We further verify if presenting all samples across $1000$ different classes affects the visualization of the correlation. According to WordNet~\cite{fellbaum2005wordnet} hierarchy, we map the original $1000$ fine-grained classes to $45$ higher hierarchical classes. Figure \ref{fig:angle_freq} exhibits the relationship between Human Selection Frequency and $\Vert x \Vert_2$ for three representative higher classes containing $58, 7, 1$ fine-grained classes respectively. Noted that there is still not any visible direct proportion between these two variables across all plots.

\clearpage
\section{Additional discussions on the Difference between AVH and Model Confidence}
\label{sec:extradis}
The difference between AVH and Model Confidence lies in the feature norm and its role during training. To illustrate the difference, we consider a simple binary classification case where the softmax score (i.e., Model Confidence) for class 1 is 
$$
\frac{\exp(\bm{w}_1\bm{x})}{\sum_i\exp(\bm{w}_i\bm{x})}=\frac{\exp(\|\bm{w}_1\|\|\bm{x}\|\cos(\theta_{\bm{w}_1,\bm{x}}))}{\sum_i\exp(\|\bm{w}_i\|\|\bm{x}\|\cos(\theta_{\bm{w}_i,\bm{x}}))}
$$ 
where $\bm{w}_i$ is the classifier weights of class $i$, $\bm{x}$ is the input deep feature and $\theta_{\bm{w}_i,\bm{x}}$ is the angle between $\bm{w}_i$ and $\bm{x}$. To simplify, we assume the norm of $\bm{w}_1$ and $\bm{w}_2$ are the same, and then the classification result is based on the angle now. Once $\theta_{\bm{w}_1,\bm{x}}$ is smaller than $\theta_{\bm{w}_2,\bm{x}}$, the network will classify the sample $\bm{x}$ as class 1. However, in order to further minimize the cross-entropy loss after making $\theta_{\bm{w}_1,\bm{x}}$ smaller than $\theta_{\bm{w}_2,\bm{x}}$, the network has a trivial solution: increasing the feature norm $\|\bm{x}\|$ instead of further minimizing the $\theta_{\bm{w}_1,\bm{x}}$. It is obviously a much more difficult task to minimize $\theta_{\bm{w}_1,\bm{x}}$ rather than increasing $\|\bm{x}\|$. Therefore, the network will tend to increase the feature norm $\|\bm{x}\|$ to minimize the cross-entropy loss, which is equivalent to maximizing the Model Confidence in class 1. In fact, this also matches our empirical observation that the feature norm keeps increasing during training. Most importantly, one can notice that AVH will stay unchanged no matter how large the feature norm $\|\bm{x}\|$ is. Moreover, this also matches our empirical result that AVH easily gets saturated while Model Confidence can keep improving. Therefore, AVH is able to better characterize the visual hardness, since it is trivial for the network to increase feature norm. This is the fundamental difference between Model Confidence and AVH.
\par
To get a more intuitive sense of how feature norm can affect the Model Confidence, we plot the value of the Model Confidence for two scenarios: $\theta_{\bm{w}_1,\bm{x}}<\theta_{\bm{w}_2,\bm{x}}$ and $\theta_{\bm{w}_1,\bm{x}}>\theta_{\bm{w}_2,\bm{x}}$. Under the case that the sample $\bm{x}$ belongs to class 1, once we have $\theta_{\bm{w}_1,\bm{x}}<\theta_{\bm{w}_2,\bm{x}}$, then we only need to increase the feature norm and can easily get nearly perfect confidence on this sample. In contrast, AVH will stay unchanged during the entire process and therefore is a more robust indicator for visual hardness than Model Confidence.

\begin{figure*}[h]
  \centering
   \setlength{\abovecaptionskip}{3pt}
   \setlength{\belowcaptionskip}{-10pt}
  \includegraphics[width=0.7\textwidth]{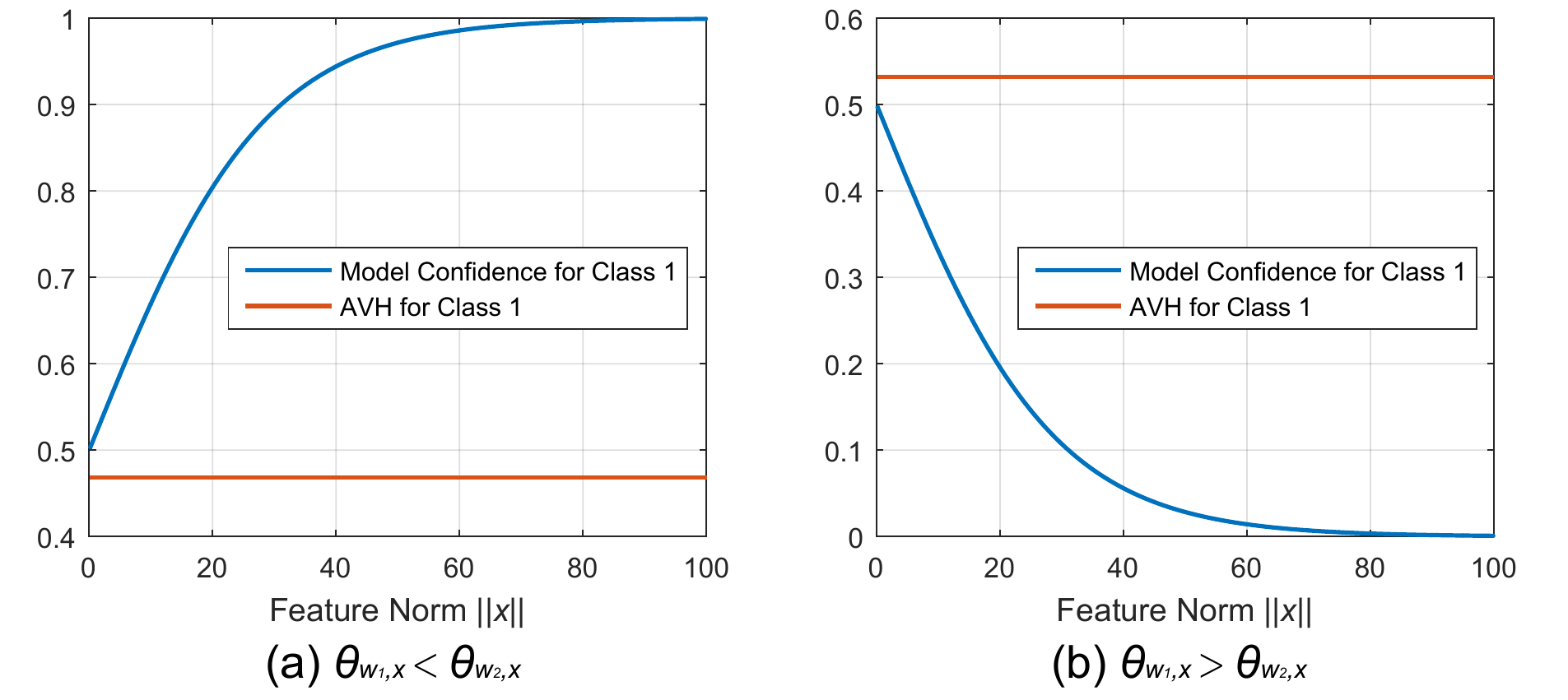}
  \caption{\footnotesize The comparison between AVH and Model Confidence when the feature norm keeps increasing. The figure is plotted according to the binary classification example discussed above. We assume $\|\bm{w}_1\|=\|\bm{w}_1\|$. When $\theta_{\bm{w}_1,\bm{x}}<\theta_{\bm{w}_2,\bm{x}}$, we use $\theta_1=\pi/4-0.05$ and $\theta_2=\pi/4+0.05$. When $\theta_{\bm{w}_1,\bm{x}}>\theta_{\bm{w}_2,\bm{x}}$, we use $\theta_1=\pi/4+0.05$ and $\theta_2=\pi/4-0.05$. Note that, unlike Model Confidence, the smaller AVH is, the more confident the network is (\ie, the easier the sample is).}\label{appendixD}
\end{figure*}

\clearpage
\newpage

\section{Experimental Details}
\label{app:F}
\begin{wrapfigure}{R}{5.5cm}
\vspace{-5mm}
  \centering
  \includegraphics[width=5.5cm]{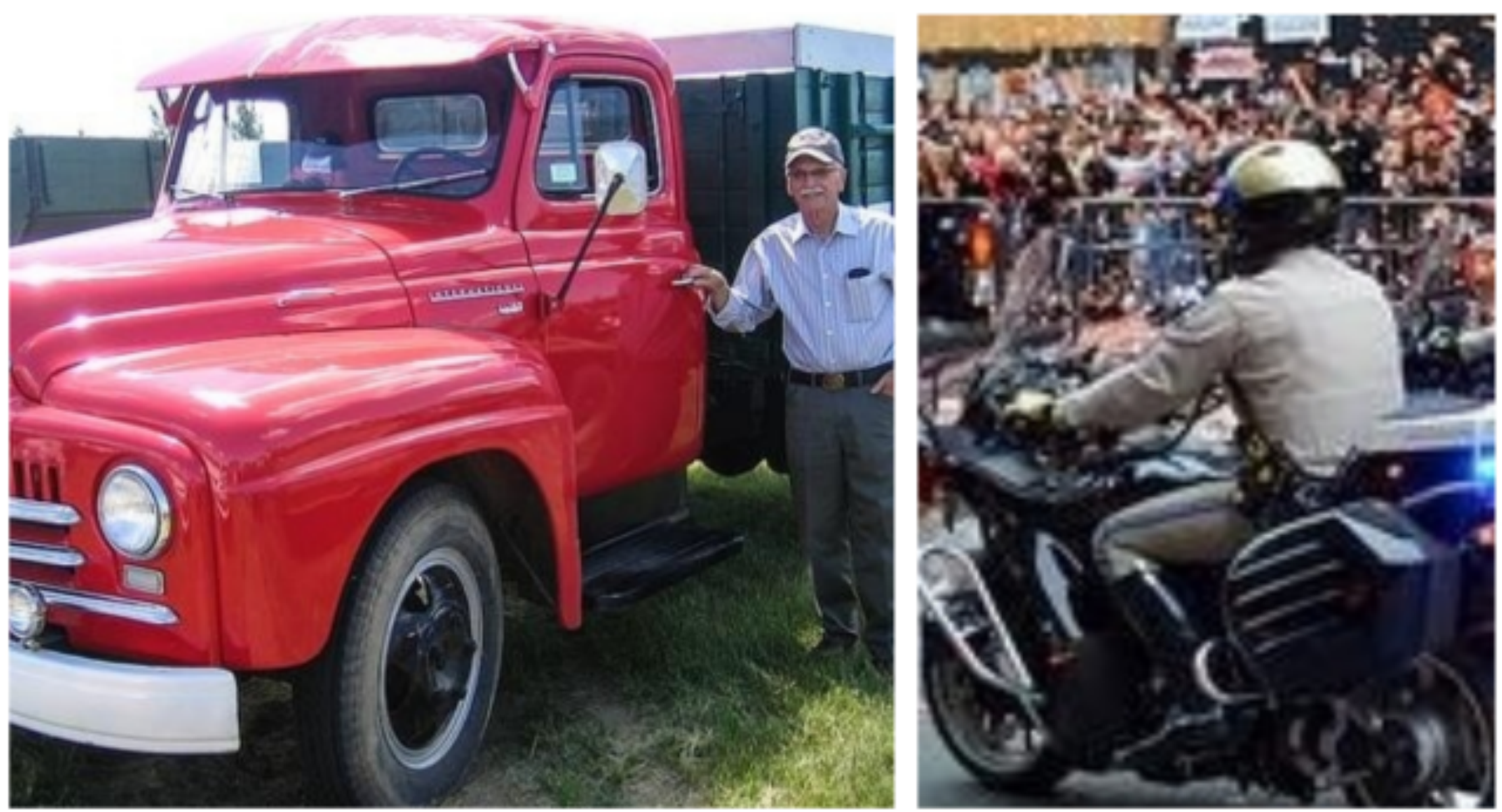}
  \caption{\footnotesize Two example images which AVH selects but softmax score do not. The left one has the true label ``Truck" but is easy to be confused with the ``Car". The right one has the true label ``Person" but is easy to be confused with the ``Motor".}\label{confusing}
  \vspace{-3mm}
\end{wrapfigure}
\textbf{Self-training and domain adaptation:} As mentioned in section 5, a major challenge of self-training is the amplification of error due to misclassified pseudo-labels. Therefore, traditional self-training methods such as CBST often use Model Confidence as the measure to select confidently labeled examples. The hope is that higher confidence potentially implies lower error rate. While this generally proves useful, the model tends to focus on the ``less informative'' samples, whereas ignoring the ``more informative'', harder ones near classier boundaries that could be essential for learning a better classifier. Figure~\ref{confusing} shows examples of what AVH selects and labels correctly but the softmax score does not select in CBST. We can see they are all visually confusing examples which can better help with the iterative self-training process when pseudo labeled correctly. The left one has the true label ``Truck" but is easy to be confused with the ``Car". The right one has the true label ``Person" but is easy to be confused with the ``Motor". 

\textbf{Domain generalization:} For domain generalization, we use the PACS benchmark dataset~\cite{li2017deeper} which contains consists of art painting, cartoon, photo and sketch domains. Each domain has the same 7 classes. Our experimental settings basically follow \cite{li2017deeper}. Specifically, we pick one domain as the unseen testing domain and train our model on the remaining three domains. The testing accuracy is evaluated on the unseen testing domain. Therefore, we will have 4 testing accuracies in total and we can use the average accuracy as the final evaluation metric. We use a convolutional neural network similar to~\cite{liu2017deep} with the detailed structure of [7$\times$7, 64] $\Rightarrow$ 2$\times$2 Max Pooling $\Rightarrow$ [3$\times$3, 64]$\times$3 $\Rightarrow$ 2$\times$2 Max Pooling $\Rightarrow$ [3$\times$3, 128]$\times$3 $\Rightarrow$ 2$\times$2 Max Pooling $\Rightarrow$ [3$\times$3, 256]$\times$3 $\Rightarrow$ 2$\times$2 Max Pooling $\Rightarrow$ 512-dim Fully Connected. For example, [3$\times$3, 64]$\times$3 denotes 3 cascaded convolution layers with 64 filters of size 3$\times$3. We use momentum SGD with momentum as $0.9$ and batch size $40$. Batch normalization and ReLU activation are used by default. Following the existing methods~\cite{li2017deeper,li2018learning,balaji2018metareg,li2019feature}, we will first pretrain our network on ImageNet with standard learning rate and decay, and then finetune on the PACS dataset with batch size $40$ and smaller learning rate ($1e-3$).

\clearpage
\section{Extensions and Applications}
\label{app:g}

\textbf{Adversarial Example: A Counter Example?}
\begin{wrapfigure}{R}{5.5cm}
\vspace{-3mm}
  \centering
  \includegraphics[width=5.5cm]{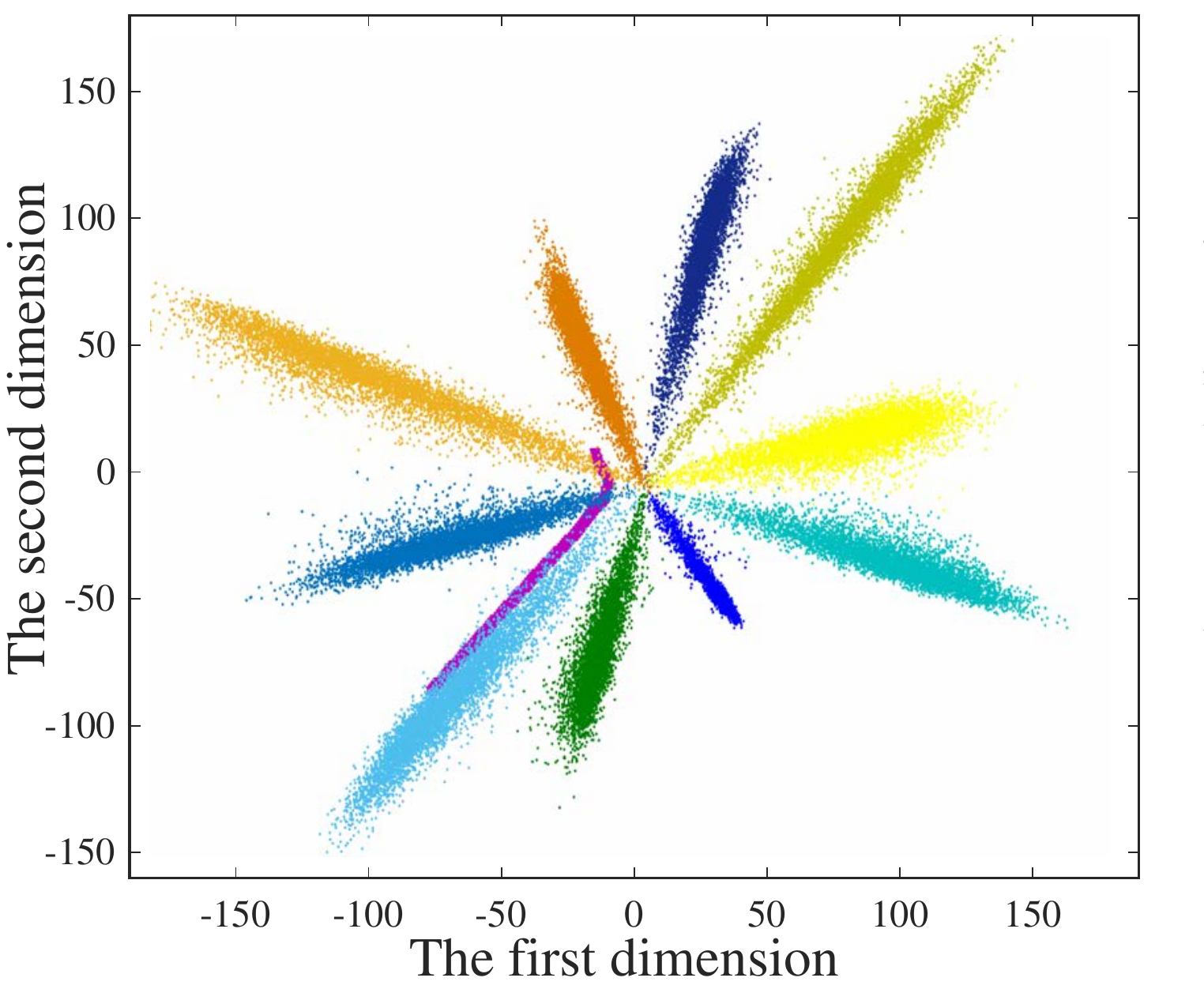}
  \caption{\footnotesize Trajectory of an adversarial example switching from one class to another. The purple line denotes the trajectory of the adversarial example.}\label{adv_mnist}
  \vspace{-3mm}
\end{wrapfigure}

Our claim about the stronger correlation between $\text{AVH}$ score and human visual hardness does not apply on non-natural images such as adversarial examples. For such examples, the human can not tell the difference visually, but the adversarial example has a worse $\text{AVH}$ than the original image, which runs counter to our claim  that AVH has strong correlation with human visual hardness. So this claim is limited to distribution of natural images. However, on a positive note, we do find that AVH is slower to change compared to the embedding norm during the dynamics of adversarial training. 

We show a special case in Figure~\ref{adv_mnist} to illustrate how the norm and the angle change when one sample switches from one class to another. Specifically, we change the sample from one class to another using adversarial perturbation. It is essentially performing gradient ascent to the ground truth class. In Figure~\ref{adv_mnist}, the purple line denotes the trajectory of an adversarial sample switching from one class to another. We can see that the sample will first shrink its norm towards origin and then push its angle away from the ground truth class. Such a trajectory indicates that the adversarial sample will first approach to the origin in order to become a hard sample for this class. Then the sample will change the angle in order to switch its label. This special example fully justifies the importance of both norm and angle in terms of the hardness of samples.

\begin{wrapfigure}{R}{5.5cm}
\vspace{-3mm}
	\begin{center}
		\begin{tabular}{ccc}
		\hspace{-0.2cm}
			\includegraphics[width=5.5cm]{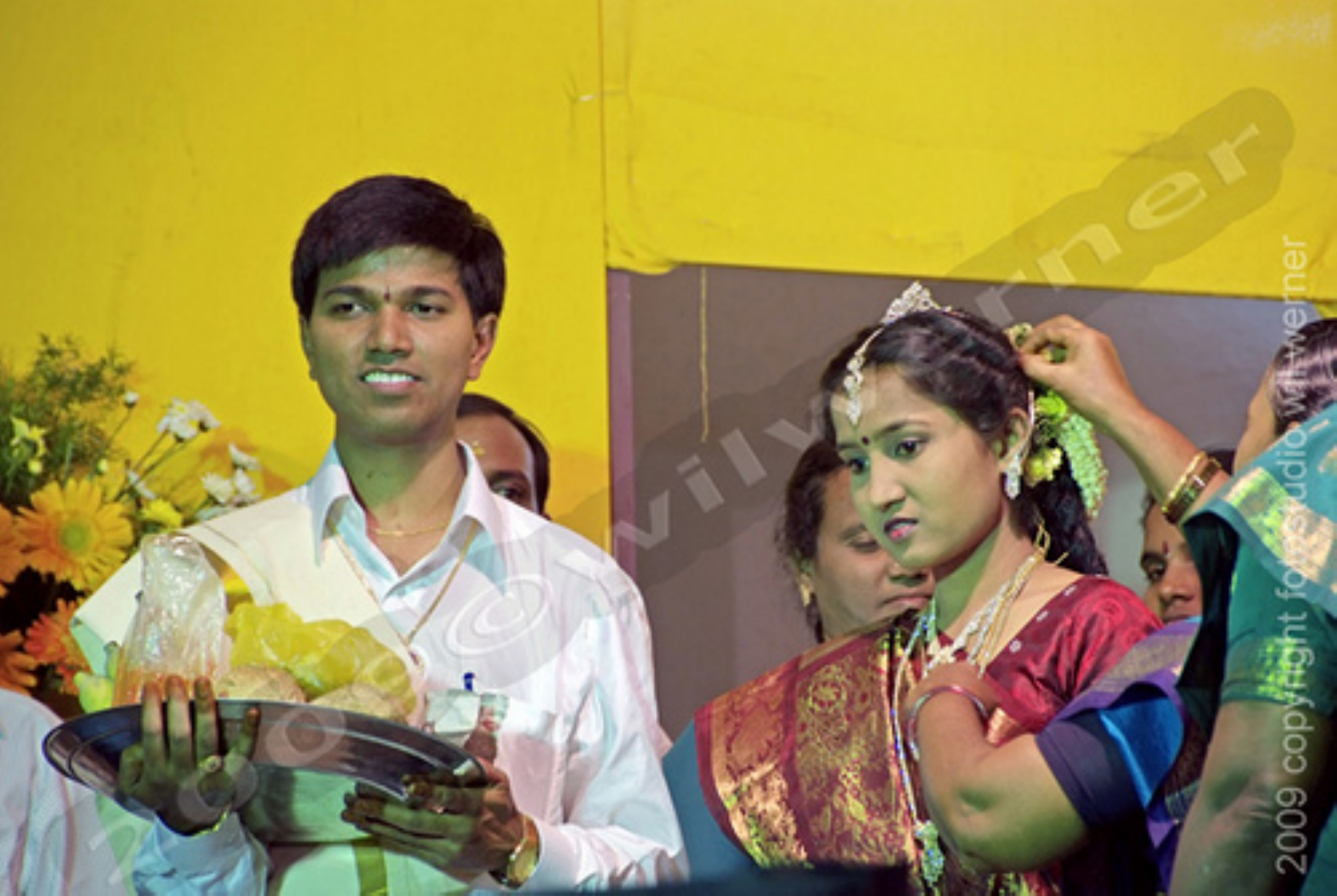}
		\end{tabular}
	\end{center}
 	\vspace{-4mm}
	\caption{\footnotesize An image of an Indian Groom from ImageNet.}
	\label{fig:groom}
  \vspace{-3mm}
\end{wrapfigure}

\par
\textbf{Measuring Human Visual Hardness is Hard:} Measuring Human Visual Hardness is non-trivial and dependent on many factors such as (i) How much are the annotators penalized for wrong answers and how much time are they given? (ii) What are the cultural and language differences that can cause annotators to be confused about the label categories. Figure~\ref{fig:groom} shows an example of groom from ImageNet dataset. Since a large contingent of Mturk users are from India, they have high confidence for this image, but the answer would be very different if asked a different population. The proxies we used in this paper, Human Selection Frequency and Image Degradation Level are best efforts.  

\textbf{Connection to deep metric learning:} Measuring the hardness of samples is also of great importance in the field of deep metric learning~\cite{oh2016deep,sohn2016improved,wu2017sampling}. For instance,   objective functions in deep metric learning consist of \eg, triplet loss~\cite{schroff2015facenet} or contrastive loss~\cite{hadsell2006dimensionality}, which requires data pair/triplet mining in order to perform well in practice. One of the most widely used data sampling strategies is semi-hard negative sample mining~\cite{schroff2015facenet} and hard negative sample mining. These negative sample mining techniques highly depend on how one defines the hardness of samples.  $\text{AVH}$ can be potentially useful in this setting.

\par
\textbf{Connections to fairness in machine learning:} Easy and hard samples can implicitly reflect imbalances in latent attributes in the dataset. For example, the CASIA-WebFace dataset~\cite{yi2014learning} mostly contains white celebrities, so the neural network trained on CASIA-WebFace is highly biased against the other races.~\cite{buolamwini2018gender}  demonstrates a performance drop of faces of darker people due to the biases in the training dataset. In order to ensure fairness and remove dataset biases, the ability to identify hard samples automatically can be very useful. We would like to test if AVH is effective in these settings.   

\par
\textbf{Connections to knowledge transfer and curriculum learning:} The efficiency of knowledge transfer~\cite{hinton2015distilling} is partially determined by the sequence of input training data. \cite{liu2017iterative} theoretically shows feeding easy samples first and hard samples later (known as curriculum learning) can improve the convergence of model.~\cite{bengio2009curriculum} also show that the curriculum of feeding training samples matters in terms of both accuracy and convergence. We plan to investigate the use of AVH metric in such settings. 

\end{appendix}